\documentclass{article}
\usepackage{amsmath,amssymb,amsthm}

    \usepackage[final,nonatbib]{neurips_2020_tda}

\usepackage[utf8]{inputenc} % allow utf-8 input
\usepackage[T1]{fontenc}    % use 8-bit T1 fonts
\usepackage{amsfonts}       % blackboard math symbols
\usepackage{booktabs}       % professional-quality tables
\usepackage{hyperref}       % hyperlinks
\usepackage{microtype}      % microtypography
\usepackage{nicefrac}       % compact symbols for 1/2, etc.
\usepackage{paralist}       % in-paragraph enumerations
\usepackage{siunitx}        % SI units
\usepackage{graphicx}     
\usepackage{grffile}
\usepackage{placeins}
\usepackage{tikz}
\usetikzlibrary{arrows.meta,bending,automata}
\tikzset{style green/.style={
    set fill color=green!50!lime!60,
    set border color=white,
  },
  style cyan/.style={
    set fill color=cyan!90!blue!60,
    set border color=white,
  },
  style orange/.style={
    set fill color=orange!80!red!60,
    set border color=white,
  },
  hor/.style={
    above left offset={-0.15,0.31},
    below right offset={0.15,-0.125},
    #1
  },
  ver/.style={
    above left offset={-0.1,0.3},
    below right offset={0.15,-0.15},
    #1
  }
}
\usepackage[customcolors]{hf-tikz}

\newtheorem{thm}{Theorem}%
\theoremstyle{definition}
\newtheorem{example}[thm]{Example}%

\newcommand{\PH}{\operatorname{PH}}
\newcommand{\PI}{\operatorname{PI}}

\title{Can neural networks learn persistent homology features?} 

\author{Guido Mont\'ufar\\
	UCLA Math / Stats and MPI MIS\\
	montufar@math.ucla.edu
	\And
	Nina Otter \\
	UCLA Math\\
	otter@math.ucla.edu
	\And
	Yuguang Wang\\
MPI MIS\\
yuguang.wang@mis.mpg.de
}

\begin{document}

\maketitle

\begin{abstract}
Topological data analysis uses tools from topology --- the mathematical area that studies shapes ---  to create representations of data. In particular, in persistent homology, one studies one-parameter families of spaces associated with data, and persistence diagrams describe the lifetime of topological invariants, such as connected components or holes, across the one-parameter family. In many applications, one is interested in working with features associated with persistence diagrams rather than the diagrams themselves. In our work, we explore the possibility of learning several types of features extracted from persistence diagrams using neural networks.
\end{abstract}

\section{Introduction}\label{S:intro}
Learning representations of data is a core component of machine learning and data science. 
Topological data analysis (TDA) computes feature representations of data inspired by topology. 
One such representation is in terms of persistence diagrams, which are multisets of points in $\mathbb{R}^2$. The computation of these diagrams is often computationally expensive \cite{otter2017roadmap}. Furthermore, it is not always easy to perform statistical analysis on the space of persistence diagrams, and for subsequent analysis it may be preferable to replace them by suitable approximations. 

In our work, we train neural networks (GNNs, CNNs) to map data to specific representations of the corresponding persistence diagrams. For the data, we focus on raster images, but the idea extends to other types of data, including point clouds. For the representations, we focus on tropical coordinates \cite{Kalisnik} and binary features of  persistence diagrams. Other interesting features can be considered, including persistence images \cite{PI} and features extracted from them, such as the number of ``blobs''. 

There are two main applications of persistence diagrams to the study of data: (i) ``homological inference'', in which one assumes that the data set is sampled from a manifold, and tries to learn properties of the manifold, such as  the number of components or holes, by computing the persistence diagrams. In such a case, one considers the points further from the diagonal to represent significant features; (ii) classification, in which one is not interested in inferring a specific homology type, but rather uses the persistence diagrams to distinguish between different data sets \cite{byttner,dunaeva+, GGM19,lawson+,leon+}. In the latter case, it might be the points at a certain distance from the diagonal that represent the significant features, see for instance \cite{Bendich2016,bubenik2020persistent}. 

In the present work, we are interested in applications of persistent homology to downstream tasks. We use digital images and filtered cubical complexes  as input of different neural networks to learn features of barcodes. We show that neural networks can indeed be trained to produce good approximations of some commonly used persistent homology features, at least in the considered data sets, MNIST and CIFAR-10. Thus, our approach can be  exploited to compute persistent homology  features for data sets for which such computations are traditionally  too expensive. In ongoing work, we take a natural next step, which is to consider a broader range of data sets and persistent homology features and to use the features computed by a neural network as an input for a downstream task, with the possibility also to fine-tune them. 

\begin{figure}[ht!]
\begin{center}
\scalebox{.9}{
\begin{tikzpicture}
[auto,
 block/.style ={rectangle, draw=yellow, thick, fill=yellow!20, text width=5em,align=center, rounded corners, minimum height=2em},
 block1/.style ={rectangle, draw=yellow, thick, fill=yellow!20, text width=5em,align=center, rounded corners, minimum height=2em},
 line/.style ={draw, thick, -latex',shorten >=2pt},
 cloud/.style ={draw=red, thick, ellipse,fill=red!20,
 minimum height=1em}]
 \node at (-0.7,1.5) {\includegraphics[width=0.08\textwidth]{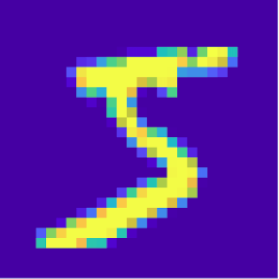}};
   \node at (0.45,1.5)
 {\includegraphics[width=0.08\textwidth]{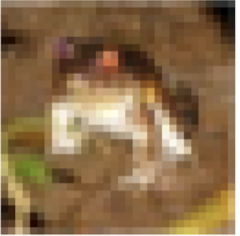}};
 \node at (2.3,1.5) {\includegraphics[width=0.08\textwidth]{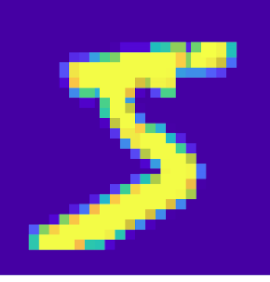}};
  \node at (3.45,1.5) {\includegraphics[width=0.08\textwidth]{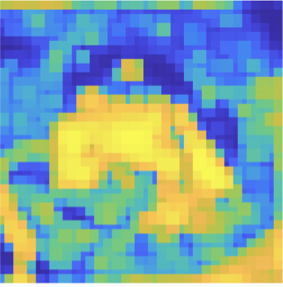}};
 \node at (5.3,1.5) {\includegraphics[width=0.08\textwidth]{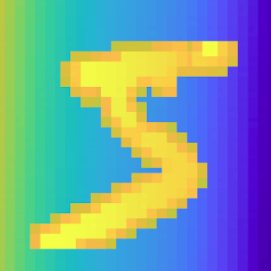}};
  \node at (6.45,1.5) {\includegraphics[width=0.08\textwidth]{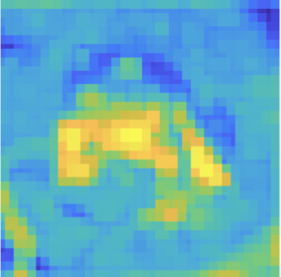}};
  \node at (8.3,1.5)
 {\includegraphics[width=0.08\textwidth]{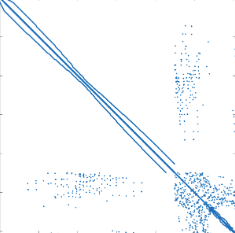}};
  \node at (9.5,1.5)
  {\includegraphics[width=0.08\textwidth]{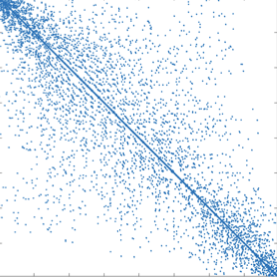}};
    \node at (11.3,1.5)
  {\includegraphics[width=0.08\textwidth]{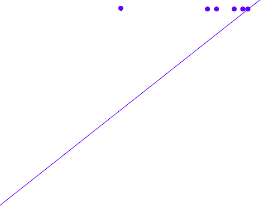}};
    \node at (12.5,1.5)
  {\includegraphics[width=0.08\textwidth]{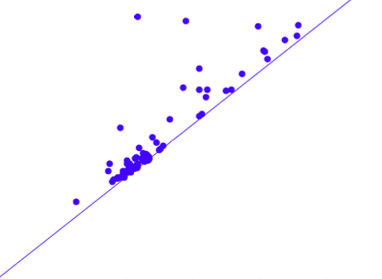}};
  \node at (9.2,-1.5)
  {\includegraphics[width=0.08\textwidth]{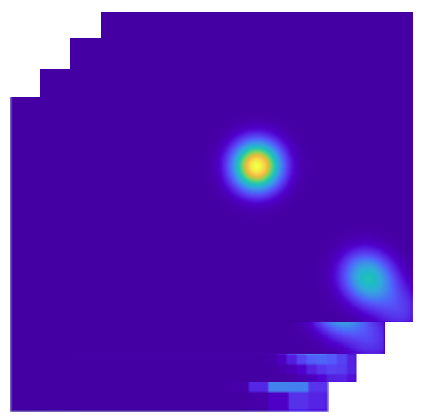}};
   \node at (10.2,-1.5)
  {\includegraphics[width=0.08\textwidth]{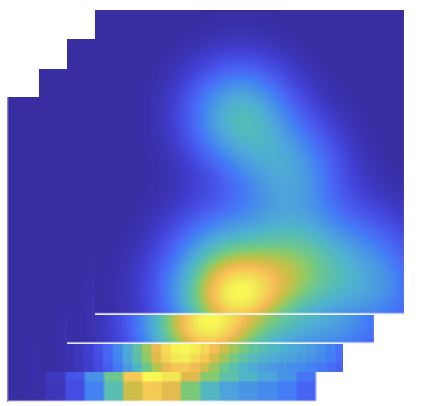}};
    \node at (9.5,-2.6)
  {\includegraphics[width=0.2\textwidth]{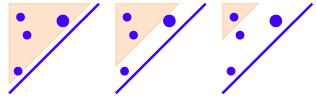}};
  
\draw (0,0) node[block] (A) {original image};
\path (3,0) node[block] (B){ cubical complex}
      (6,0) node[block] (C){ filtered cubical complex}
      (9,0) node[block] (D){ boundary matrix} 
      (12,0) node[block] (E){ persistence diagram}
      (12,-2) node[block] (F){ features};
\draw[->, ultra thick] (A.east) -- (B);
\draw[->, ultra thick] (B.east) -- (C);
\draw[->, ultra thick] (C.east) -- (D);
\draw[->, ultra thick] (D.east) -- (E);
\draw[->, ultra thick] (E) edge  node[right] {$F$} (F);
\draw[->, ultra thick] (A) |-  %edge [bend right=30] 
node[below] {$F \circ \PH$\;} (F.west);
\end{tikzpicture}
}
\end{center}
\caption{%Pipeline for computing and learning features of persistence diagrams. 
Persistent homology computation pipeline for two images from MNIST and CIFAR-10. 
Top row, from left to right: original image, cubical complex, filtered cubical complex, symmetrized boundary matrix, persistence diagram in dimension $1$. 
Bottom row: features extracted from persistence diagrams, such as persistence images for dimension $1$ at different resolutions levels, and indicator functions of points at a certain distance from the diagonal.} 
\label{F:pipeline}
\end{figure}
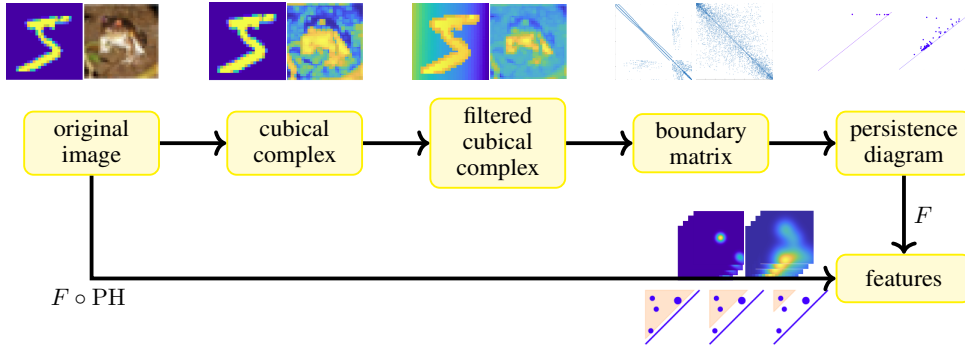

\section{Methods}
Let $\mathcal{X}$ be a space of 2D images (e.g., $\mathbb{R}^{n\times n}$), $\mathcal{D}$ the space of persistence diagrams, and $\mathcal{Y}$ a space of features that one can extract from persistence diagrams. 
We want to obtain the function ${F\circ \PH \colon \mathcal{X}  \to \mathcal{D} \to \mathcal{Y} }$ as a neural network trained from data.

To each image, we associate first a cubical complex, a space built out of points, edges, squares, in which we label each cell by a grey value \cite{wagner2012efficient}. We then filter the cells of the cubical complex by increasing grey values, and using the order induced by this filtration we construct a boundary matrix that stores information about adjacency of cells of co-dimension $1$. 
By reducing this matrix using standard methods, one can read off birth-death pairs of topological features: we obtain a persistence diagram in homological degree $0$ that gives information about the lifetime of components across the filtration, and a persistence diagram in homological degree $1$ for the lifetime of holes. For details see Appendix~\ref{app:barcodes}
Each persistence diagram is a multiset of points in $\mathbb{R}^2$, where the two coordinates of each point correspond to birth-death times of a topological feature. An alternative representation of the birth-death pairs is as intervals in what is usually called a barcode. We illustrate the different steps in this pipeline in Figure~\ref{F:pipeline} and provide further details in Appendix~\ref{A:pipeline}.

Once the persistence diagram is computed, the extraction of standard hand-crafted features is usually trivial. We may consider different types of such features, including (1) tropical coordinates, which capture information such as the mean distance from the diagonal, or the sum of the distances of the two points furthest from the diagonal \cite{Kalisnik}, which are standard quantities of interest in TDA; (2) persistence images \cite{PI}; (3) Fourier coefficients of persistence images; (4) indicator function of points at a certain distance from the diagonal; and (5) number of blobs in a persistence image. 
We note that features (1)--(2) are known to be stable, in the sense that the map $F\colon \mathcal{D}\to \mathcal{Y}$ is Lipschitz with respect to suitable choices of distances on the space of diagrams and the space of features \cite{Kalisnik, PI}.
On the other hand, features (4)--(5) are not stable, as they are discrete invariants associated with the images.  We provide more details about these features in Appendix~\ref{app:features}. 
In the present work, we focus on features (1) and (4). The binary features that we consider here can be fine-tuned for specific classification tasks, for instance  when one is interested in studying points at a certain distance from the diagonal, as discussed in Section \ref{S:intro}. We note that in  \cite{pi-net} the authors explored how feature (2) can be learned from time-series data. 

We train the neural network using different types of input data, corresponding to different stages in the above pipeline. We find that using the original images as inputs often gives better results than using cubical complexes. 
We should point out that the trained network may represent a function that is only an approximation of the features. This approximation may still be sensitive to perturbations of the input, which is an important topic on itself, beyond the scope of our work. The theory for the stability of the features that we consider here is a requirement for any investigation of stability questions.

\section{Experiments}
\begin{figure}[h!]
\centering
%MNIST - Predicting a bar \\
  \begin{tikzpicture}
\node at (-9,0) {\includegraphics[height=.2\textwidth]{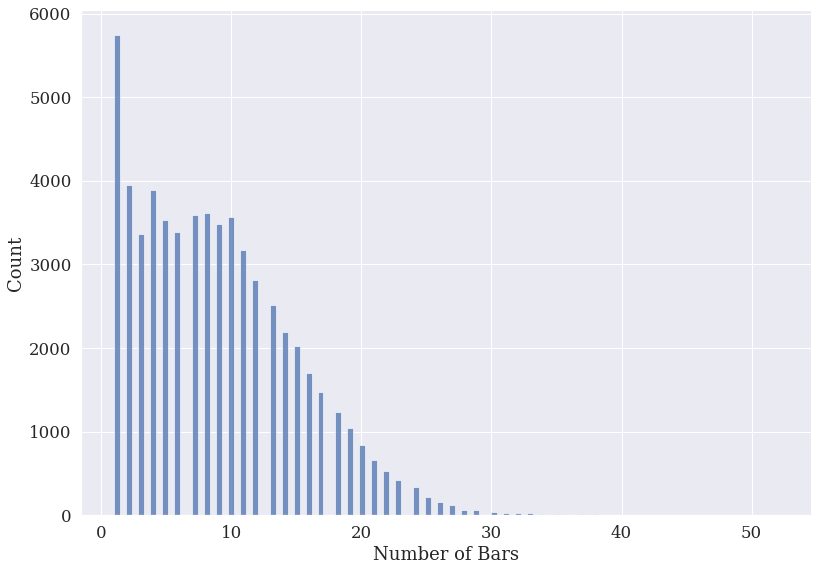}}; 
\node at (-4.5,0) {\includegraphics[height=.2\textwidth]{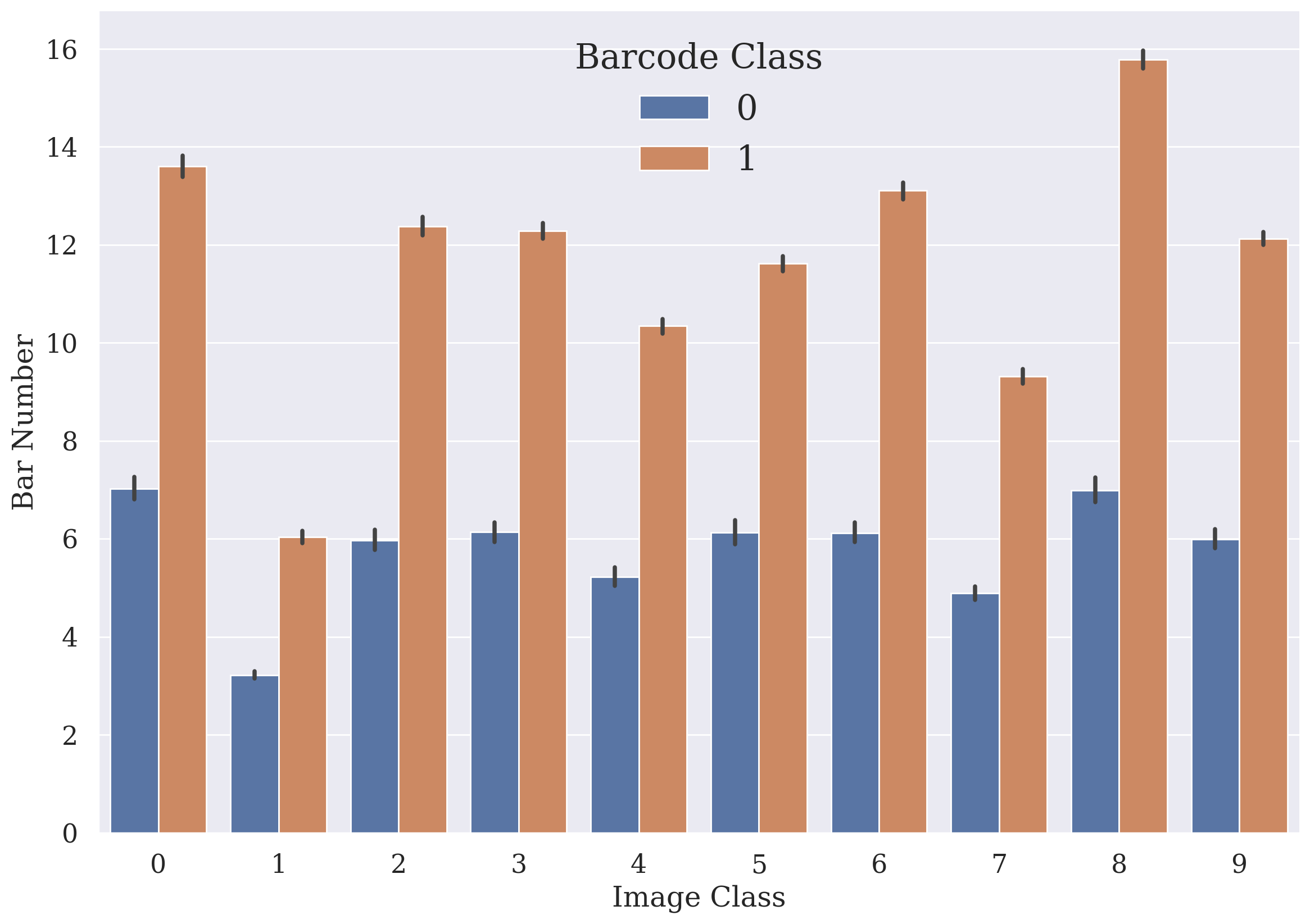}}; 
\node at (0,0) {\includegraphics[height=.22\textwidth]{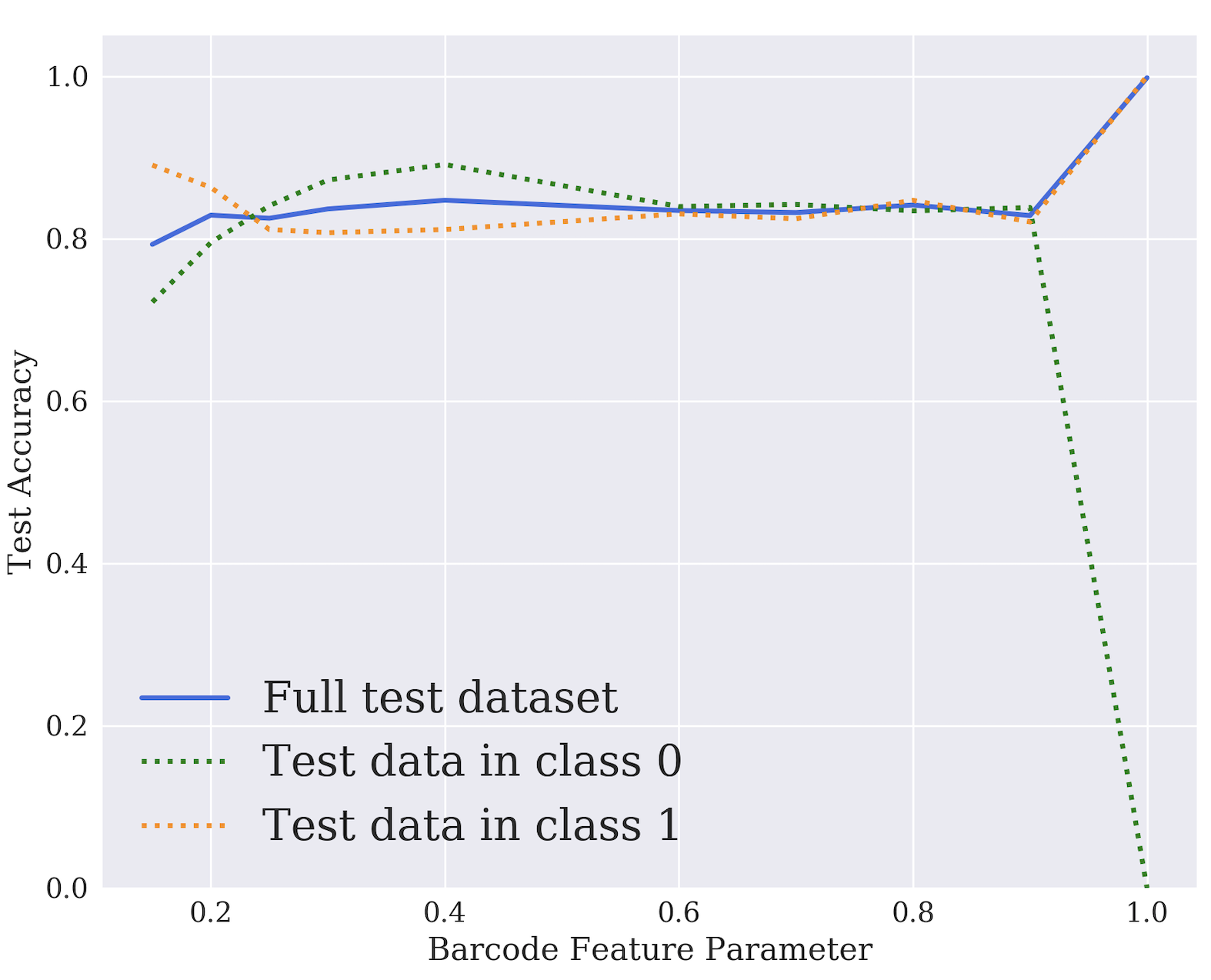}}; 
\node at (-0.9,-.495) {\includegraphics[width=.043\textwidth]{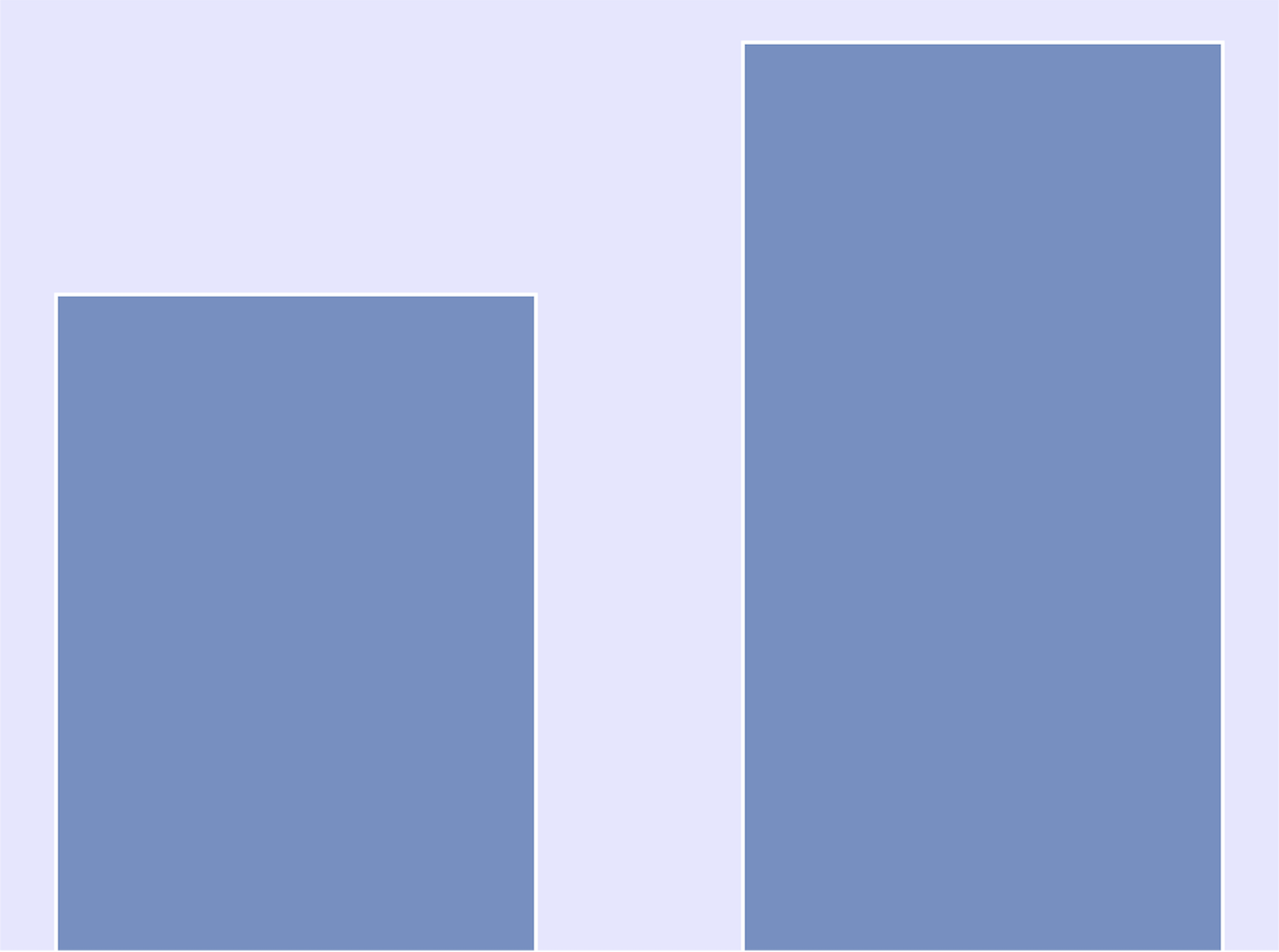}}; 
\node at (-0.9,-.12) {\tiny$\theta=.15$}; 
\node at (0.1,-.495) {\includegraphics[width=.043\textwidth]{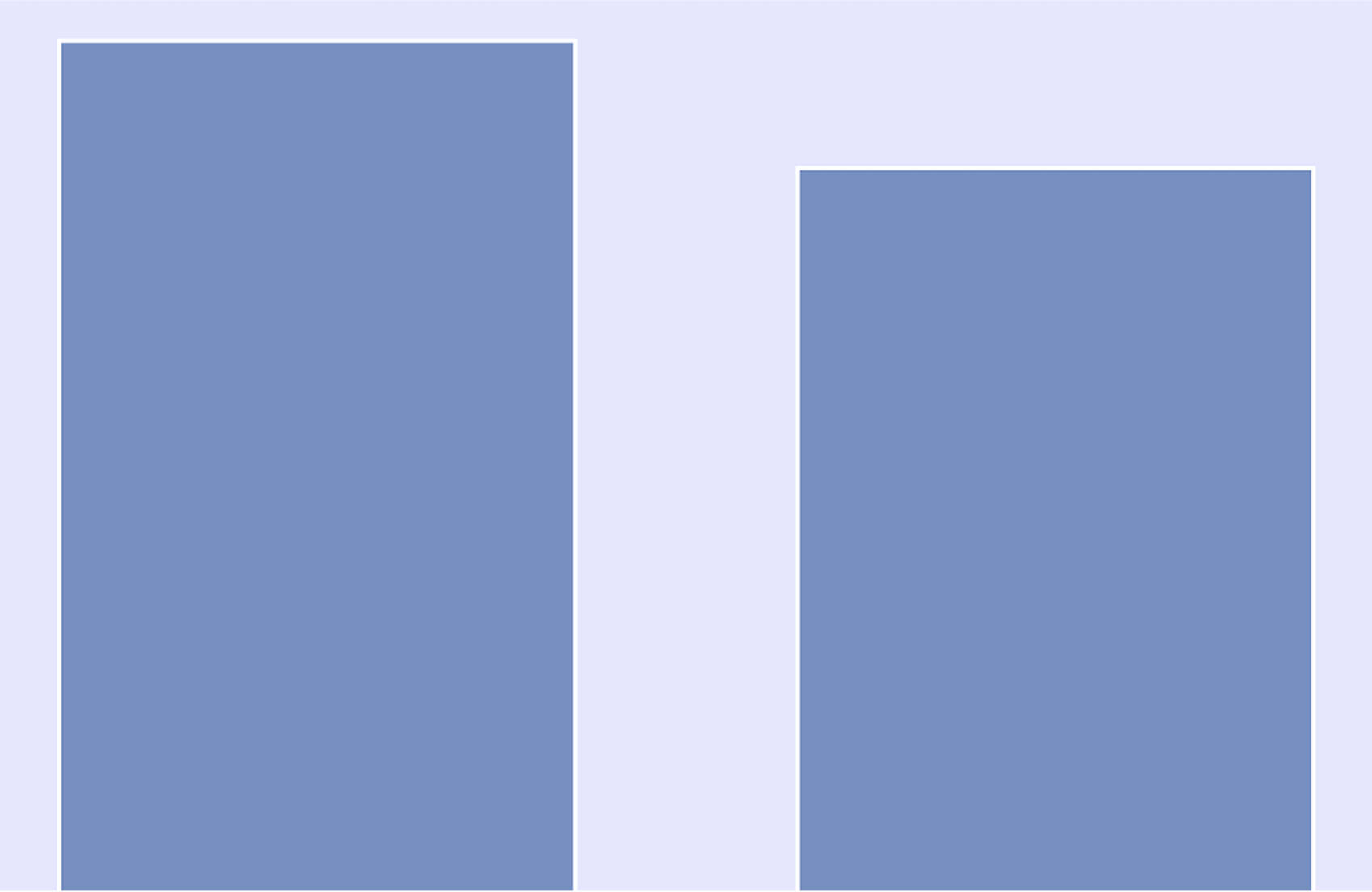}}; 
\node at (0.1,-.12) {\tiny$\theta=.3$}; 
\node at (1.1,-.495) {\includegraphics[width=.043\textwidth]{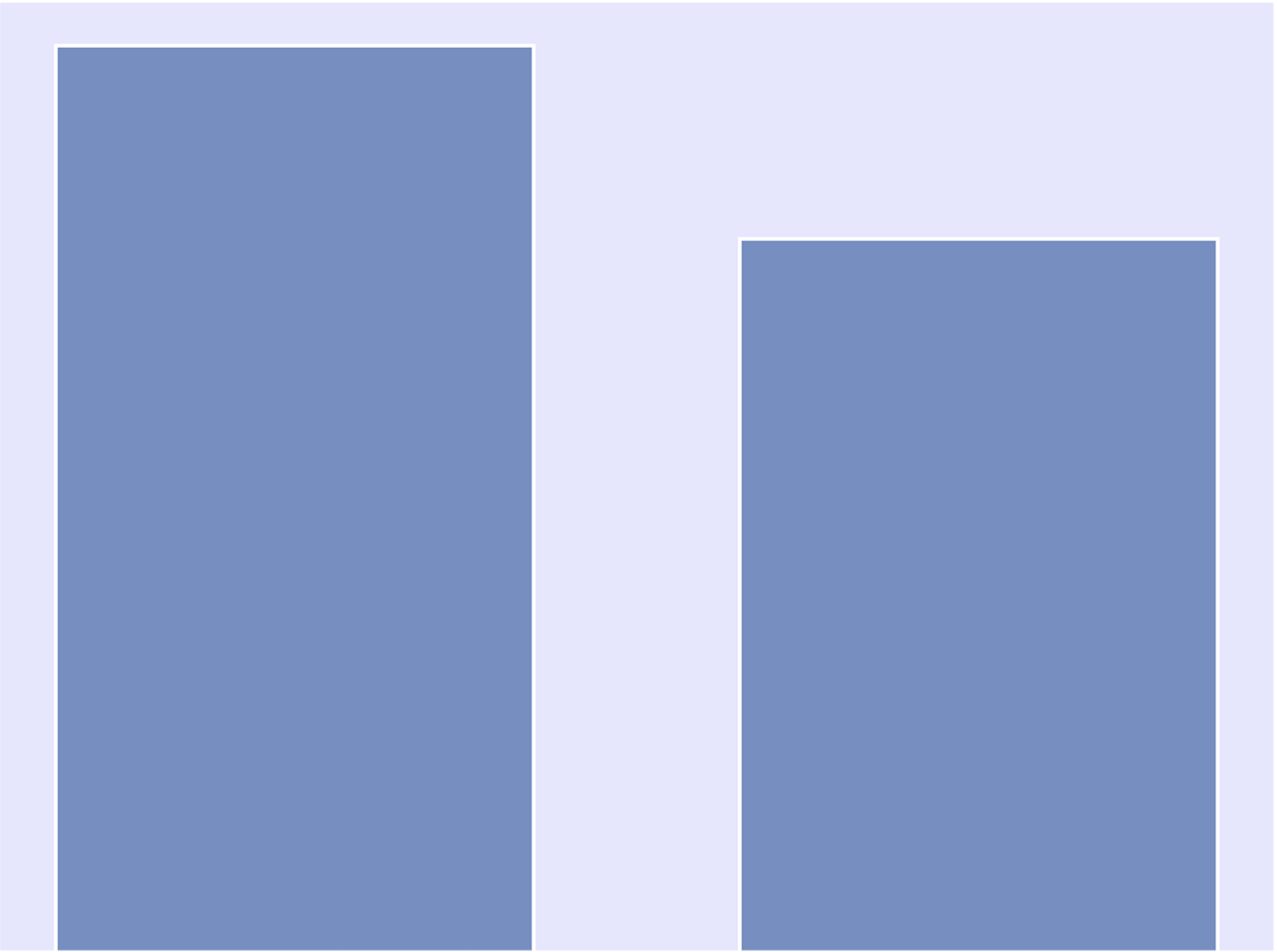}}; 
\node at (1.1,-.12) {\tiny$\theta=.8$}; 
\node at (-4.25,1.75) {MNIST}; 
\end{tikzpicture}
\\ 
%CIFAR-10 - Predicting a bar \\
\begin{tikzpicture}
\node at (-9,0) {\includegraphics[height=.2\textwidth]{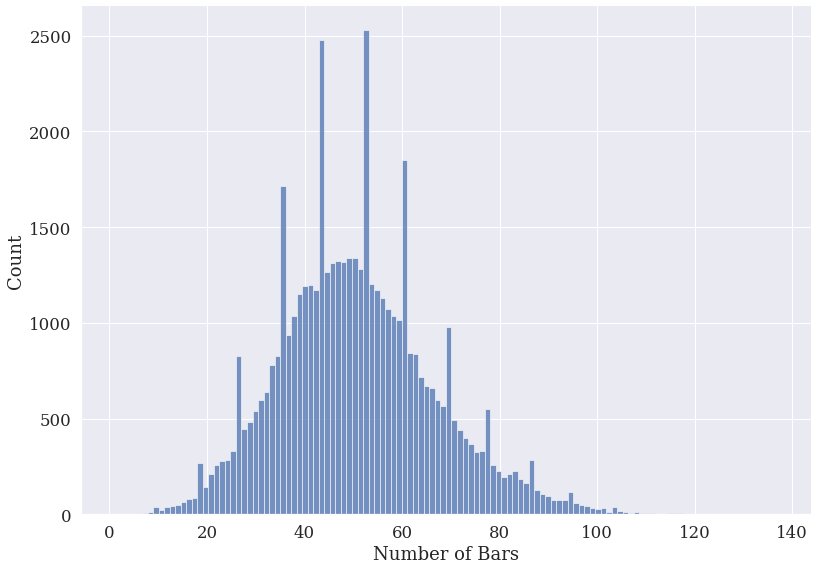}};
\node at (-4.5,0) {\includegraphics[height=.2\textwidth]{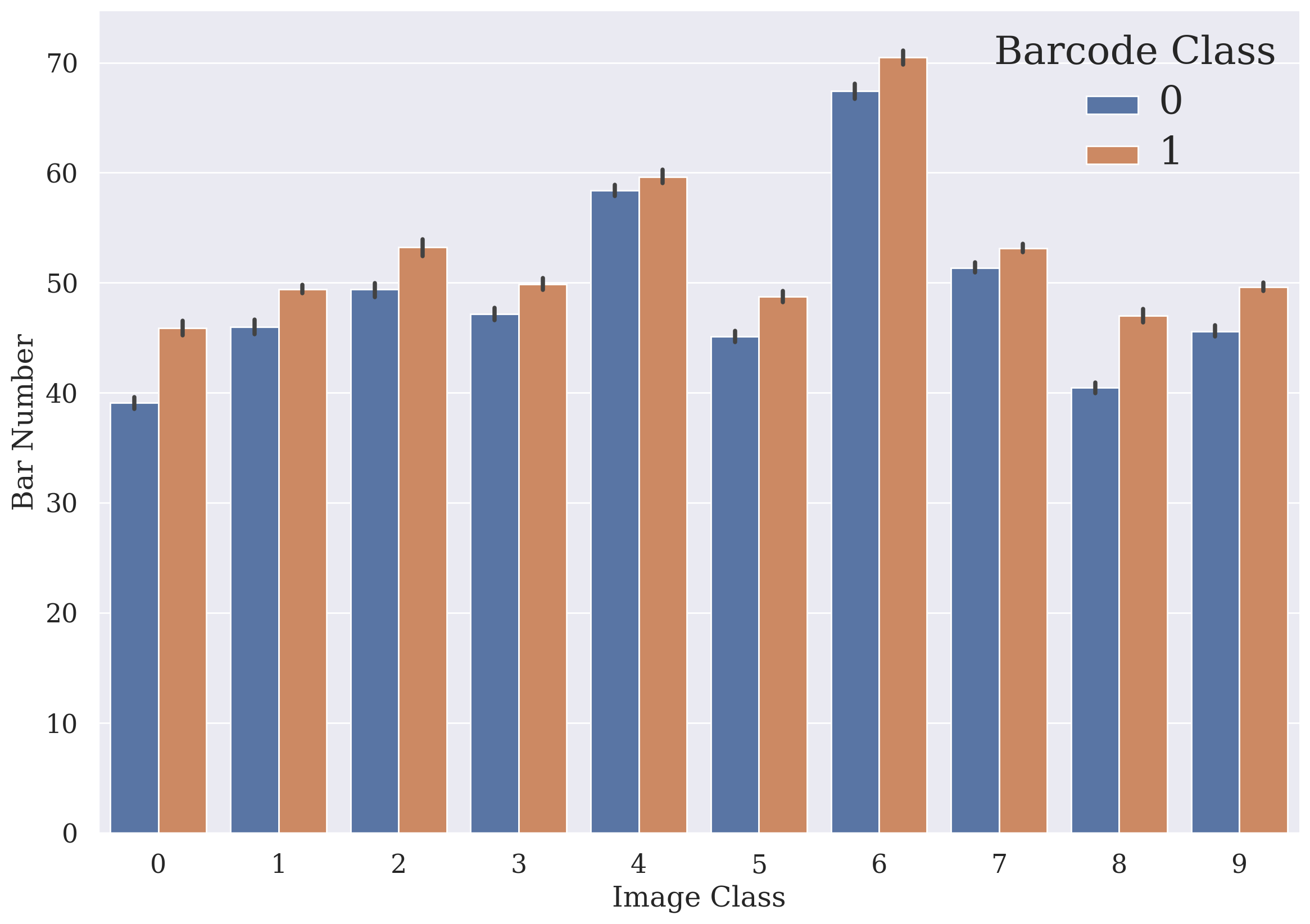}};
\node at (0,0) {\includegraphics[height=.22\textwidth]{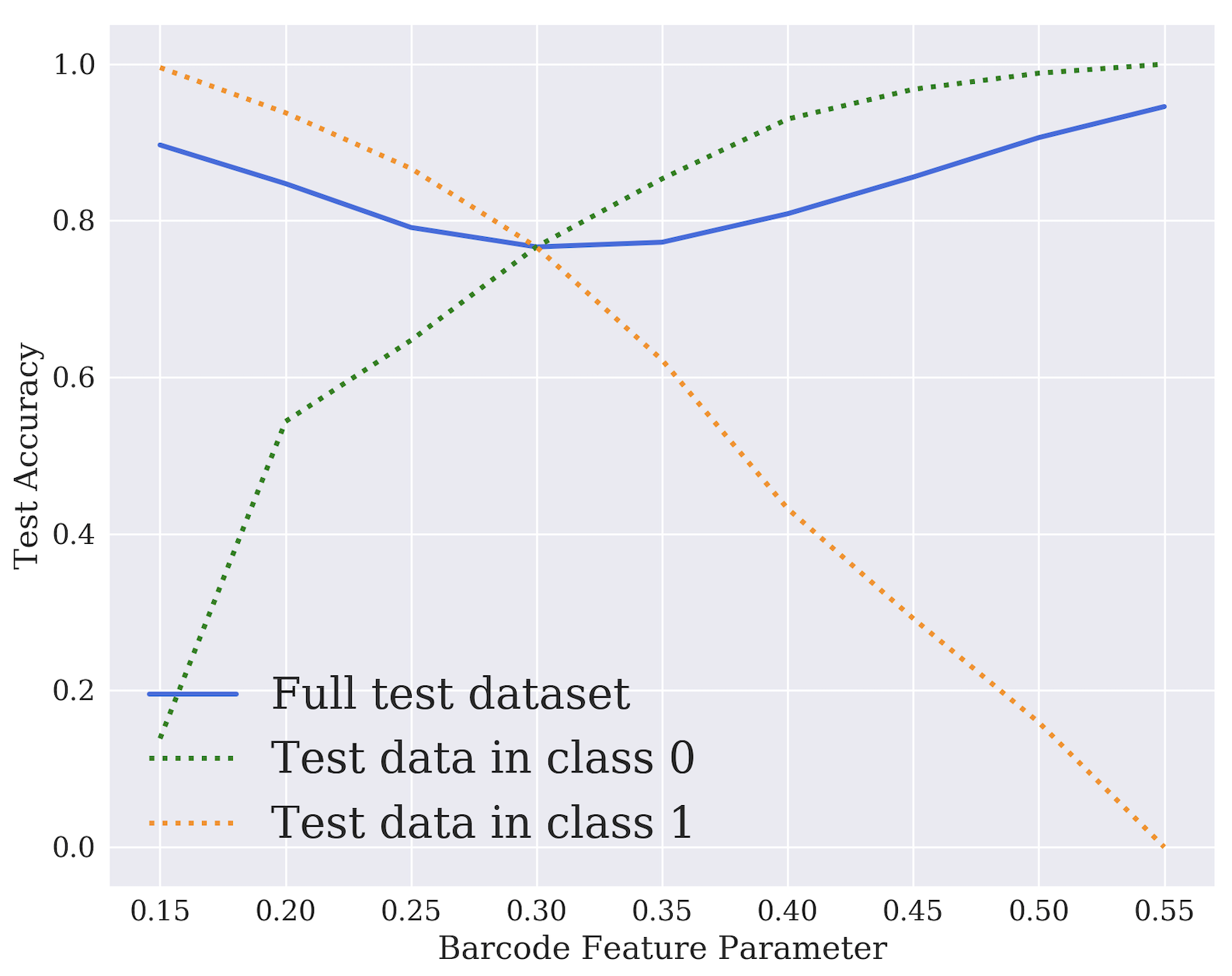}}; 
\node at (-0.9,-.495) {\includegraphics[width=.043\textwidth]{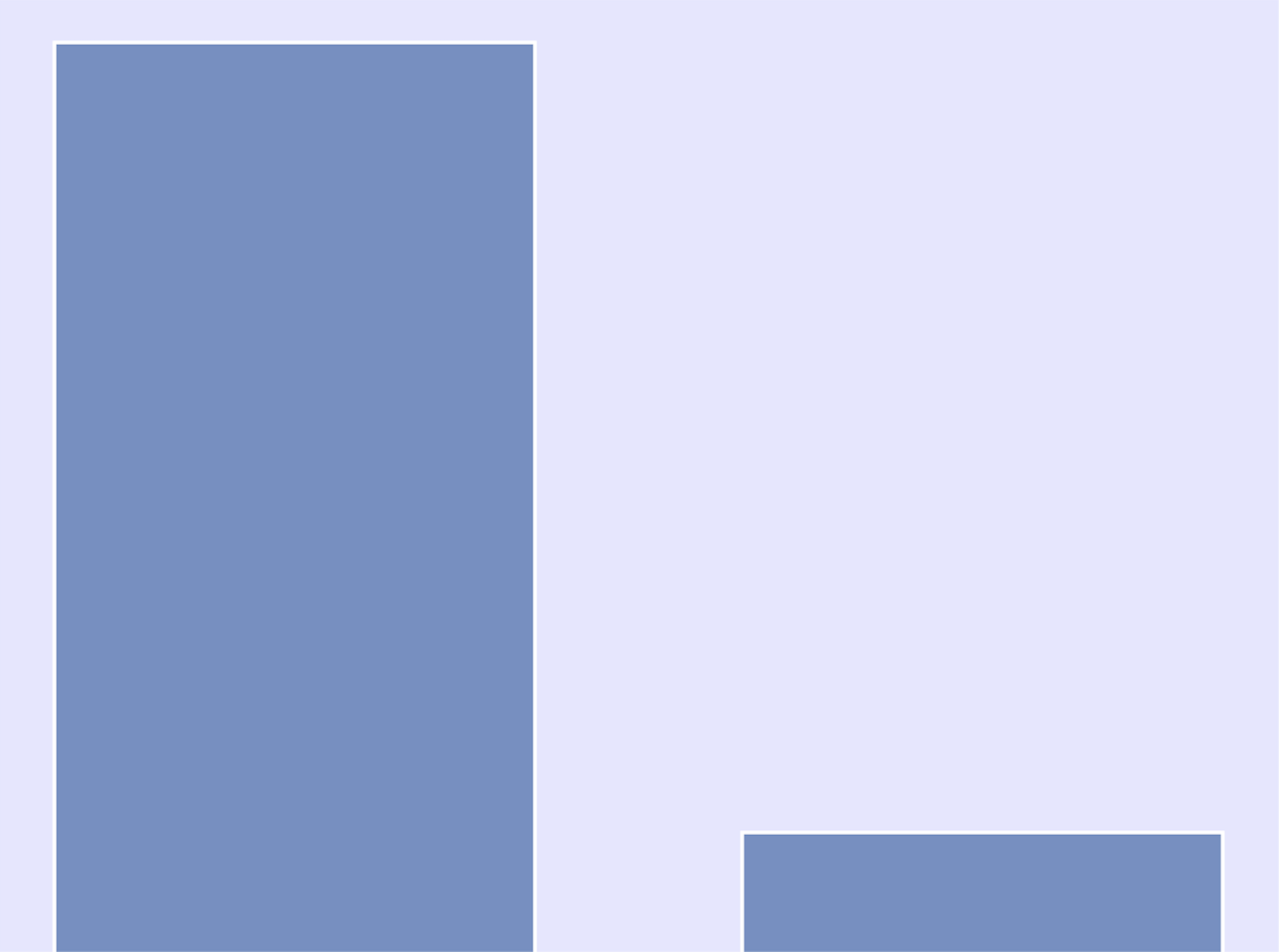}}; 
\node at (-0.9,-.12) {\tiny$\theta=.15$}; 
\node at (0.1,-.495) {\includegraphics[width=.043\textwidth]{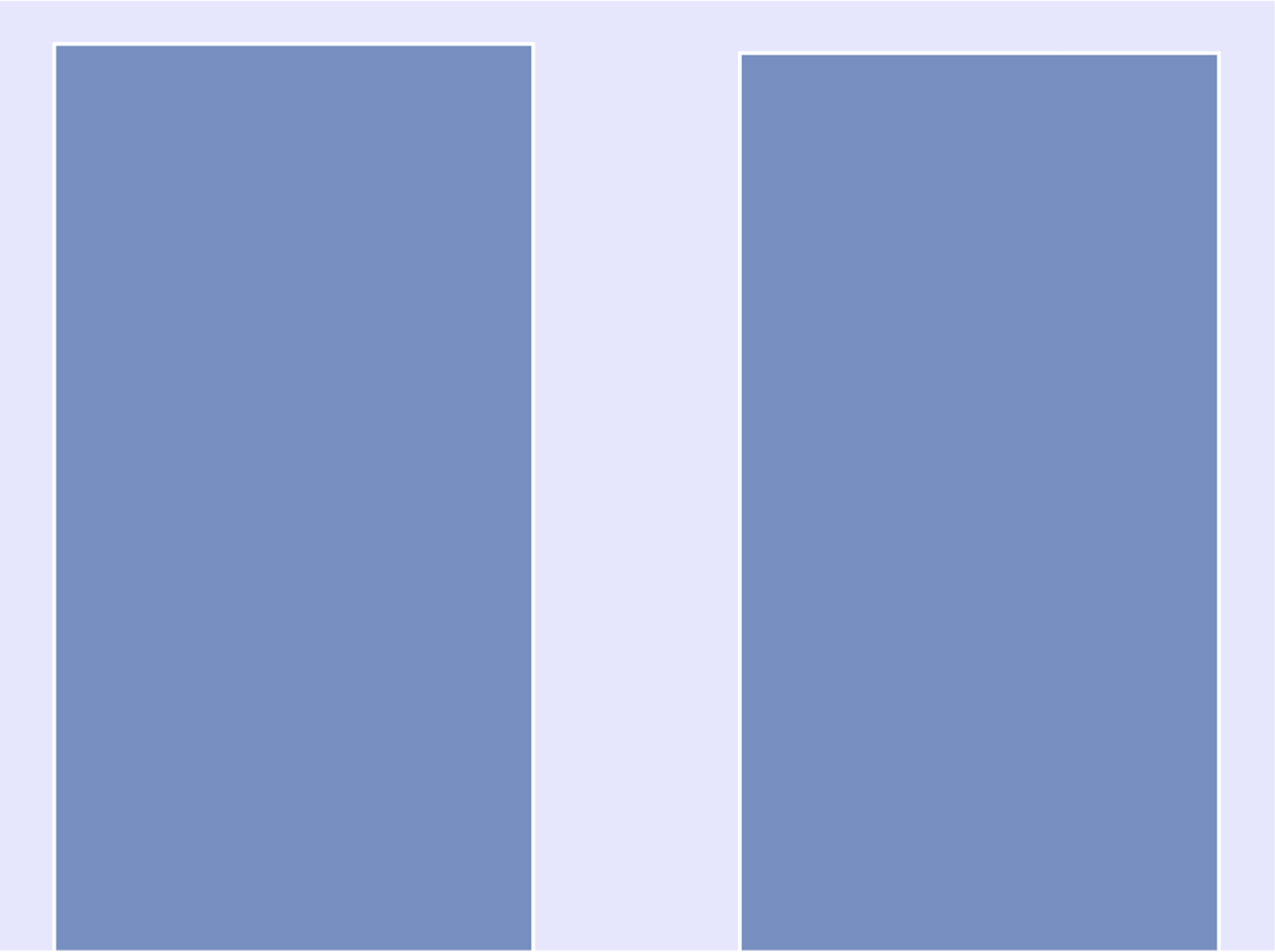}}; 
\node at (0.1,-.12) {\tiny$\theta=.3$}; 
\node at (1.1,-.495) {\includegraphics[width=.043\textwidth]{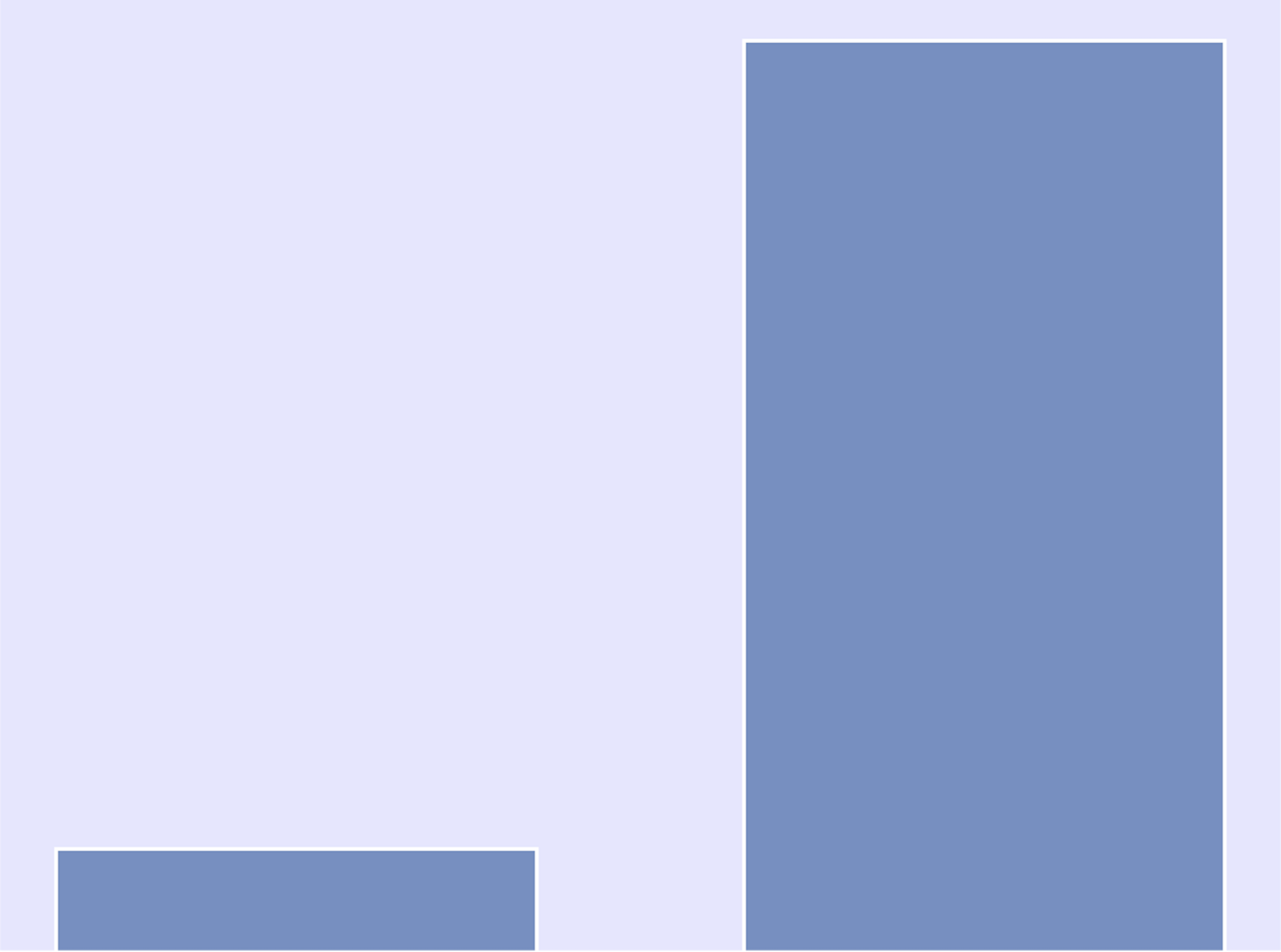}}; 
\node at (1.1,-.12) {\tiny$\theta=.5$}; 
\node at (-4.25,1.75) {CIFAR-10}; 
\end{tikzpicture}
\caption{Left: Histogram for the number of bars in barcodes. 
Middle: Average number of bars in the barcodes of images in each class of MNIST and CIFAR-10 intersected with the two classes determined by our persistence feature with parameter $\theta=0.3$.
Right: Test accuracy of a CNN for predicting whether the persistence diagram of an input image contains a bar of a certain length, for $10$ values of the feature parameter $\theta$. 
The blue, green and orange curves represent the test accuracy on the full test data, on the test data in class 0, and test data in class 1, respectively. 
Inserts show the distribution of images in the two classes defined by the feature, for three values of $\theta$.
}
\label{F:histo Cifar}
    \label{fig:cifar10_auc}
    \label{fig:MNIST_auc}
\end{figure}
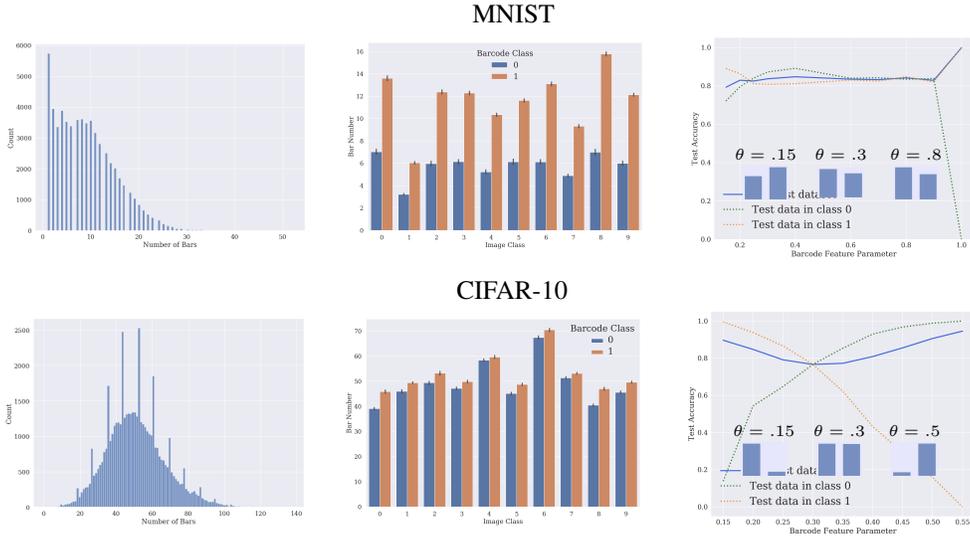

We start by taking a look at the statistics of the persistence diagrams for images. The left panel of Figure~\ref{F:histo Cifar} shows the histogram of the number of bars in barcodes for images in MNIST \cite{lecun-mnisthandwrittendigit-2010} and CIFAR-10 \cite{krizhevsky2009learning}. 
We then group the barcodes into two classes by whether there is at least one bar whose length is larger than $0.3$. The mid panel shows the average number of bars in barcodes in these two classes for each of the 10 different image classes in MNIST and CIFAR-10. 
These figures indicate that this topological feature has different statistics over the different image classes.

\subsection{Mapping images to binary features of their persistence diagrams}\label{SS:images}
Here we take original images as inputs and train a network to predict properties of the barcodes. We use CNNs with architecture LeNet-5 \cite{lecun1998gradient} for both MNIST and CIFAR-10. We consider 10 binary features indicating the presence of bars of a certain length in the barcode.

For MNIST, the features indicate the presence of bars of length in the intervals $[0.1, \theta]$, with feature parameter values $\theta = 0.15, 0.2, 0.25, 0.3, 0.4, 0.6, 0.7, 0.8, 0.9, 1$. For CIFAR-10, the features indicate the presence of bars of length at least $\theta$, with $\theta = 0.15, 0.19, \ldots, 0.55$ (equally spaced). We train all features as separate binary classification tasks using early stopping with patience 30 and decreasing the learning rate with patience 5 based on the validation set error. 

Figure~\ref{fig:MNIST_auc} shows the test results of the trained LeNet-5 for the ten binary classification tasks on MNIST and CIFAR-10. 
We report the test accuracy on the full test data sets, and also on test data restricted to class 0 and class 1. As the figure shows, the overall accuracies for MNIST images are all above 79\%, varying slightly with the feature parameter. 
For $\theta=1$, most images have feature label 1, hence predicting the label is easy and the overall test accuracy is close to 1.
For CIFAR-10, the overall test accuracies are at least 75\% for all values of the feature parameter. 
For $\theta$ taking the extreme values 0.15 and 0.55, the test data is highly imbalanced, and the test accuracy is above 95\%.
Training curves and additional details are presented in Appendix~\ref{app:image-to-bin}.

%\FloatBarrier

\subsection{Mapping cubical complexes to binary features of the persistence diagrams}

In this set of experiments, we consider inputs taken at an intermediate stage of the computational pipeline, namely cubical complexes (CC) and filtered cubical complexes (FCC). We consider the same labels as in Section~\ref{SS:images} indicating bars of a certain length. 
We model this as a graph classification task with the CC or FCC as an adjacency matrix of the input graph. We view the CC  or FCC matrix as the adjacency matrix of a graph, where the cells are the nodes of the graph. We use graph neural network (GNN) models with 3 combined layers of GCN or GIN convolutions plus TopK pooling and a two-layer MLP. For both CC and FCC, the GCN model has good performance on classifying the bar feature. In contrast, GIN model does not perform well, partly due to CC and FCC matrices are non-symmetric, and then the input graphs are directed. More details on this experiment, including data processing and implementation, are provided in Appendix~\ref{app:experiments-CC-FCC}.

\subsection{Mapping images to tropical coordinates of their persistence diagrams}\label{sec:regress_tropical}

\begin{figure}[h]
    \centering
MNIST - regress tropical coordinates of barcodes\\[2mm]
\begin{tabular}{ccccc}
   \includegraphics[width=.16\textwidth]{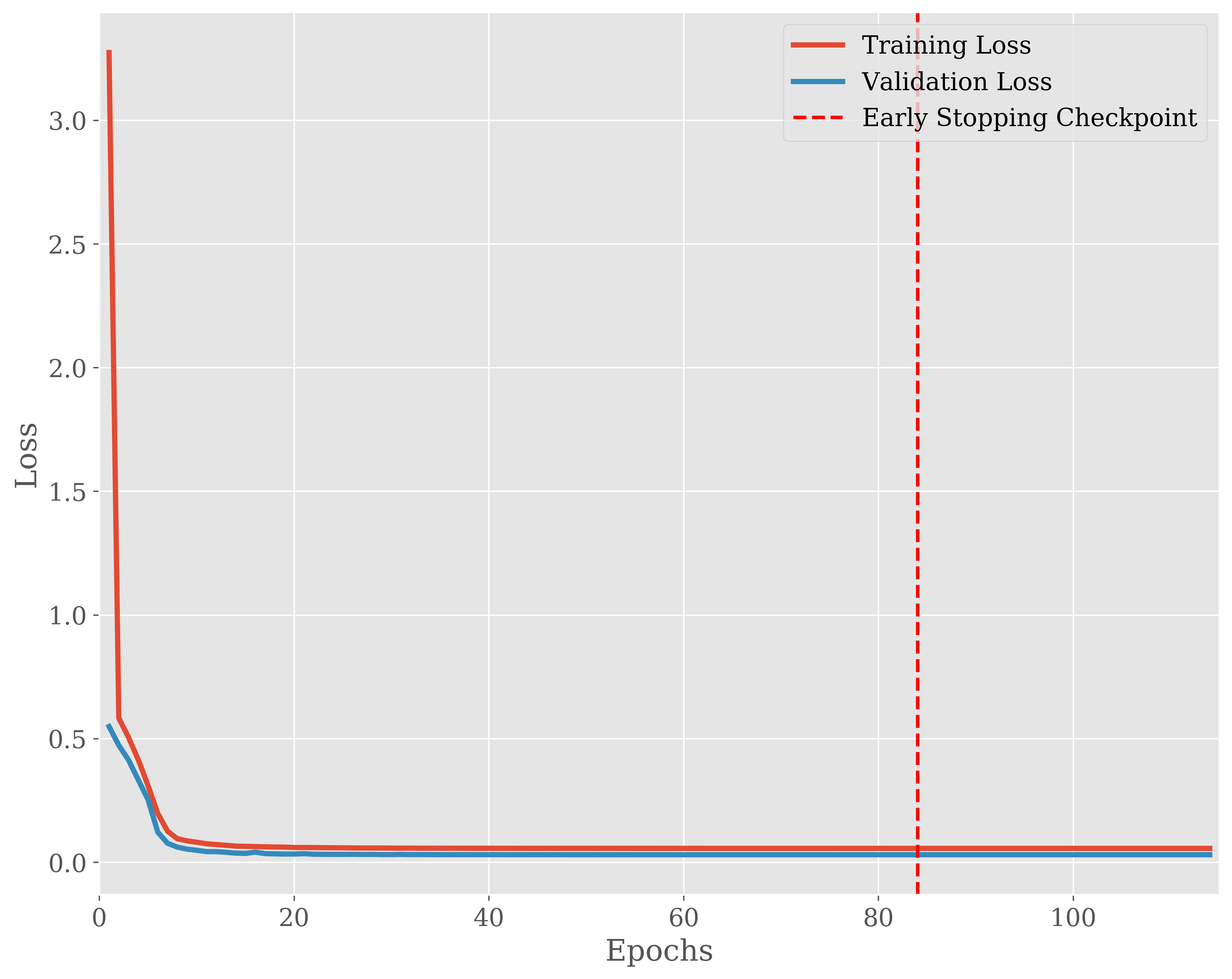}
    &\includegraphics[width=.16\textwidth]{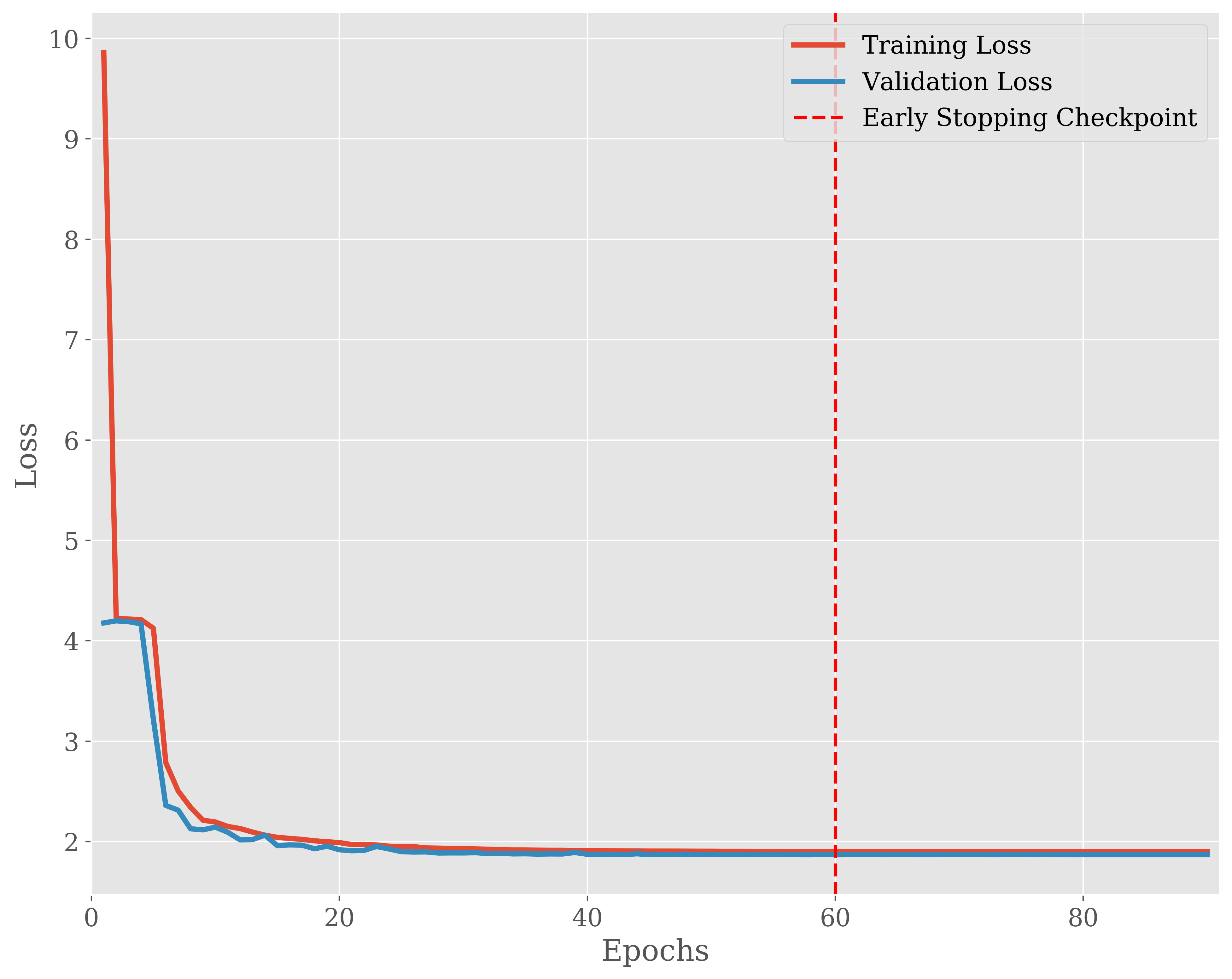}
    &\includegraphics[width=.16\textwidth]{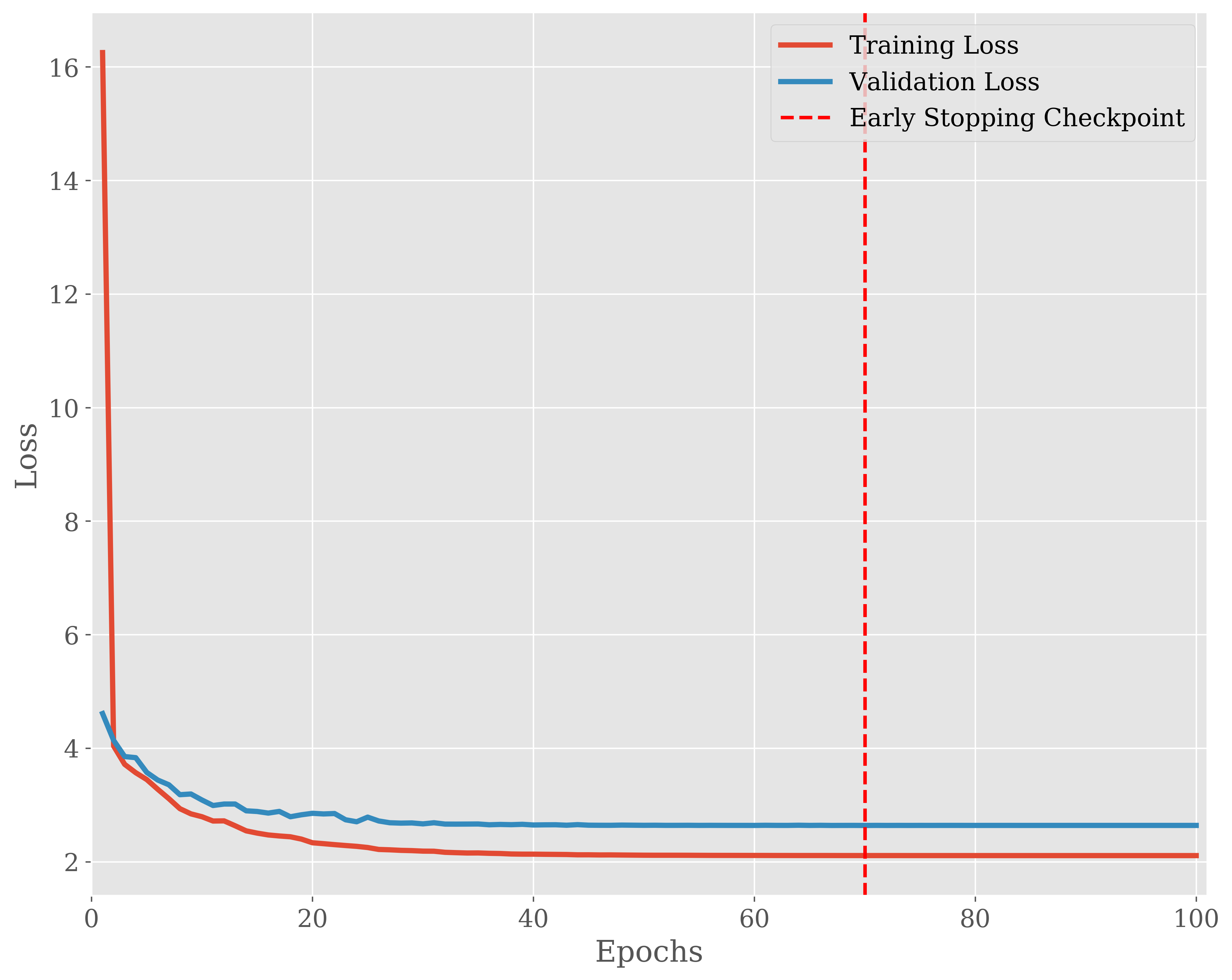}
    &\includegraphics[width=.16\textwidth]{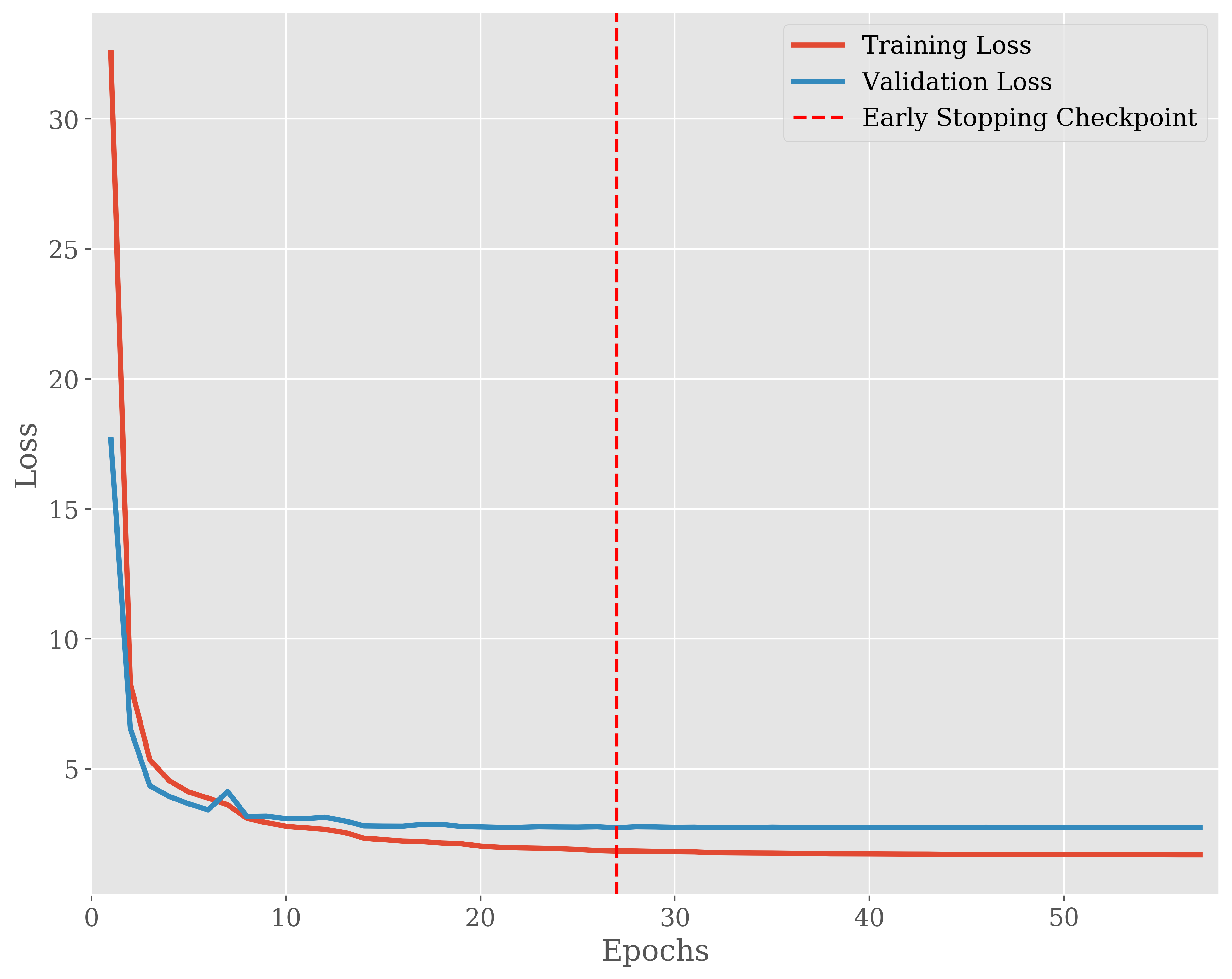}
    &\includegraphics[width=.16\textwidth]{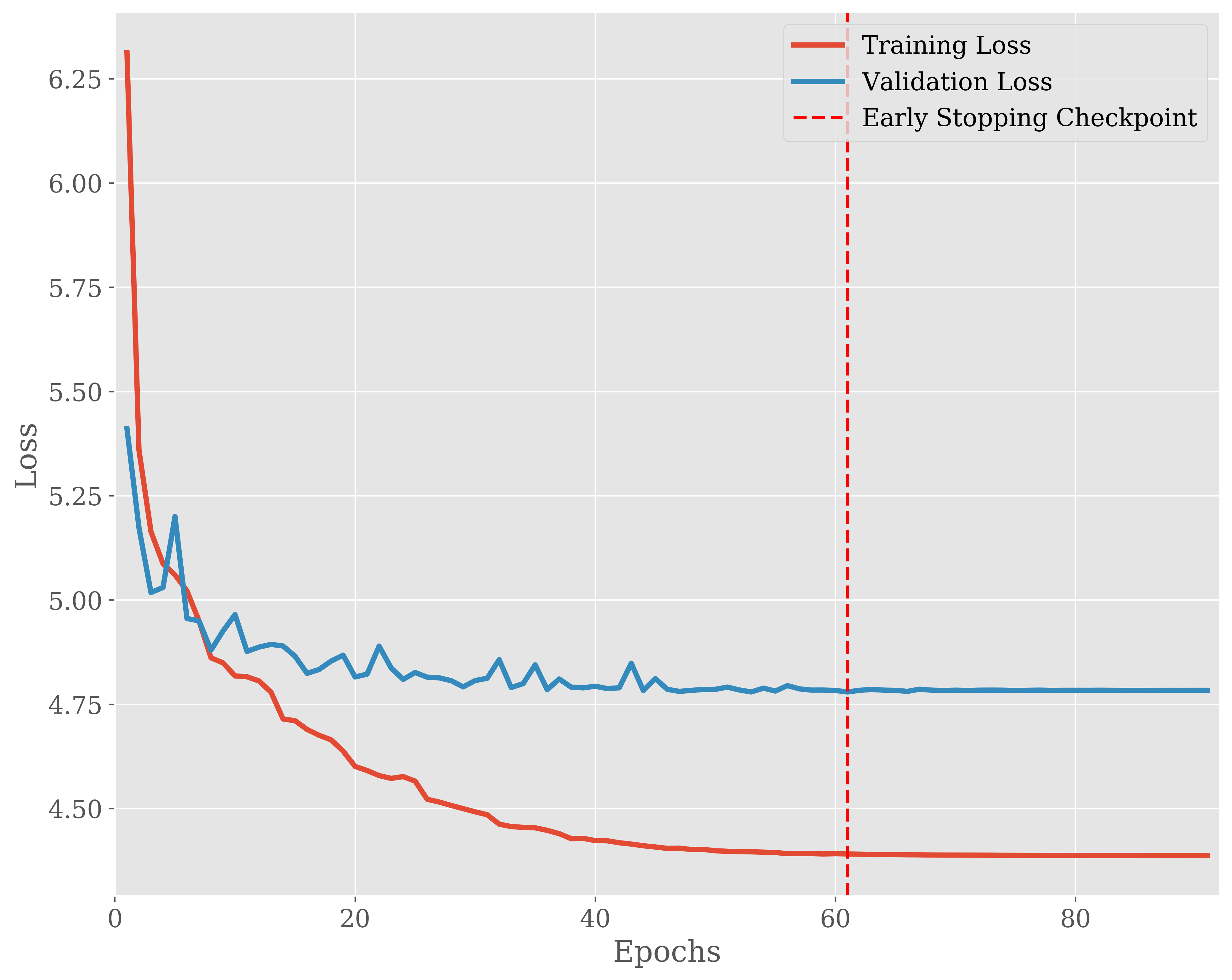}\\
    \includegraphics[width=.16\textwidth]{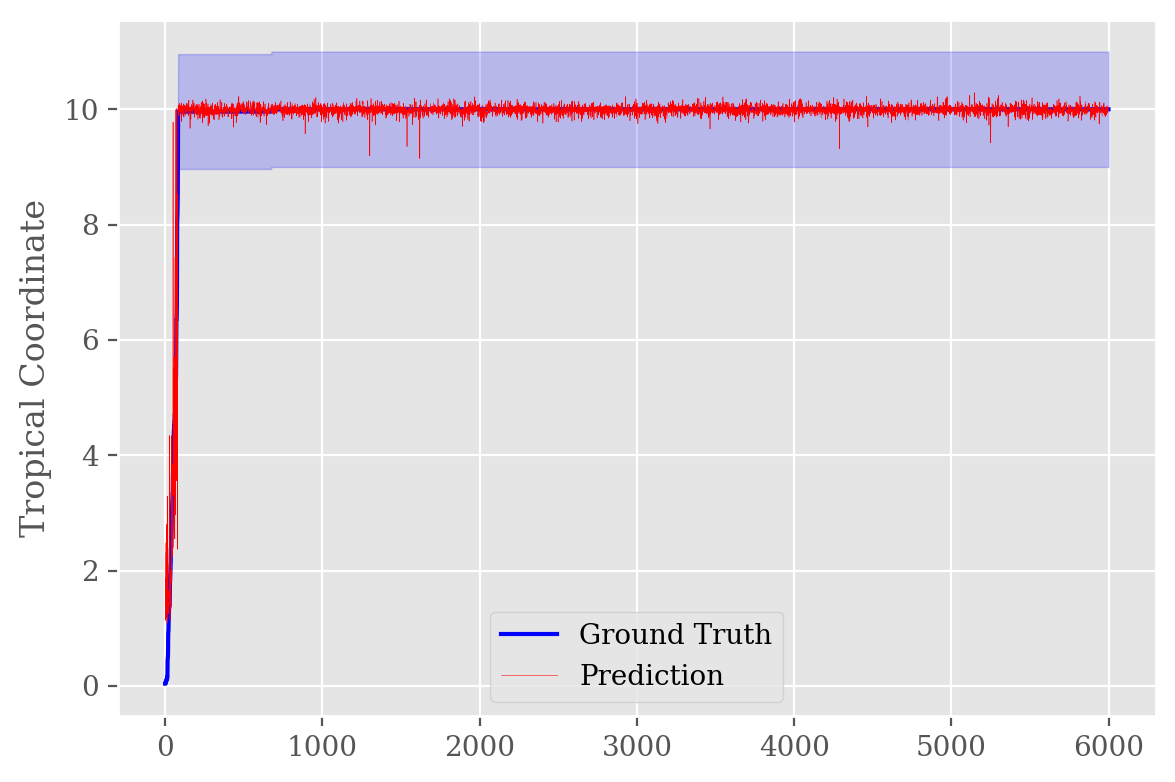}
    &\includegraphics[width=.16\textwidth]{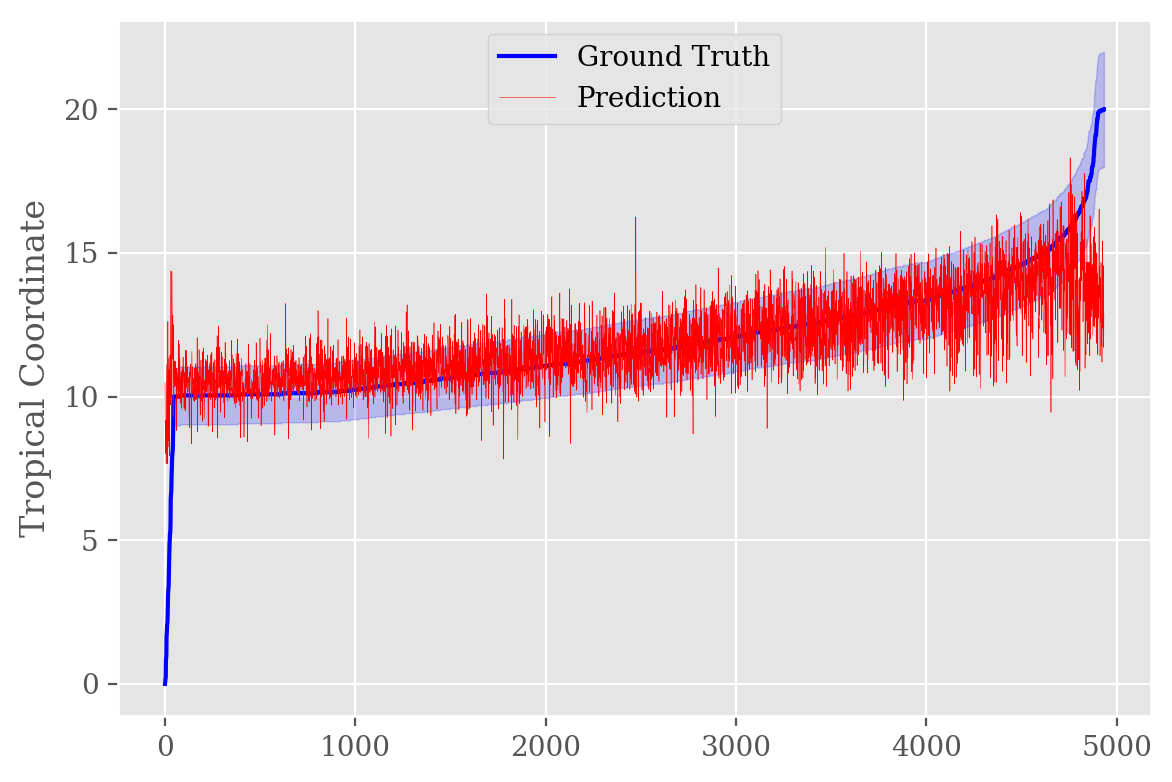}
    &\includegraphics[width=.16\textwidth]{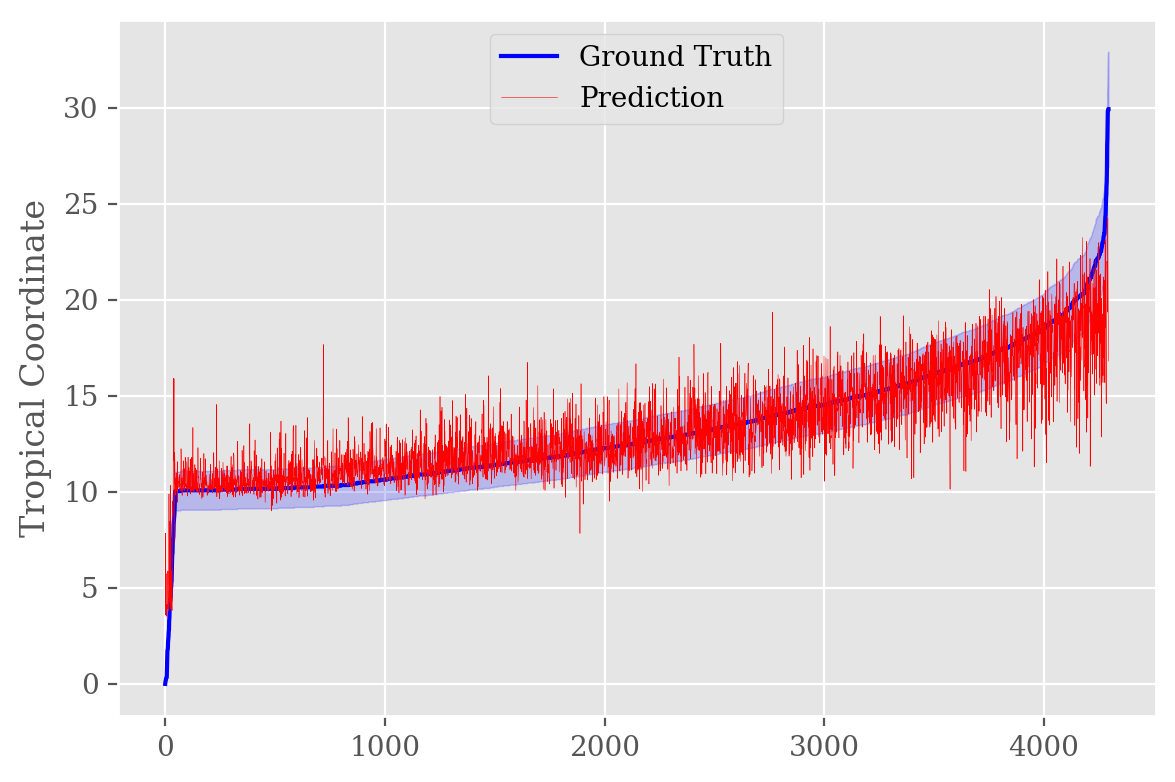}
    &\includegraphics[width=.16\textwidth]{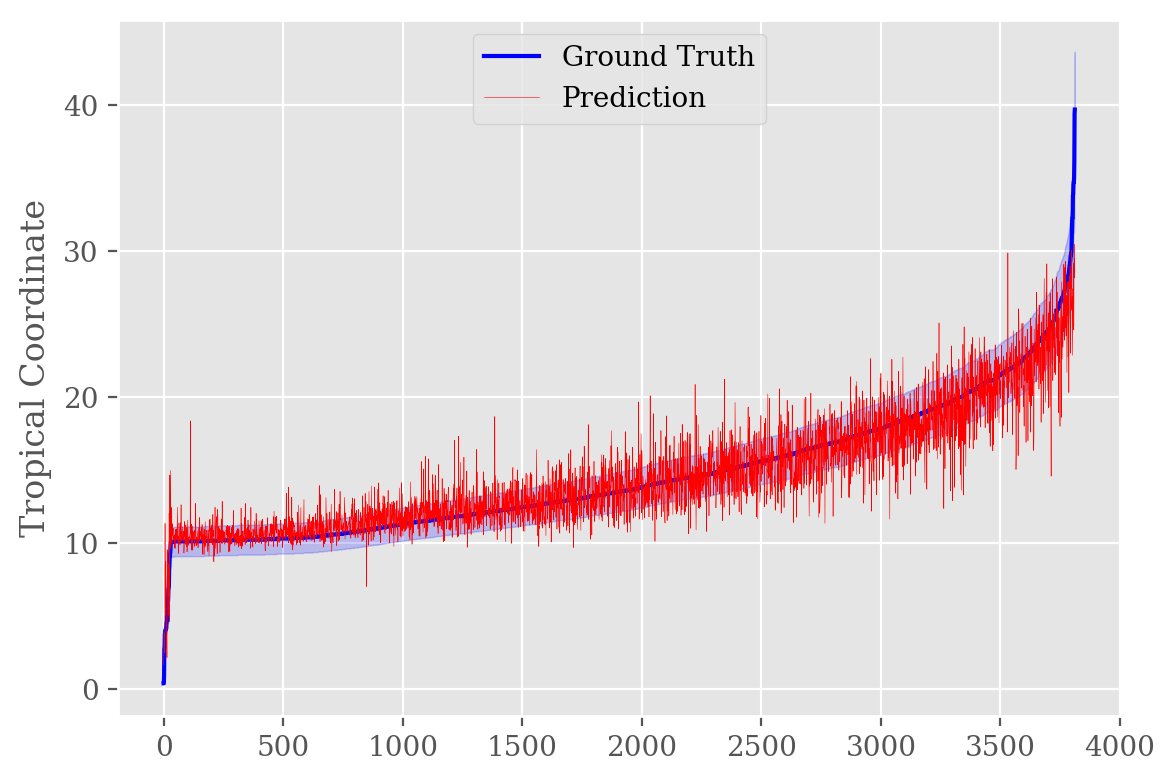}
    &\includegraphics[width=.16\textwidth]{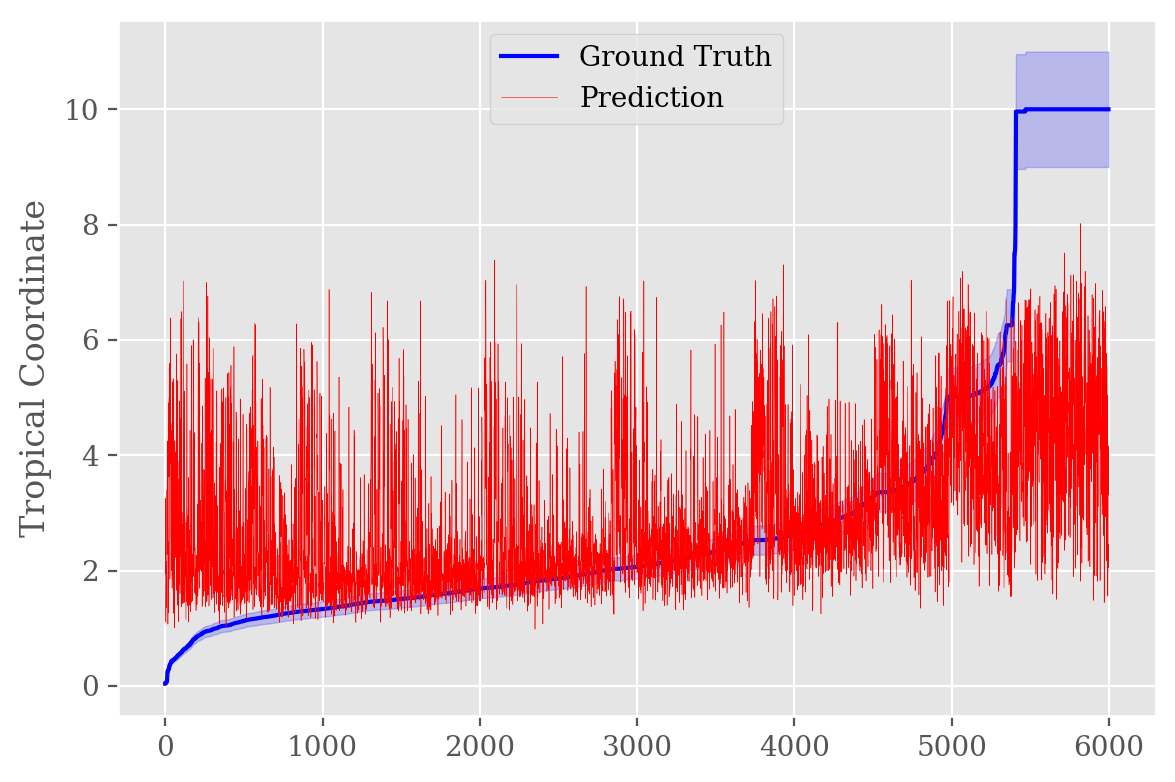}
\end{tabular}\\
CIFAR-10 - regress tropical coordinates of barcodes\\[2mm]
\begin{tabular}{ccccc}
   \includegraphics[width=.16\textwidth]{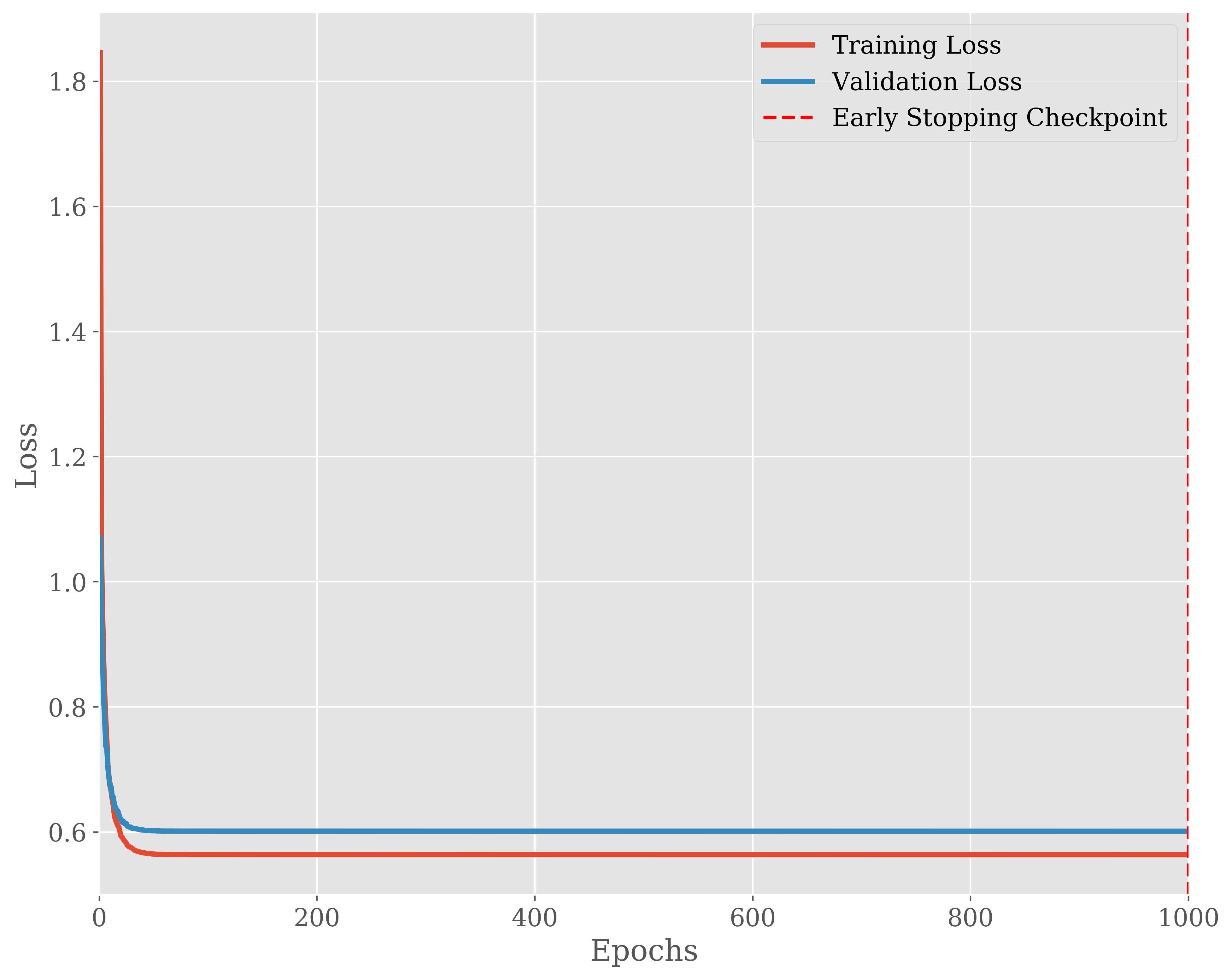}
    &
   \includegraphics[width=.16\textwidth]{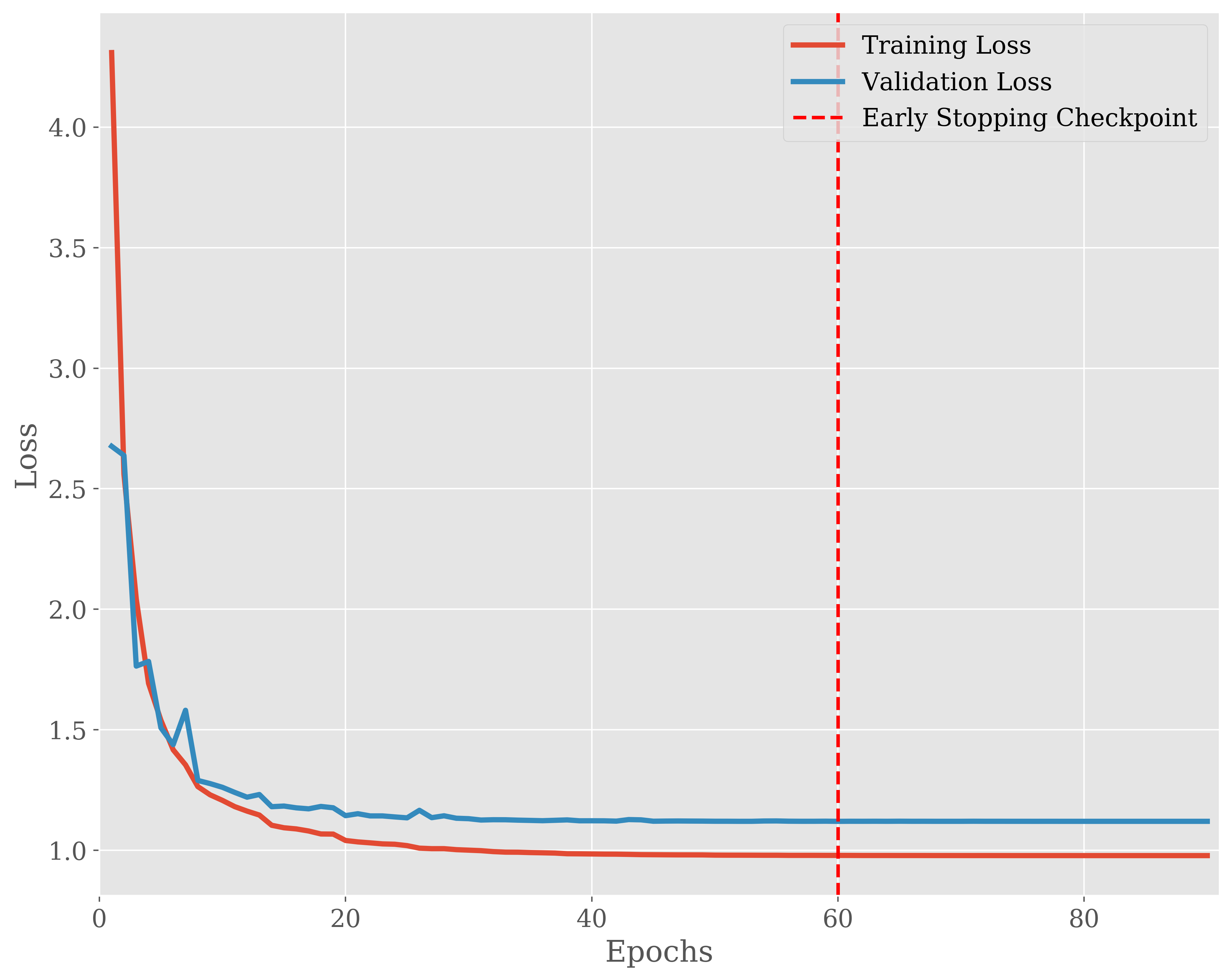}
    & 
   \includegraphics[width=.16\textwidth]{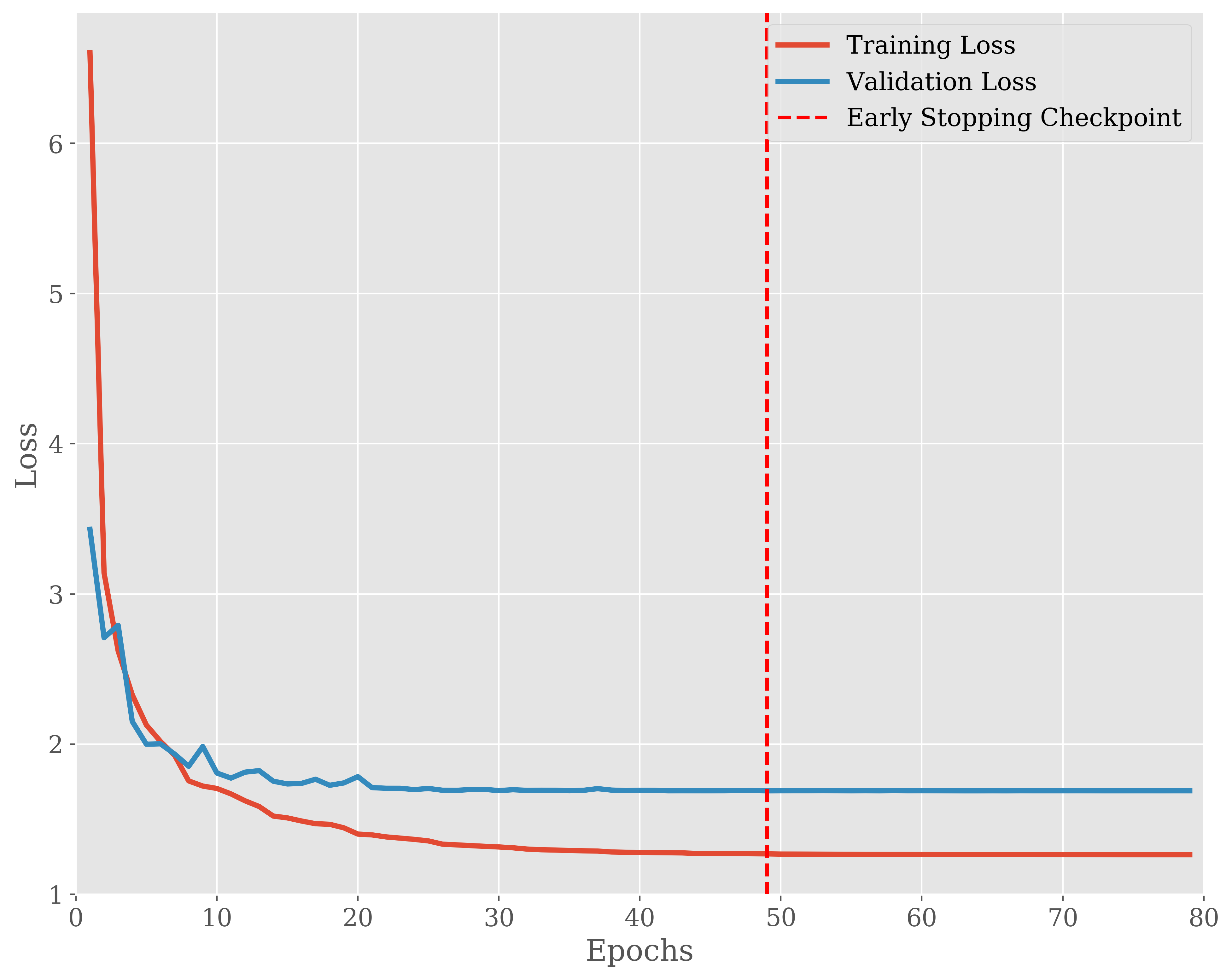}
    & 
   \includegraphics[width=.16\textwidth]{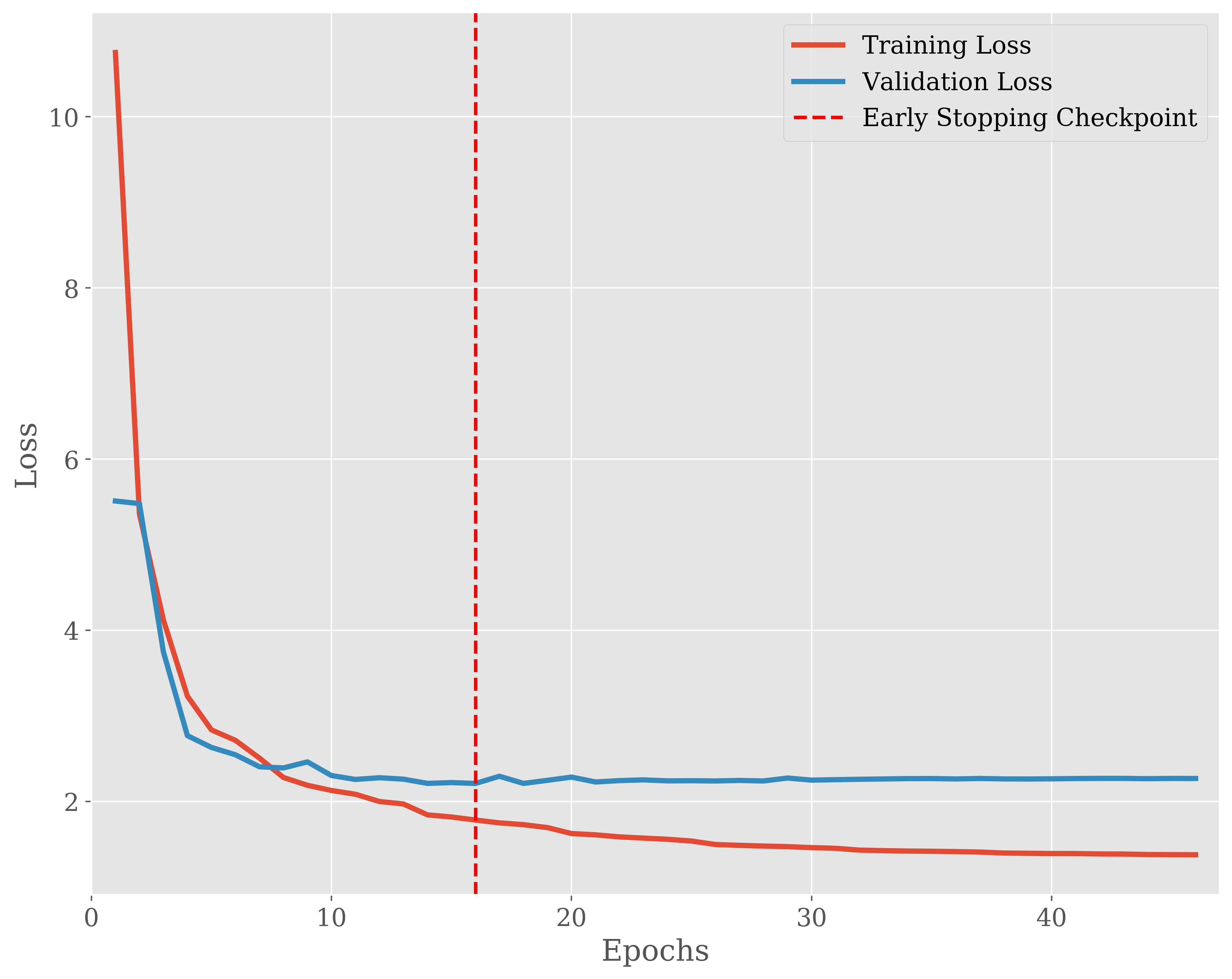}
    &
   \includegraphics[width=.16\textwidth]{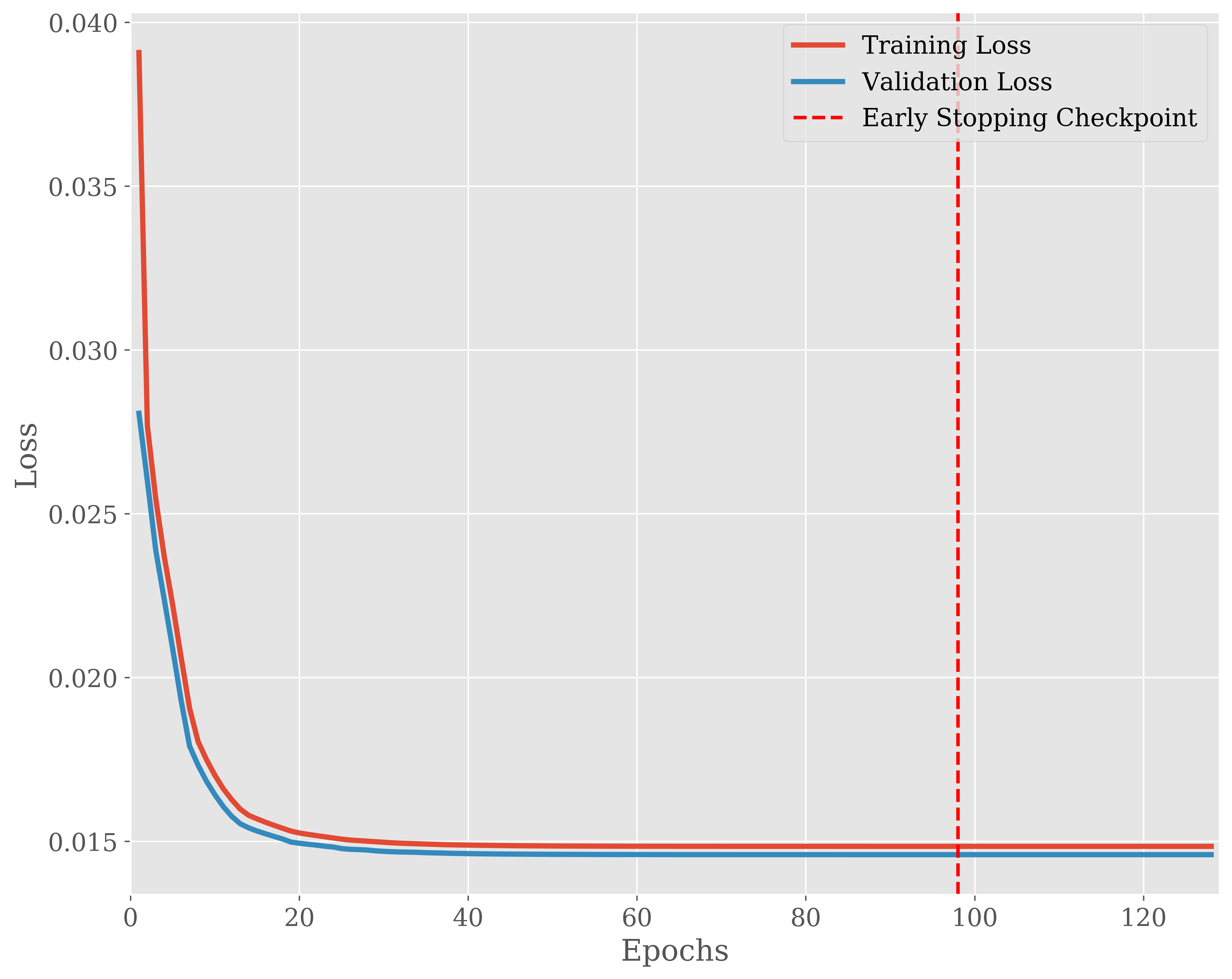}\\
   \includegraphics[width=.16\textwidth]{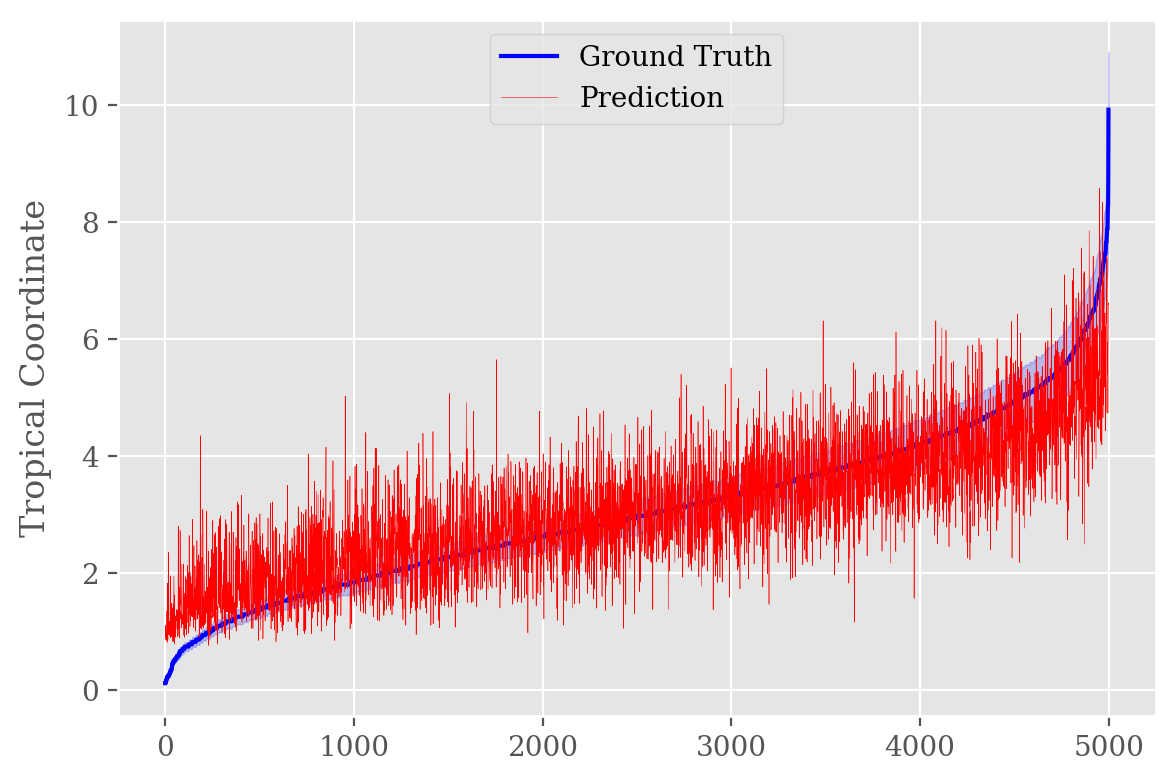}
    &
   \includegraphics[width=.16\textwidth]{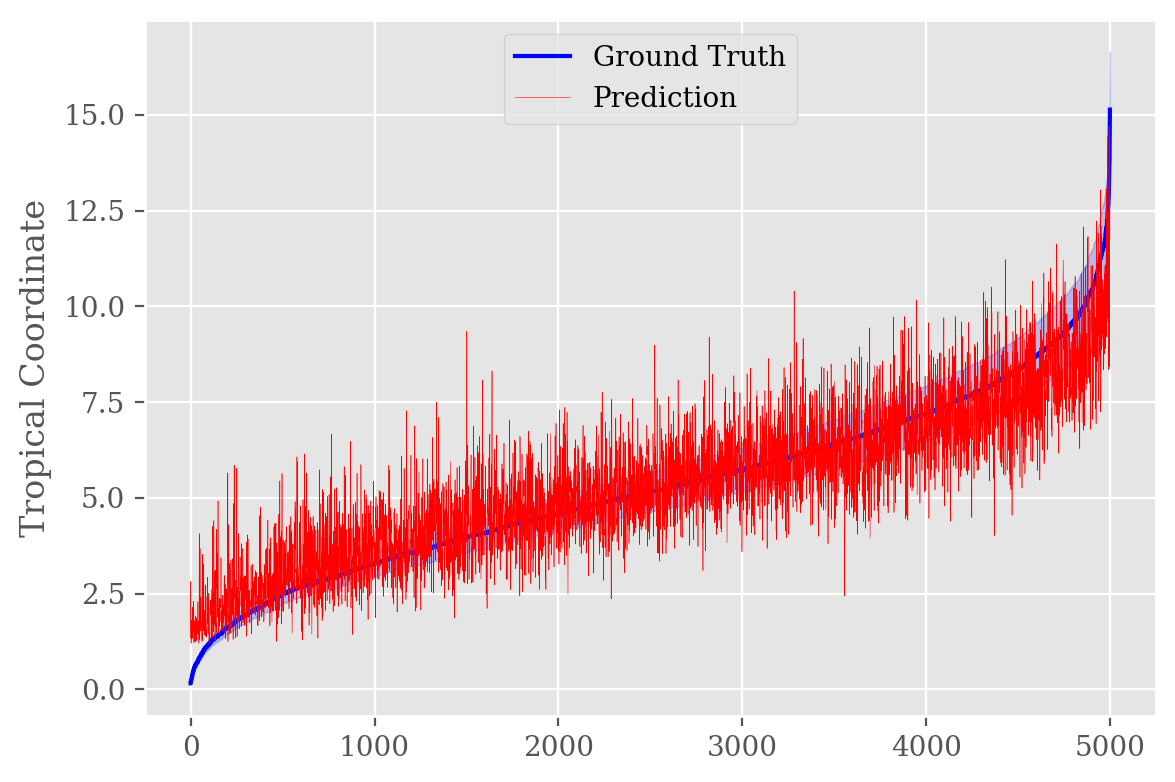}
    & 
   \includegraphics[width=.16\textwidth]{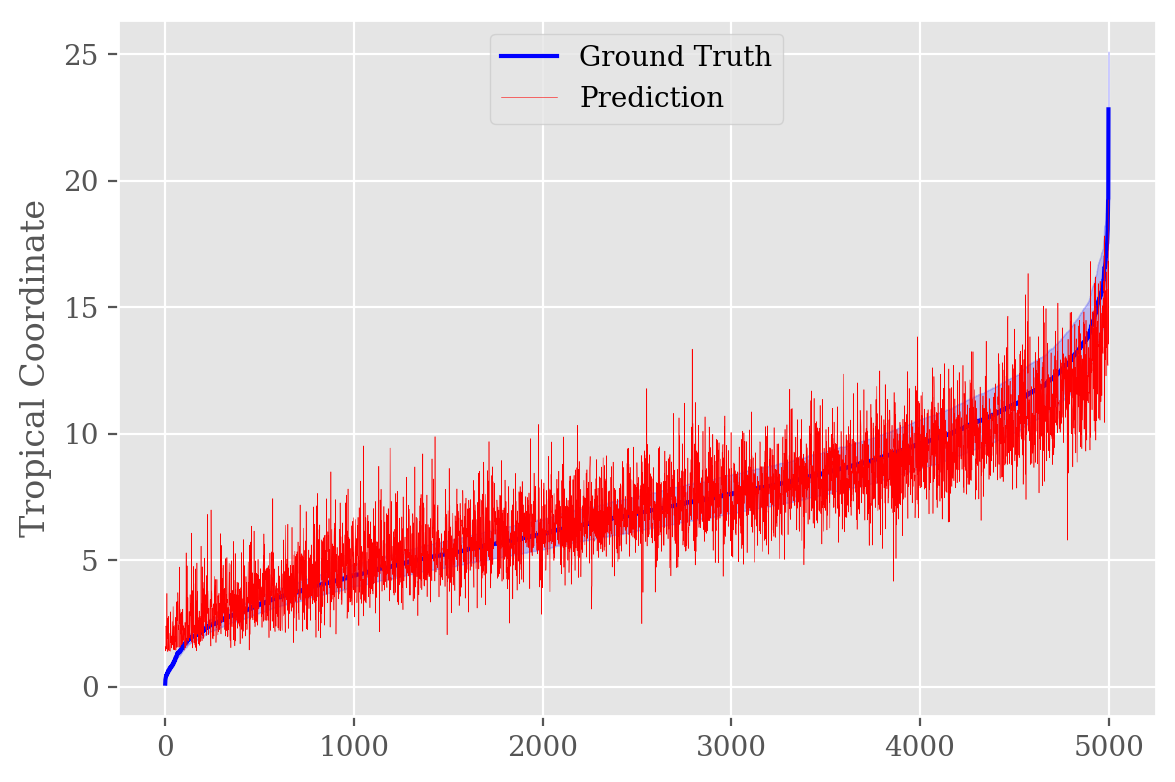}
    & 
   \includegraphics[width=.16\textwidth]{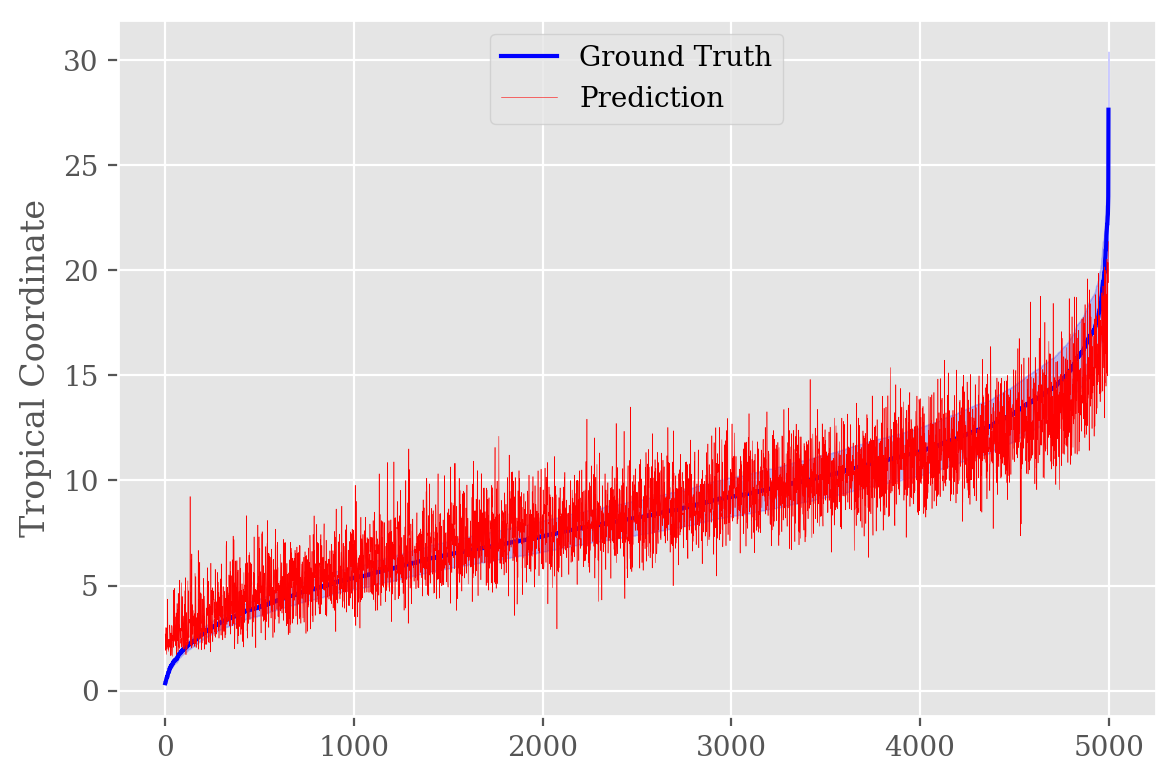}
    &
   \includegraphics[width=.16\textwidth]{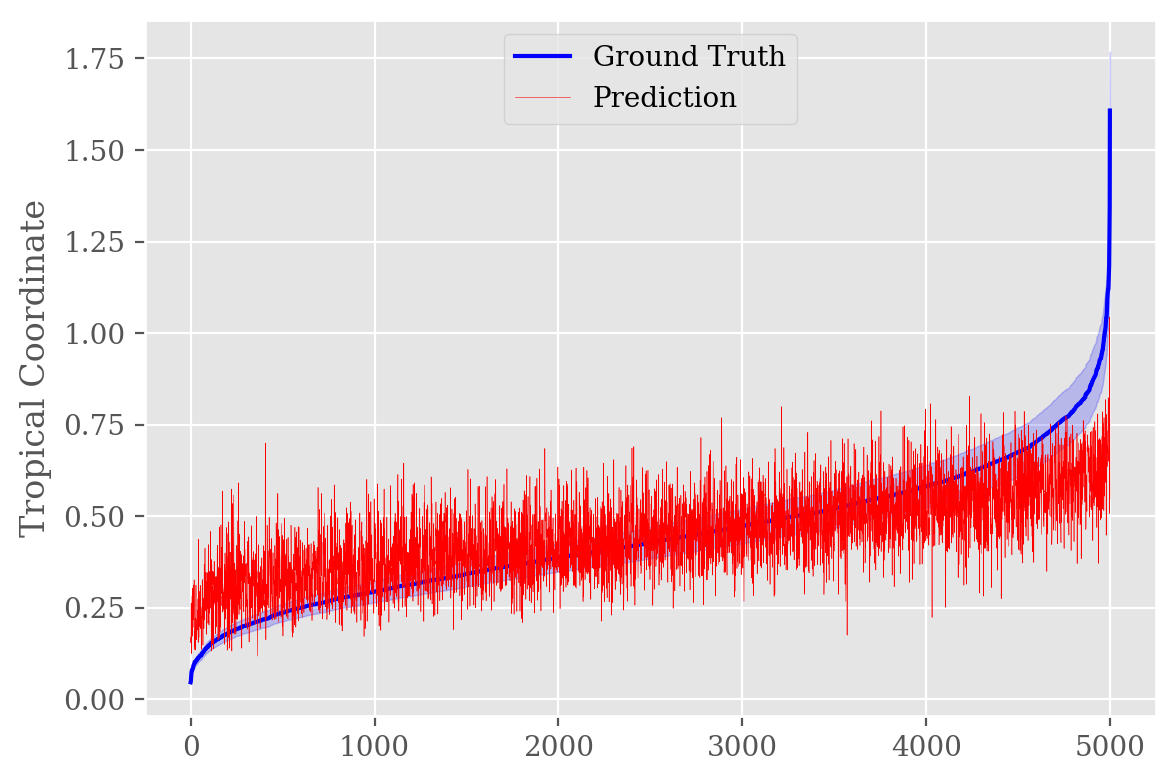}
\end{tabular}
    \caption{For MNIST and CIFAR-10:  
    First row: Training and validation losses during training a CNN for $\max_i\{d_i\}, \max_{i<j}\{d_i+d_j\}, \max_{i<j<k}\{d_i+d_j+d_k\}, \max_{i<j<k<l}\{d_i+d_j+d_k+d_l\}$ and mean length of bars. 
    Second row: Predicted tropical coordinates (red) vs ground truth (blue). The $x$-axis is the order index of all image samples, where it is in the increasing order of ground truth values.}
        \label{fig:mnist-cifar-regress}
\end{figure}

In this experiment, we seek to learn the map from original images to four tropical coordinates and the mean bar-length of their persistence diagrams. We model this as a mean square error regression task with tropical coordinates scaled by a factor of 10 to improve numerical computation. As in Section~\ref{SS:images}, we use a CNN model with LeNet-5 architecture for MNIST and CIFAR-10. 
In Figure~\ref{fig:mnist-cifar-regress}, the first and third rows show the training and validation losses during training. 
The second and fourth rows show the predictions (in red) of the trained networks on the test data set, compared with the ground truth (in blue). 
As we observe, training is fairly quick, and in most cases, the CNNs can predict the tropical coordinates with relatively good precision.
Predictions of the mean bar length are less accurate in our experiment. We think that hyperparameter tuning can further improve the results. 
More details on this experiment are provided in Appendix~\ref{regression_experiment}.  

%%
%CONCLUSIONS
%%
\section{Conclusions and outlook}
Using the MNIST and CIFAR-10 data sets,  we show that we can train neural networks to compute several types of features of persistence diagrams. 
In our experiments (details in Appendix~\ref{app:compute-time}) a trained CNN can produce approximate values of persistence diagram features in $0.005$s, which take $3$s to compute with traditional methods. Two typical applications of TDA are homological inference and classification. In ongoing work, we are working on substituting the exact features with the approximations obtained by neural networks and evaluating how competitive they are.
We suggest that neural networks (and GPUs) can be exploited to compute topological features for data sets that are traditionally computationally expensive, such as large 3D point clouds and 3D images.

\FloatBarrier

\subsection*{Acknowledgment}
This material is based in part on work conducted during the Collaborate@ICERM on ``Geometry of Data and Networks'' while the authors were in residence at ICERM, supported by the National Science Foundation under Grant No.\ DMS-1439786. 
This project has received funding from the European Research Council (ERC) under the European Union's Horizon 2020 research and innovation programme (grant agreement no 757983).

%\newpage 

\bibliographystyle{plain}
\bibliography{scnn}

\appendix

\section*{Supplementary material}

\section{Persistent homology computation pipeline} 

\subsection{Persistence barcodes and diagrams}
\label{app:barcodes}

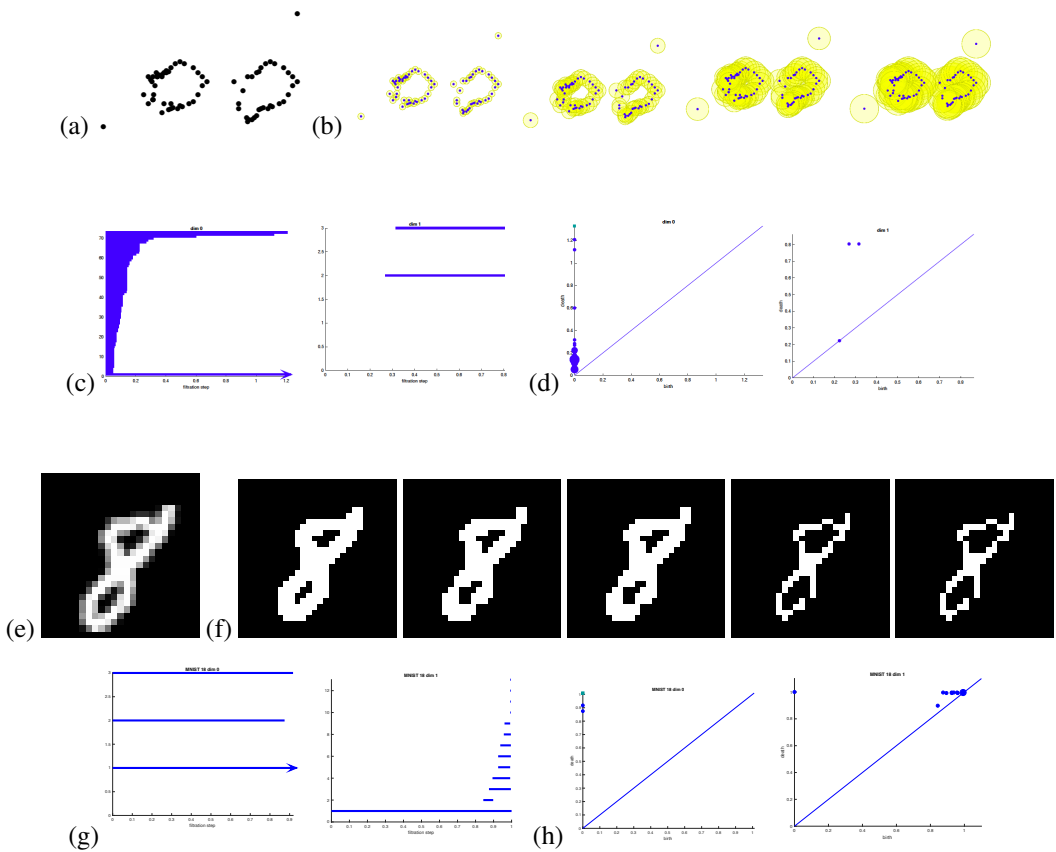
\begin{figure}[h!]
\centering
(a)
\begin{tikzpicture}[scale=0.6, every node/.style={scale=0.5}]]
\foreach \position in
{
(2,2),(2.1,2.3),(1.9,2.5),
(2.2,2.6), (2.3,2.6),(2.4,2.7),
(2.45,2.72),(2.5,2.8),(2.6,2.9),
(2.7,2.95),(2.8,2.9),
(3,2.8),(3.1,2.7),(3.2,2.6),
(3.3,2.5),(3.2,2.4),(3.1,2.2),
(3,2.05),(2.9,2),(2.75,1.95),
(2.7,1.97),(2.65,1.9),(2.55,1.92),
(2.5,1.9),(2.45,1.88),
(2.4,1.75),
(2.35,1.7),(2.3,1.65),(2.2,1.6),
(2.22,1.65),(2.18,1.7),(2,1.9),
(0,2),(0.1,2.1),(-0.1,2.2),
(0.2,2.6), (0.3,2.6),(0.4,2.7),
(0.45,2.72),(0.5,2.8),(0.6,2.9),
(0.7,2.95),(0.8,2.9),
(1,2.8),(1.1,2.7),(1.2,2.6),
(1.3,2.5),(1.2,2.4),(1.1,2.2),
(1,2.05),(0.9,2),(0.75,1.95),
(0.7,1.97),(0.65,1.9),(0.55,1.92),
(0.5,1.9),(0.45,1.88),
(0.4,2.75),
(0.35,2.7),(0.3,2.65),(0.2,2.6),
(0.22,2.65),(0.18,2.7),(0,2.6),
(0.1,2.1),(0.12,2.2),(0.3,1.8),
(0.35,2),(0.5,2.),(0.05,2.55),
(0.08,2.5),(0.1,2.6),(0.15,2.4),
(3.3,4),(-1,1.5)
}
\node at \position []{$\bullet$};
\end{tikzpicture}
(b)
\includegraphics[width=0.15\textwidth]{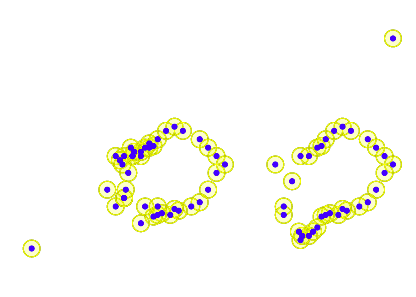}
\includegraphics[width=0.15\textwidth]{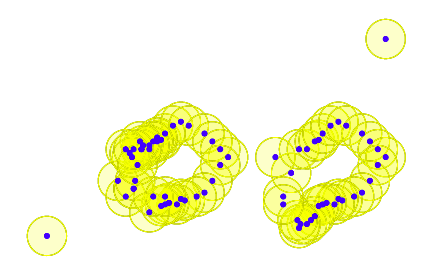}
\includegraphics[width=0.15\textwidth]{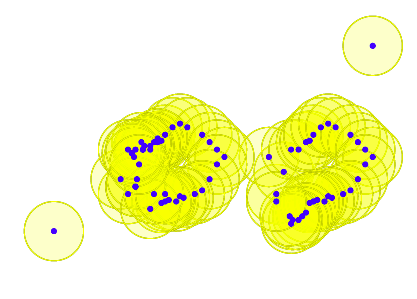}
\includegraphics[width=0.15\textwidth]{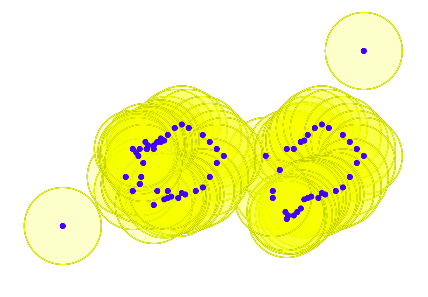}\\
\vspace{0.8cm}
(c)\includegraphics[width=0.2\textwidth]{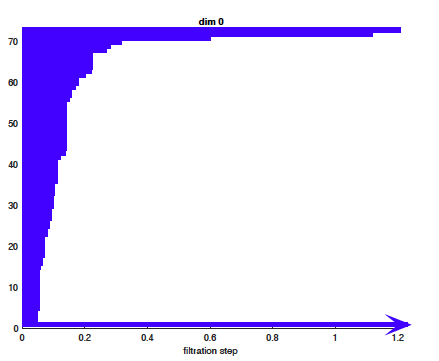}
\includegraphics[width=0.2\textwidth]{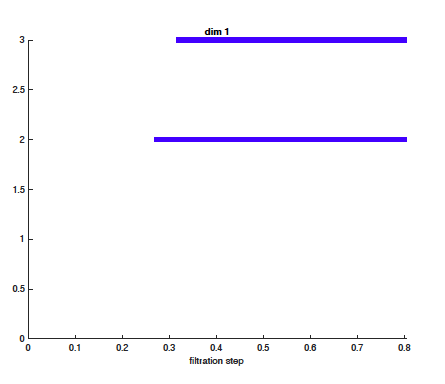}
(d)\includegraphics[width=0.2\textwidth]{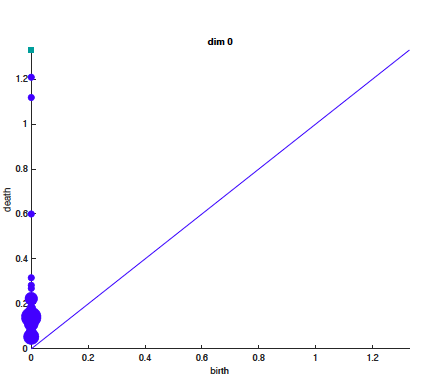}
\includegraphics[width=0.2\textwidth]{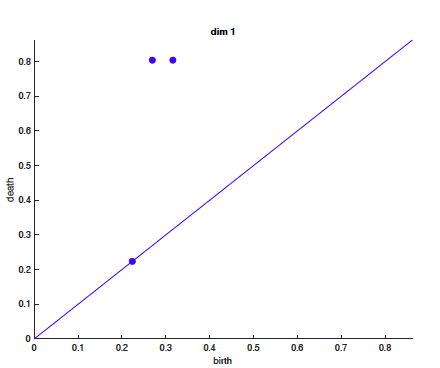}
\vspace{1cm}

(e) \includegraphics[width=0.15\textwidth]{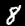}
(f) \includegraphics[width=0.15\textwidth]{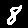}
\includegraphics[width=0.15\textwidth]{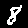}
\includegraphics[width=0.15\textwidth]{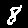}
\includegraphics[width=0.15\textwidth]{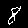}
\includegraphics[width=0.15\textwidth]{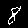}\\

(g)\includegraphics[width=0.2\textwidth]{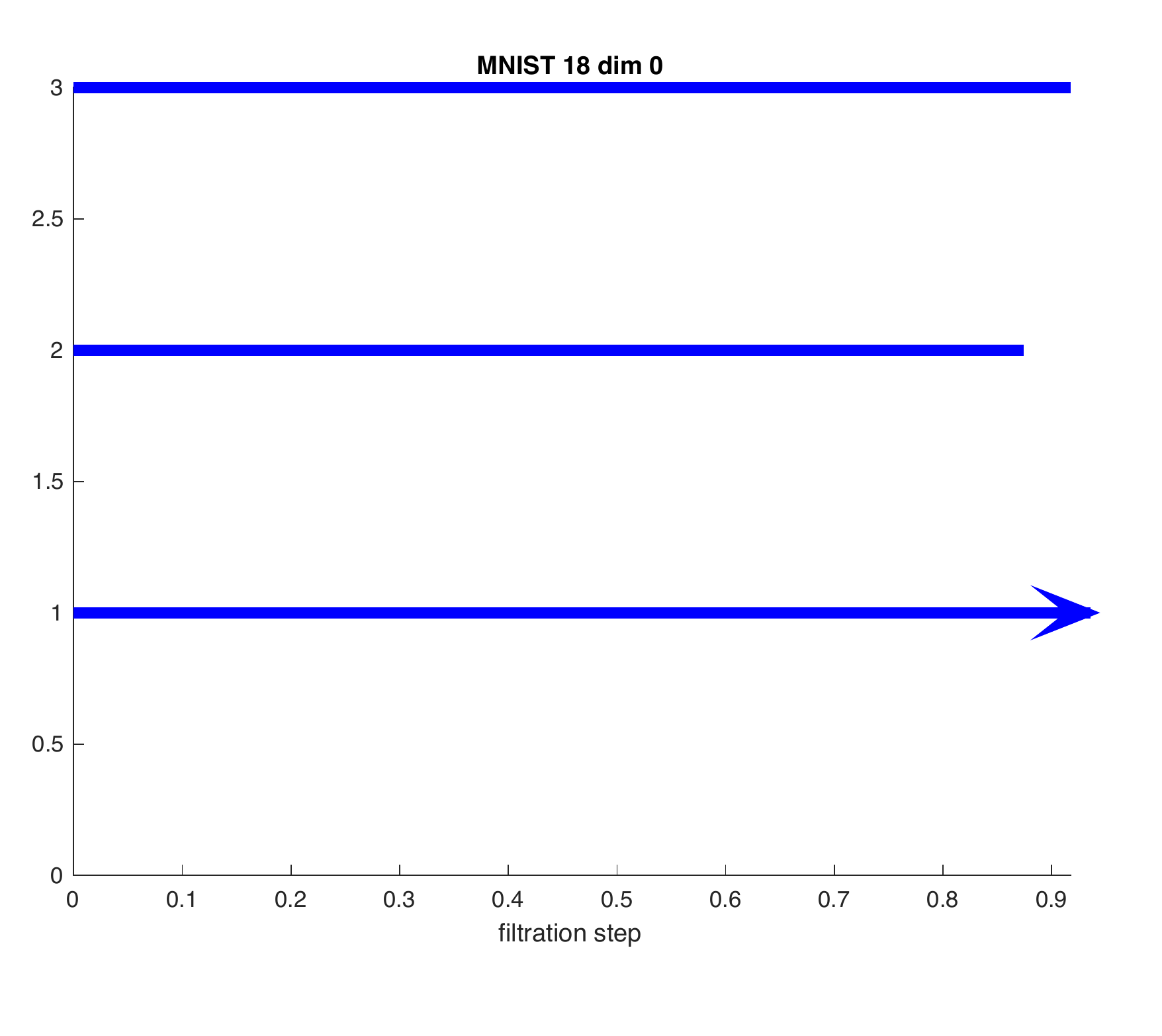}
\includegraphics[width=0.2\textwidth]{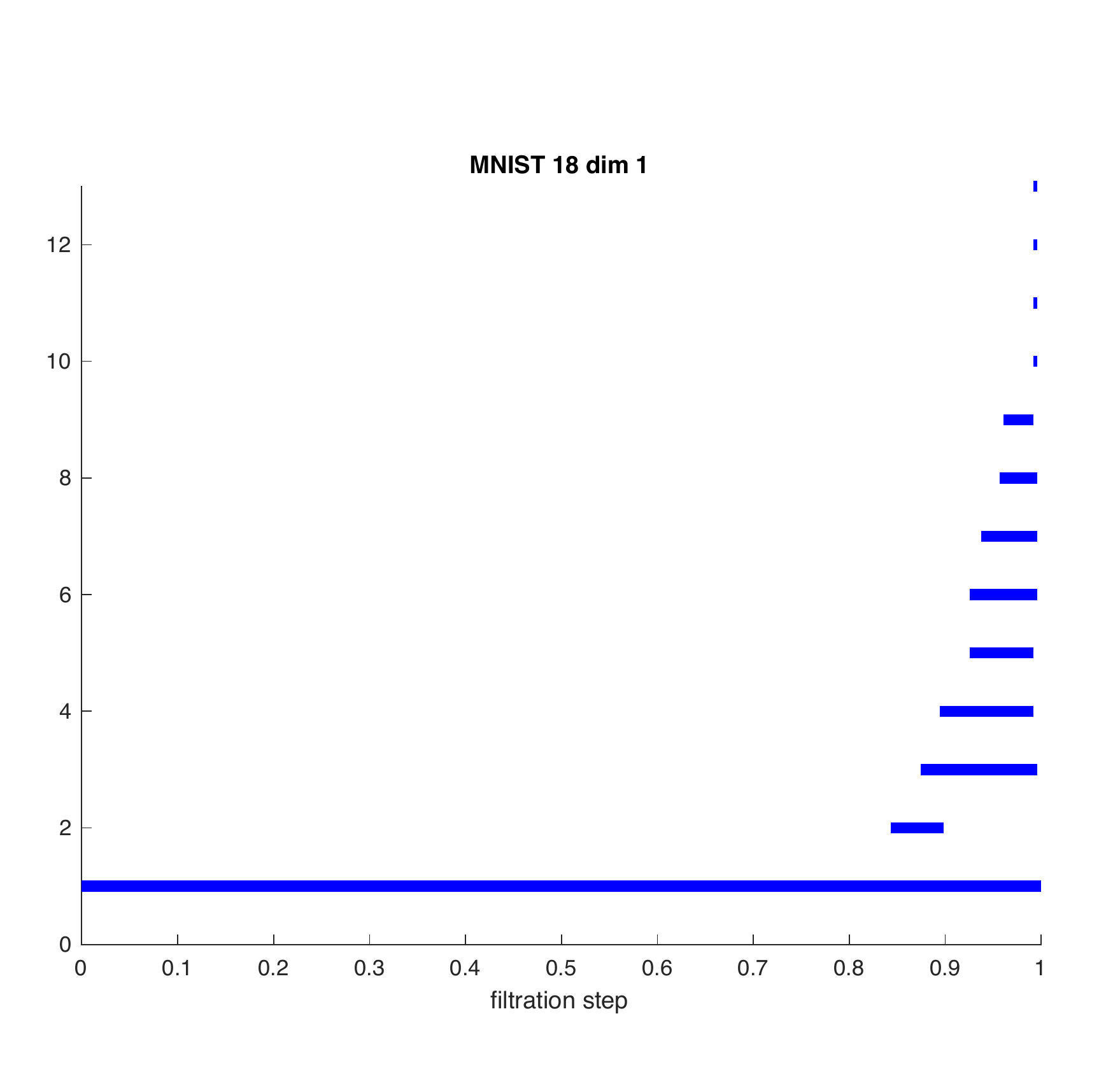}
(h)\includegraphics[width=0.2\textwidth]{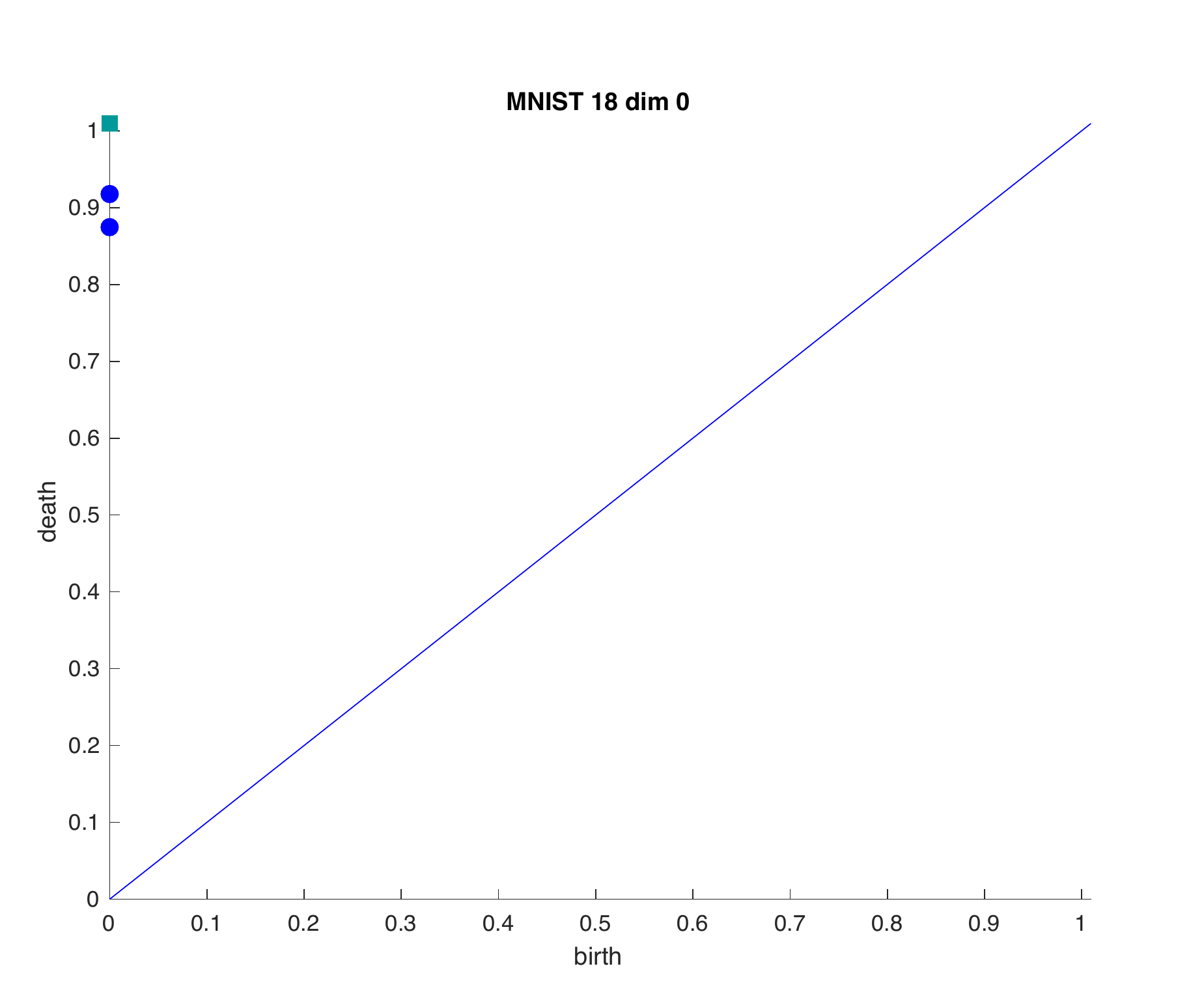}
\includegraphics[width=0.2\textwidth]{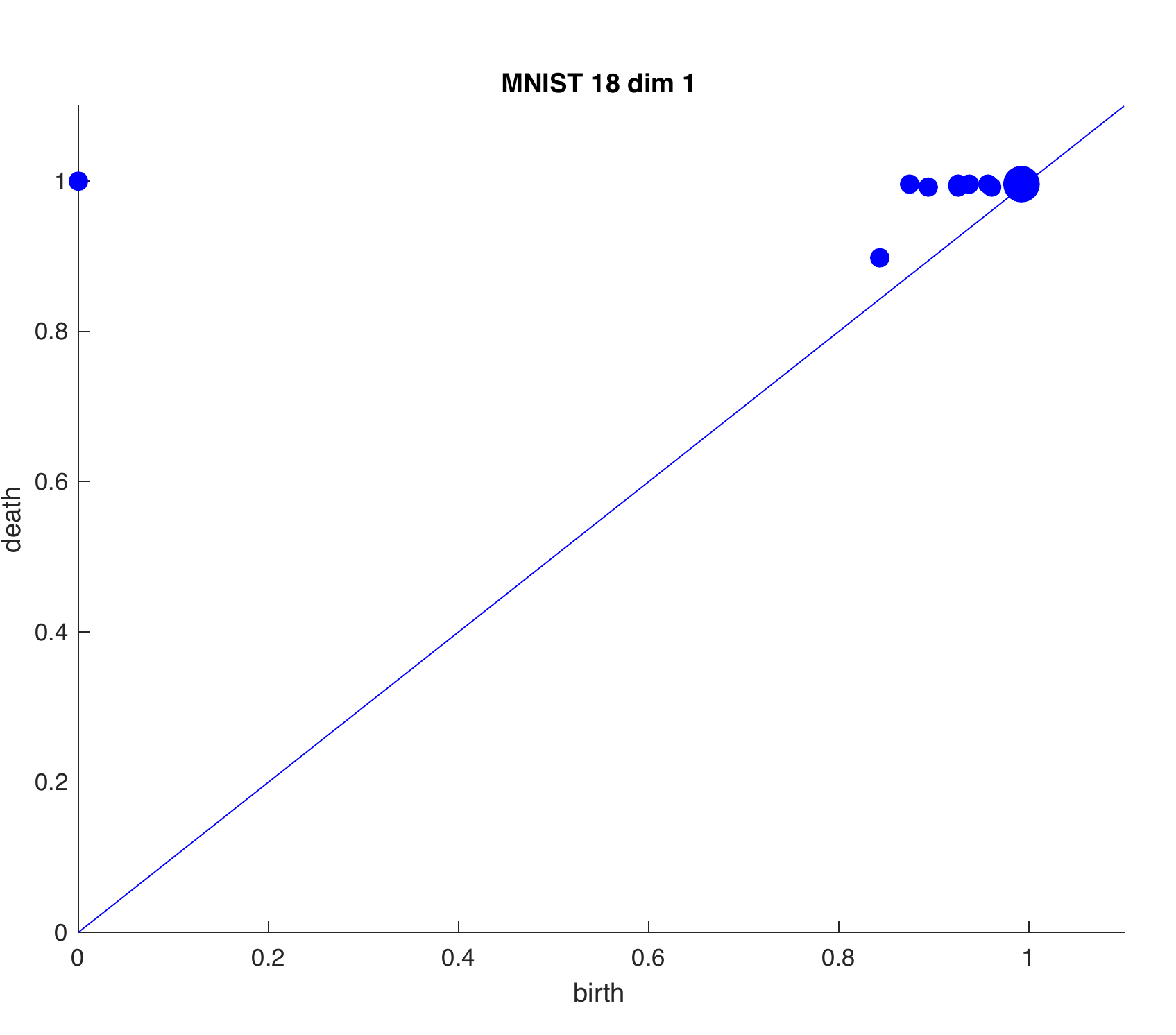}

\caption{(a) A finite metric space. (b) A filtration of nested spaces obtained by taking the union of  balls with increasing radius values around the points. (c) Barcodes describing the lifetime of components (left) and holes (right). (d) The corresponding persistence diagrams. (e) A grey-scale digital image. (f) A filtration of nested spaces obtained by thresholding pixels by increasing grey values. (g) Barcodes describing the lifetime of components and holes, and (h) the corresponding persistence diagrams.
}
\label{F:PH}
\end{figure}

Persistent homology is one of the most successful methods in topological data analysis. Given a finite metric space as in Figure~\ref{F:PH}(a), one considers a ``thickening" of the metric space at different distance scales, which gives a nested sequence (a so-called filtration) of spaces, and then analyses the evolution (``persistence") of topological features across this filtration (see Figure~\ref{F:PH}(b)). Given a grey-scale digital image, one can associate to it a filtration by thresholding the pixels or voxels by increasing grey values (see Figure~\ref{F:PH}(f)).

The topological features that one examines include connected components, holes and voids. The barcode is an algebraic invariant that summarises how topological features of a certain dimension evolve across the nested sequence: the left endpoint of an interval represents the birth of a feature, while its right endpoint represents the death of the same feature. When a feature is still ``alive" at the largest radius or grey-scale value that one considers, the lifetime interval is infinite (see Figure~\ref{F:PH}(c) and (g)). 

An alternative way to represent a barcode is what is called a ``persistence diagram'': this is a multiset of points in $\mathbb{R}^2$ where we represent an interval $[a,b)$ in the barcode by a point $(a,b)$ (see Figure~\ref{F:PH}(d) and (h)).

\subsection{Images, cubical complexes, boundary matrices}
\label{A:pipeline}
In practice, one cannot work with filtrations of spaces such as the ones in Figure \ref{F:PH} and instead needs combinatorial approximations of such spaces. Thus, to each of the thresholded images in the filtration in Figure \ref{F:PH}(f) we associate a combinatorial approximation of it called cubical complex, which is a space built out of vertices, edges, and squares.
In ongoing work we are considering  alternative ways to associate a filtration to a digital image. For instance, one could threshold pixels by decreasing grey values, thus obtaining a filtration consisting of  ``positive'' versions of the images in Figure \ref{F:PH}(f), or by fixing a certain threshold for grey values, and then creating a filtration for the thresholded image by swiping through the pixels to capture spatial information, in a similar way as done in \cite{ACC16}.

In the present work, to  each digital image we associate a filtered cubical complex using the algorithm from \cite{wagner2012efficient}. 
We  then train the neural network using different types of input data, corresponding to the following  steps in the PH pipeline (see Example \ref{E:input data}): 

\begin{enumerate} 
\item Original image. 

\item \label{item cc} Cubical complex (CC): we associate a cubical complex with the image as follows. We represent pixels by vertices, we join vertices corresponding to adjacent pixels by an edge, and we join quadruples of vertices by squares if the corresponding pixels are pairwise adjacent. We then extend the grey values to the cells (i.e., vertices, edges and squares) in the complex by associating a cell with the maximum grey value of the pixels corresponding to it.  We represent such a cubical complex with a 
matrix where each entry corresponds to a cell, and each entry stores the (extended) grey value of the corresponding cell.

\item \label{item fcc} Filtered cubical complex (FCC): we put a total order on the cells of the cubical complex in the previous item in such  a way that the following two conditions are satisfied: (i) for each cell its faces have to appear before the cell in the order; and (ii) if a cell has smaller grey value than another cell, then it has to appear before it  in the order. 
We represent such a filtered cubical complex by a
%$55\times55$
matrix in which each entry corresponds to a cell, and each entry stores the order of the corresponding cube.

\item \label{I:bm}  Boundary matrix: we use the total order from the previous item to label rows and columns of a matrix by the cells of the cubical complex. The matrix stores adjacency information between a cell and its faces of codimension $1$:  the $(i,j)$th entry of the matrix contains a $1$ if the $i$th cell in the order is a face of codimension $1$ of the  $j$th cell in the order.
In our experiments, we symmetrize the boundary matrix, and we interpret it as and adjacency matrix of a graph in which vertices correspond to cells and edges encode adjacency information between cells whose dimension differs by $1$.
\end{enumerate}

\begin{example}\label{E:input data}
\begin{enumerate}
\item Original image. 
Consider the $2\times 2$ image given by the following array of grey values
\[
\begin{pmatrix}%{cc}
1 & 3\\
3 & 2
\end{pmatrix}. 
\]

\item Image extended to the cubical complex. 

The grey values extended to the cubical complex for the above image are: 
\[
\begin{pmatrix}%{ccc}
1  &   3  &   3\\
3  &   3  &   3\\
3  &   3   &  2
\end{pmatrix} 
\qquad \qquad
%. 
%
\begin{array}{ccccc}
    \tikzmarkin[hor=style orange]{v1} 1 \tikzmarkend{v1} & \tikzmarkin[hor=style green]{e12}    & 3 & \tikzmarkend{e12} & \tikzmarkin[hor=style orange]{v2} 3 \tikzmarkend{v2} \\
      \tikzmarkin[hor=style green]{e13} &\tikzmarkin[hor=style cyan]{square} & & & \tikzmarkin[hor=style green]{e24}\\
      3 & & 3 & & 3\\
     \tikzmarkend[hor=style green]{e13} & & &\tikzmarkend[hor=style cyan]{square} & \tikzmarkend[hor=style green]{e24}\\
    \tikzmarkin[hor=style orange]{v3} 3 \tikzmarkend{v3}  &   \tikzmarkin[hor=style green]{e34} & 3  &  \tikzmarkend[hor=style green]{e34}& \tikzmarkin[hor=style orange]{v4} 2 \tikzmarkend{v4} %\\
  \end{array}
\]

\item The filtered cubical complex.
Here we choose the total order in which cells whose faces are already in the filtration are listed before other cells of lower dimension. For instance, the edge between the vertices labelled by $3$ and $2$ as the 4th place in the order, whereas the vertex in the top right corner appears at the 6th place. Another possible way to totally order the cells would be given by listing cells according to first grey value and then dimension; thus, one would list first all cells with grey value 1, in increasing order of dimension, then all cells with grey value $2$, and so on. In our example, the order is given as: 
\begin{alignat}{2}
\begin{pmatrix}%{ccc}
1  &   8   &  6\\
5  &   9  &   7\\
3  &   4  &   2
\end{pmatrix}
 & \notag \qquad \qquad
\begin{array}{ccccc}
    \tikzmarkin[hor=style orange]{vv1} 1 \tikzmarkend{vv1} & \tikzmarkin[hor=style green]{ee12}    & 8 & \tikzmarkend{ee12} & \tikzmarkin[hor=style orange]{vv2} 6 \tikzmarkend{vv2} \\
      \tikzmarkin[hor=style green]{ee13} &\tikzmarkin[hor=style cyan]{ssquare} & & & \tikzmarkin[hor=style green]{ee24}\\
      5 & & 9 & & 7\\
     \tikzmarkend[hor=style green]{ee13} & & &\tikzmarkend[hor=style cyan]{ssquare} & \tikzmarkend[hor=style green]{ee24}\\
    \tikzmarkin[hor=style orange]{vv3} 3 \tikzmarkend{vv3}  &   \tikzmarkin[hor=style green]{ee34} & 4  &  \tikzmarkend[hor=style green]{ee34}& \tikzmarkin[hor=style orange]{vv4} 2 \tikzmarkend{vv4}
  \end{array}
\end{alignat}
\item The boundary matrix.  In our example we obtain the following matrix:

\[
B = 
\begin{pmatrix}%{ccccccccc}
0  &  0  &   0   &  0   &  1  &   0  &   0  &   1   &  0 \\
0  & 0 & 0  & 1 & 0  &   0   &  1  &   0  &   0 \\
0 & 0 & 0  &   1   &  1  &   0  &   0 &    0  &   0 \\
0  & 0  & 0 & 0 & 0 &  0    & 0 &    0& 1\\
0  & 0    & 0 &    0 &  0 &    0   & 0 &   0 &  1\\
0  & 0 & 0 & 0 & 0 &     0&     1    & 1 &  0\\
0  & 0  & 0 & 0  &   0    & 0    & 0    & 0 &1\\
0  & 0  &  0 & 0  &   0   &  0   &  0  &   0 & 1\\
0  & 0  & 0 & 0  & 0 & 0 & 0 & 0 & 0
\end{pmatrix} .
\]
In our experiments we use the symmetrized boundary matrix
$
B_{\rm sym} = B + B^{\top} \, .
$
\end{enumerate}
\end{example}

Given a  boundary matrix as in item \ref{I:bm}, one can reduce it using standard linear-algebra methods. One can then  read off from the reduced matrix the birth-death pairs that constitute the intervals in the barcode in homological degree $0$ (i.e., for components) and in homological degree $1$ (i.e., for holes). See Figure \ref{F:PH} for an illustration of such barcodes.  We point the reader to \cite[Section 5.3]{otter2017roadmap} for a discussion of different algorithms to reduce a boundary matrix, and of how to read off the intervals from the reduced matrix. We note that while we train the neural network using the optimized filtrations of cubical complexes described in items \ref{item cc}--\ref{item fcc}, and introduced in \cite{wagner2012efficient}, to compute the persistence diagrams we use the software library GUDHI \cite{gudhi}, which implements a different algorithm. However, the persistence diagrams obtained with the two types of filtrations are the same.

\subsection{Features from persistence diagrams}
\label{app:features}

In recent years, many ways to vectorize persistence diagrams have been studied. In our work, we consider mainly two of them: tropical coordinates and persistence images. In addition to these, in future work we will  study also persistence landscapes \cite{B15}, which are vectorizations of persistence diagrams  that have been widely used in applications of persistent homology.

\paragraph{Tropical coordinates} 
Tropical coordinates are expressions involving coordinates $(x_i,x_j)$ of points in a diagram, 
distances $d_k$ of points from  the diagonal, standard addition and $\max$. 
They were introduced in \cite{Kalisnik} as a stable version of vectorisations of persistence diagrams \cite{ACC16}. 
Tropical coordinates that we consider in our work include the following: 

\begin{alignat}{4}
&\notag(1) \max_j d_j,\quad
(2) \max_{i<j}\{d_i+d_j\},\quad
(3) \max_{i<j<k}\{d_i+d_j+d_k\},\; \\
&\notag(4) \max_{i<j<k<l}\{d_i+d_j+d_k+d_l\},\quad
(5) \operatorname{mean} \{ d_j\} \, .
\end{alignat}

\paragraph{Persistence images}
\label{SS:PI}
Persistence images \cite{PI} are obtained from persistence diagrams by a weighted sum of kernels at the birth-death locations in $2$D. This representation depends on three parameters that need to be specified: (i) Resolution, (ii) Probability distribution, (iii) Weighting function. 

Let $BC\in\mathbb{R}^{N\times 2}$ be a list of birth-dead coordinates (barcode). Let $T\colon \mathbb{R}^2\to\mathbb{R}^2; (x,y) \mapsto (x,y-x)$ be the linear transformation mapping the upper half of the positive quadrant to the positive quadrant. Let $\phi_u \colon \mathbb{R}^2\mapsto \mathbb{R}$ be a density function with mean $u\in\mathbb{R}^2$. We take simply a Gaussian with mean $u$ and covariance matrix $a I$, where $a >0$ is a hyperparameter. 
Let $f\colon \mathbb{R}^2\mapsto \mathbb{R}$ be a weight function defined by 
\begin{equation*}f (u_1,u_2) = \left\{\begin{array}{ll}0, & u_2\leq 0\\
u_2/b,& 0< u_2< b\\
1,& u_2\geq b,
\end{array}\right.
\end{equation*} 
where $b$ is a hyperparameter. 
Then the persistence image of the barcode $BC$ obtained with hyperparameters $a,b$ is the function $\PI\colon \mathbb{R}^2\to\mathbb{R}$ defined by 
$$
\PI(v) = \sum_{u\in T(BC)} f(u) \phi_u(v). 
$$
Usually we will consider only the values at points in a discrete grid $v\in\{p_1,\ldots, p_{n}\} \times \{q_1,\ldots, q_{n}\}$. Here $n$ corresponds to the resolution of the persistence image. 
We can also consider the value obtained by averaging over cells around the grid points, i.e.\ 
$\PI'(v) = \frac{1}{|p_{i+1}-p_{i}|\cdot |q_{j+1}-q_j|}\int_{[p_{i},p_{i+1})\times [q_{j} ,q_{j+1})} \PI(u) du$, where $v=(p_i,q_j)$.

\paragraph{Features from persistence images} 
A persistence image is a high-dimensional representation (depending on the specified resolution). 
For downstream tasks, it can be convenient to consider a few salient features of the image, instead of simply reducing the resolution. 
An example of notable features are discrete Fourier coefficients obtained as 
$$
\phi_k = \sum_v \psi_k(v) \PI(v), 
$$
where $\psi_k\colon \mathbb{R}^2\to\mathbb{R}$; \;$\exp(-2\pi i v\cdot k)$, and we take $v$ in the summation over regular grids of an image. To be consistent with 2D discrete Fourier transform in Matlab routine, we do not normalised the Fourier basis.  

Another example of a feature is the number of ``blobs'' in the persistence image. Here we regard the persistence image as a real-valued function on $\mathbb{R}^2$ and define a blob simply as a local maximizer, which can be weighted by the volume of its basin.

\section{Further details on the experiments}

\subsection{Mapping images to binary features of their persistence diagrams}
\label{app:image-to-bin}
\begin{figure}[h]
    \centering
MNIST - Training and validation losses for predicting bar\\[1mm]
\begin{tabular}{ccc}
  \includegraphics[width=.3\textwidth]{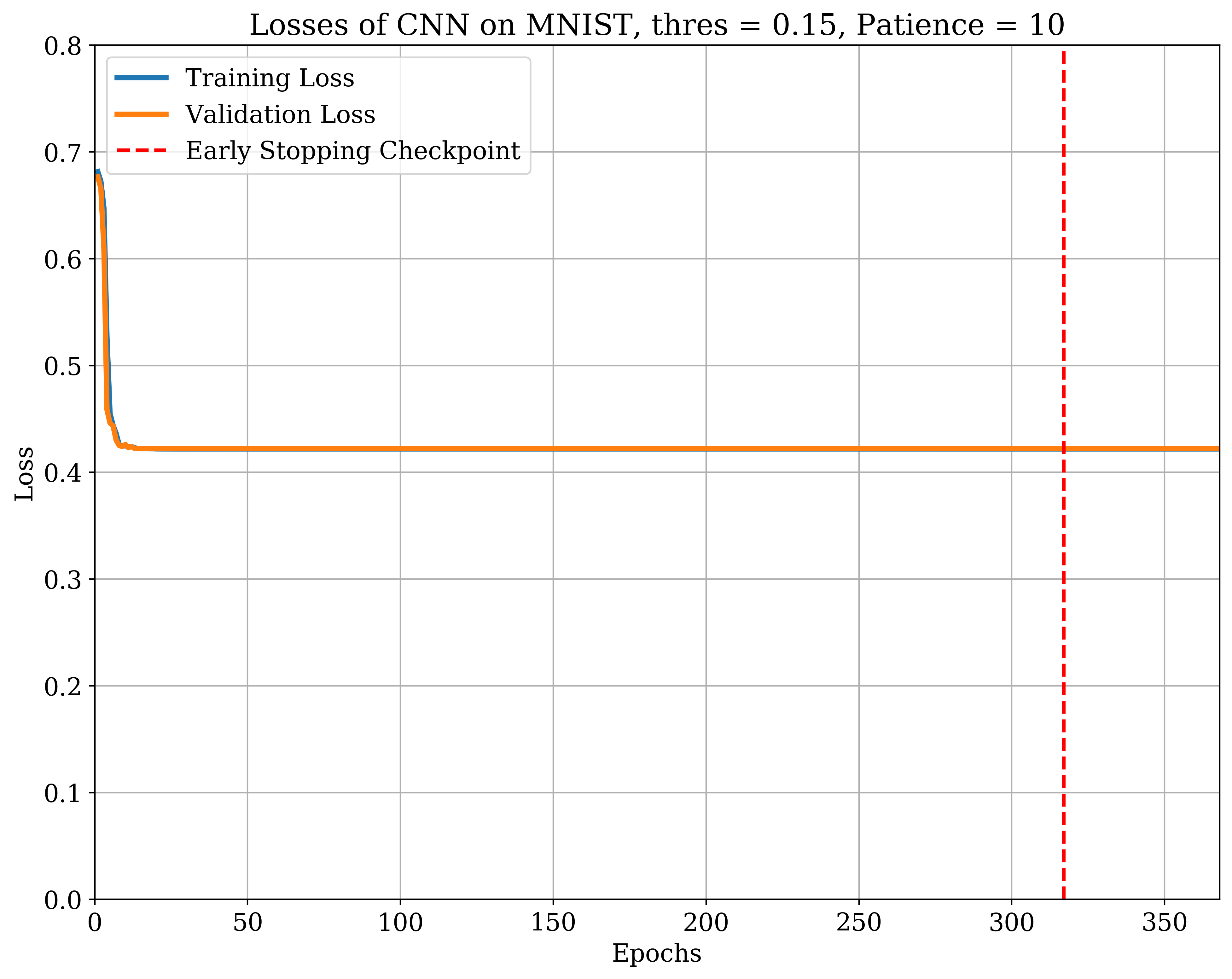} 
    &\includegraphics[width=.3\textwidth]{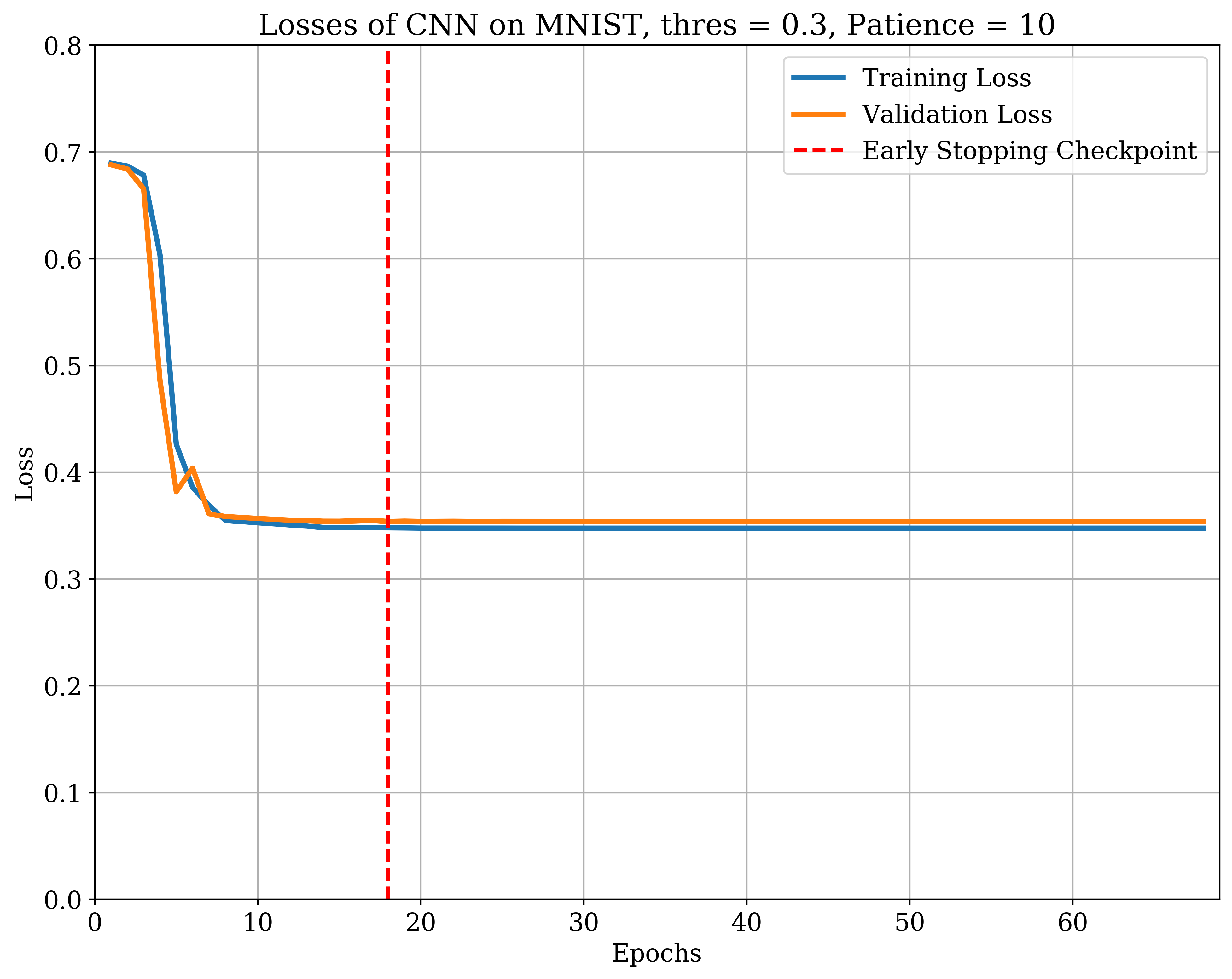}
    &\includegraphics[width=.3\textwidth]{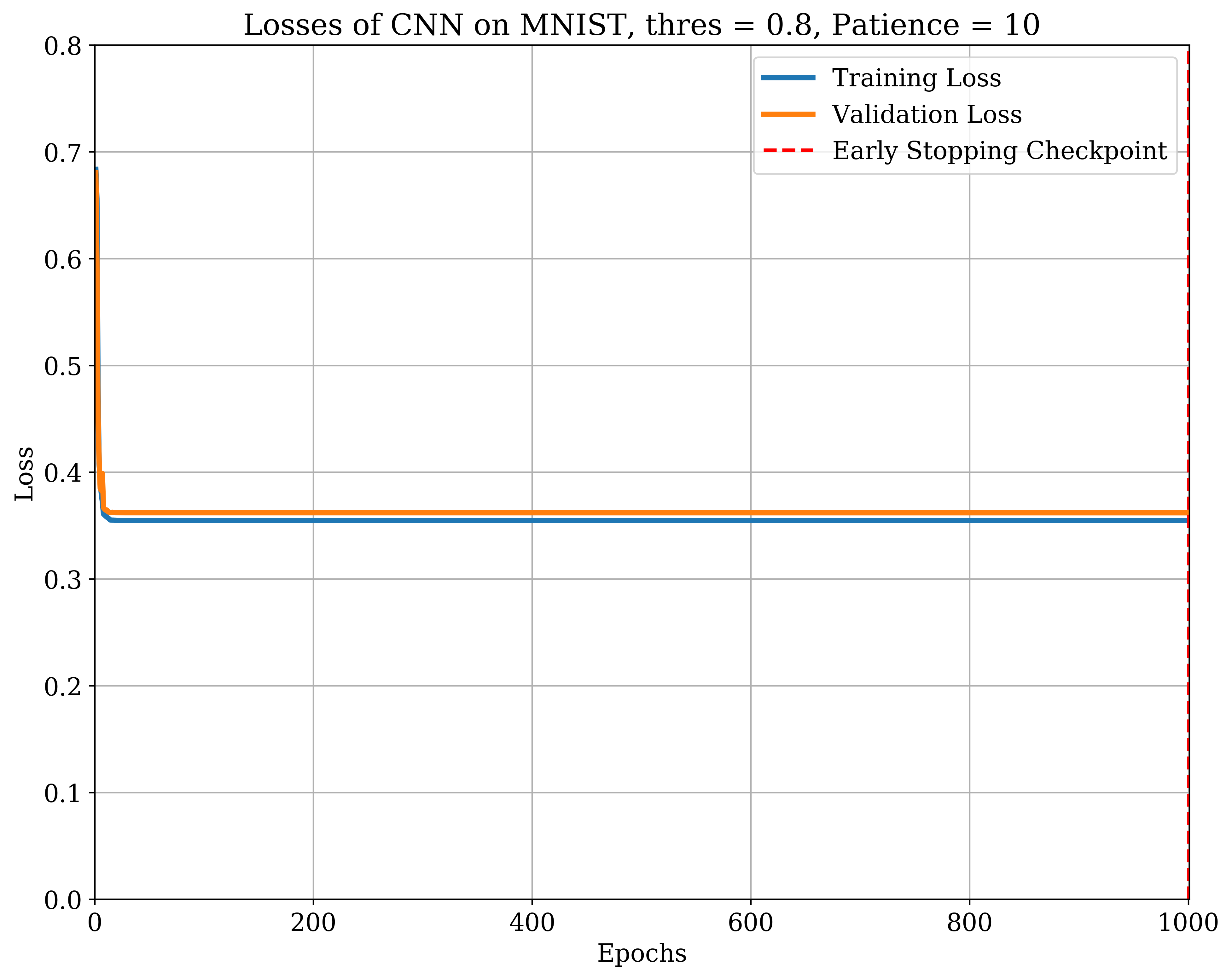}
    \vspace{4mm}
\end{tabular}
MNIST - Validation accuracy for predicting bar\\[1mm]
\begin{tabular}{ccc}
  \includegraphics[width=.3\textwidth]{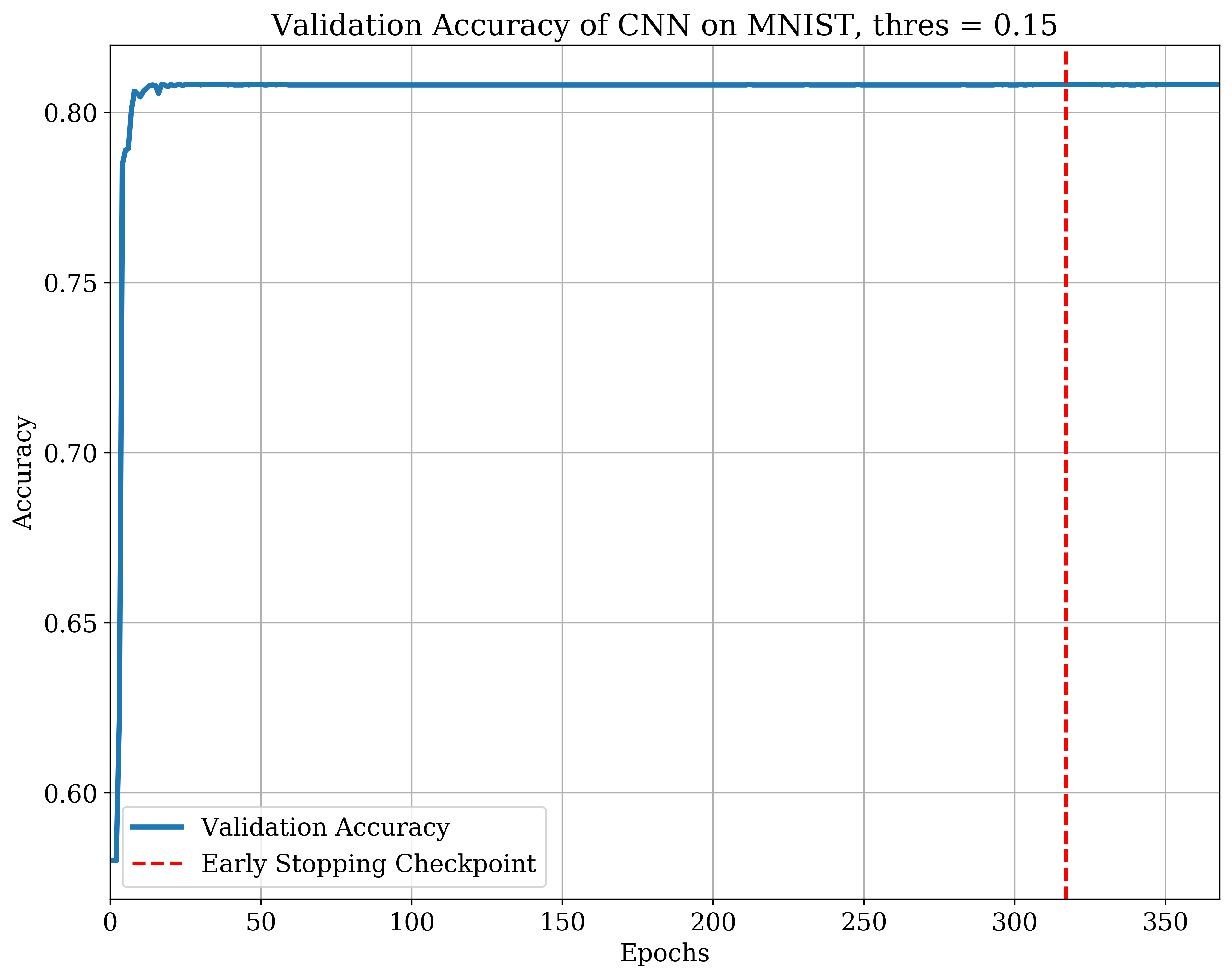} 
    &\includegraphics[width=.3\textwidth]{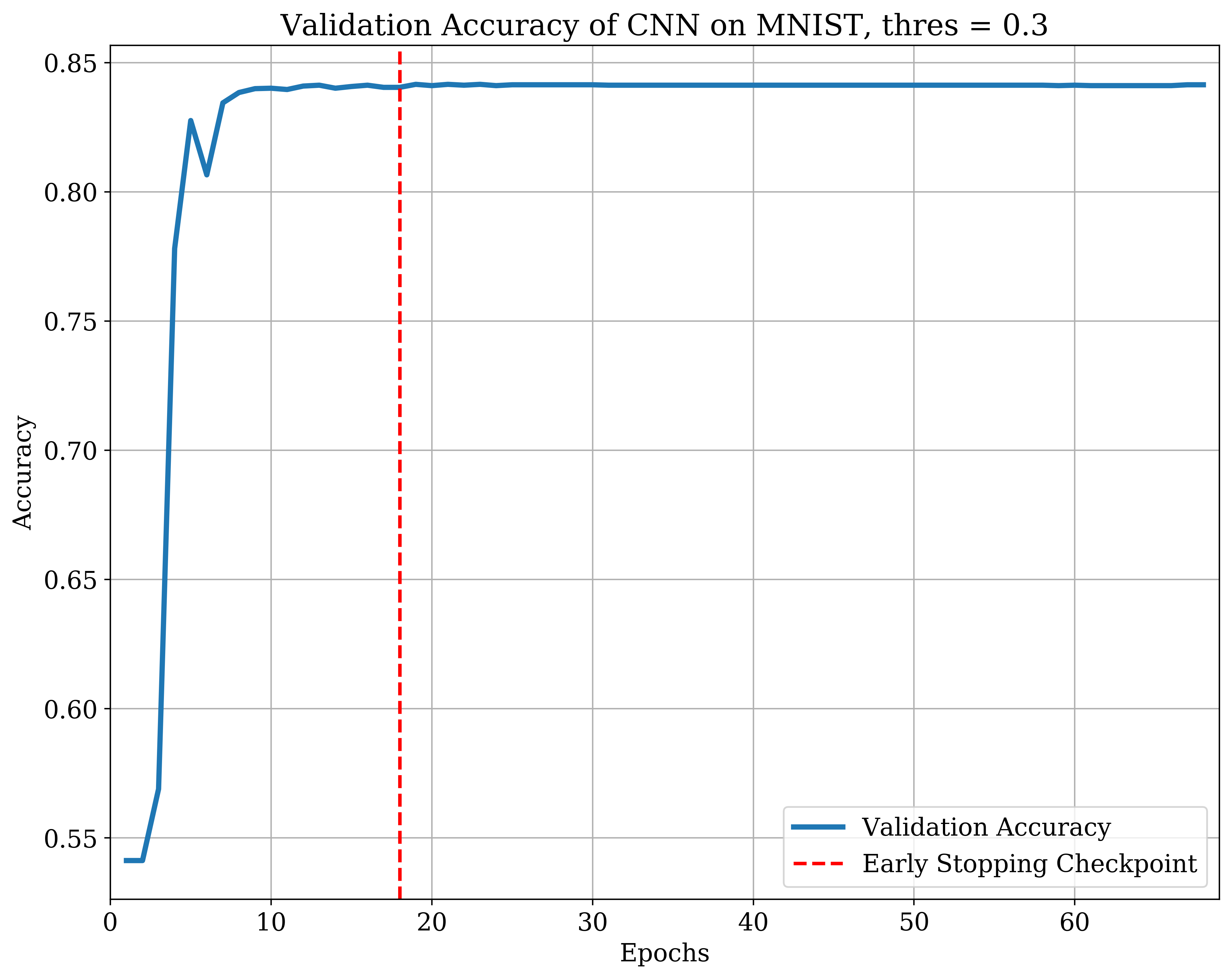}
    &\includegraphics[width=.3\textwidth]{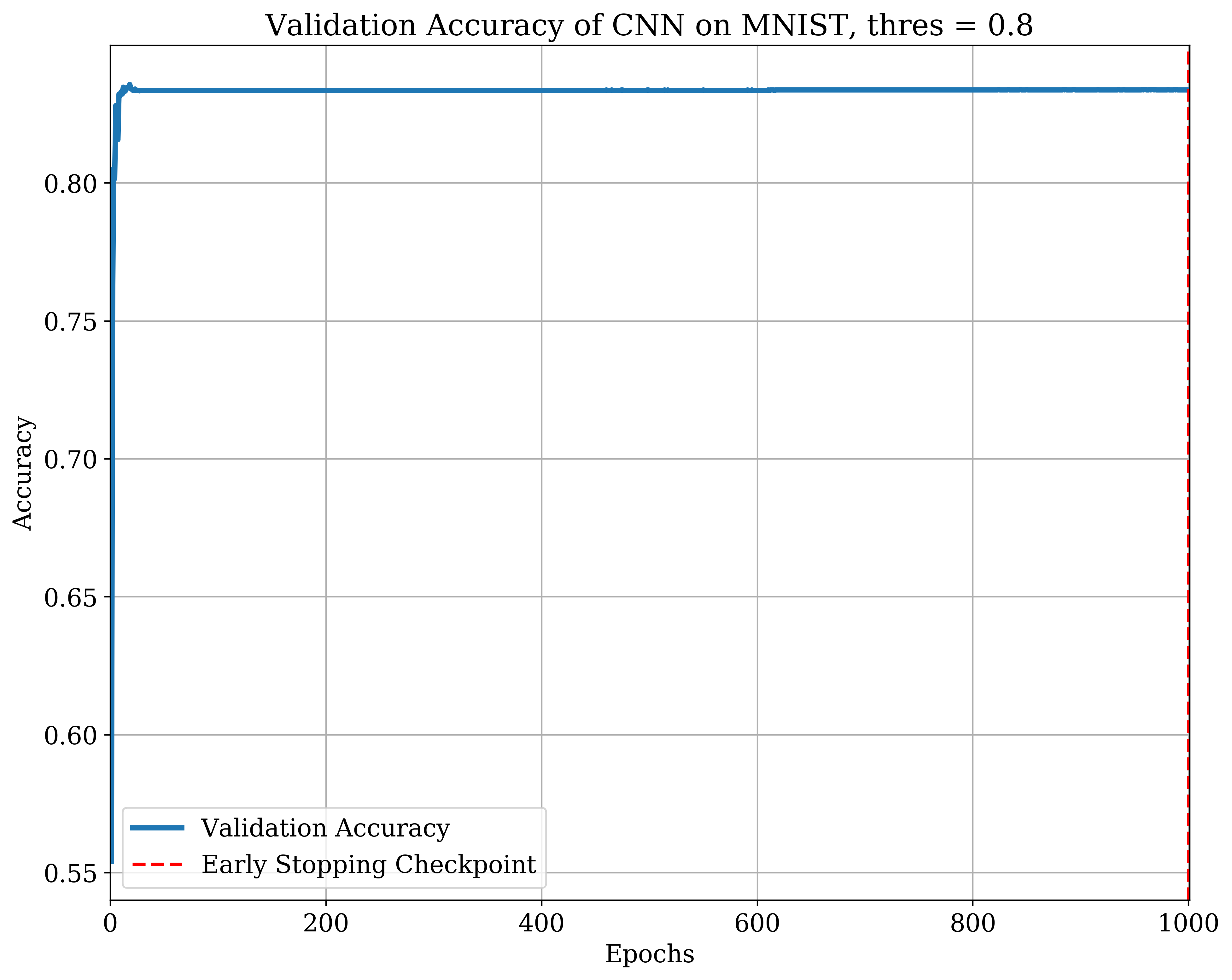}
\end{tabular}
\caption{Top row: training and validation losses of CNN for predicting a bar on MNIST, where threshold is $0.15, 0.3, 0.8$.
Bottom row: test accuracy of CNN for predicting a bar for corresponding thresholds.}\label{fig:MNIST_classification_train_val_loss_accuracy}
\end{figure}

\begin{figure}[h]
\centering
MNIST - Sample distribution in each class\\[1mm]
\begin{tabular}{ccc}
  \includegraphics[width=.3\textwidth]{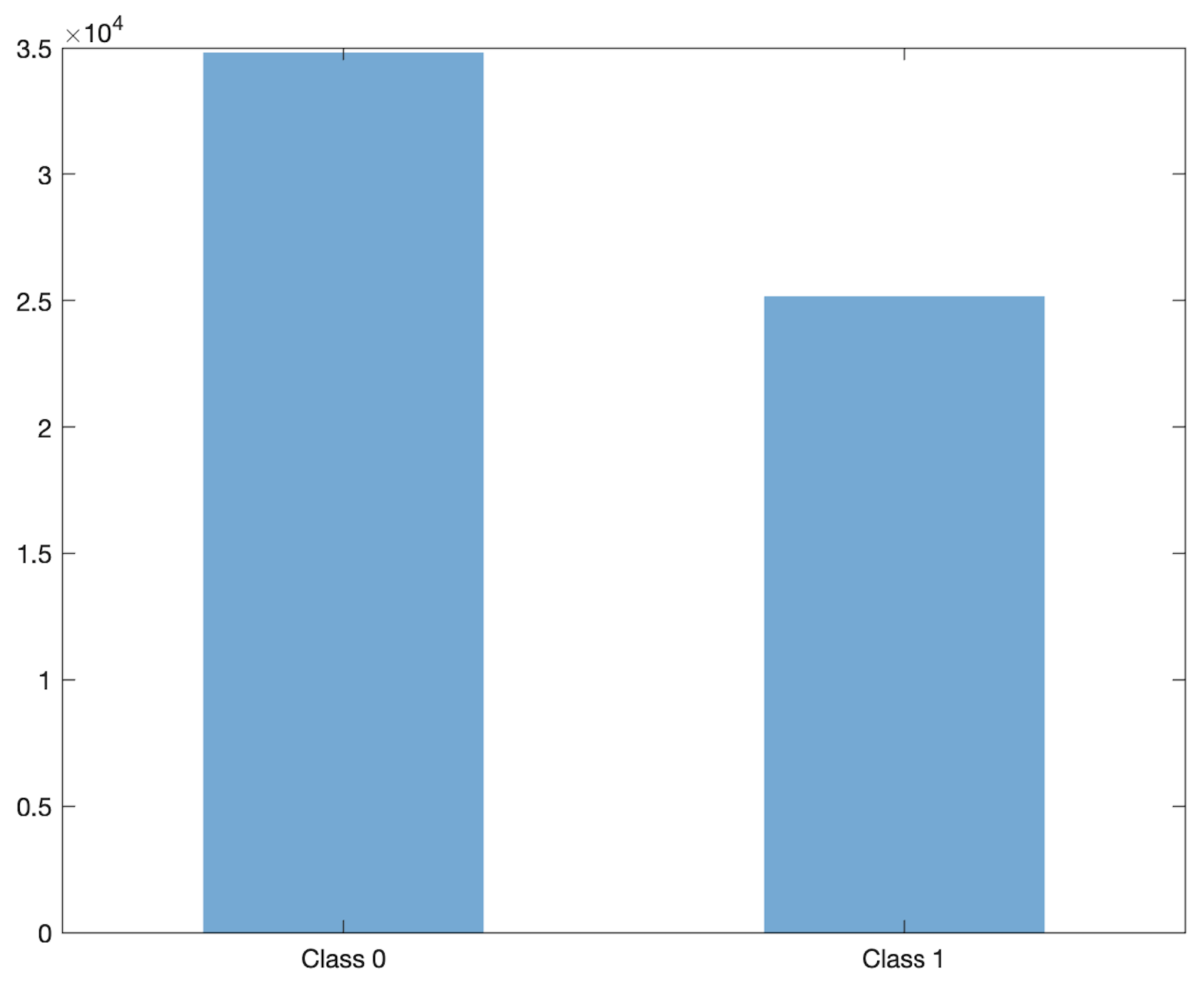} 
    &\includegraphics[width=.3\textwidth]{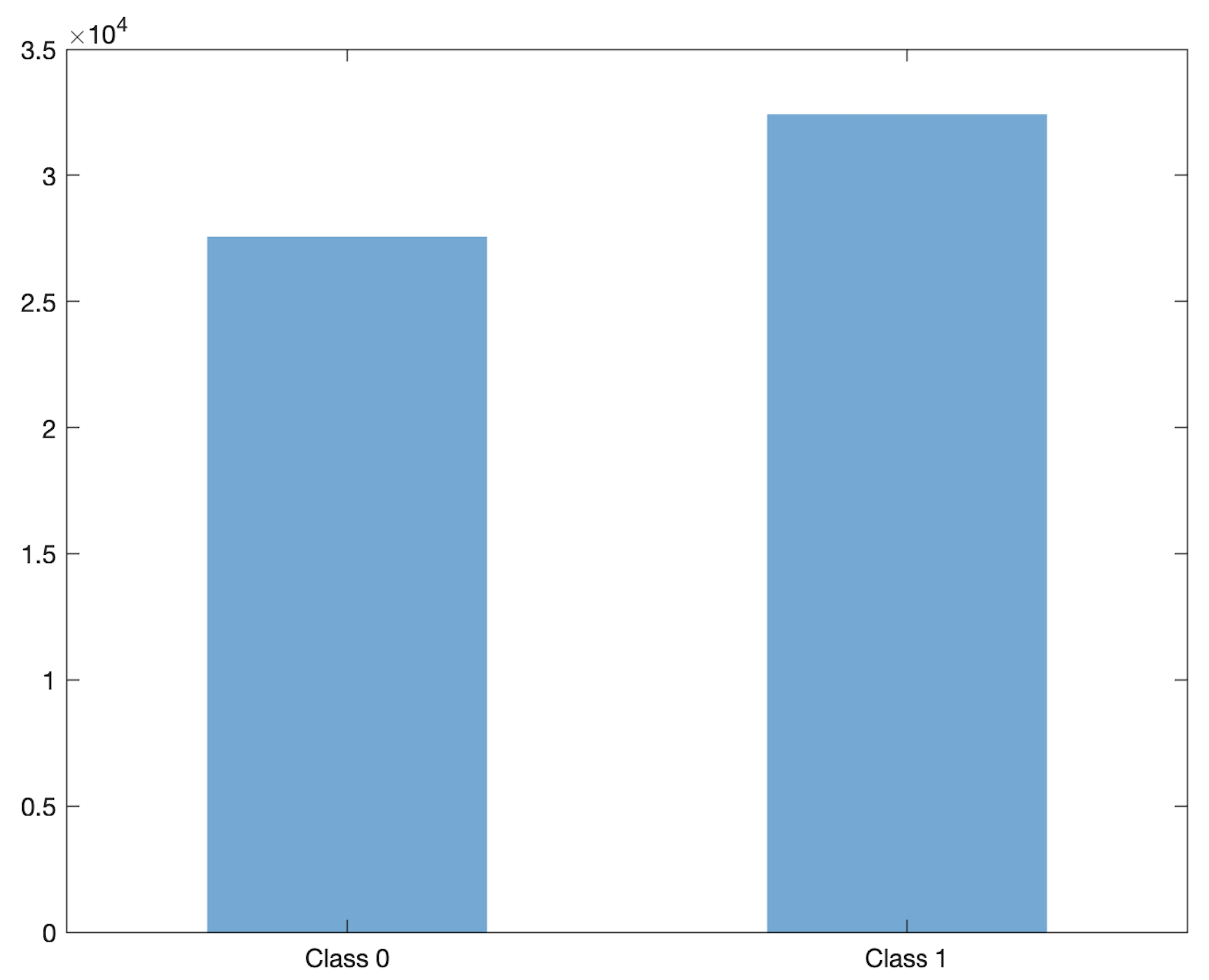}
    &\includegraphics[width=.3\textwidth]{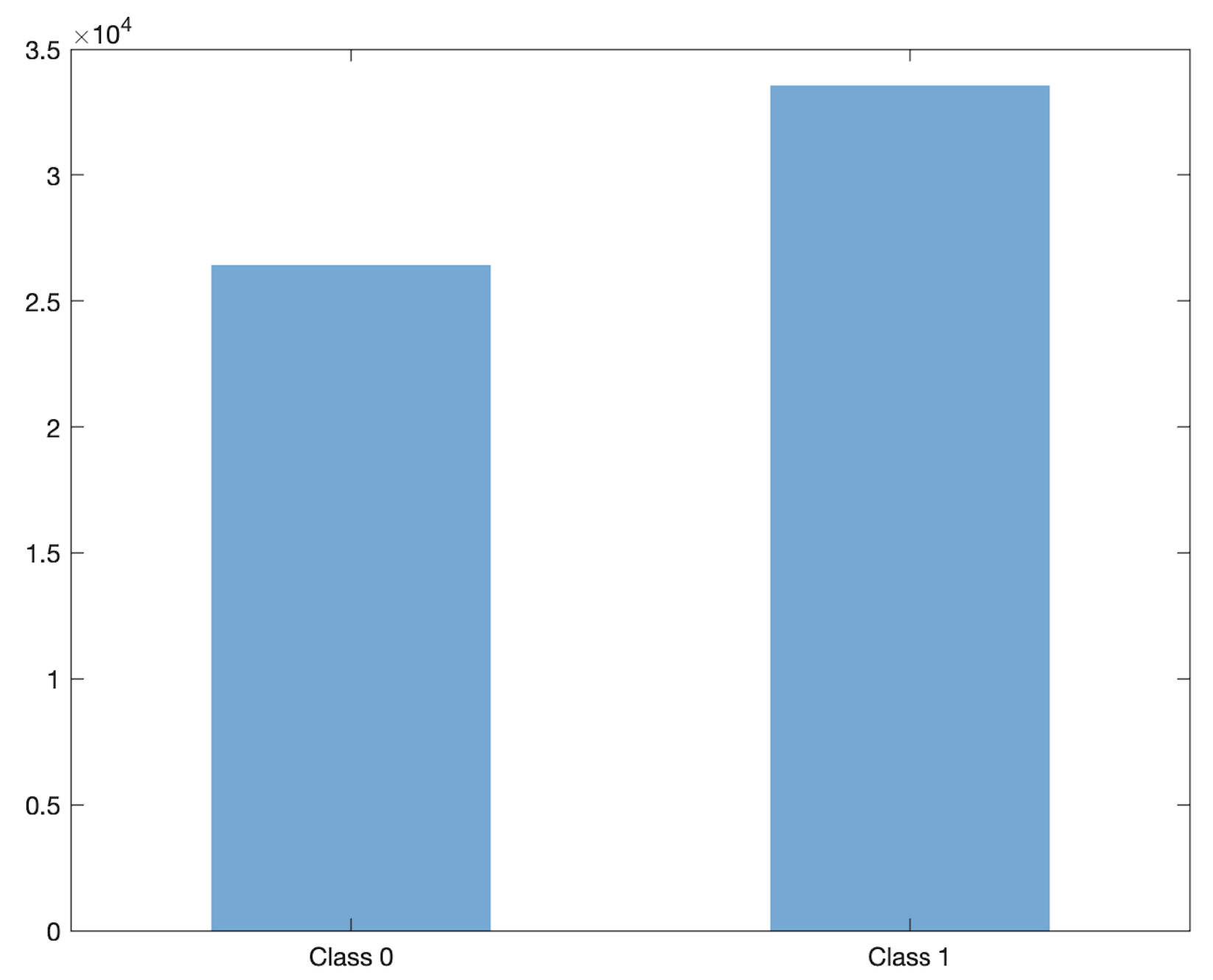}
    \vspace{4mm}
\end{tabular}
MNIST - Samples in each PH class\\[1mm]
\begin{tabular}{cccccccc}
Class 0 & Class 1 && Class 0 & Class 1 && Class 0 & Class 1\\
  \includegraphics[width=.12\textwidth]{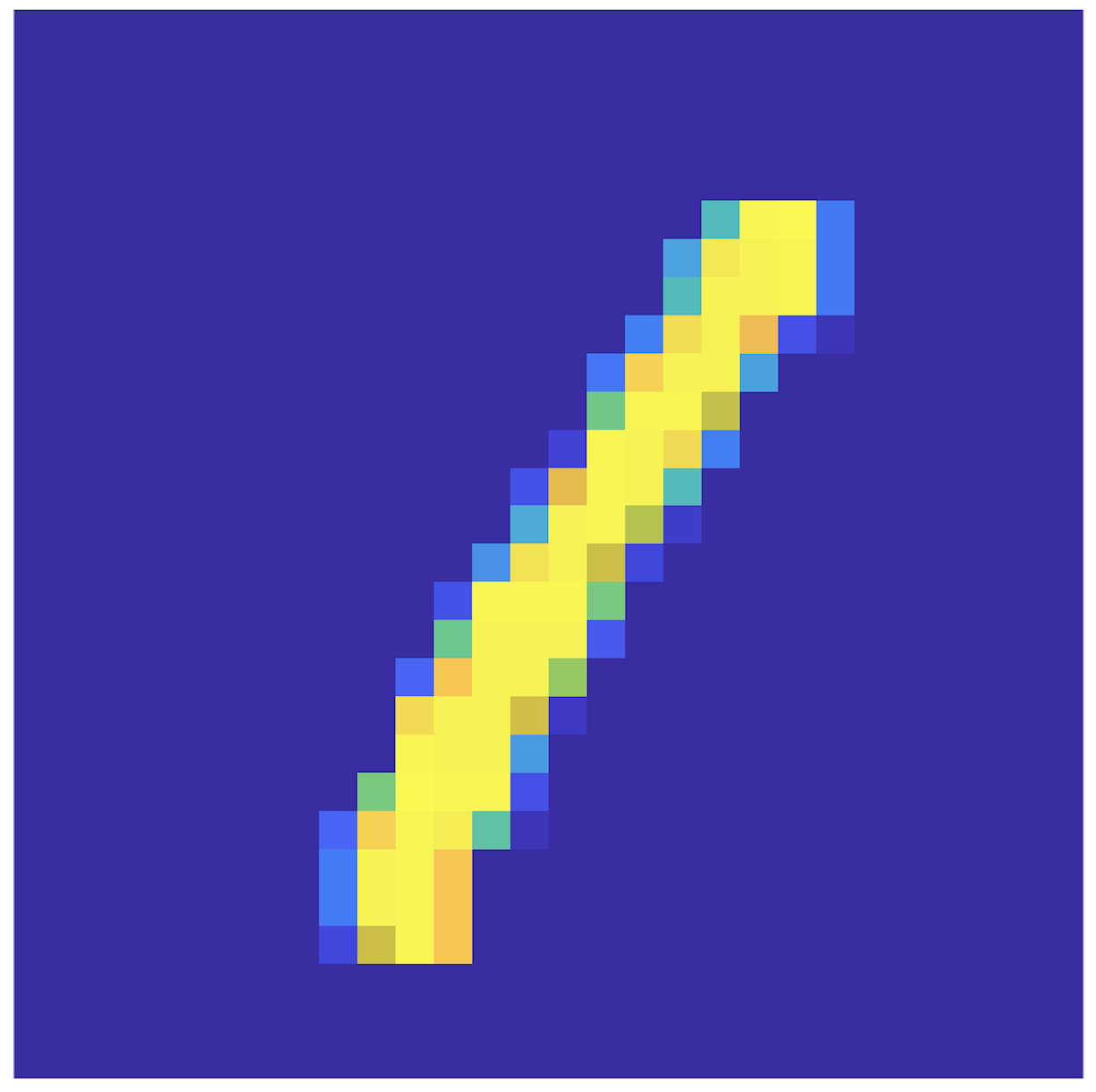}
  &\includegraphics[width=.12\textwidth]{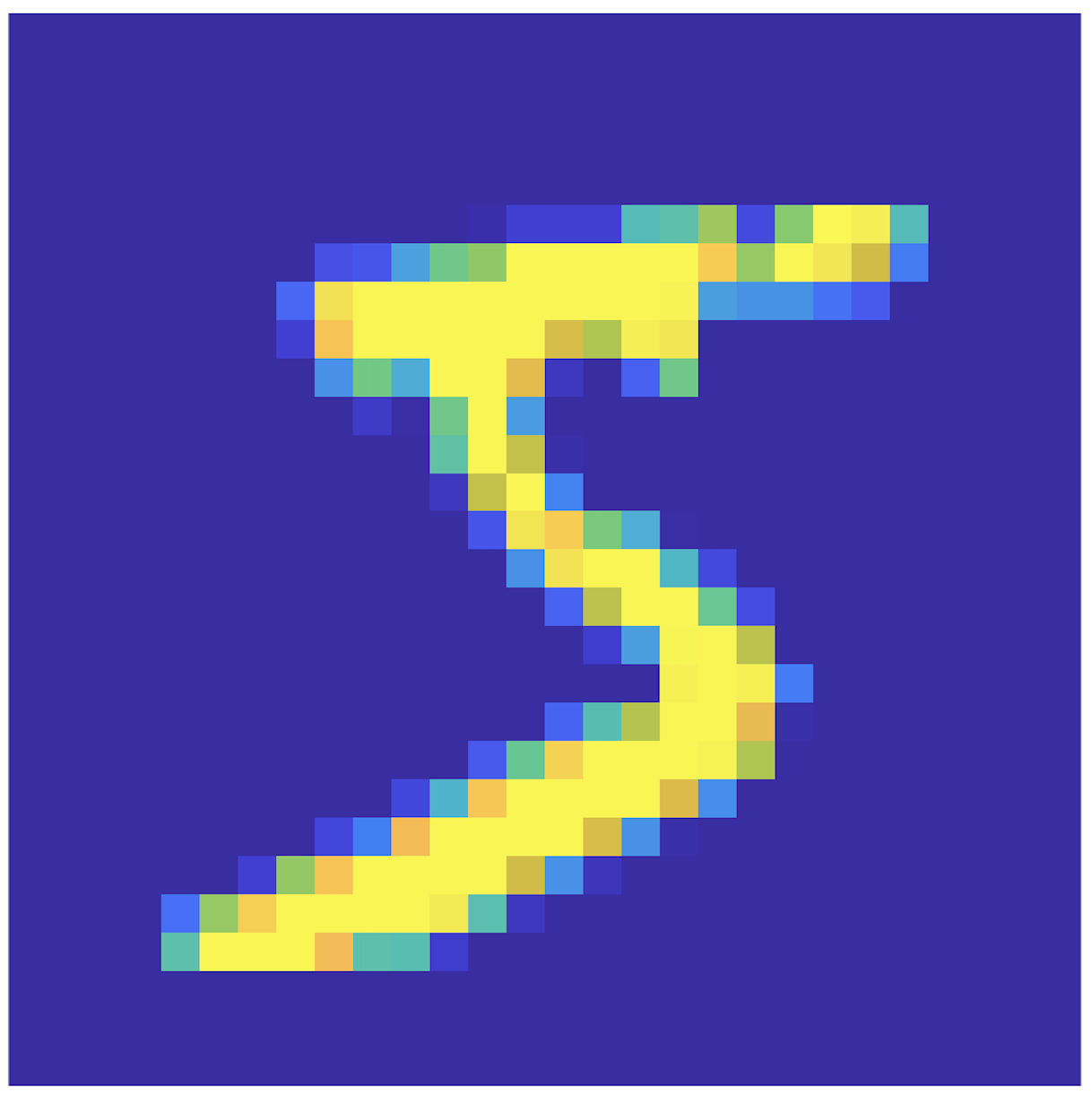}
    &&
    \includegraphics[width=.12\textwidth]{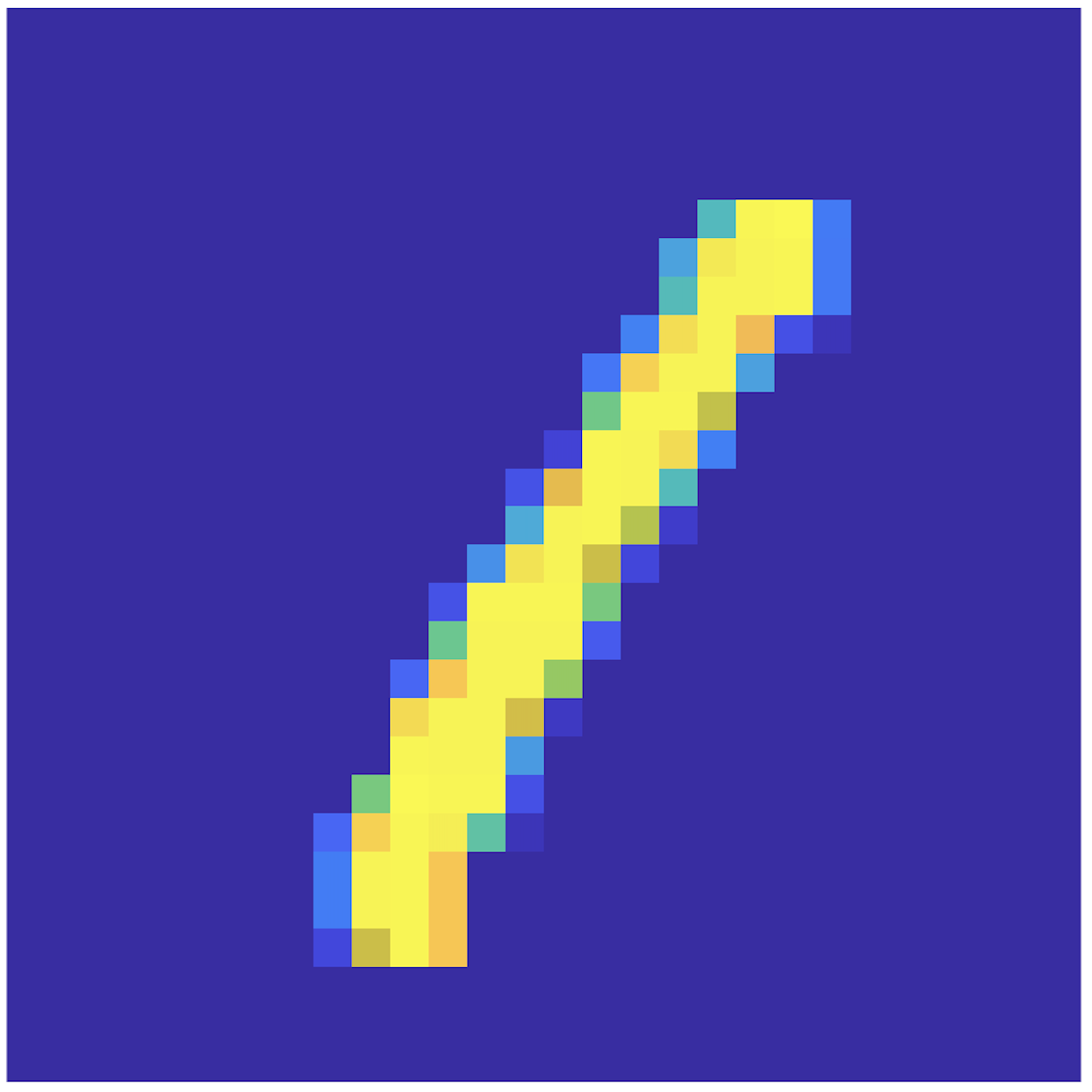}
    &\includegraphics[width=.12\textwidth]{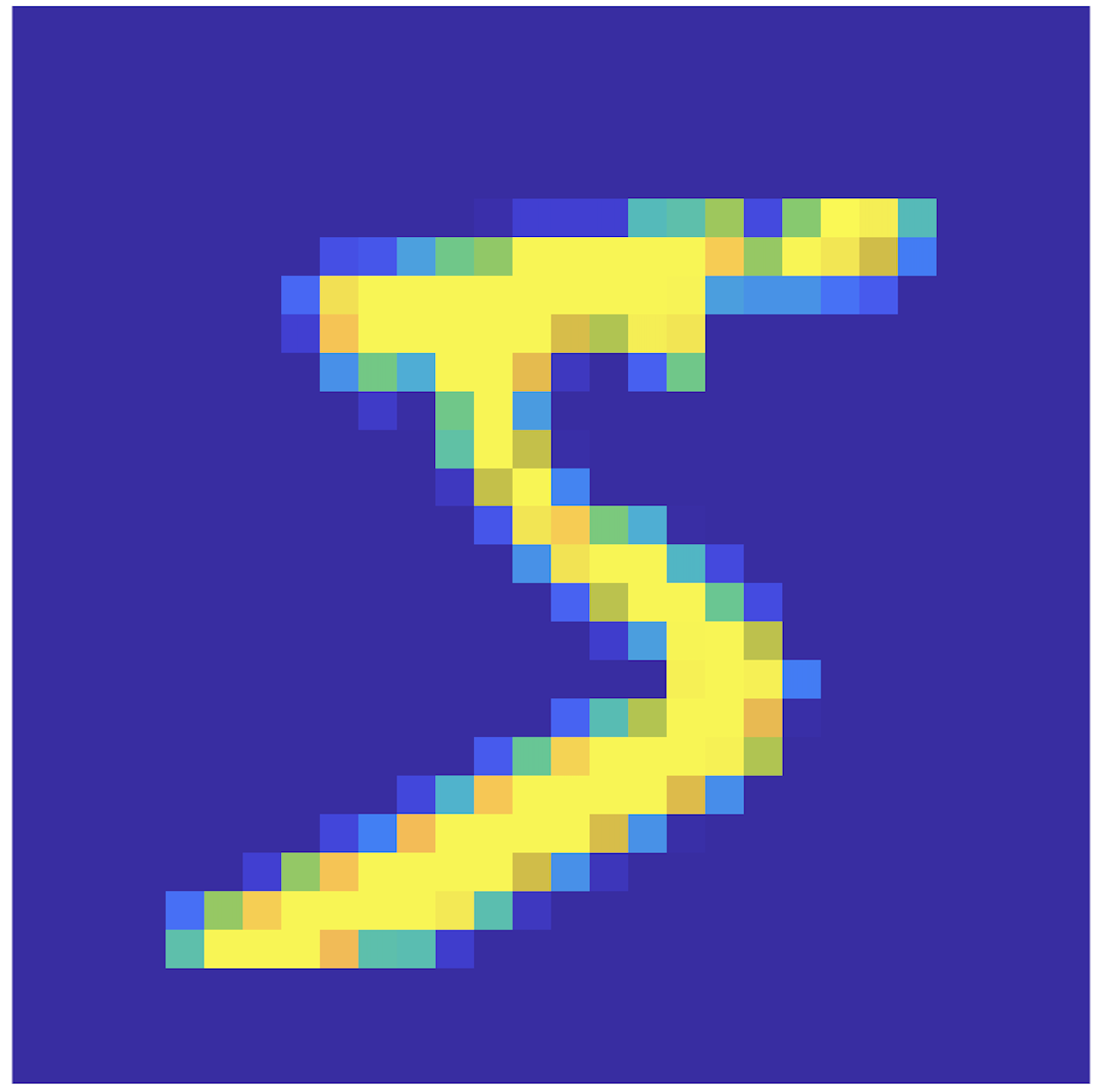}
    &&\includegraphics[width=.12\textwidth]{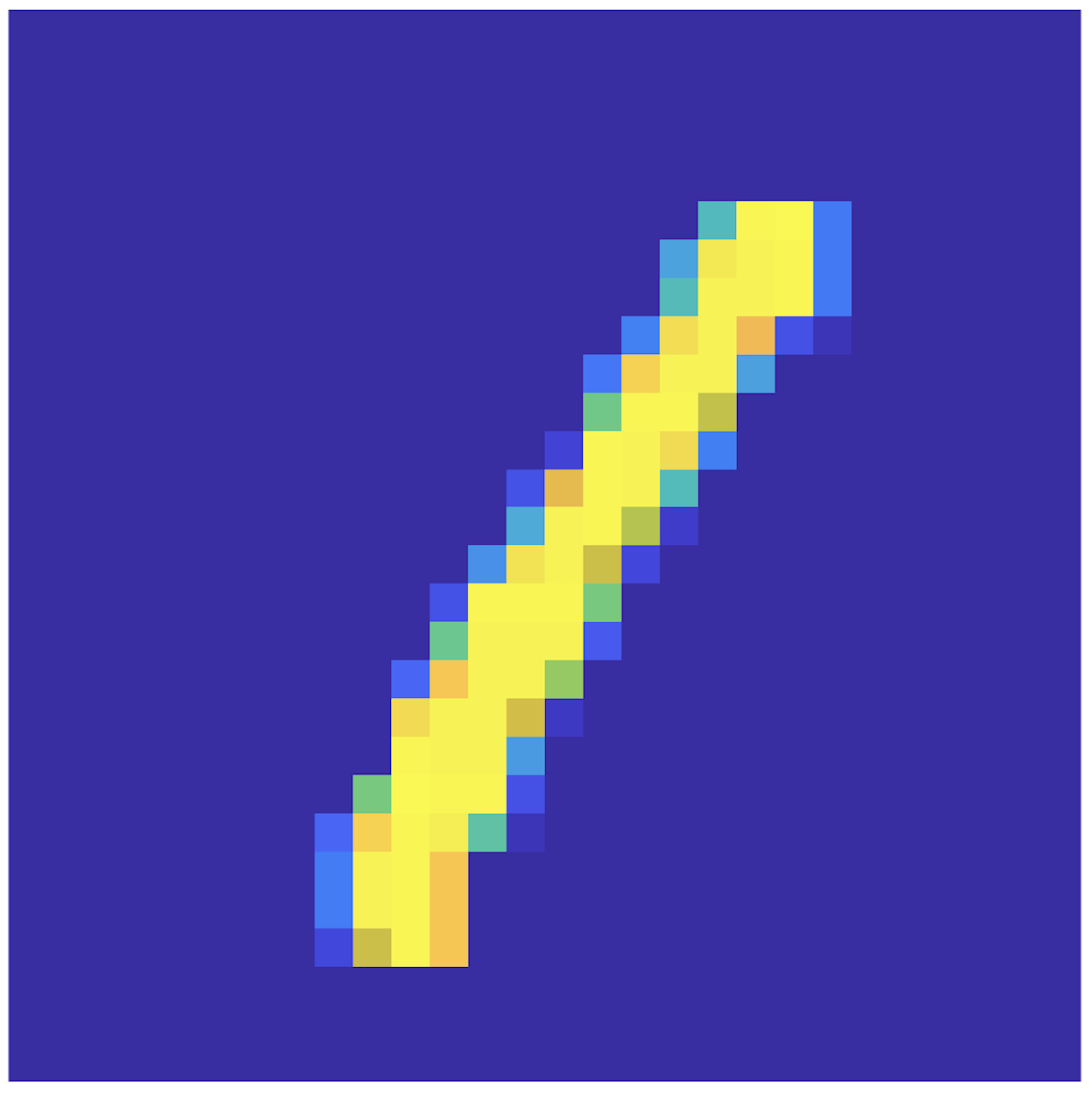}
    &\includegraphics[width=.12\textwidth]{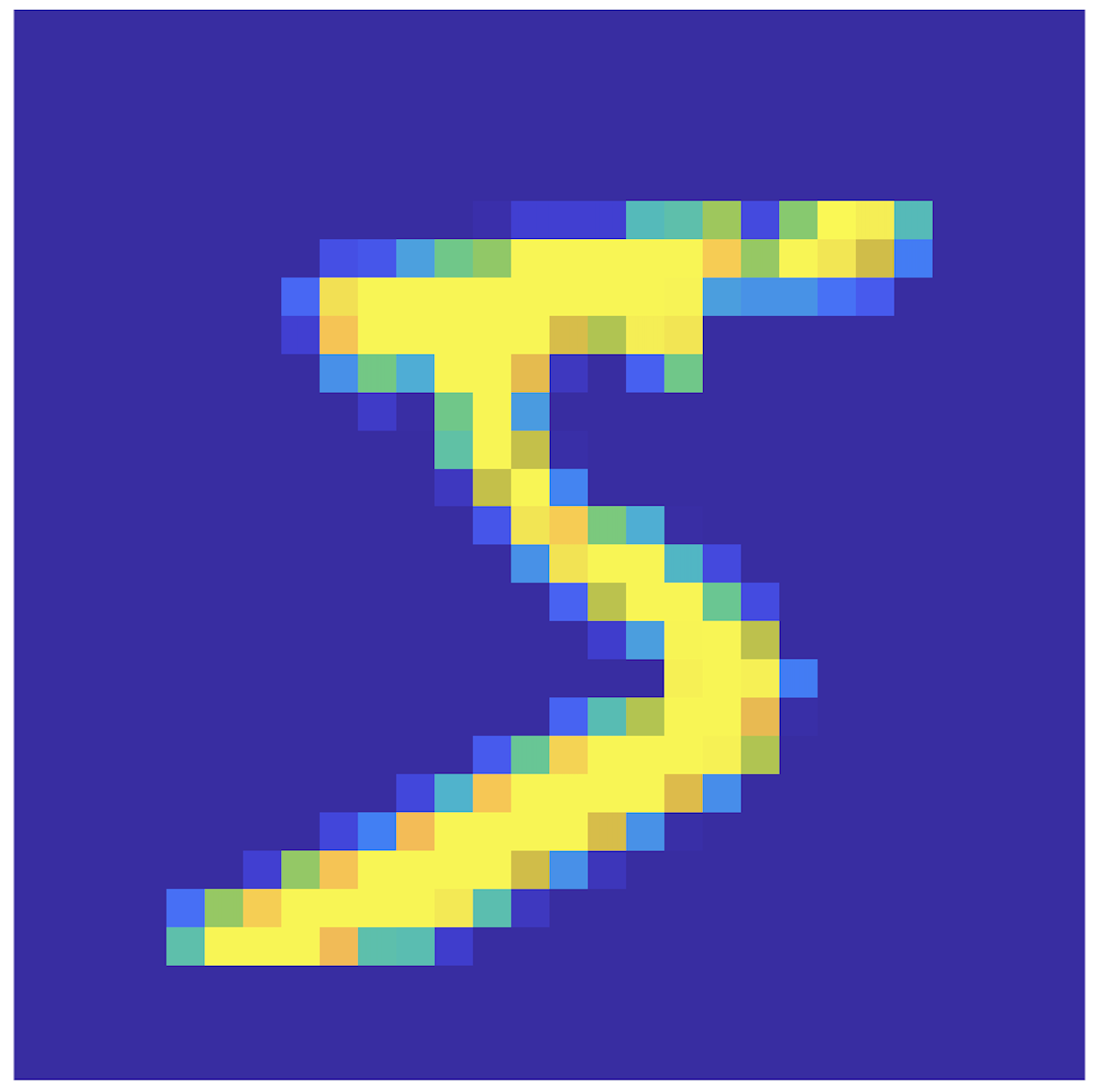}\\
  \includegraphics[width=.12\textwidth]{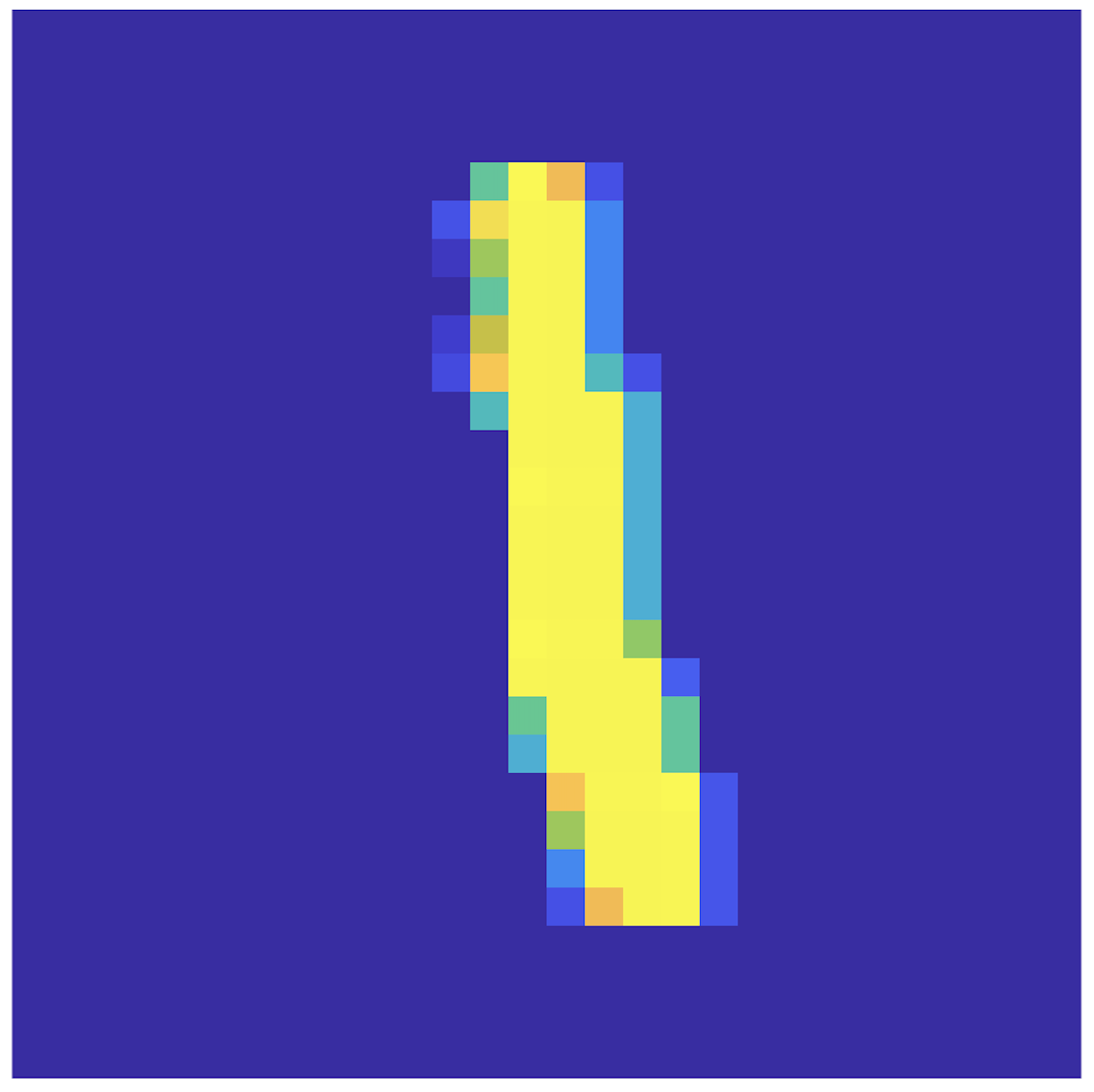}
  &\includegraphics[width=.12\textwidth]{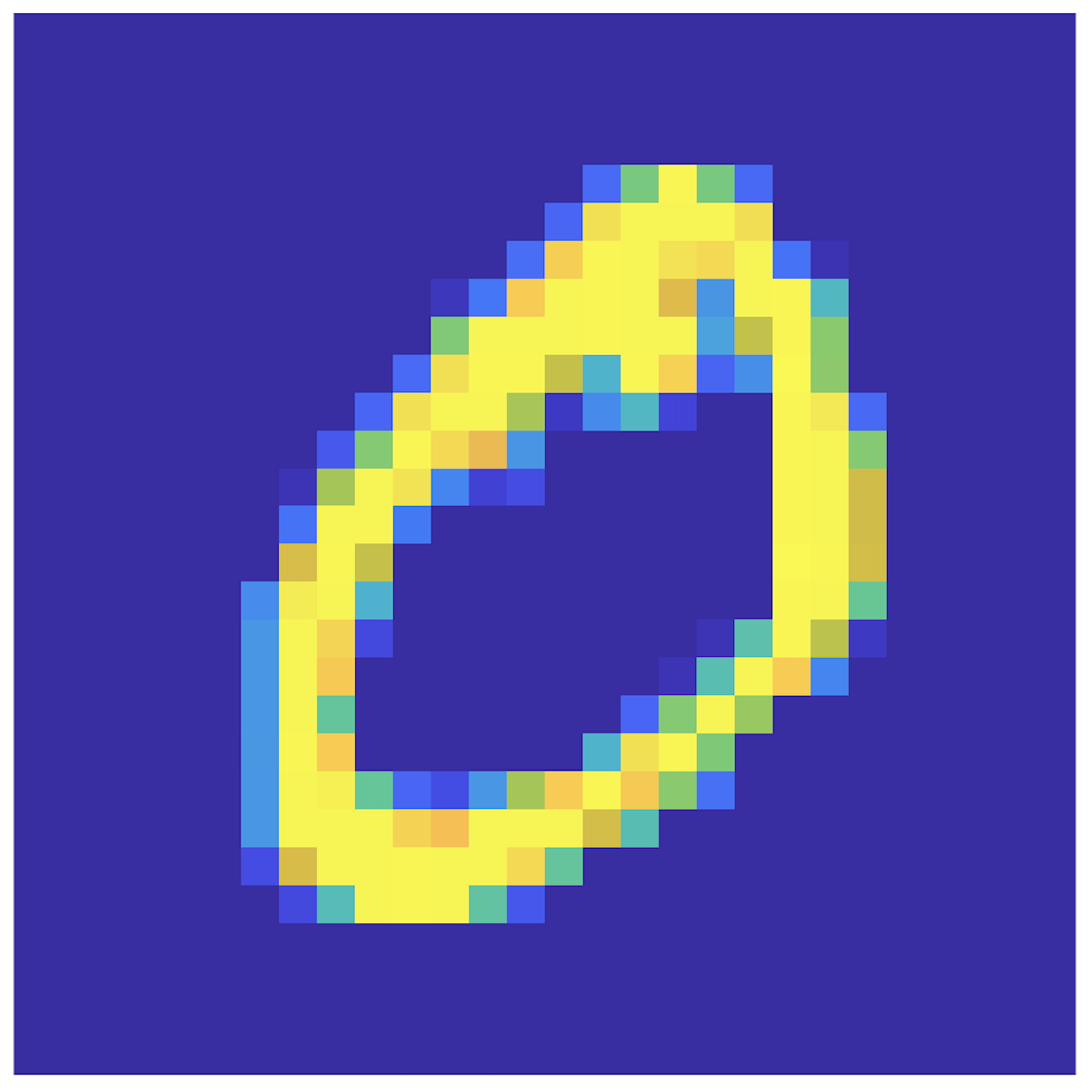}
    &&
    \includegraphics[width=.12\textwidth]{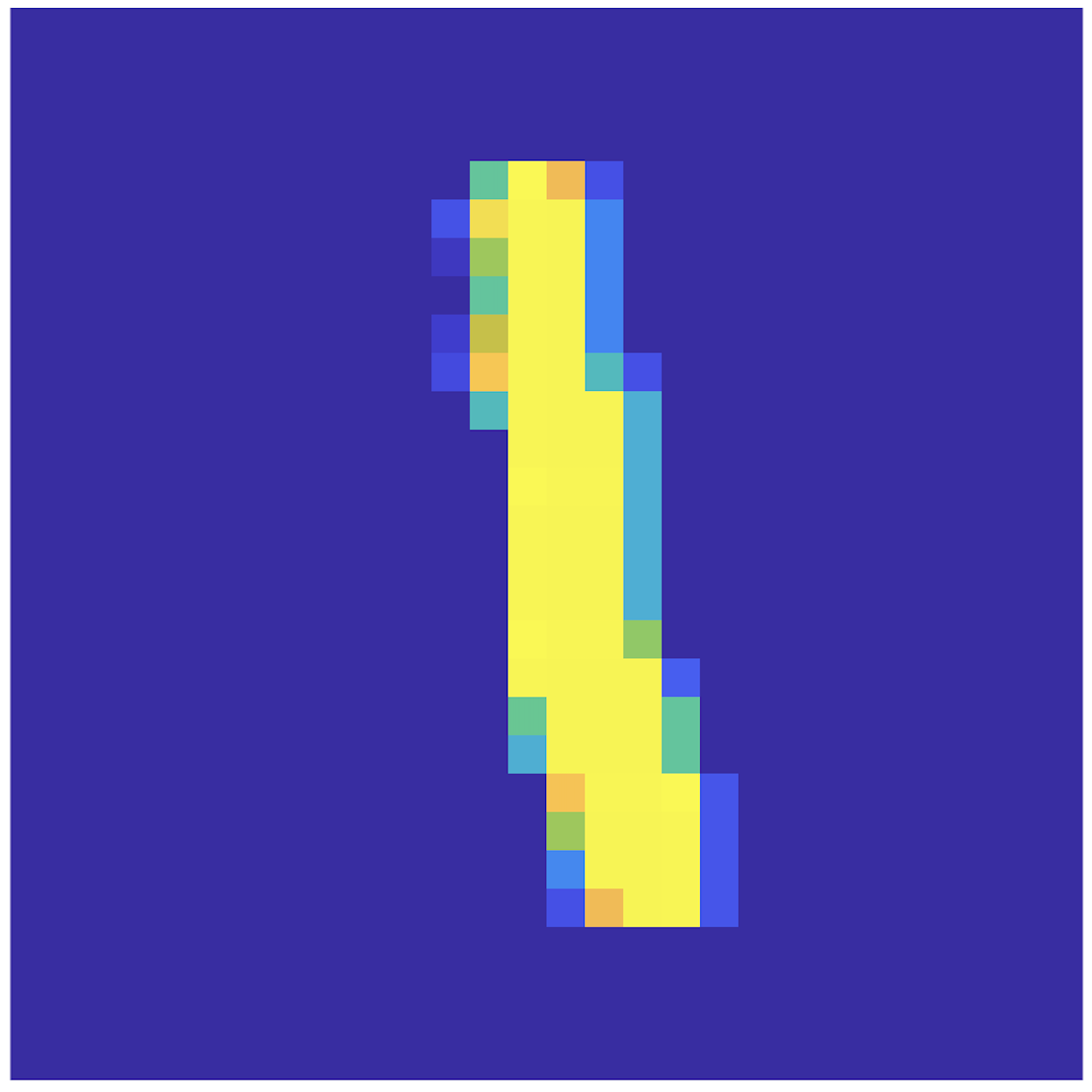}
    &\includegraphics[width=.12\textwidth]{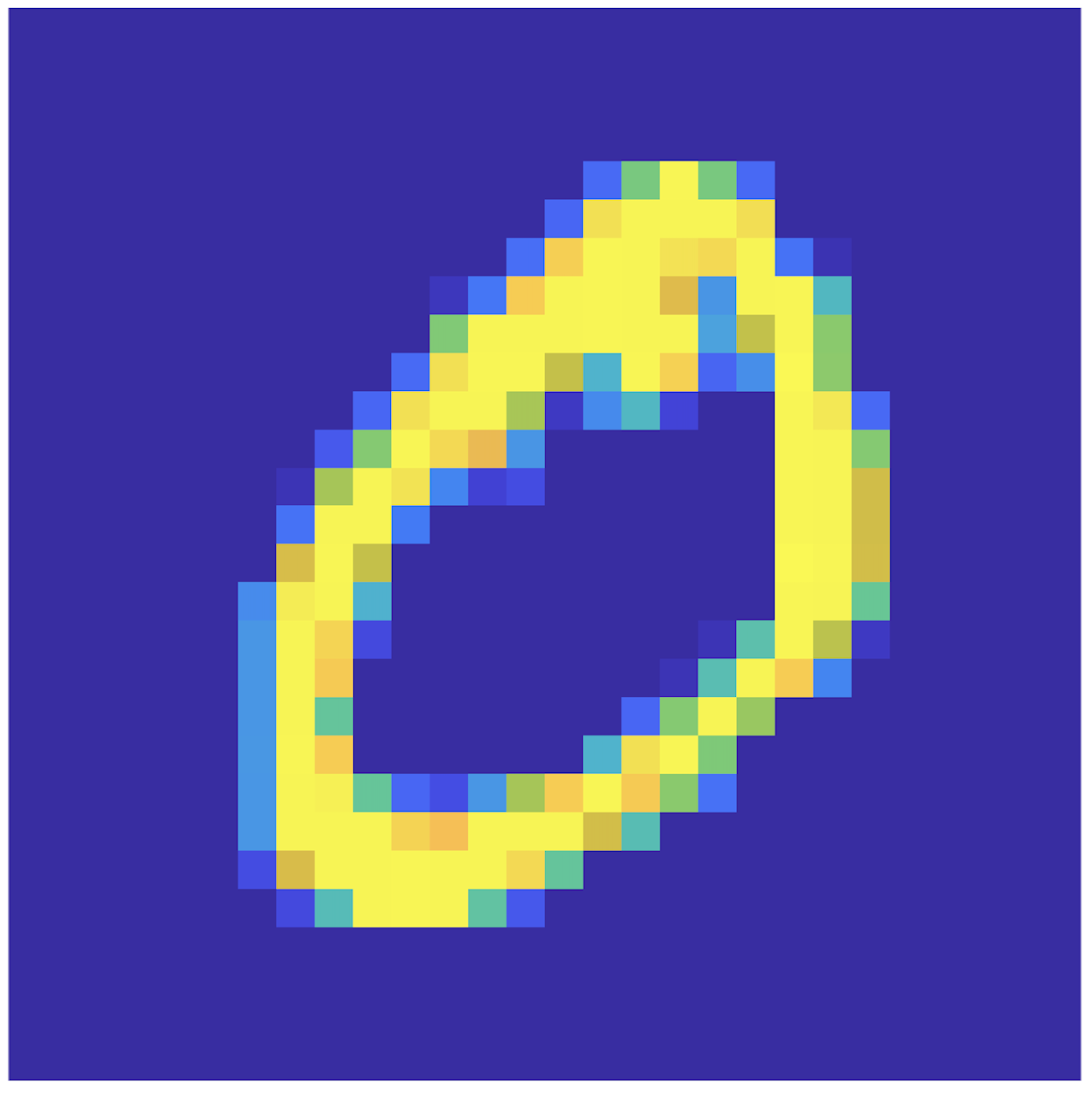}
    &&\includegraphics[width=.12\textwidth]{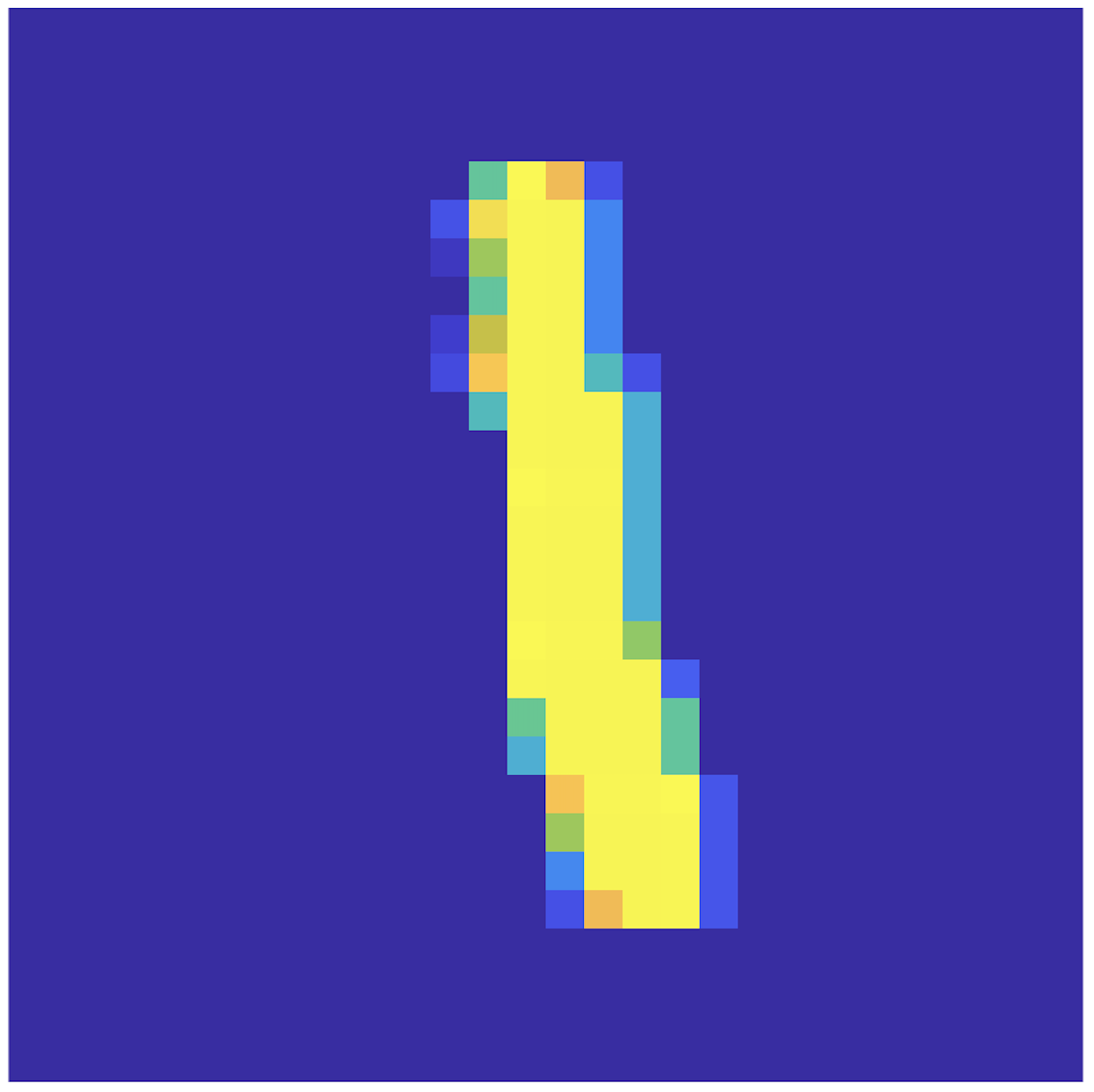}
    &\includegraphics[width=.12\textwidth]{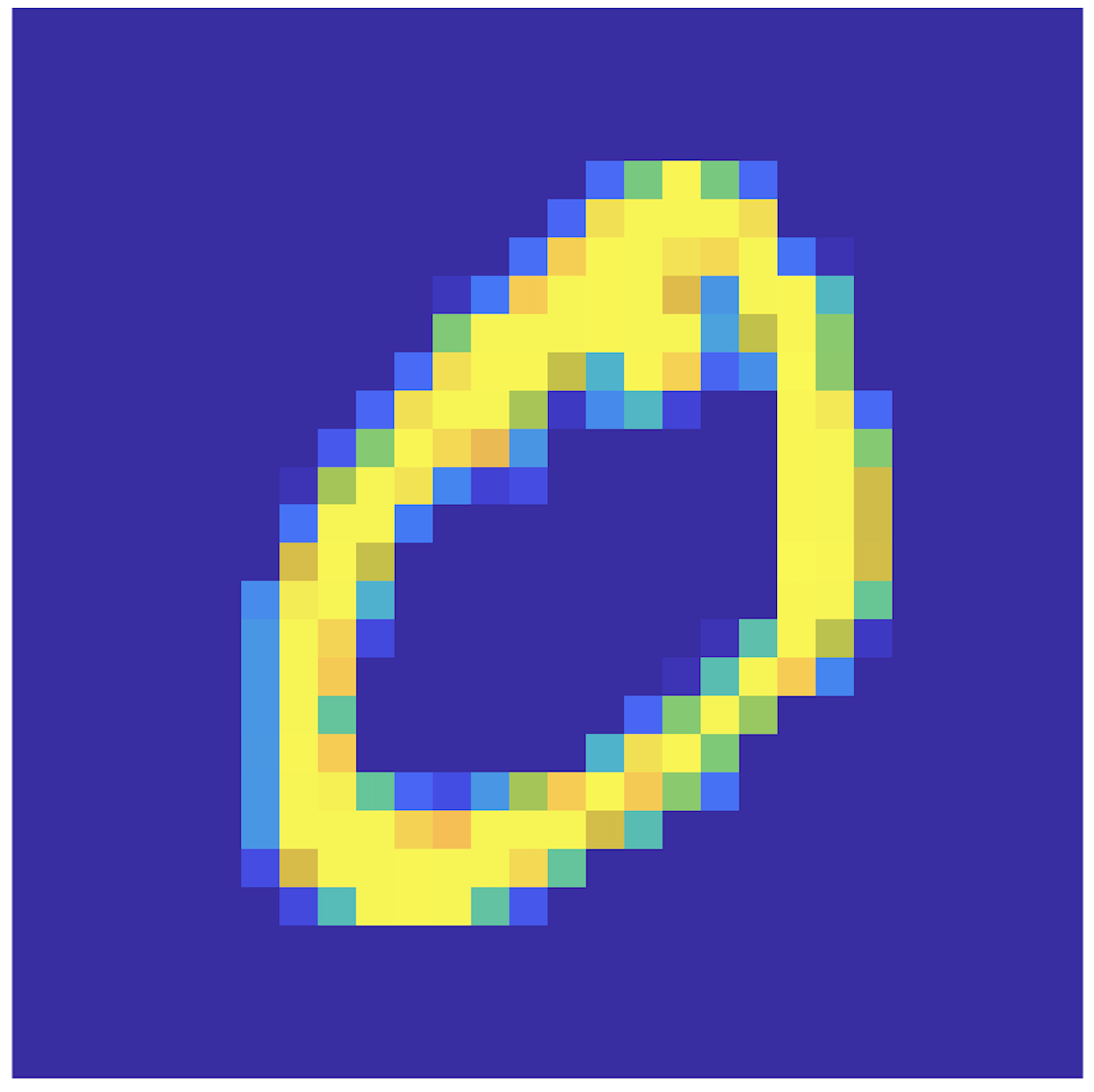}\\
  \includegraphics[width=.12\textwidth]{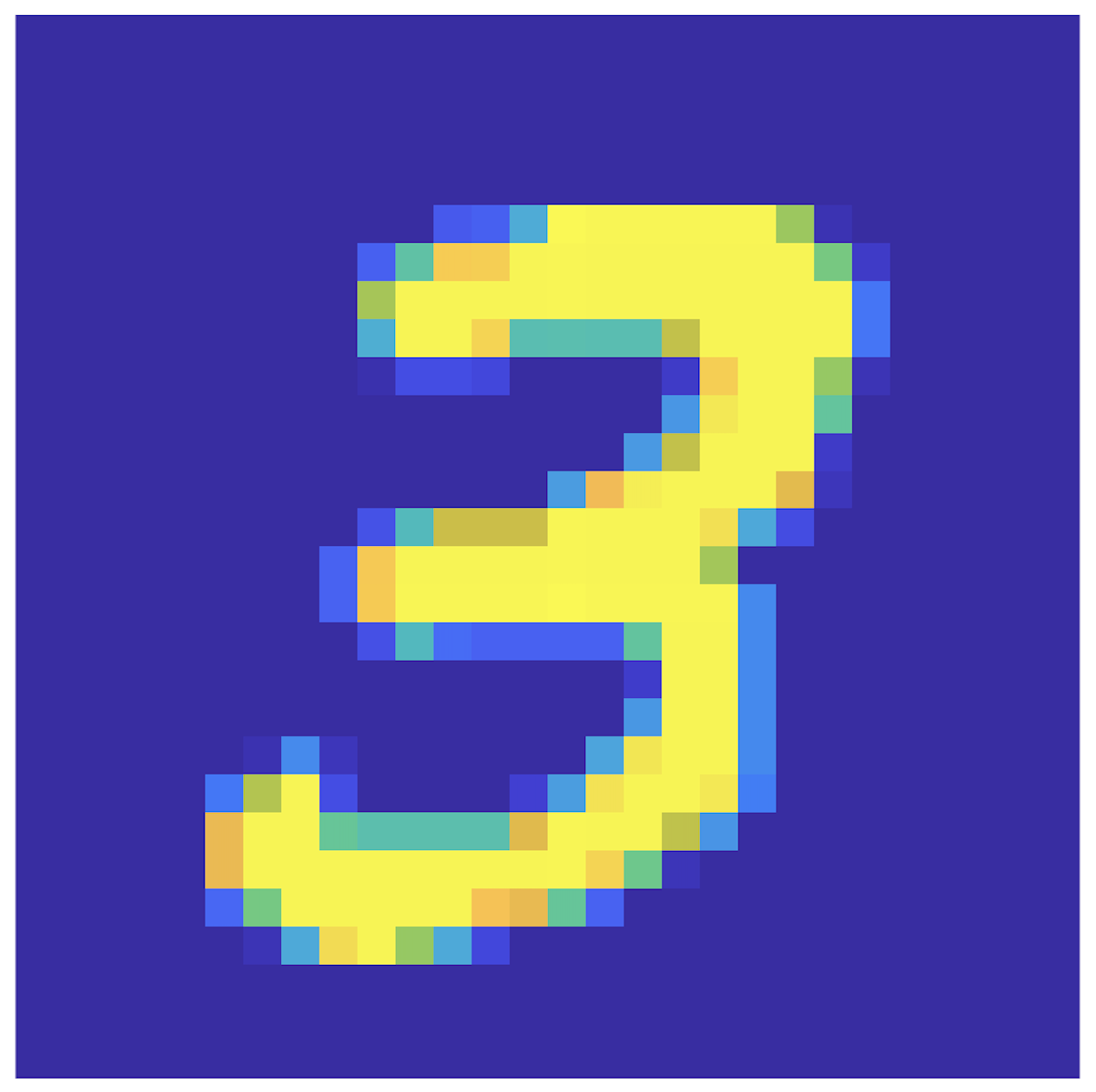}
  &\includegraphics[width=.12\textwidth]{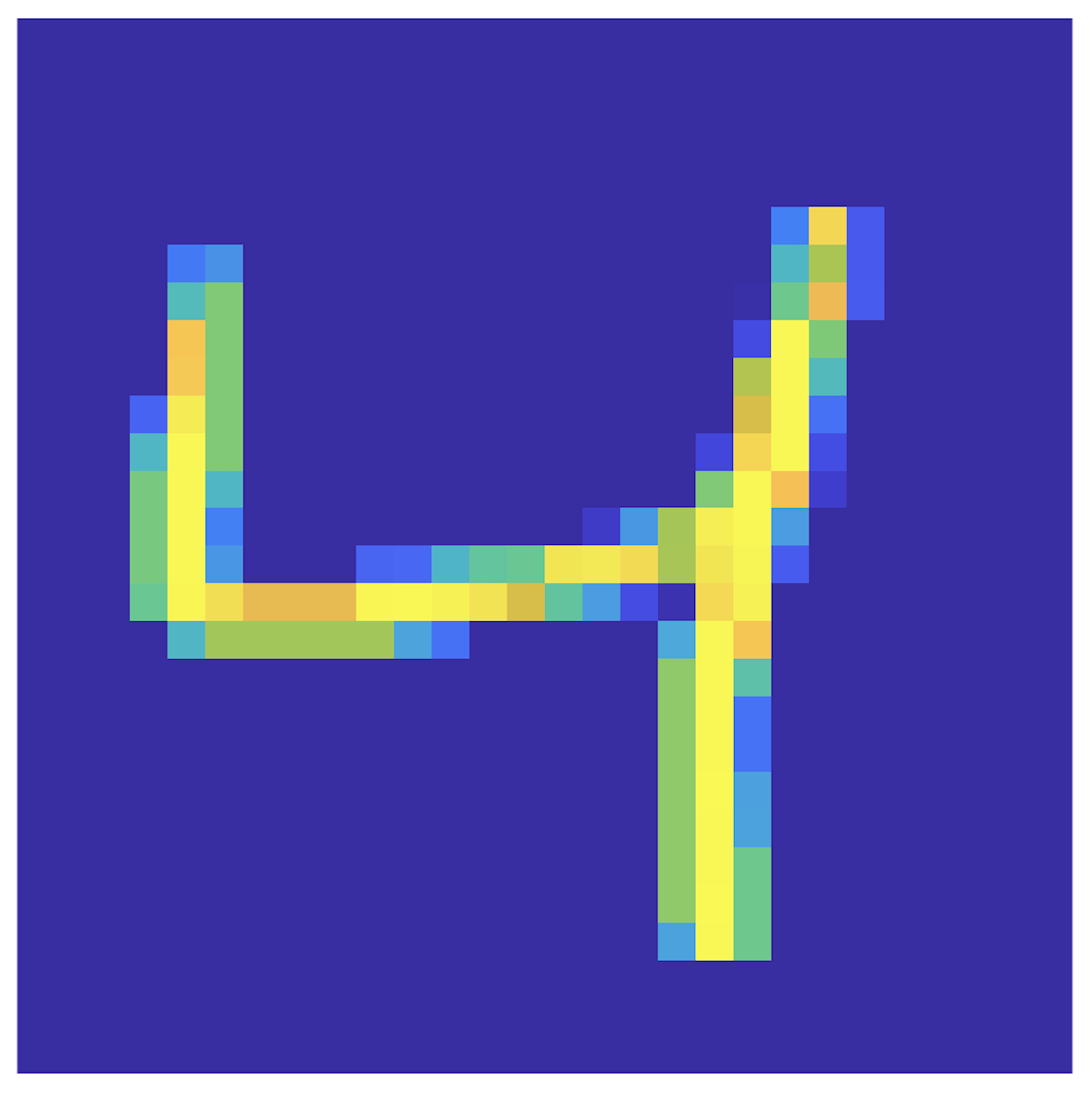}
    &&
    \includegraphics[width=.12\textwidth]{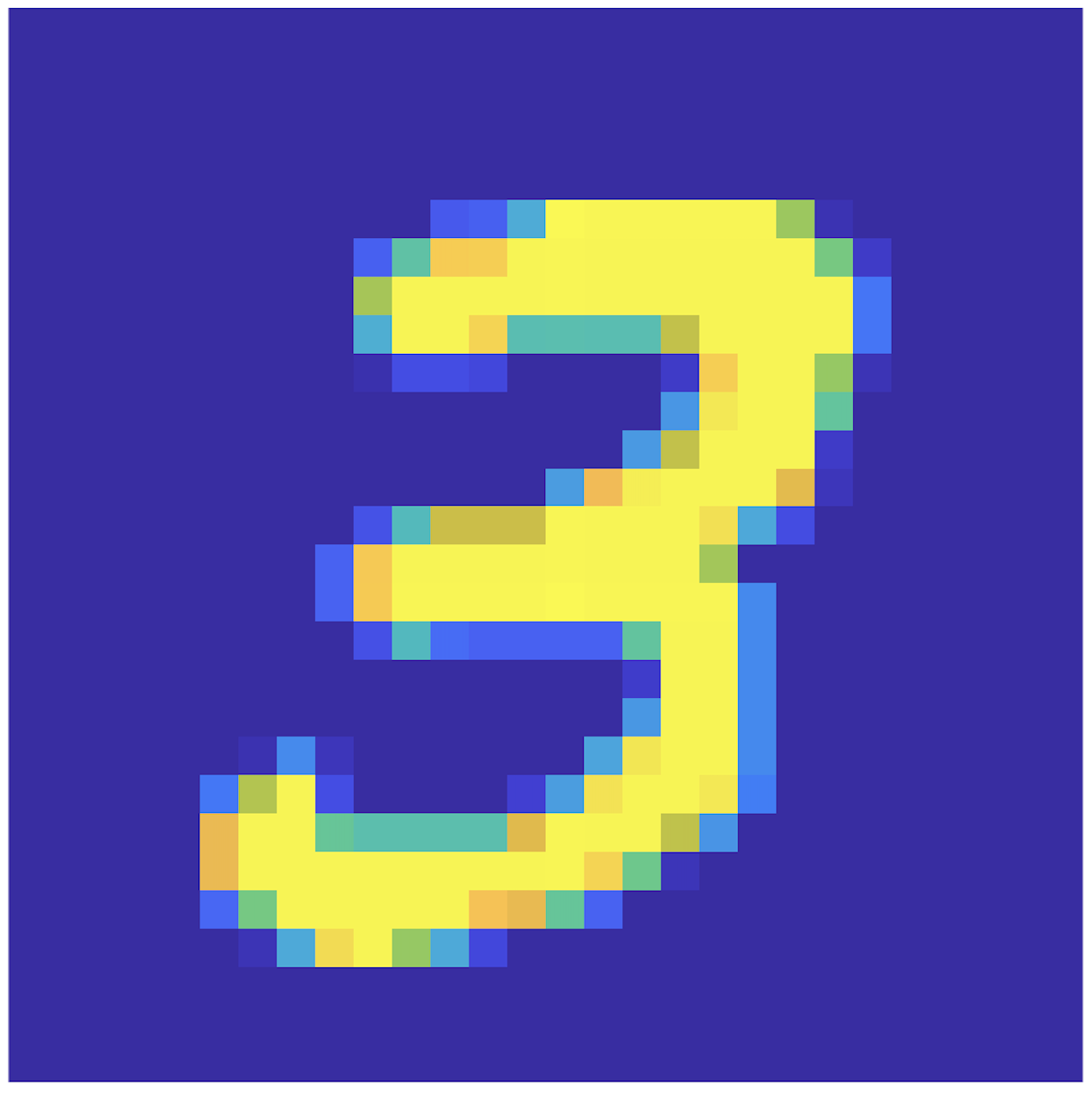}
    &\includegraphics[width=.12\textwidth]{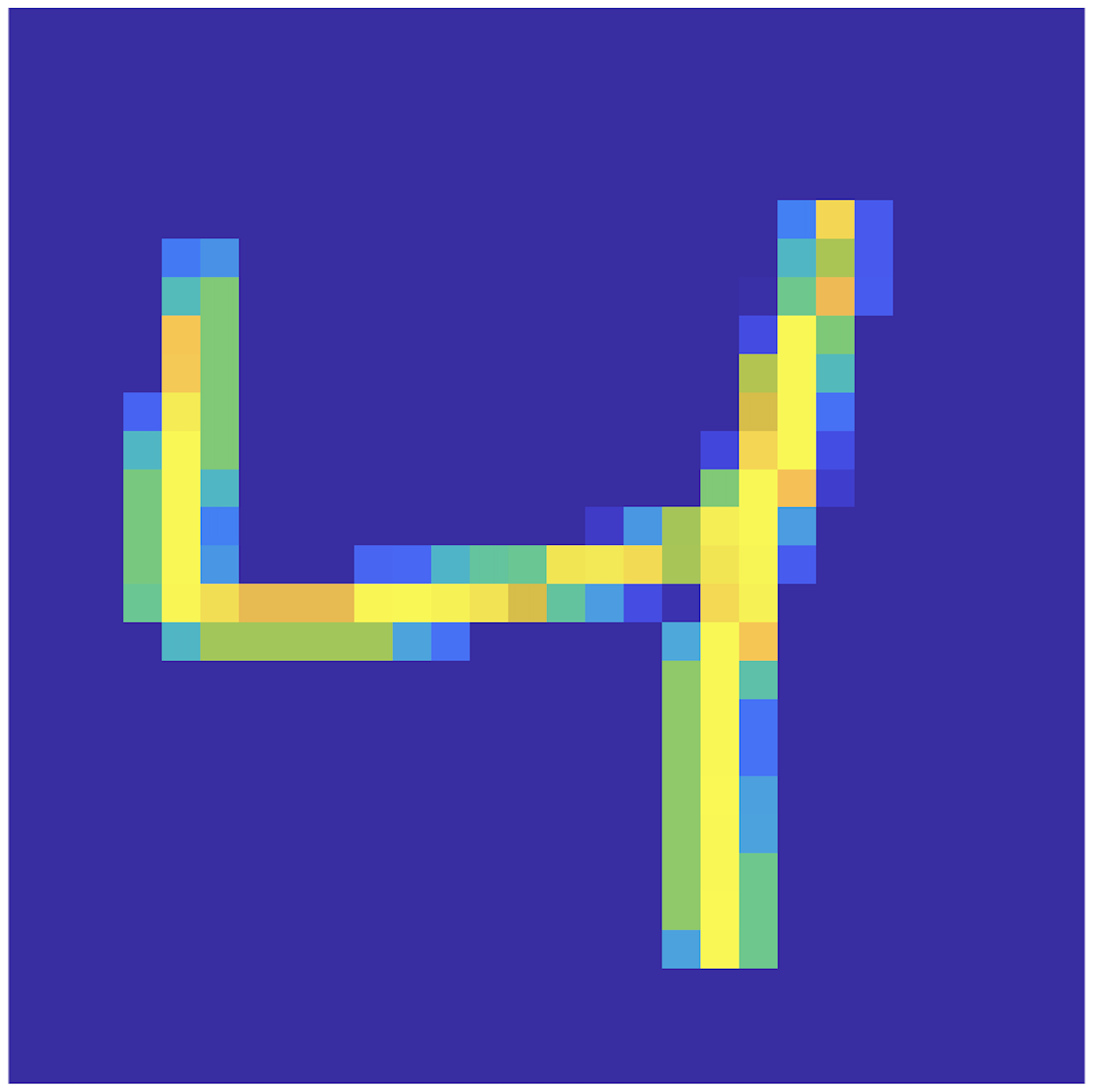}
    &&\includegraphics[width=.12\textwidth]{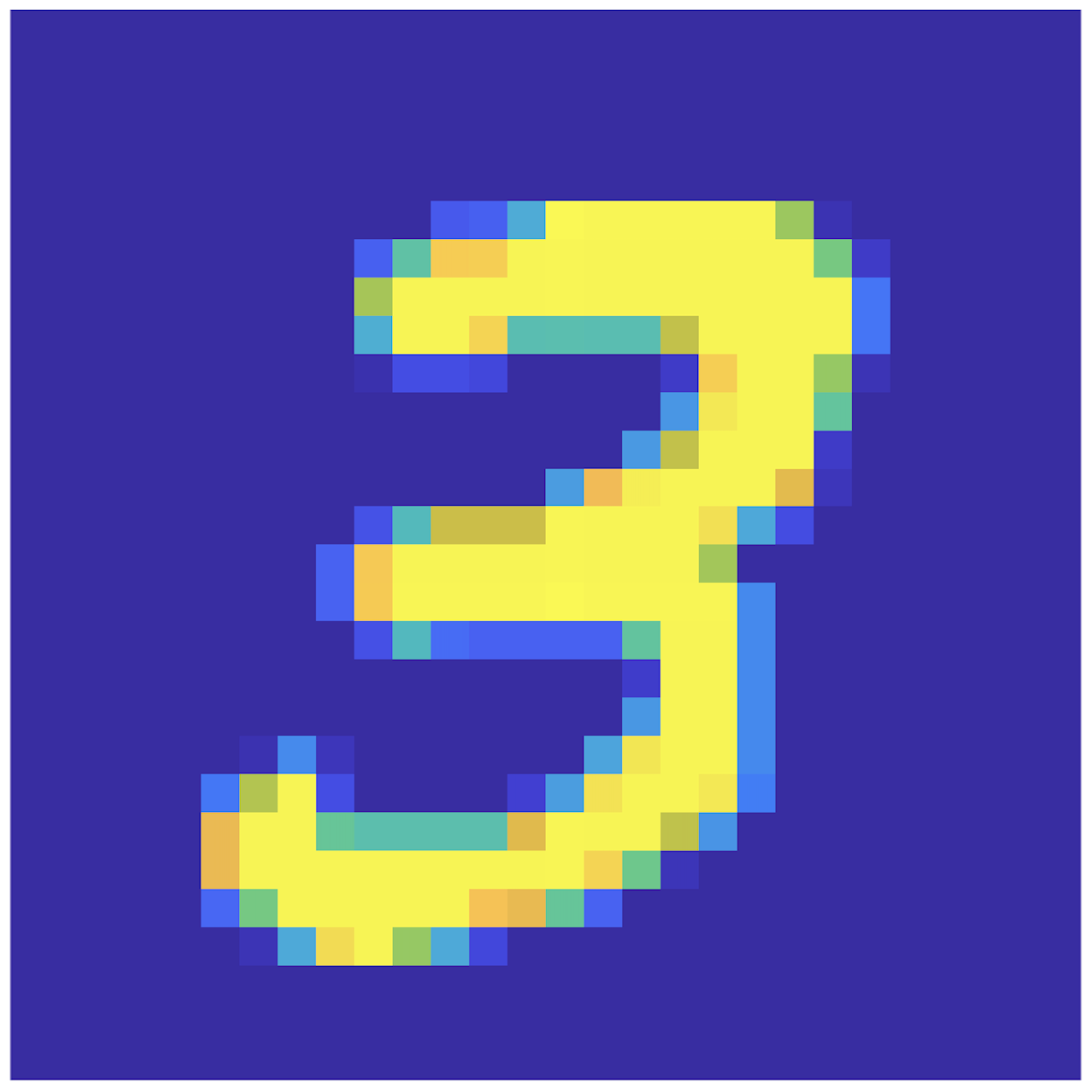}
    &\includegraphics[width=.12\textwidth]{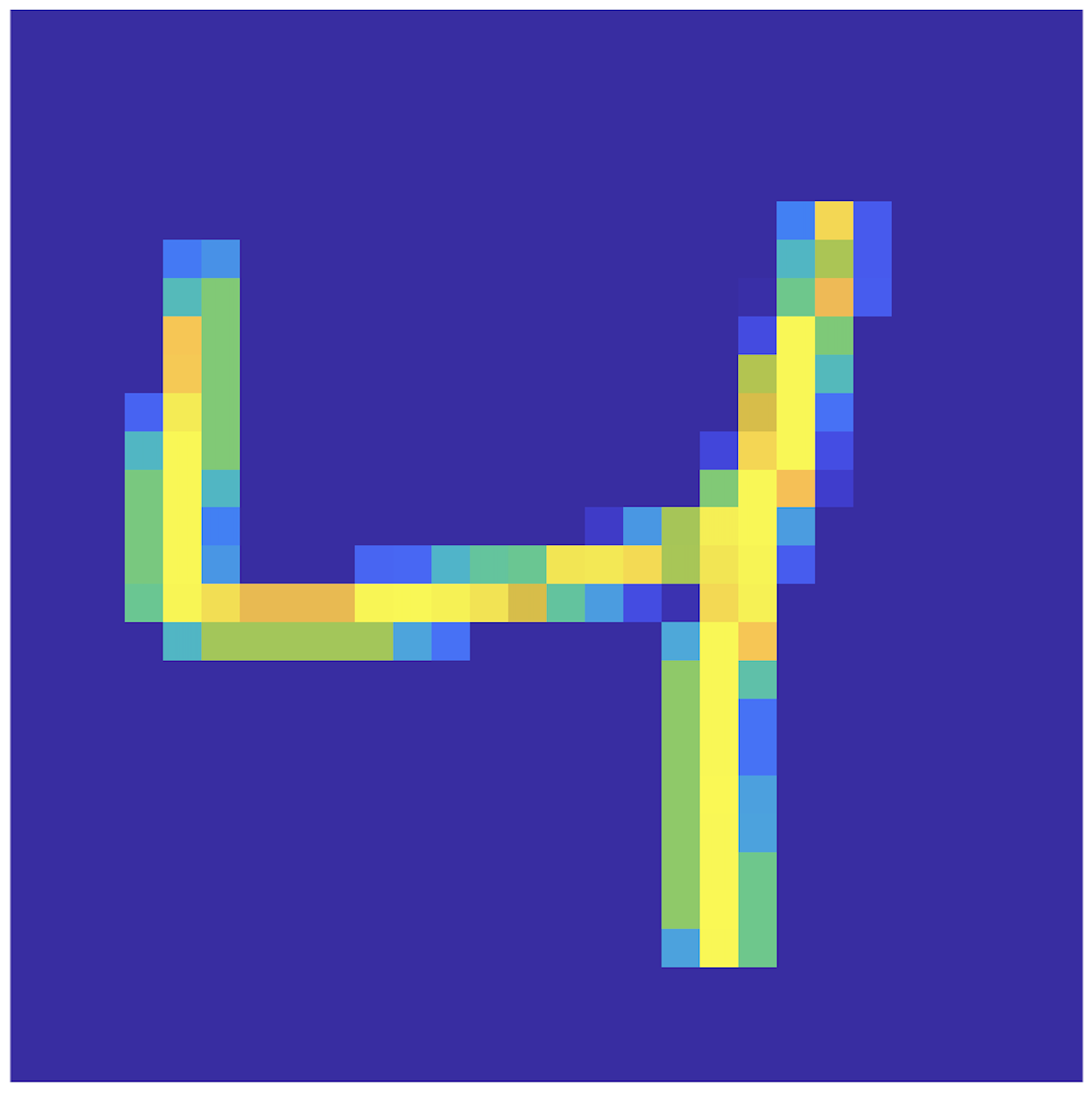}
    \vspace{4mm}
\end{tabular}
MNIST -  Distribution for number of bars for each bin and each image class\\[1mm]
\begin{tabular}{ccc}
  \includegraphics[width=.3\textwidth]{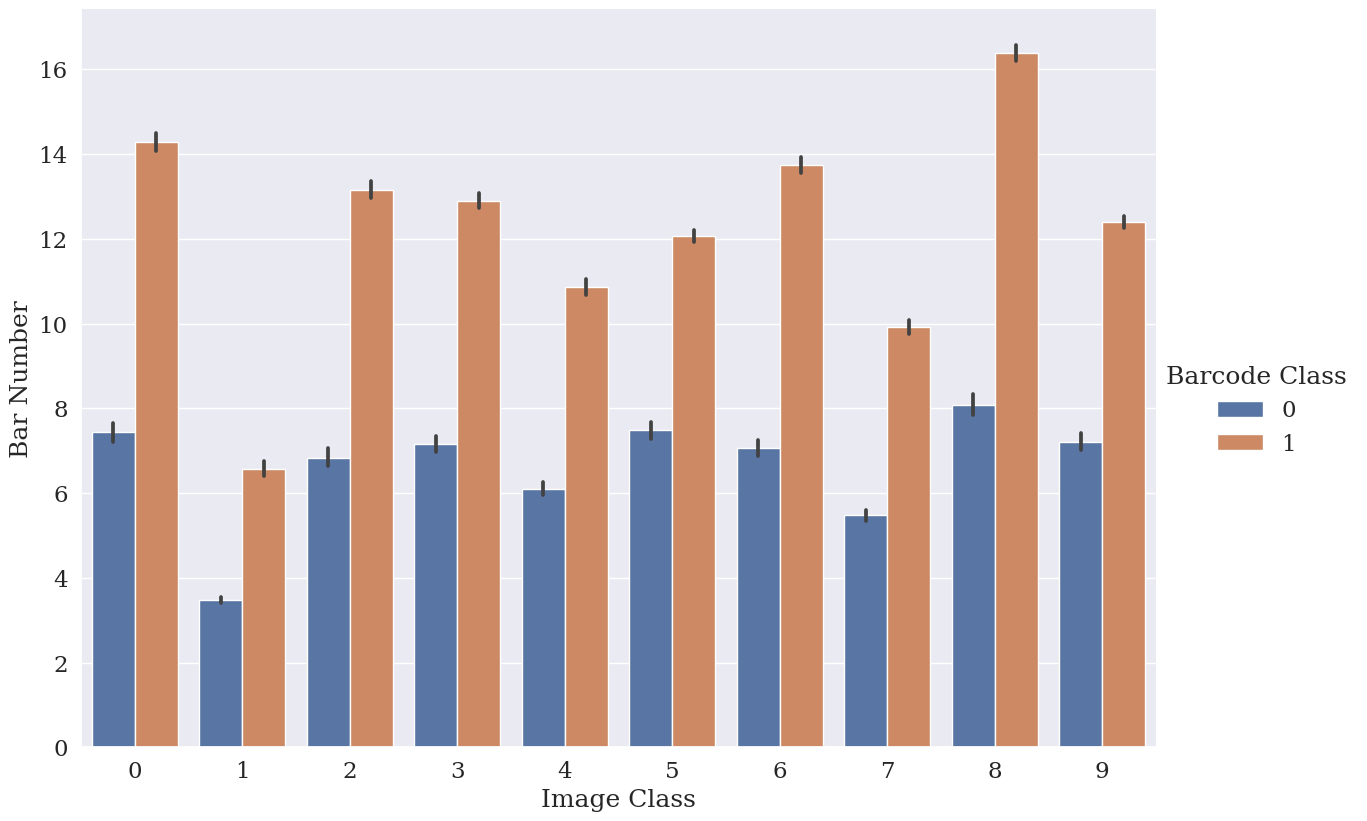} 
    &\includegraphics[width=.3\textwidth]{figures/MNIST_thres0.3.png}
    &\includegraphics[width=.3\textwidth]{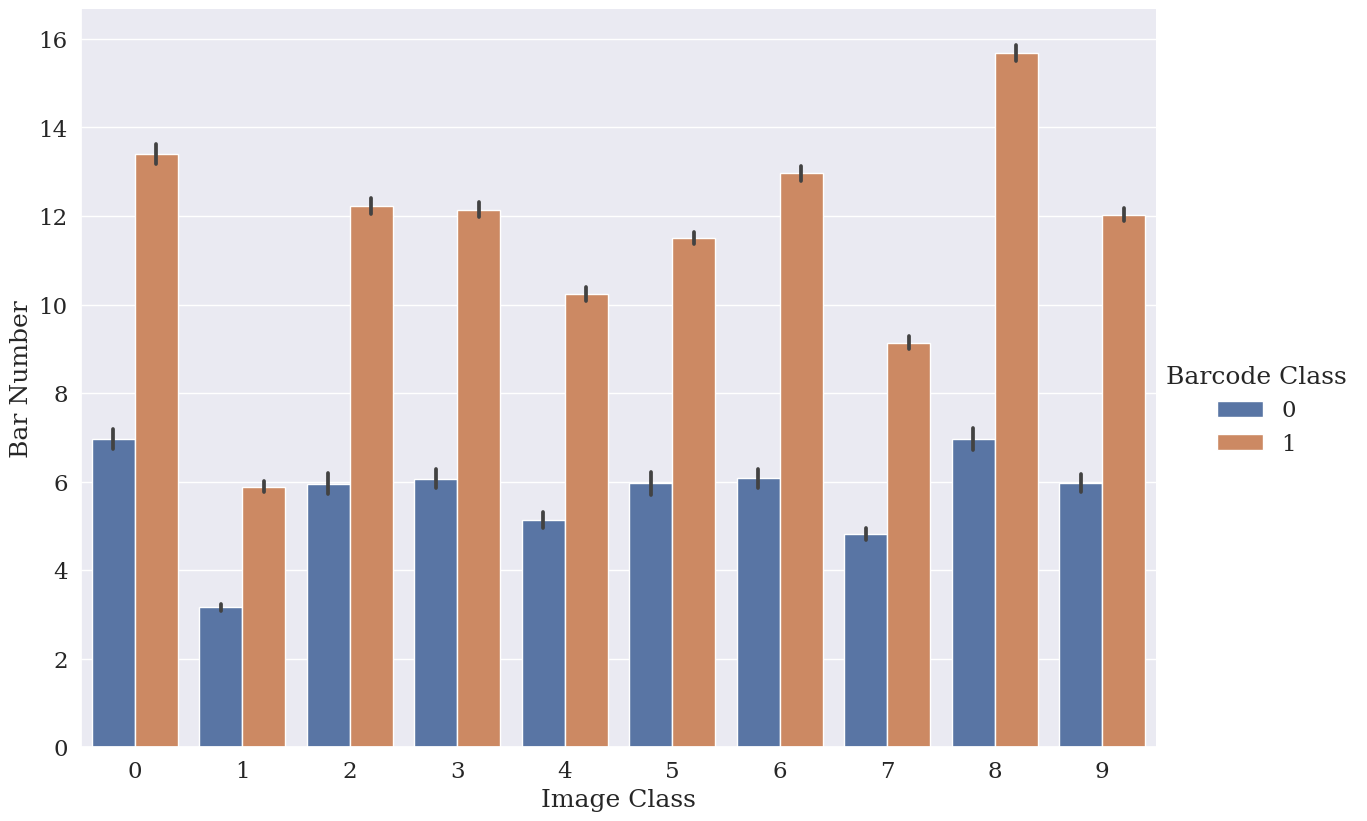}
\end{tabular}
    \caption{Data analysis for MNIST. For columns from left to right, the threshold is 0.15, 0.3 and 0.8.
    Top row: sample distribution in each class for the corresponding thresholds.
    Middle row: image samples in each PH class for the associated thresholds.
    Bottom row: number of bars of two bin for image class (0,1,2...,9).
    }
    \label{fig:MNIST_classification_data_analysis}
\end{figure}

\begin{figure}[h]
    \centering
CIFAR-10 - Training and validation losses for bar prediction\\[1mm]
\begin{tabular}{ccc}
  \includegraphics[width=.3\textwidth]{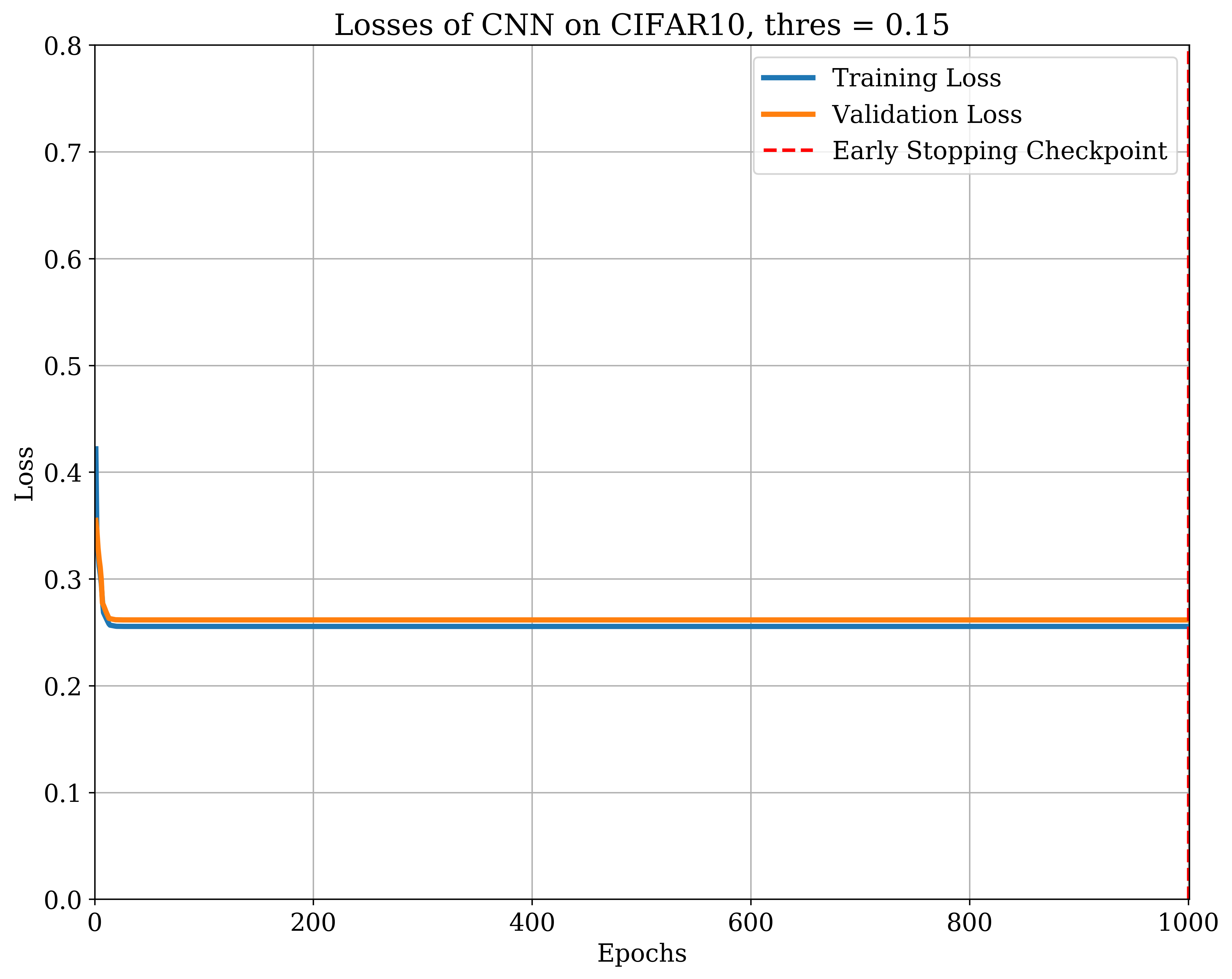} 
    &\includegraphics[width=.3\textwidth]{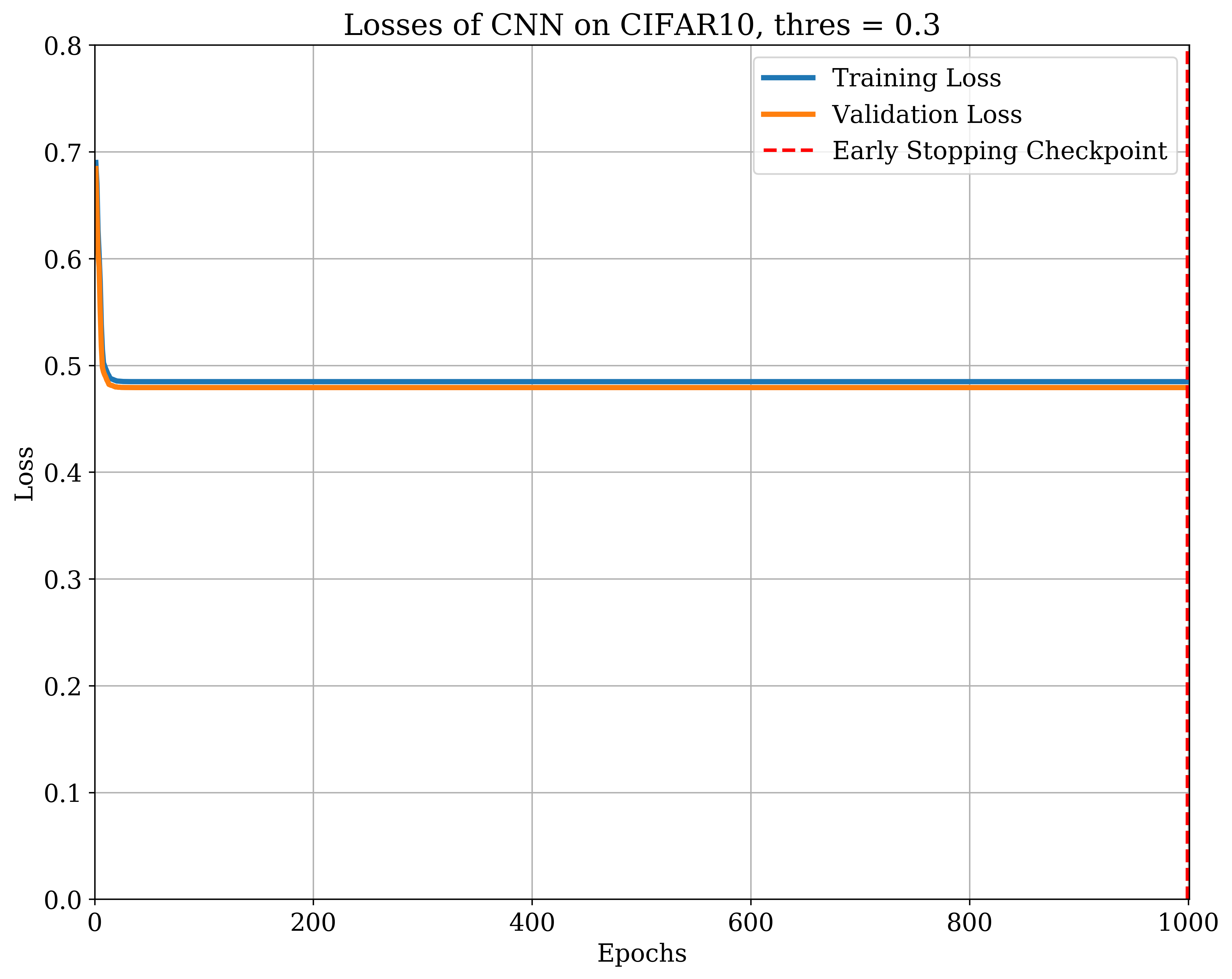}
    &\includegraphics[width=.3\textwidth]{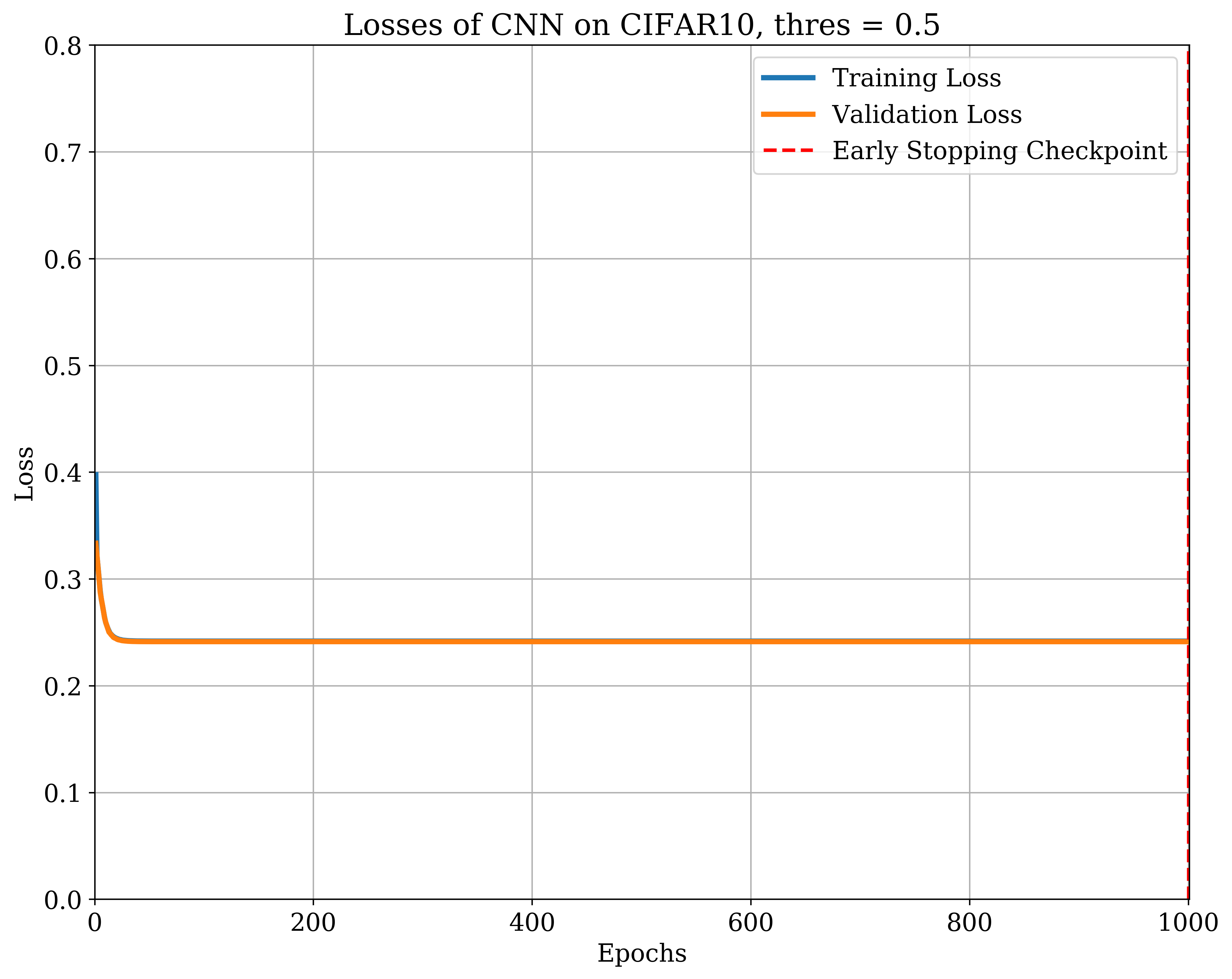}
    \vspace{4mm}
\end{tabular}
CIFAR-10 - Validation accuracy for bar prediction\\[1mm]
\begin{tabular}{ccc}
  \includegraphics[width=.3\textwidth]{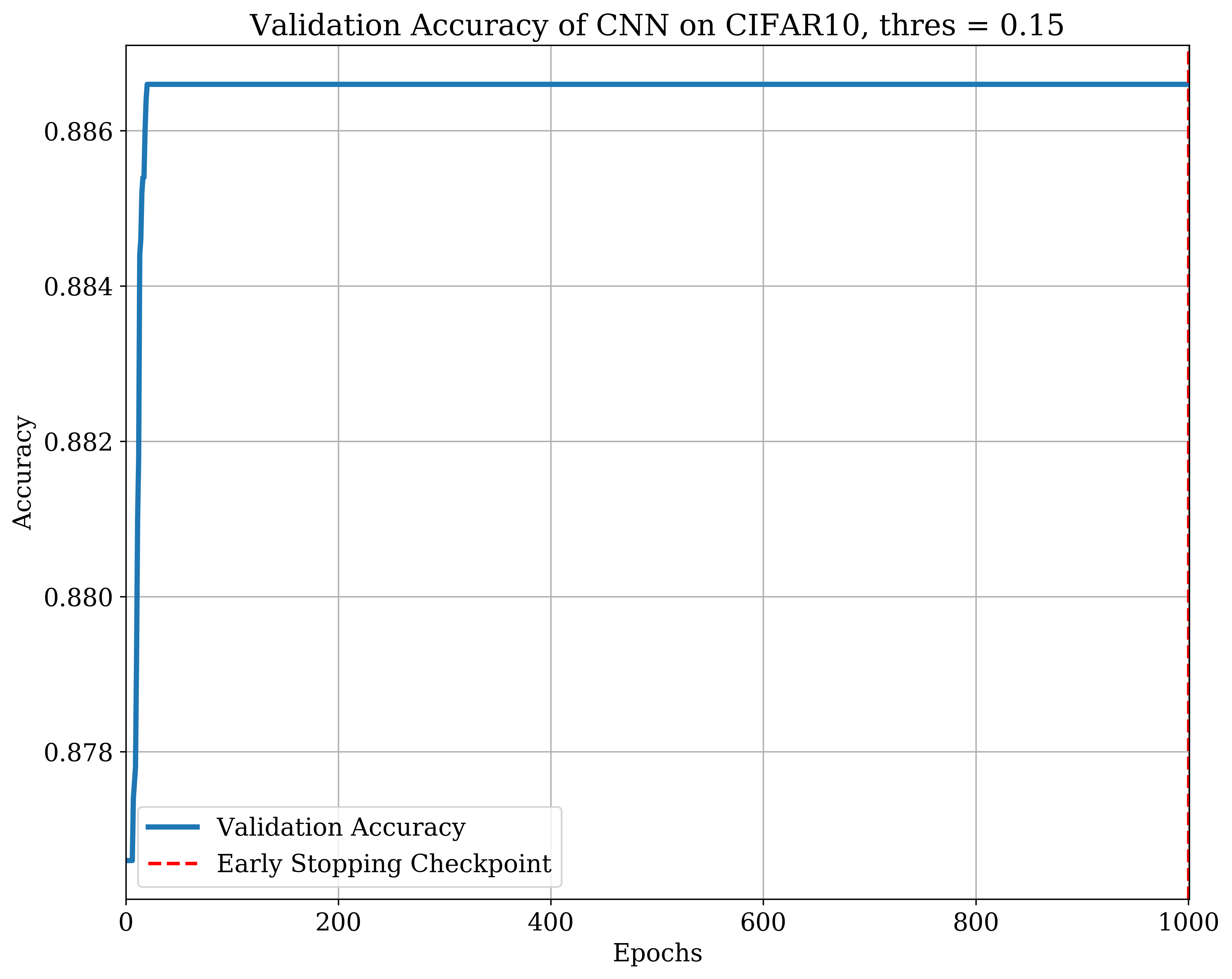} 
    &\includegraphics[width=.3\textwidth]{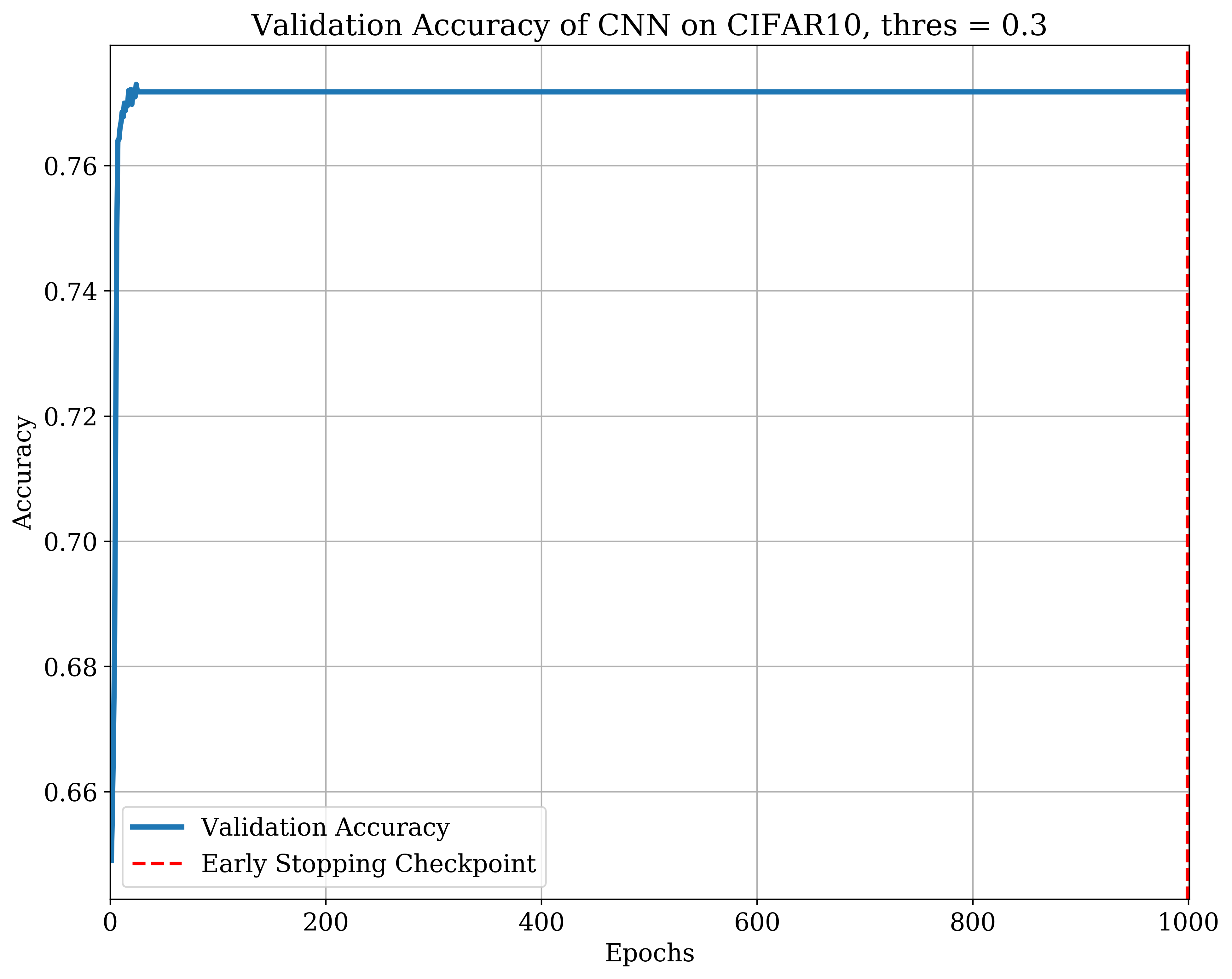}
    &\includegraphics[width=.3\textwidth]{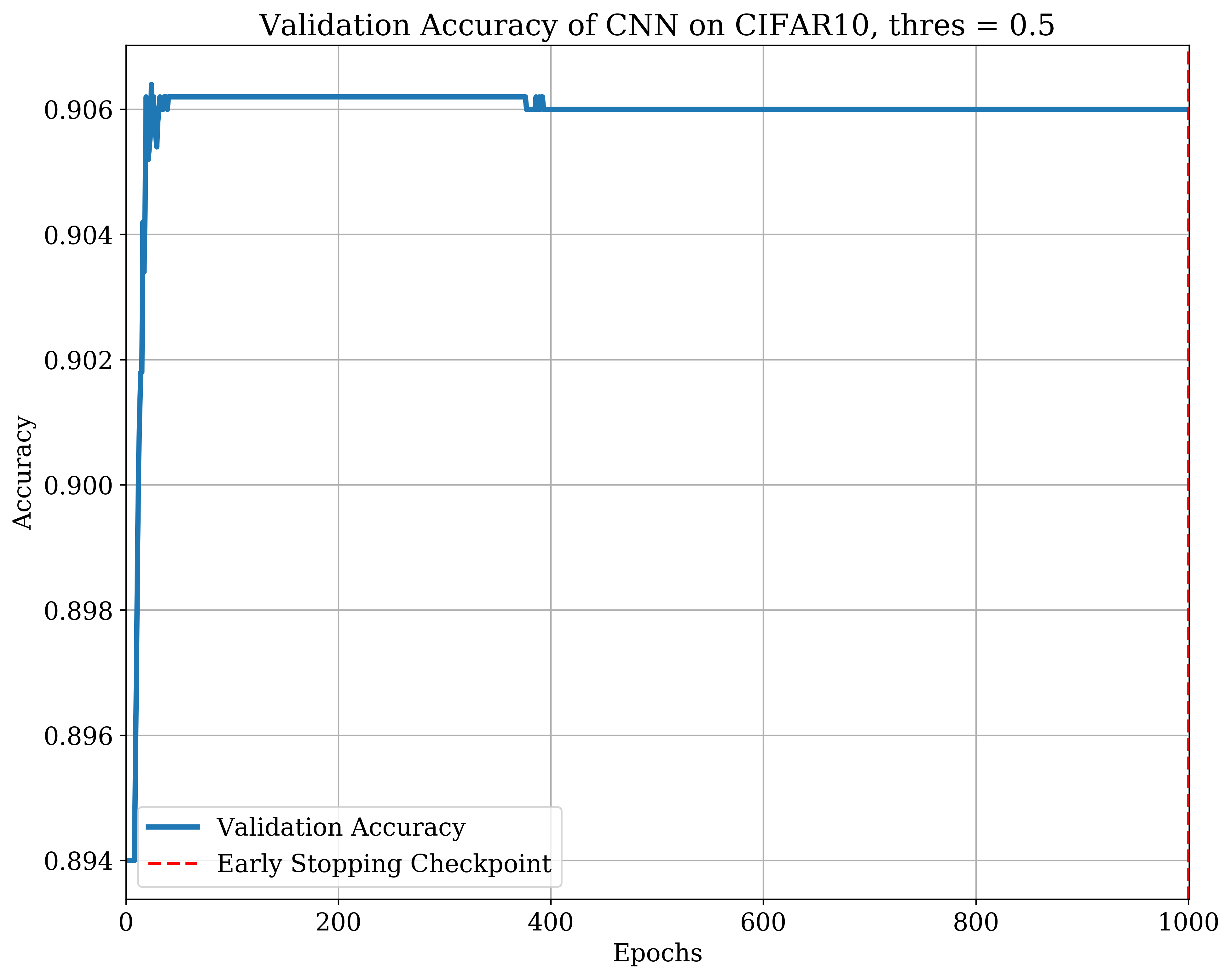}
\end{tabular}
\caption{Top row: training and validation losses of CNN for predicting a bar on CIFAR-10, where threshold is $0.15, 0.3, 0.5$.
Bottom row: test accuracy of CNN for predicting a bar for the corresponding thresholds.}\label{fig:cifar10_classification_train_val_loss_accuracy}
\end{figure}

\begin{figure}[h]
\centering
CIFAR-10 - Sample distribution for bar prediction task\\[1mm]
\begin{tabular}{ccc}
  \includegraphics[width=.3\textwidth]{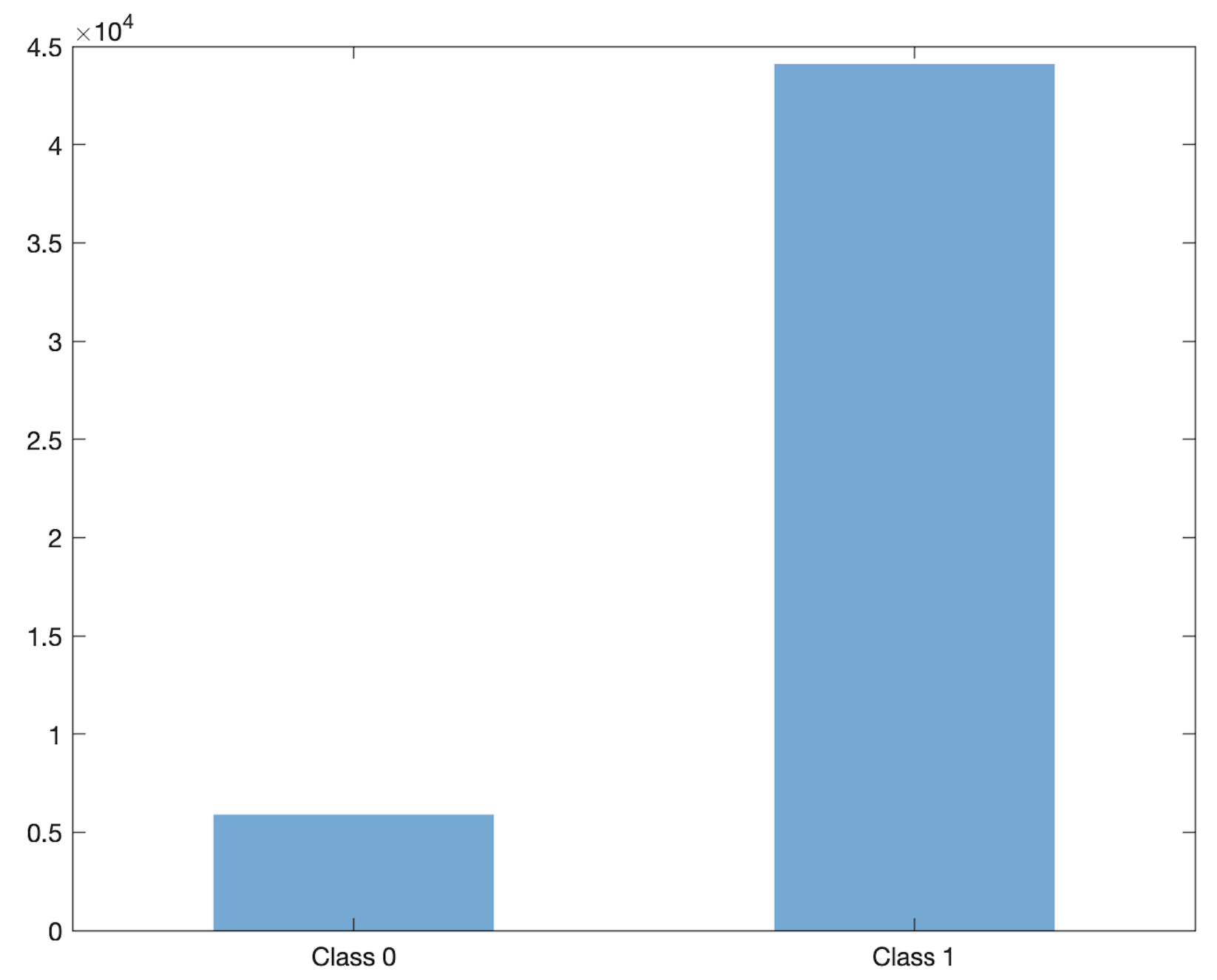} 
    &\includegraphics[width=.3\textwidth]{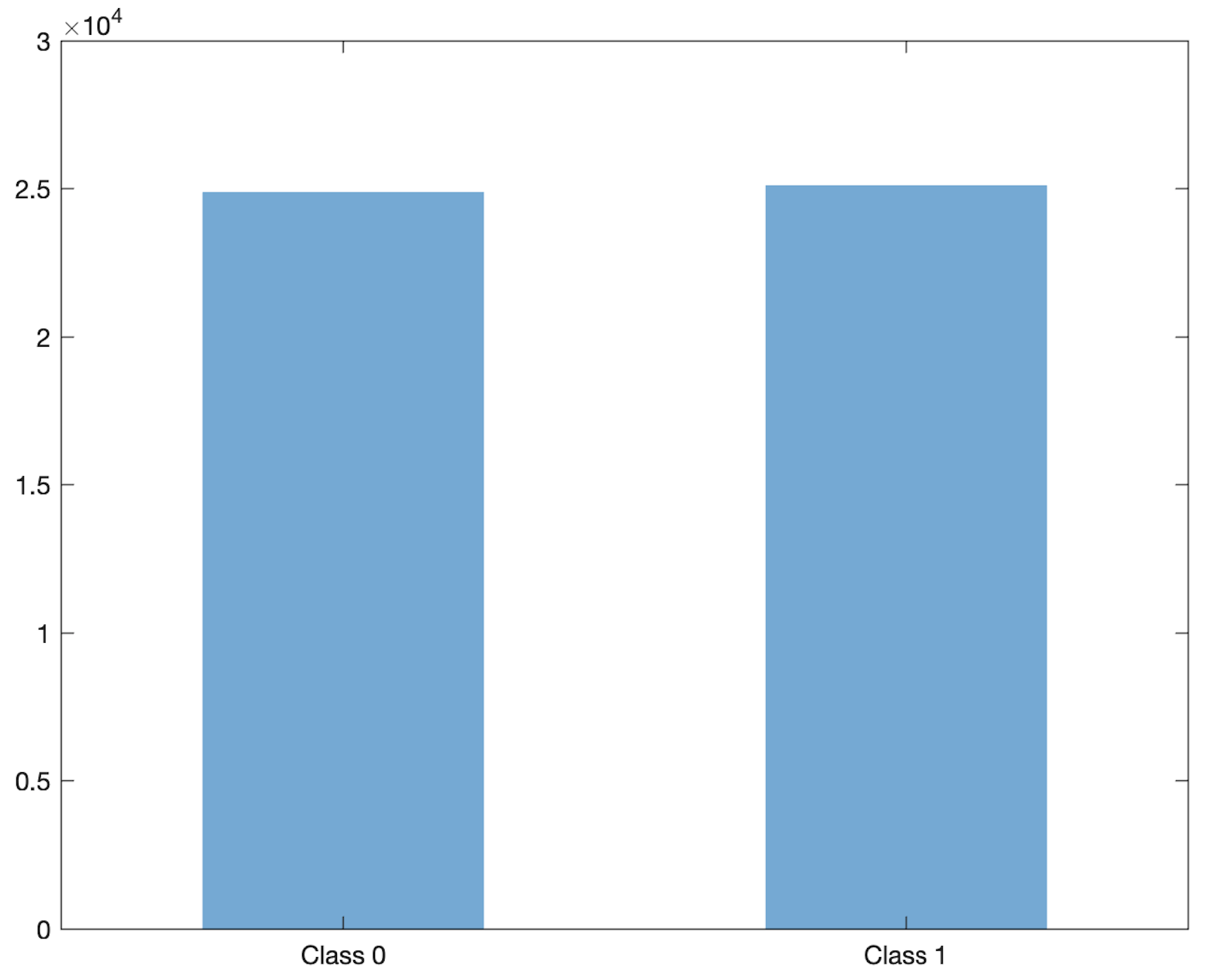}
    &\includegraphics[width=.3\textwidth]{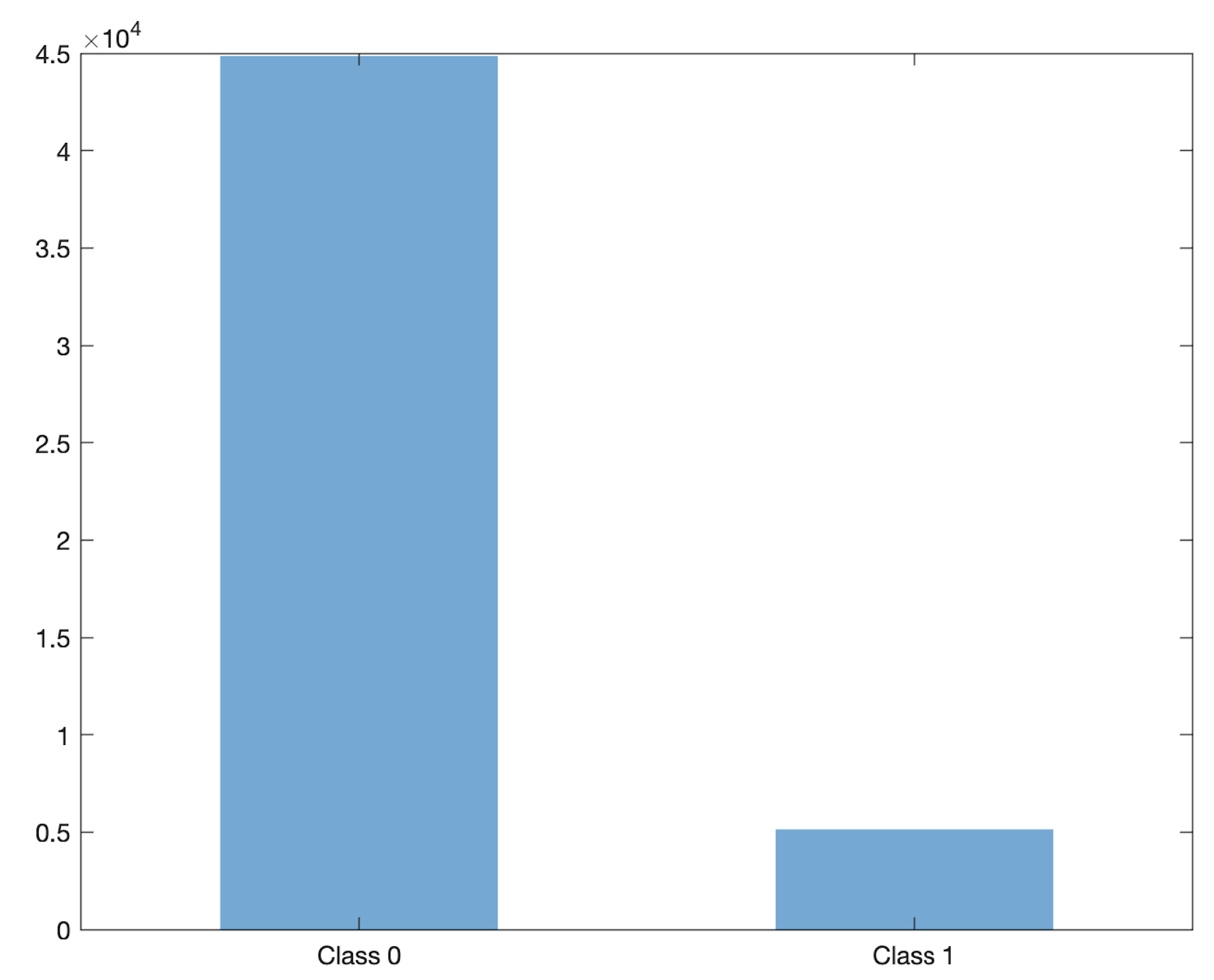}\vspace{4mm}
\end{tabular}
CIFAR-10 - Samples in each PH class\\[1mm]
\begin{tabular}{cccccccc}
Class 0 & Class 1 && Class 0 & Class 1 && Class 0 & Class 1\\
  \includegraphics[width=.12\textwidth]{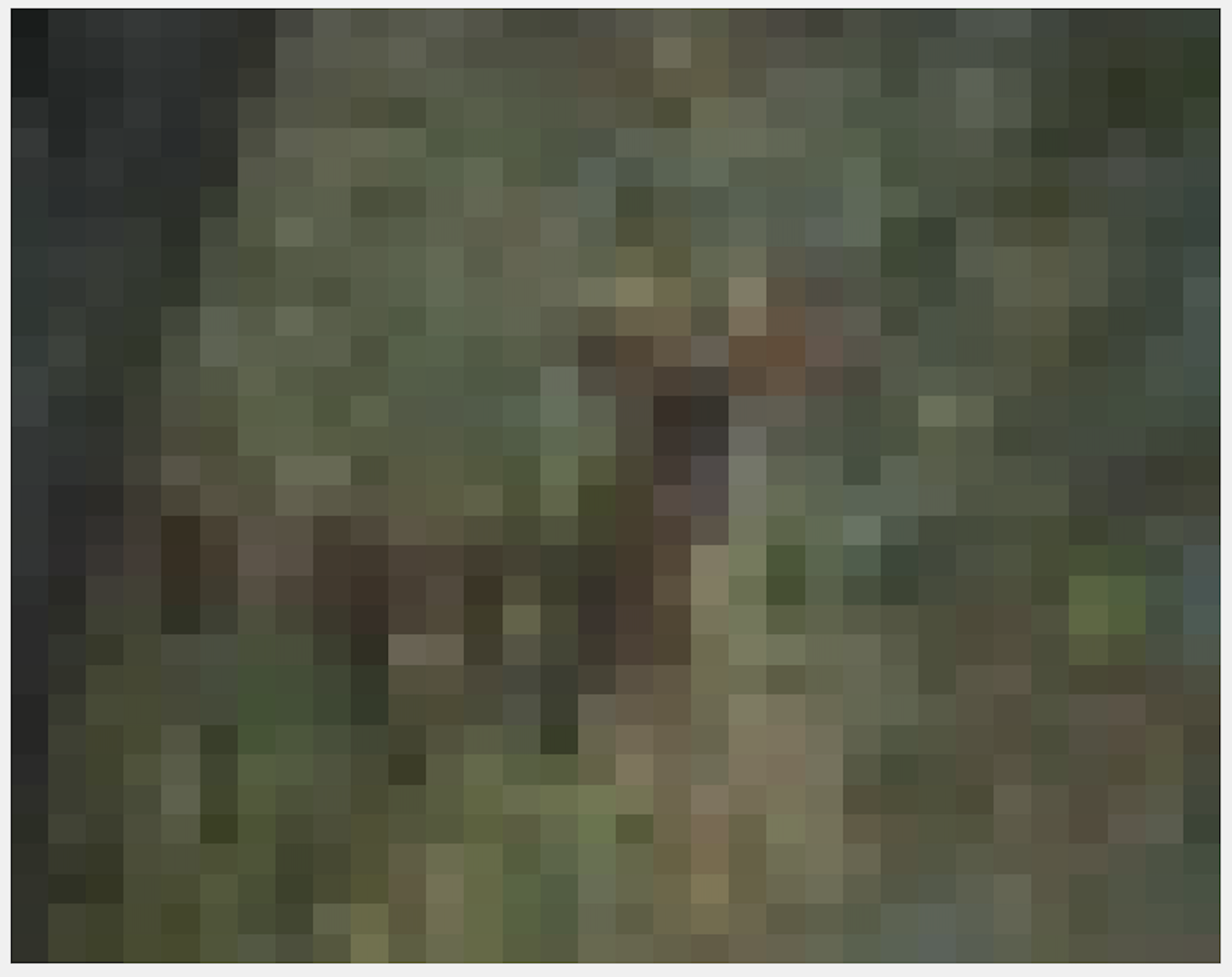}
  &\includegraphics[width=.12\textwidth]{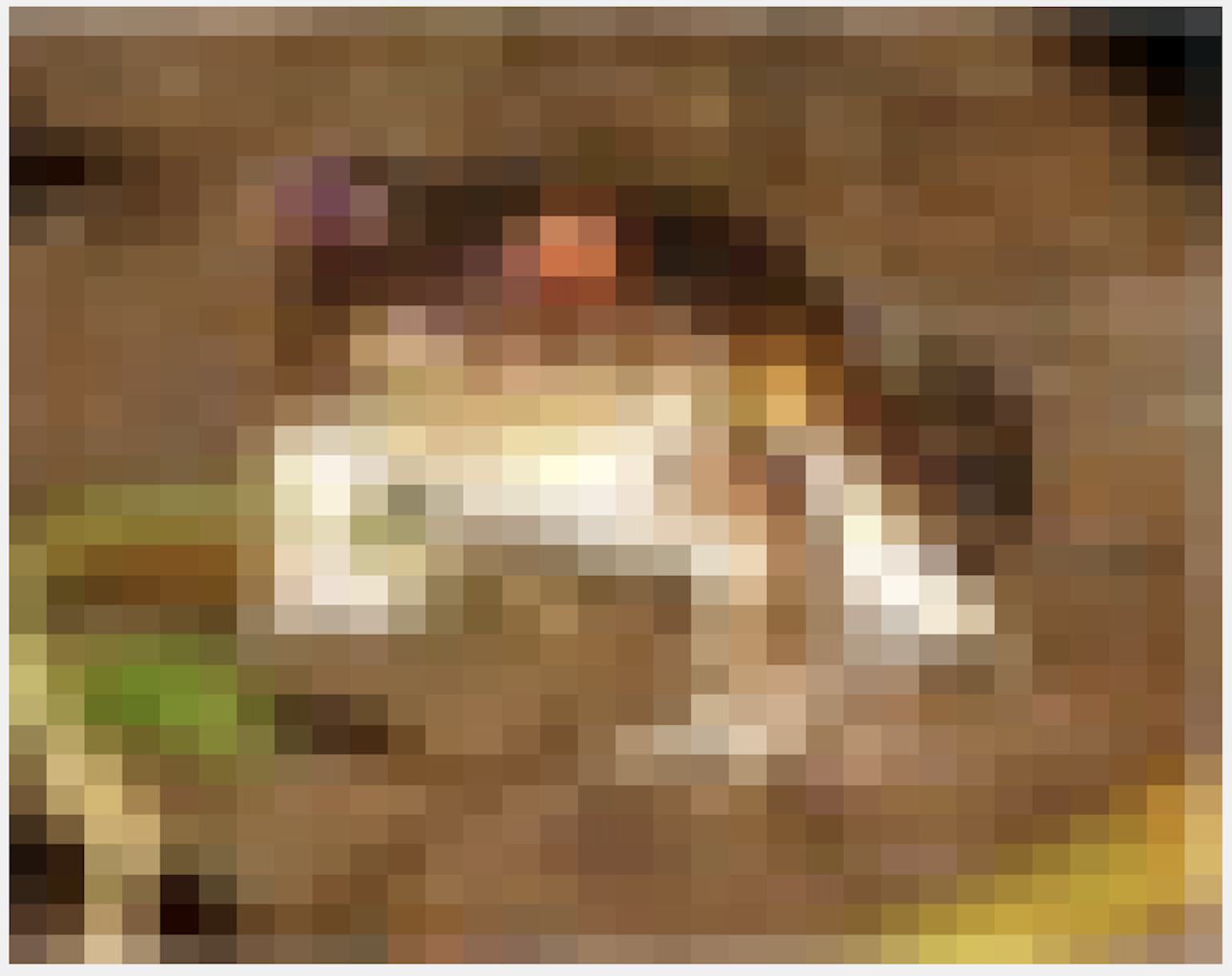}
    &&
    \includegraphics[width=.12\textwidth]{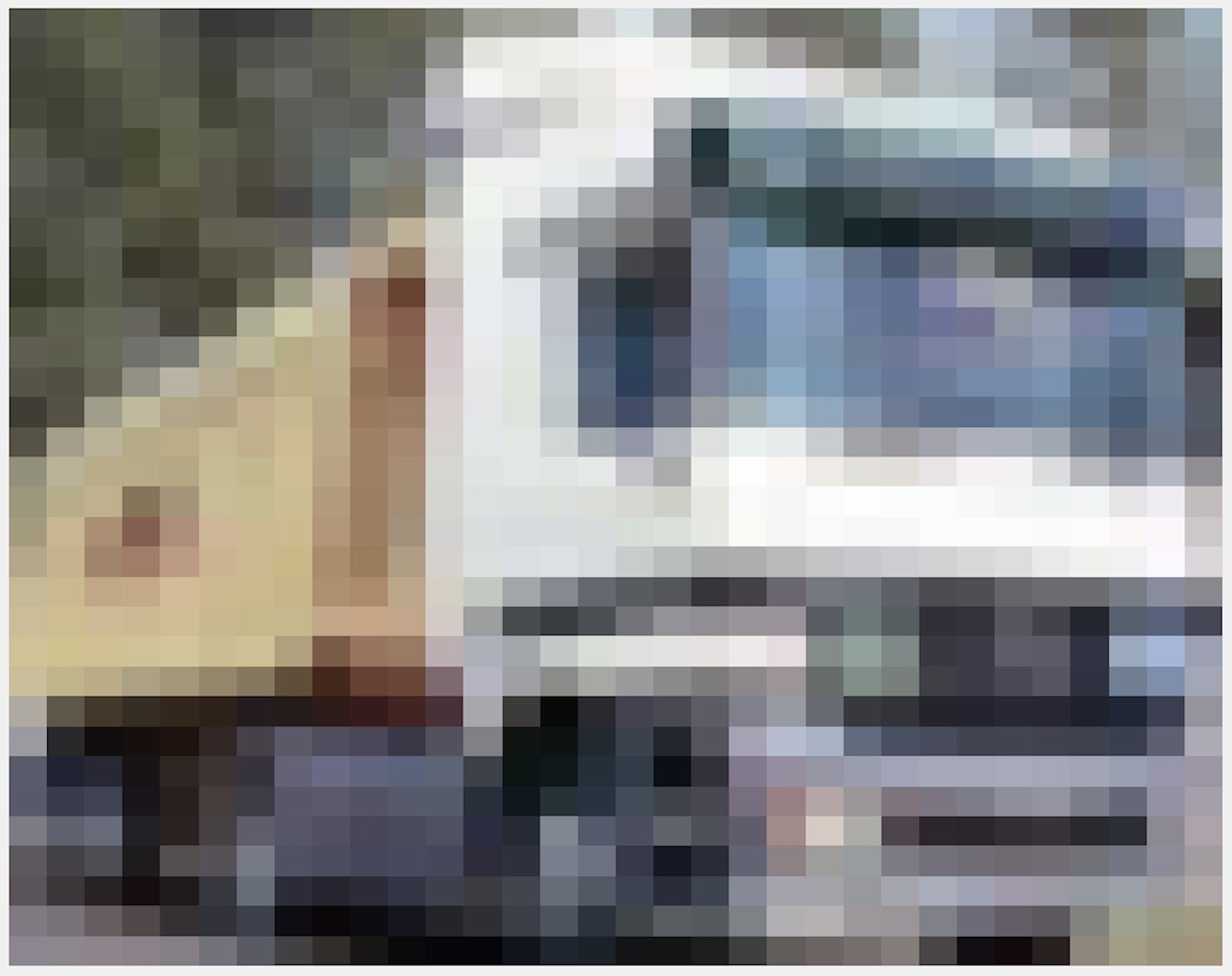}
    &\includegraphics[width=.12\textwidth]{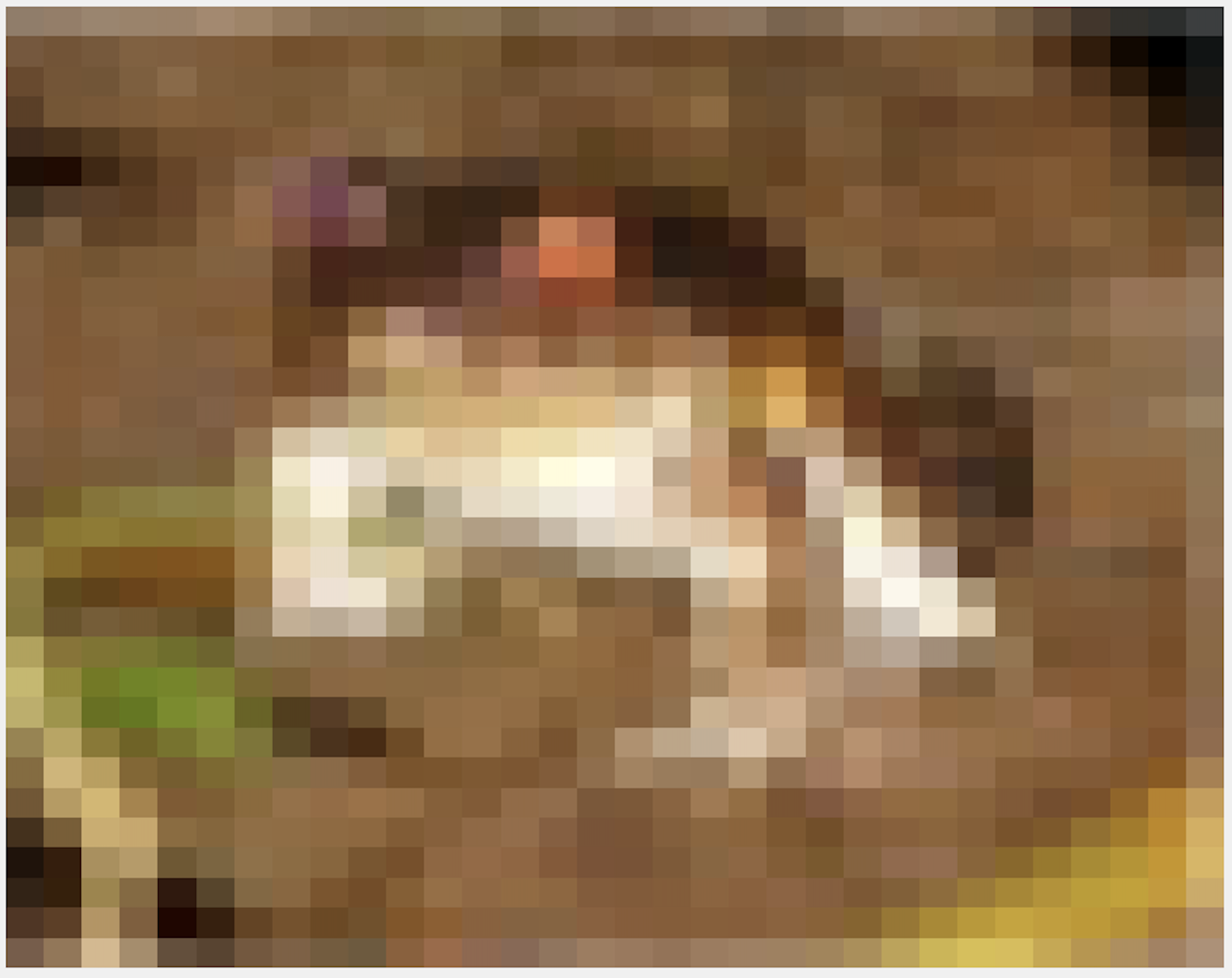}
    &&\includegraphics[width=.12\textwidth]{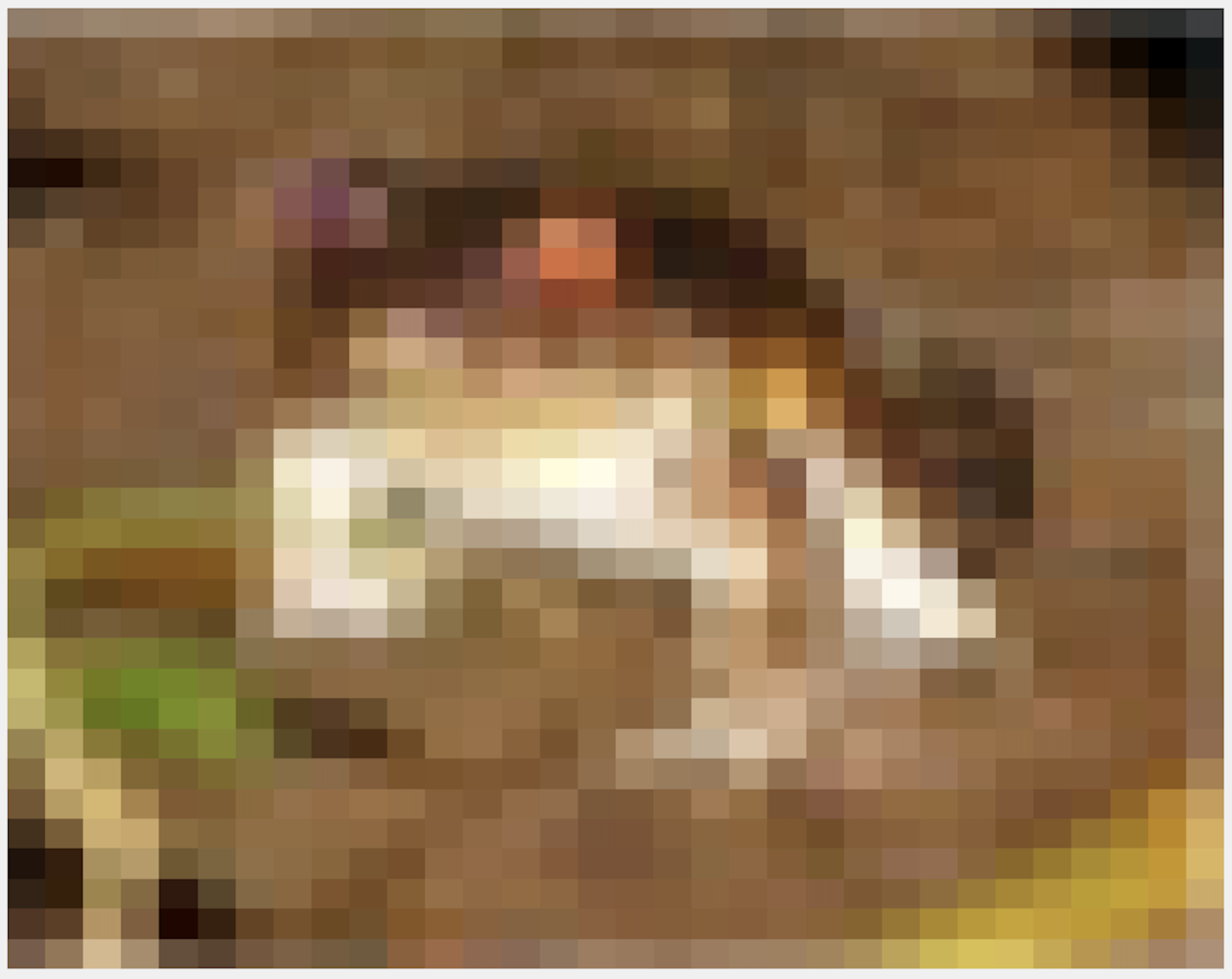}
    &\includegraphics[width=.12\textwidth]{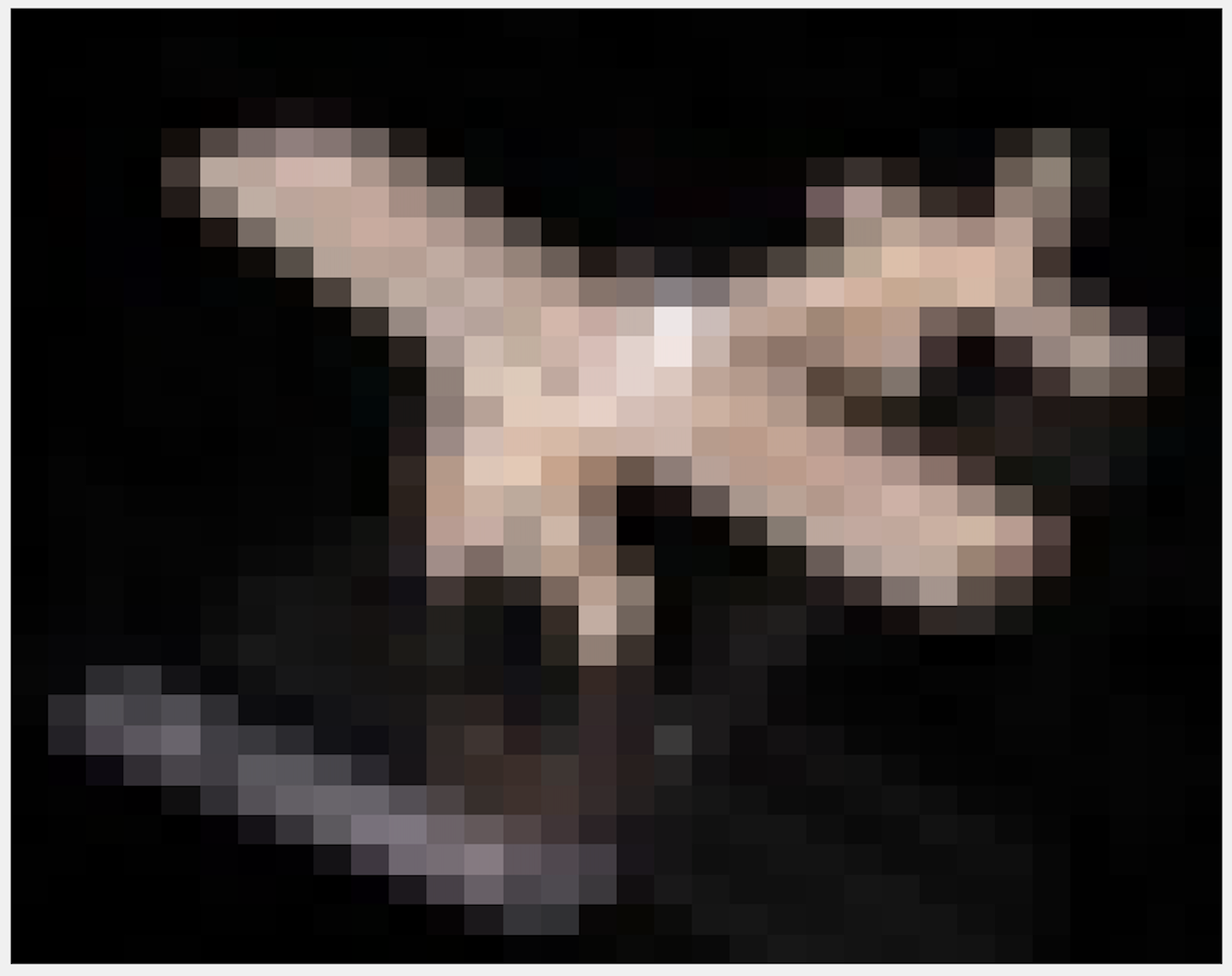}\\
  \includegraphics[width=.12\textwidth]{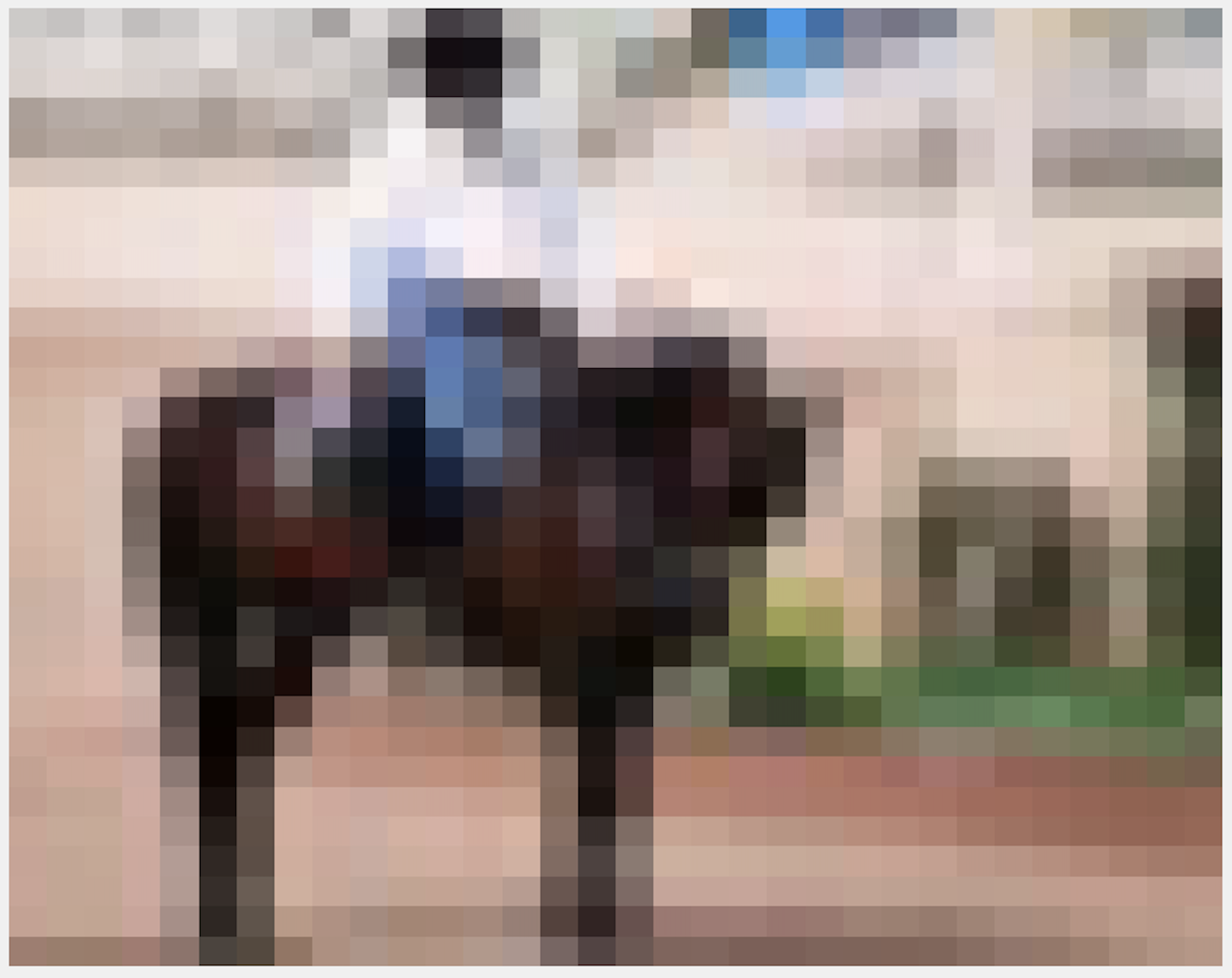}
  &\includegraphics[width=.12\textwidth]{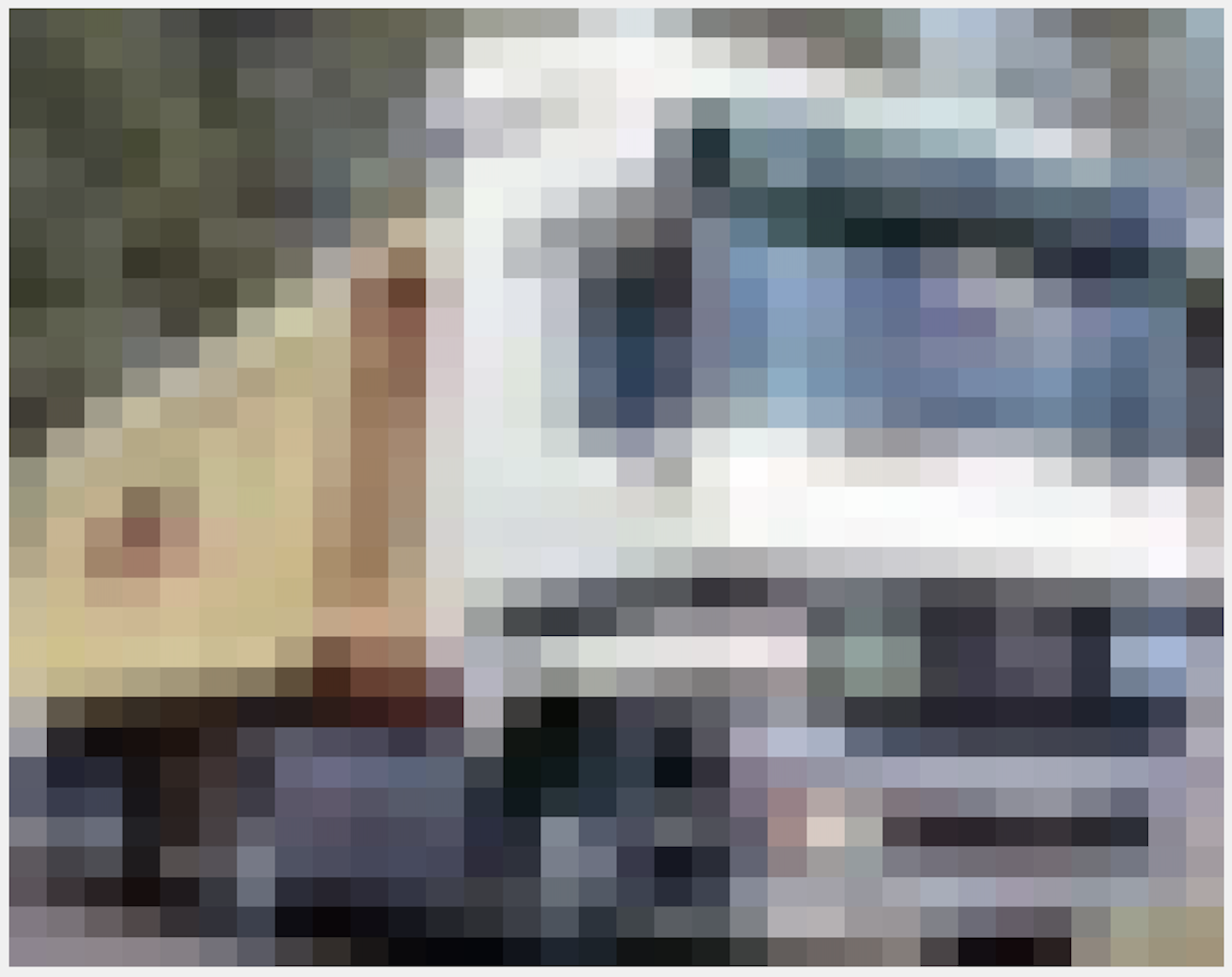}
    &&
    \includegraphics[width=.12\textwidth]{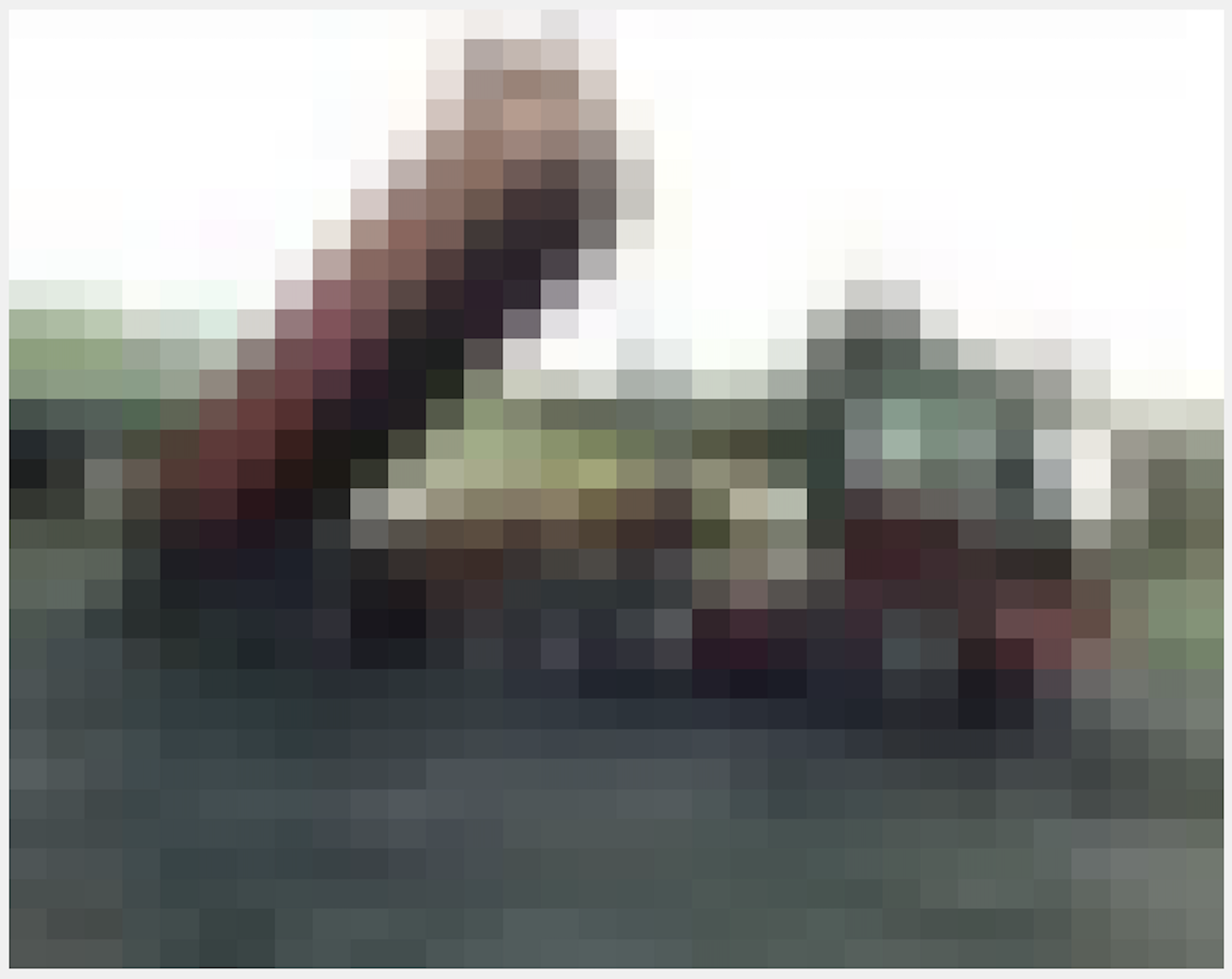}
    &\includegraphics[width=.12\textwidth]{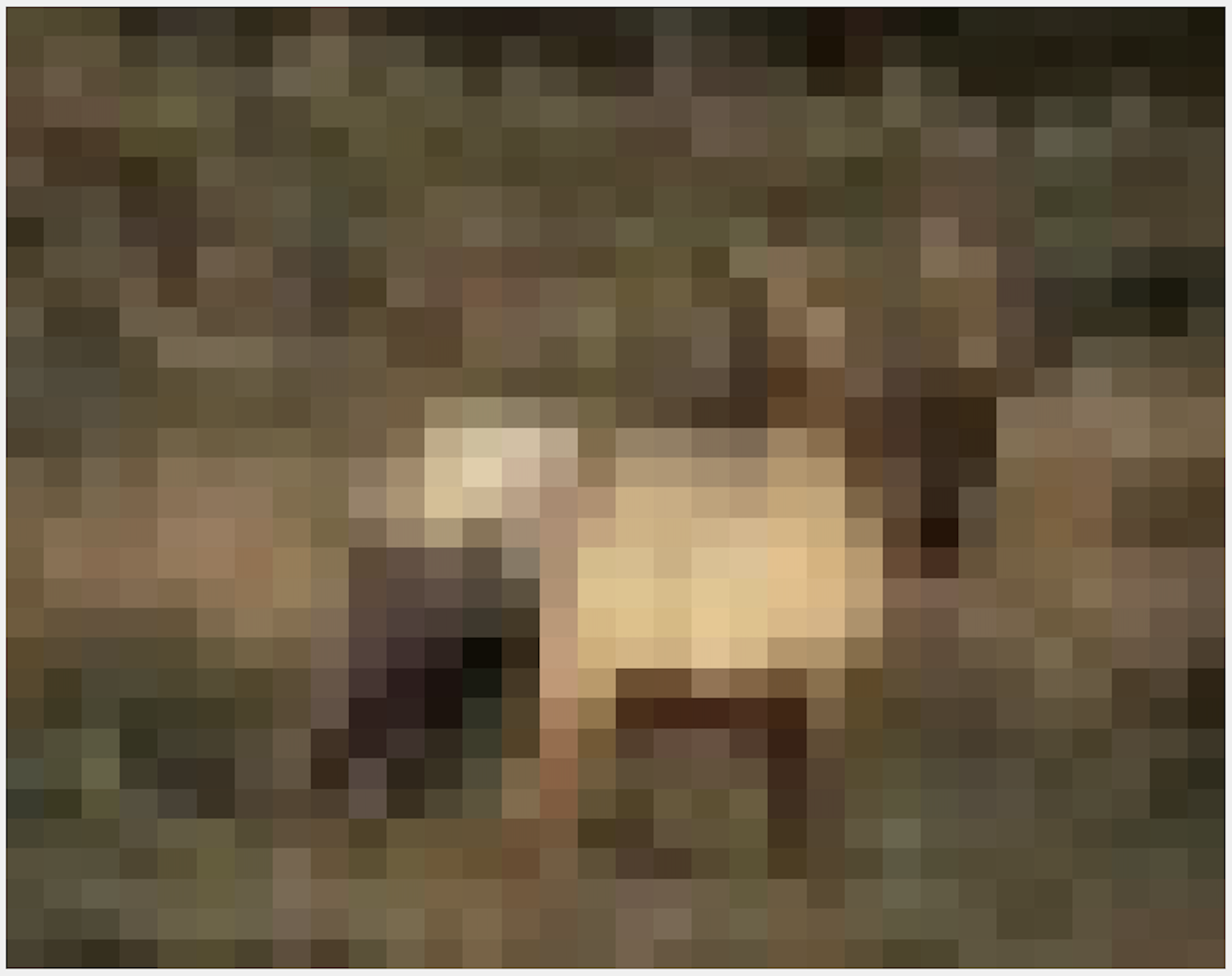}
    &&\includegraphics[width=.12\textwidth]{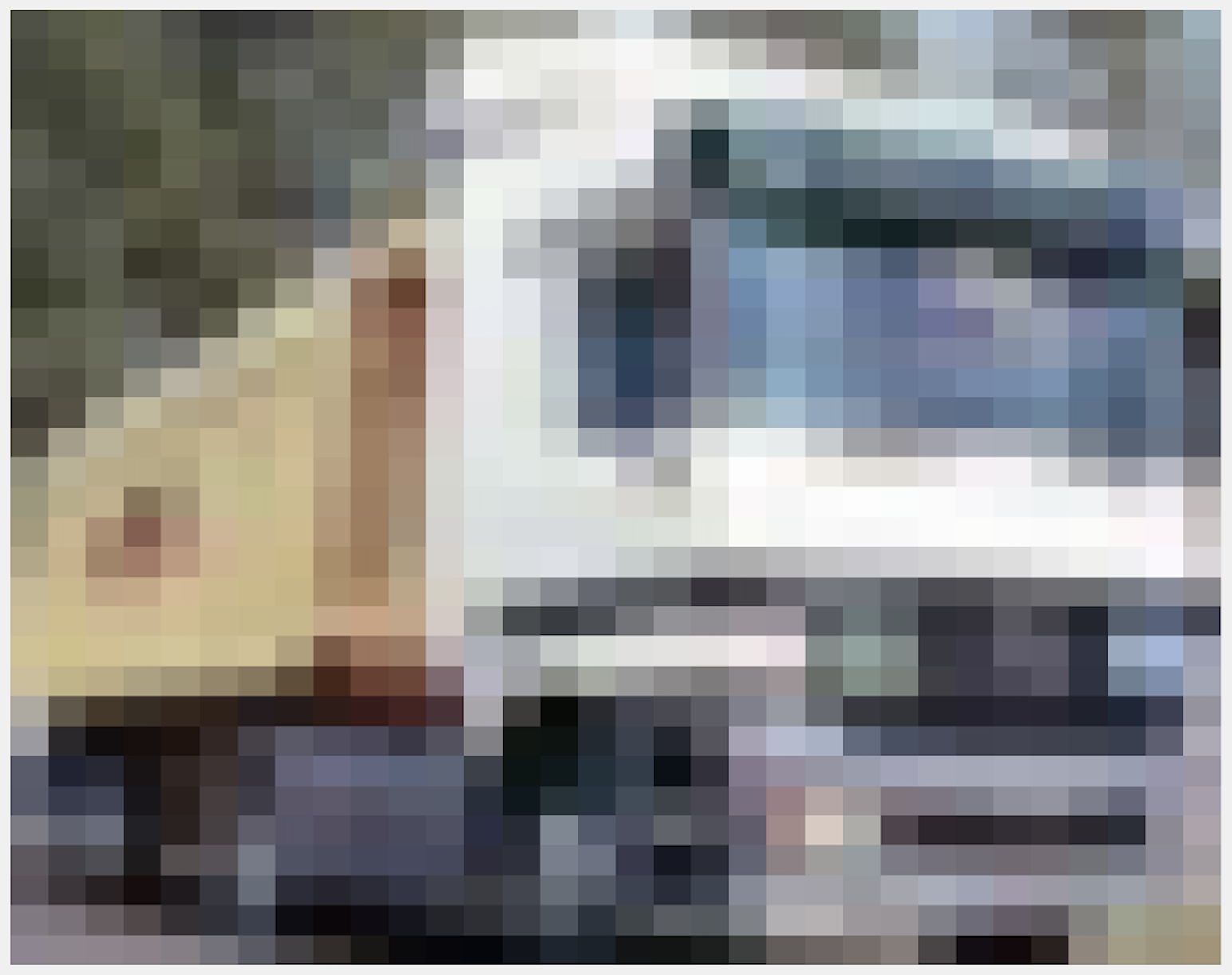}
    &\includegraphics[width=.12\textwidth]{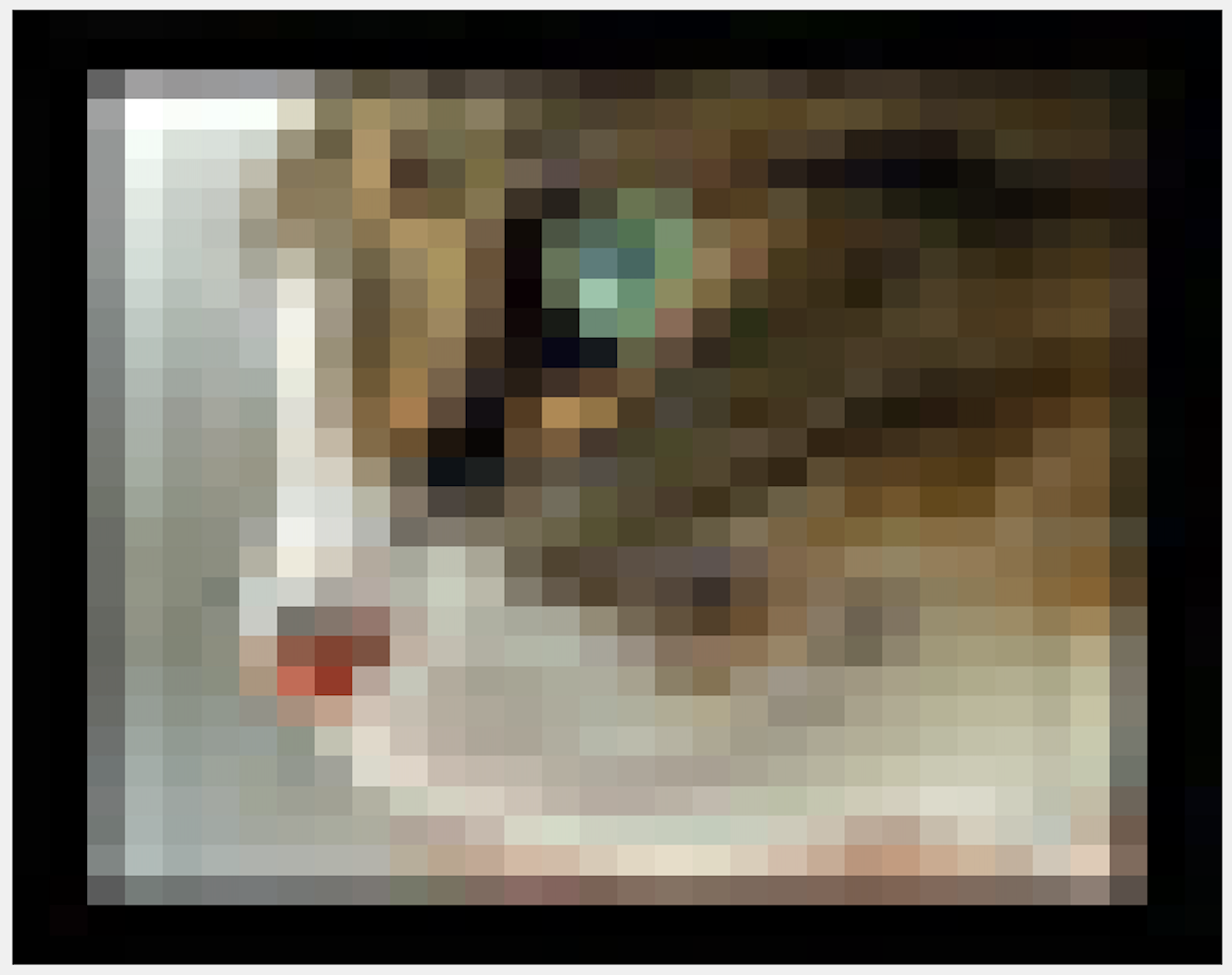}\\
  \includegraphics[width=.12\textwidth]{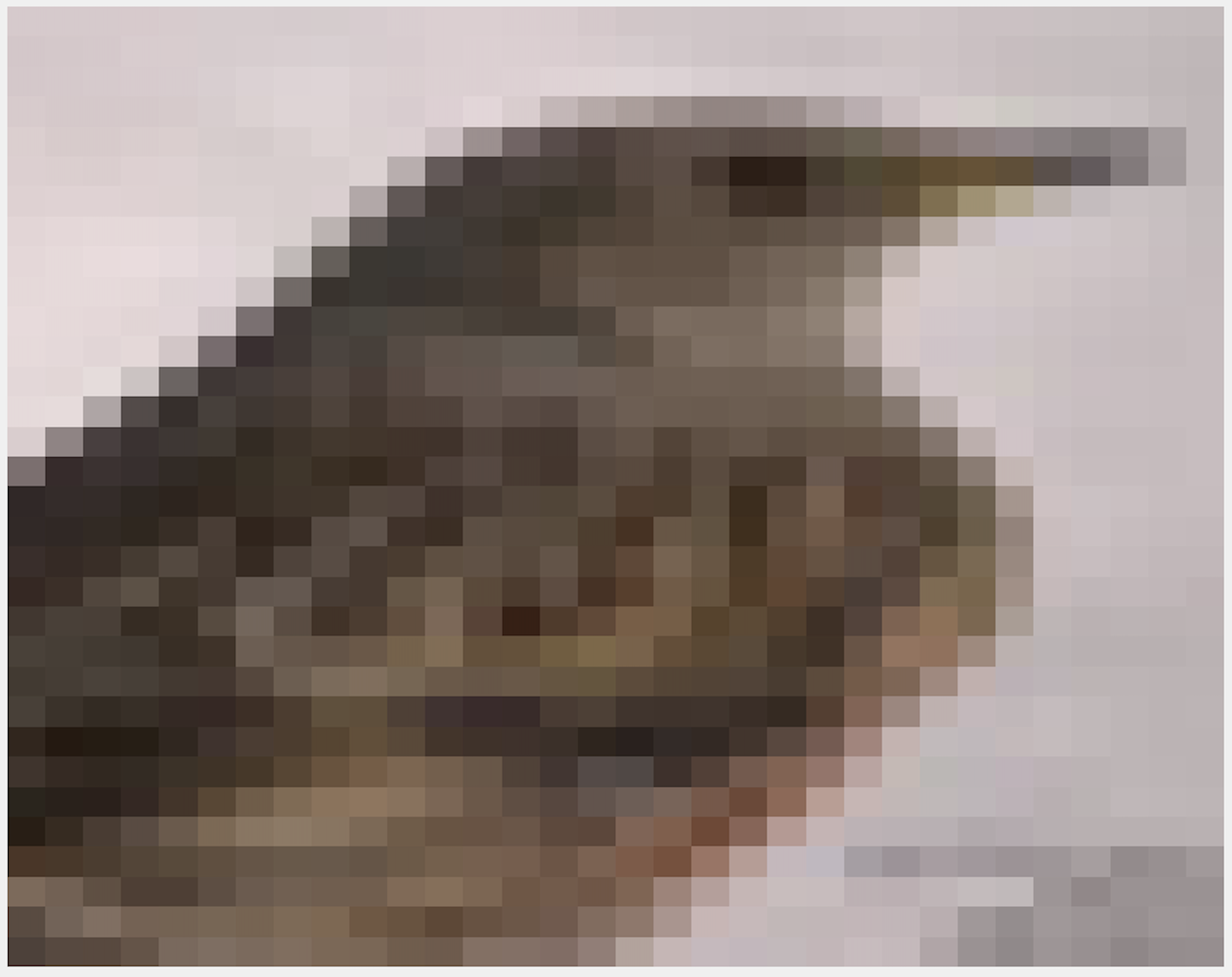}
  &\includegraphics[width=.12\textwidth]{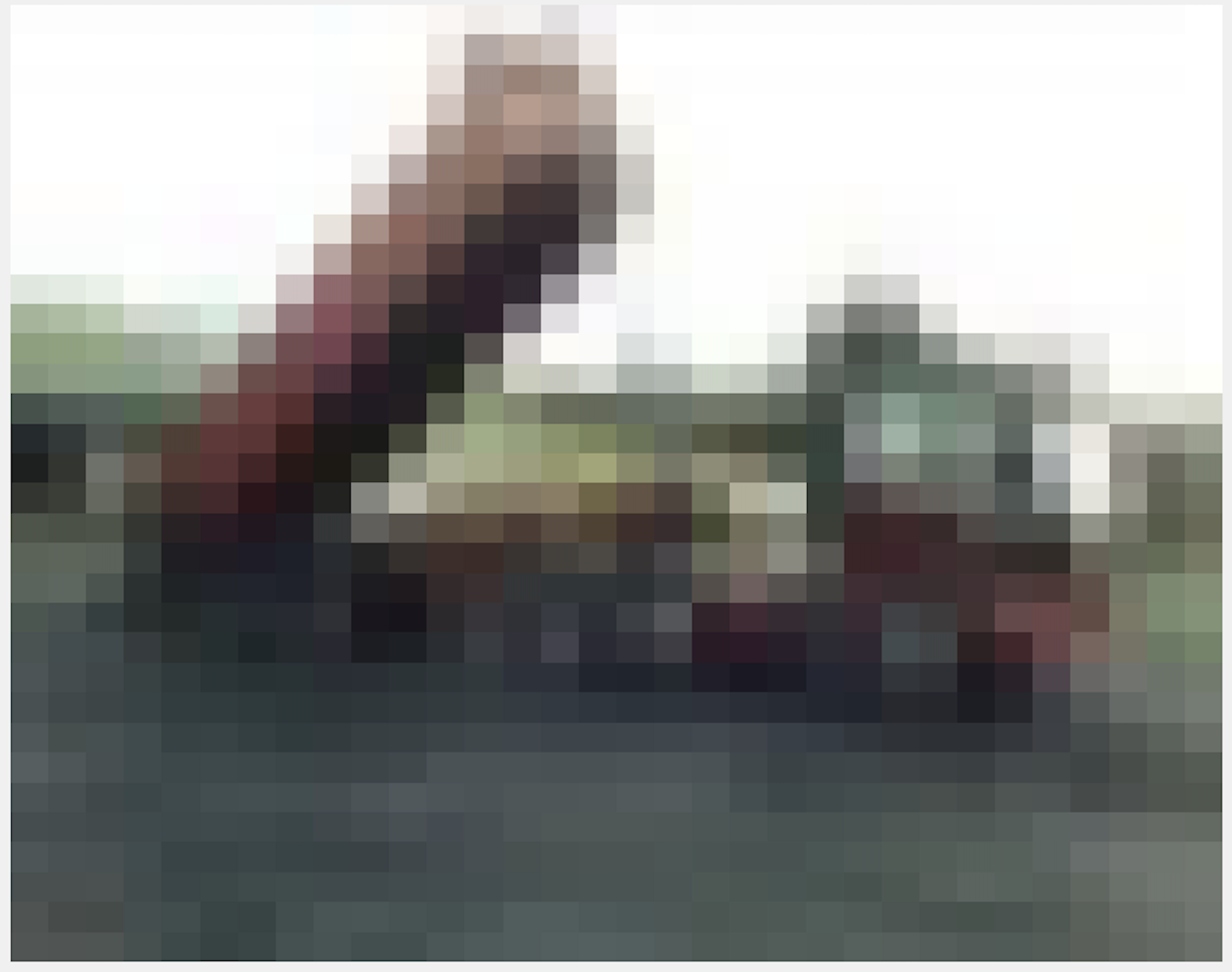}
    &&
    \includegraphics[width=.12\textwidth]{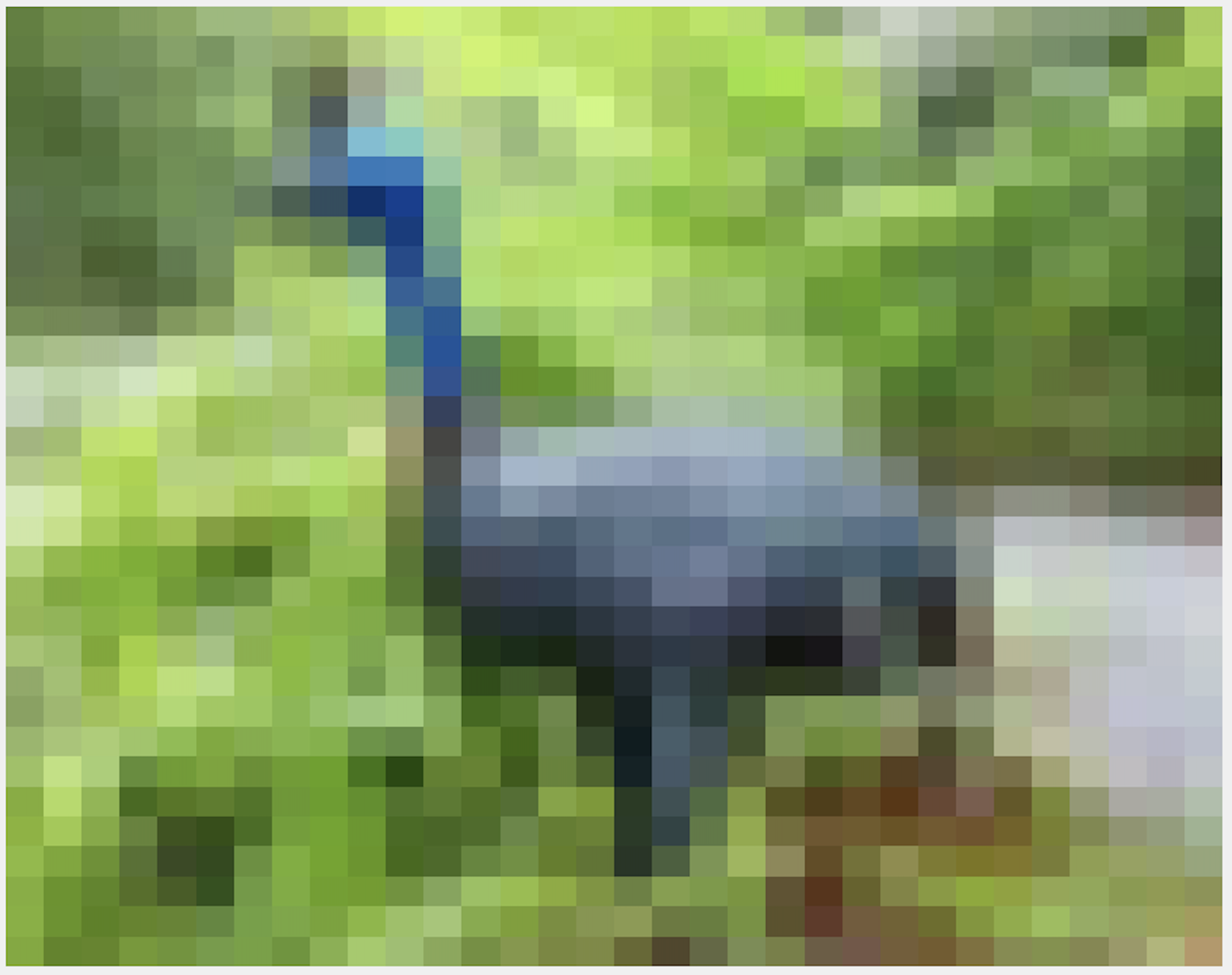}
    &\includegraphics[width=.12\textwidth]{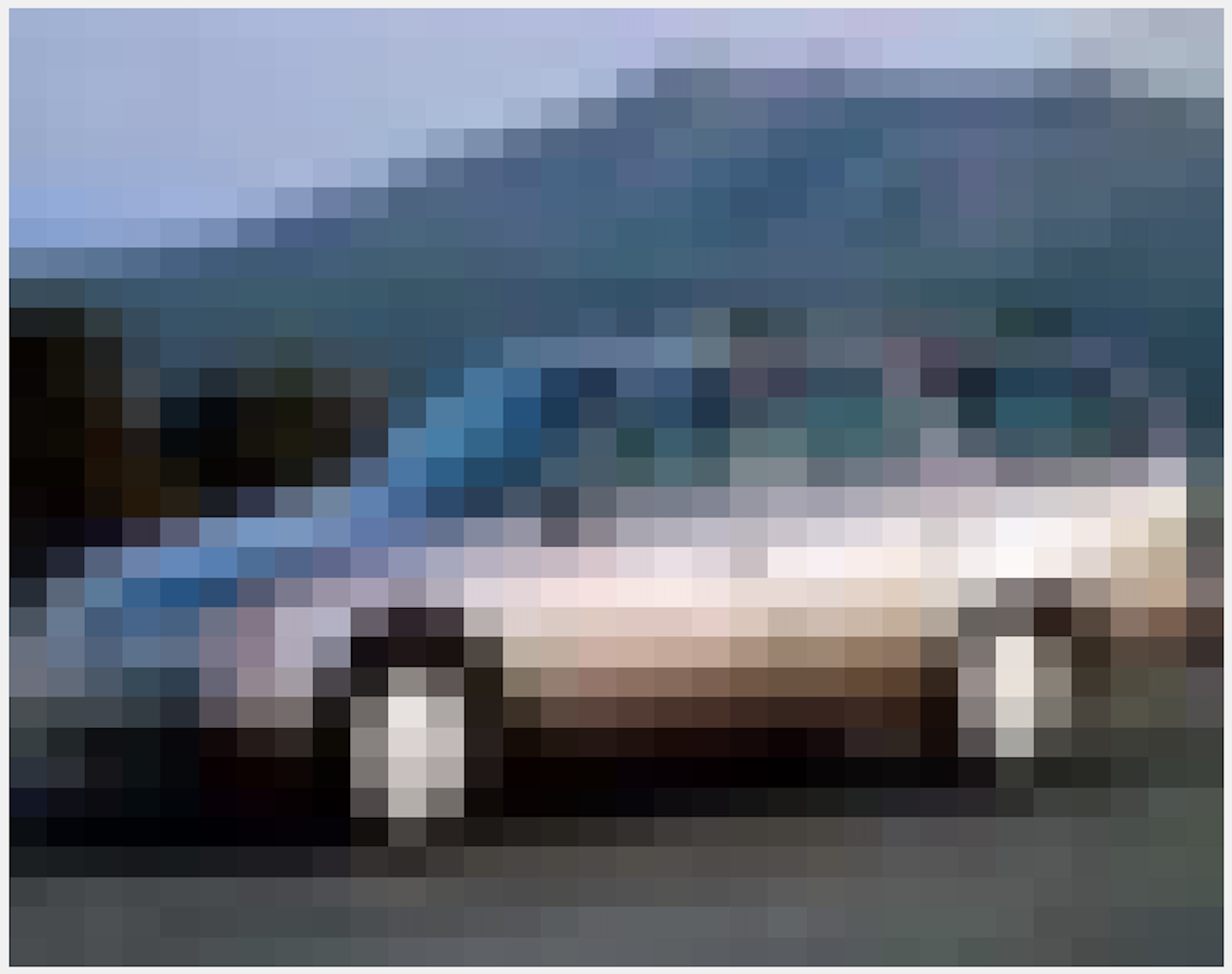}
    &&\includegraphics[width=.12\textwidth]{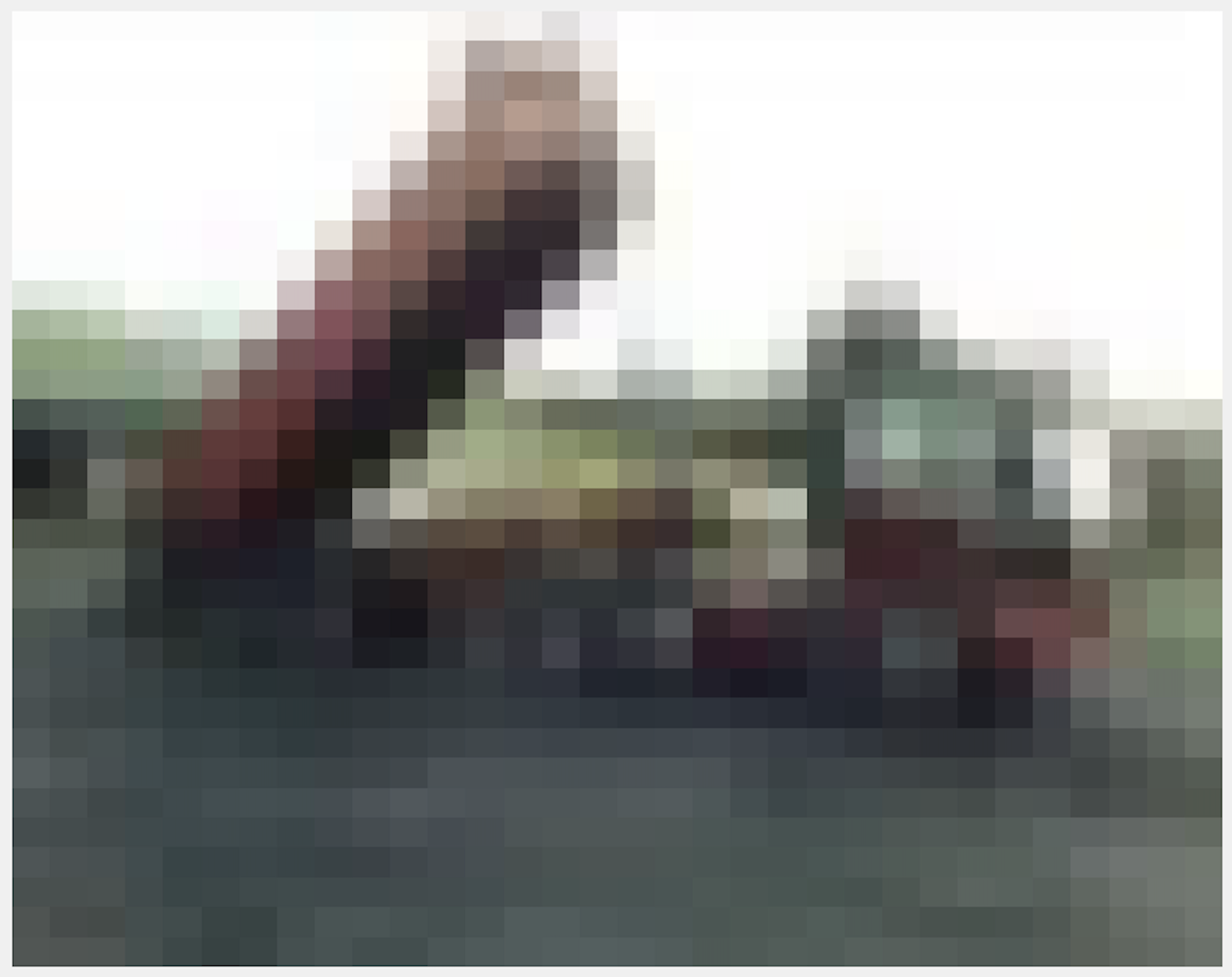}
    &\includegraphics[width=.12\textwidth]{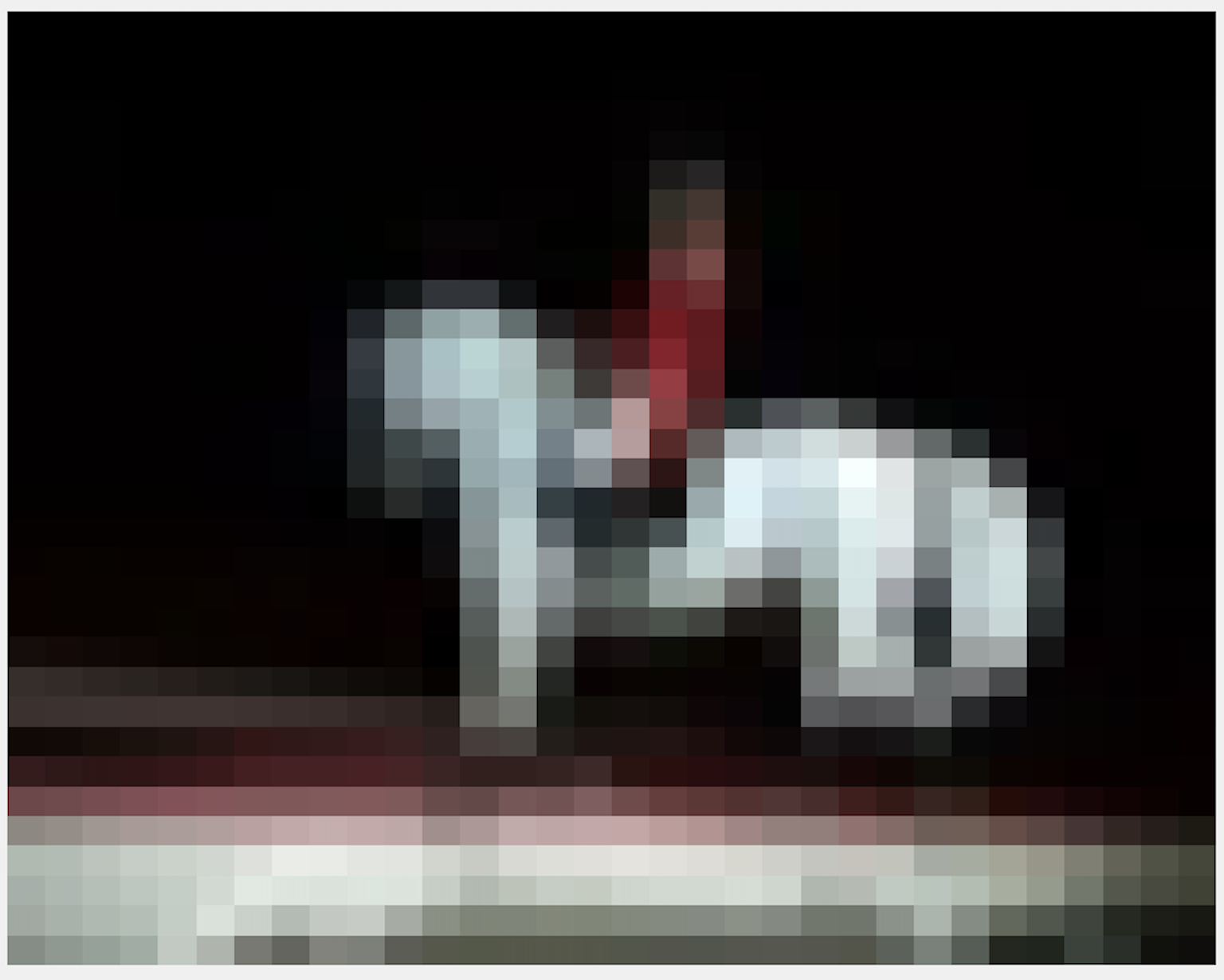}
    \vspace{4mm}
\end{tabular}
CIFAR-10 -  Distribution for number of bars for each bin and each image class\\[1mm]
\begin{tabular}{ccc}
  \includegraphics[width=.3\textwidth]{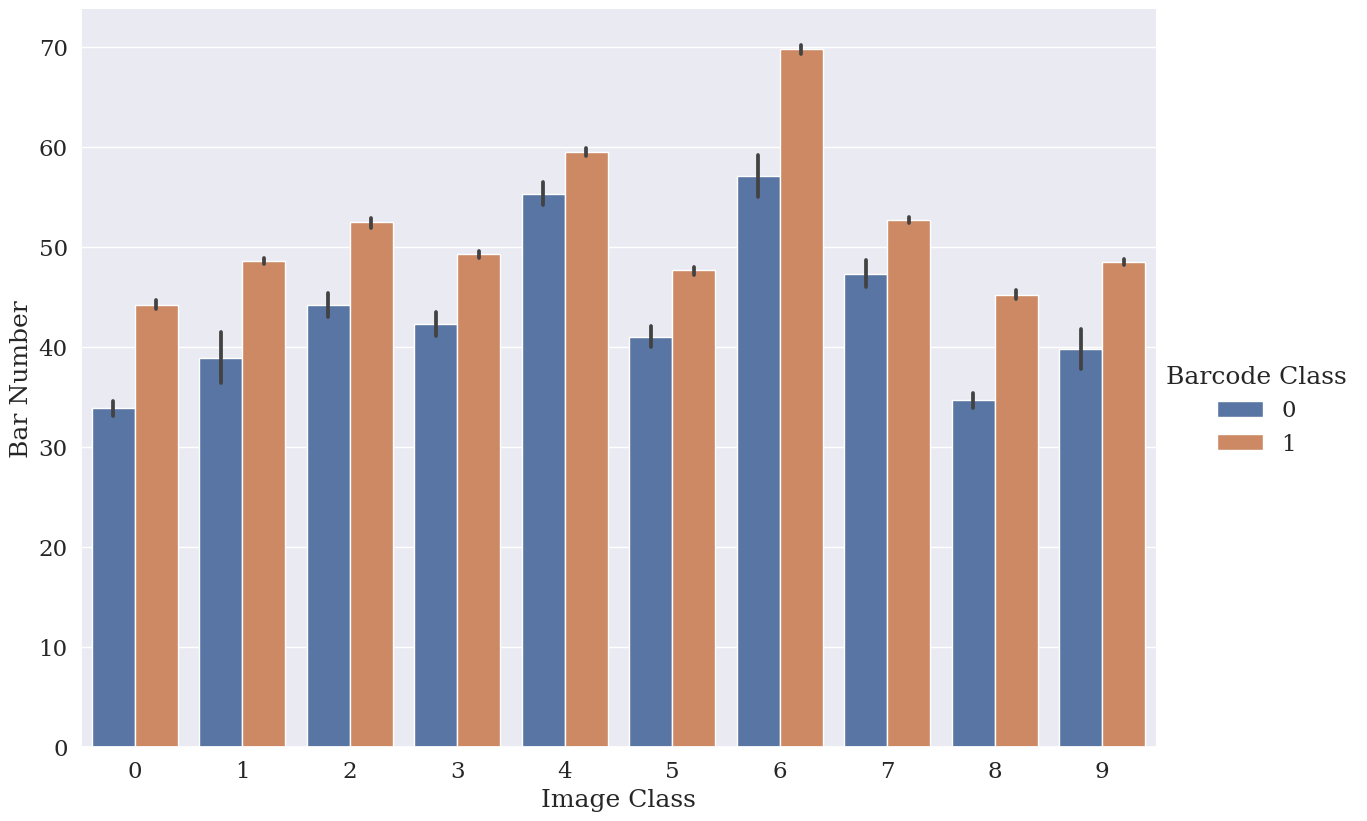} 
    &\includegraphics[width=.3\textwidth]{figures/cifar10_thres0.3.png}
    &\includegraphics[width=.3\textwidth]{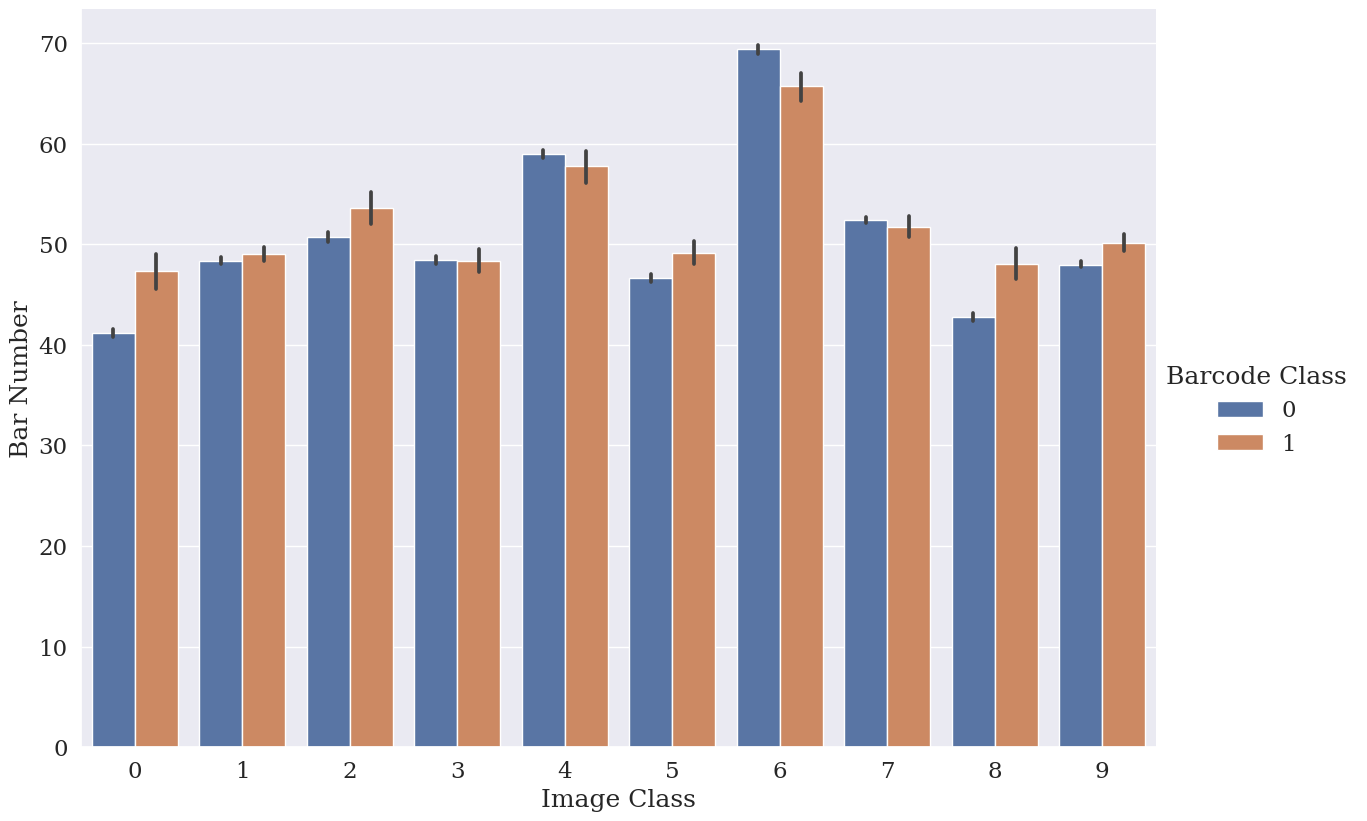}
\end{tabular}
    \caption{Data analysis for CIFAR-10. For columns from left to right, the threshold is 0.15, 0.3 and 0.5.
    Top row: sample distribution in each class for the corresponding thresholds.
    Middle row: image samples in each PH class for the associated thresholds.
    Bottom row: the number of bars of two bins for image class (0,1,2...,9).
    }
    \label{fig:cifar10_classification_data_analysis}
\end{figure}

Here we add figures to the experiments from Section~\ref{SS:images}.1. 
Examples of the training curves for MNIST and CIFAR-10 are shown in Figures~\ref{fig:MNIST_classification_train_val_loss_accuracy} and \ref{fig:cifar10_classification_train_val_loss_accuracy}.
Figures~\ref{fig:MNIST_classification_data_analysis} and \ref{fig:cifar10_classification_data_analysis} show the histogram of the number of bars in barcodes for three thresholds, image samples in two PH classes for the corresponding thresholds and average bar length in two PH classes per image class.

\subsection{Mapping cubical complexes to binary features of persistence diagrams}
\label{app:experiments-CC-FCC}

Here we provide details about the experiments in which we consider inputs taken at an intermediate stage of the computational pipeline, namely cubical complexes (CC) and filtered cubical complexes (FCC). 
This type of data is more naturally handled by graph neural networks.

We view the cubical complex matrix as the adjacency matrix of a graph, and the cells are the nodes of the graph. The label attached to each graph is 0 or 1. Here 0 or 1 indicates whether the bar of the corresponding diagram lies in the interval $[0.1,0.3]$. The percentages of the 0 and 1 classes take up 46\% and 54\%. 
The task is then to train a GNN model to predict class labels from the graph of the cubical complex. We split the data set of 59,986 graph samples into $8:1:1$ for training, validation and test, respectively. We call the graph data set \textbf{CC-MNIST}. 

On the other hand, we can also use the filtered cubical complex $F$ to map the entries of the corresponding cubical complex matrix $E$ to the entries of the adjacency matrix. That is, the $(i,j)$th entry of the adjacency matrix is $E(F_{i,j})$ which is the entry of $E$ located at $F_{i,j}$. We call the resulting graph data set \textbf{FCC-MNIST}. 

We use two graph neural network (GNN) models both with three blocks of a graph convolutional layer plus a pooling layer, followed by three fully connected layers. We use GCNConv \cite{KiWe2017} and GINConv \cite{xu2018how} and combine with TopKPooling \cite{gao2019graph,cangea2018towards} for the 2-graph classification task. In the experiment, the hyperparameters are set as follows: learning rate 0.001, weight decay rate 5e-4, pooling ratio 0.5, number of hidden neurons 128, and the maximum number of training epochs 100. 

\begin{figure}[h!]
    \centering
MNIST CC - predict a bar of some length
\begin{tabular}{ccc}
  \includegraphics[width=.3\textwidth]{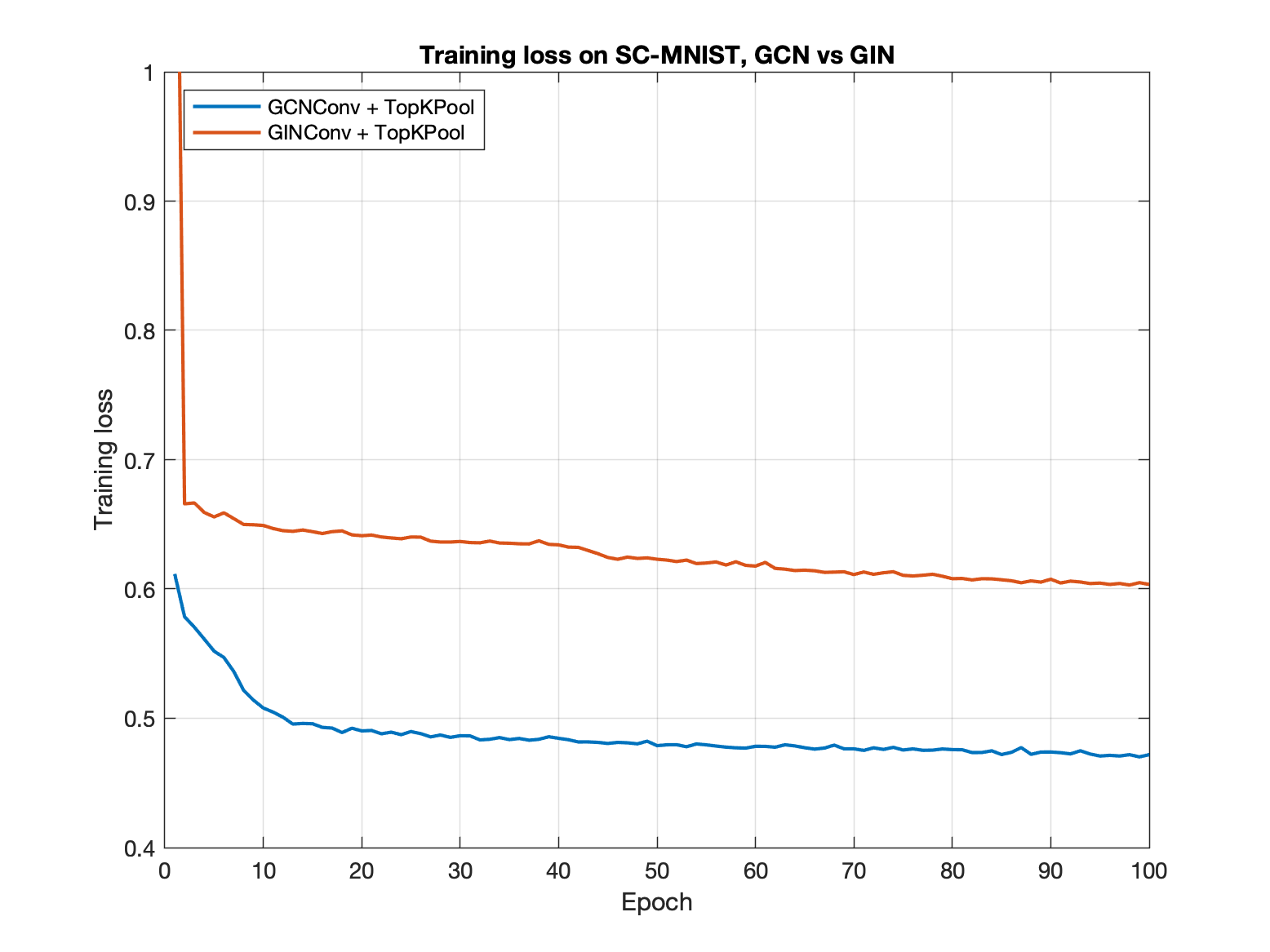}
  \includegraphics[width=.3\textwidth]{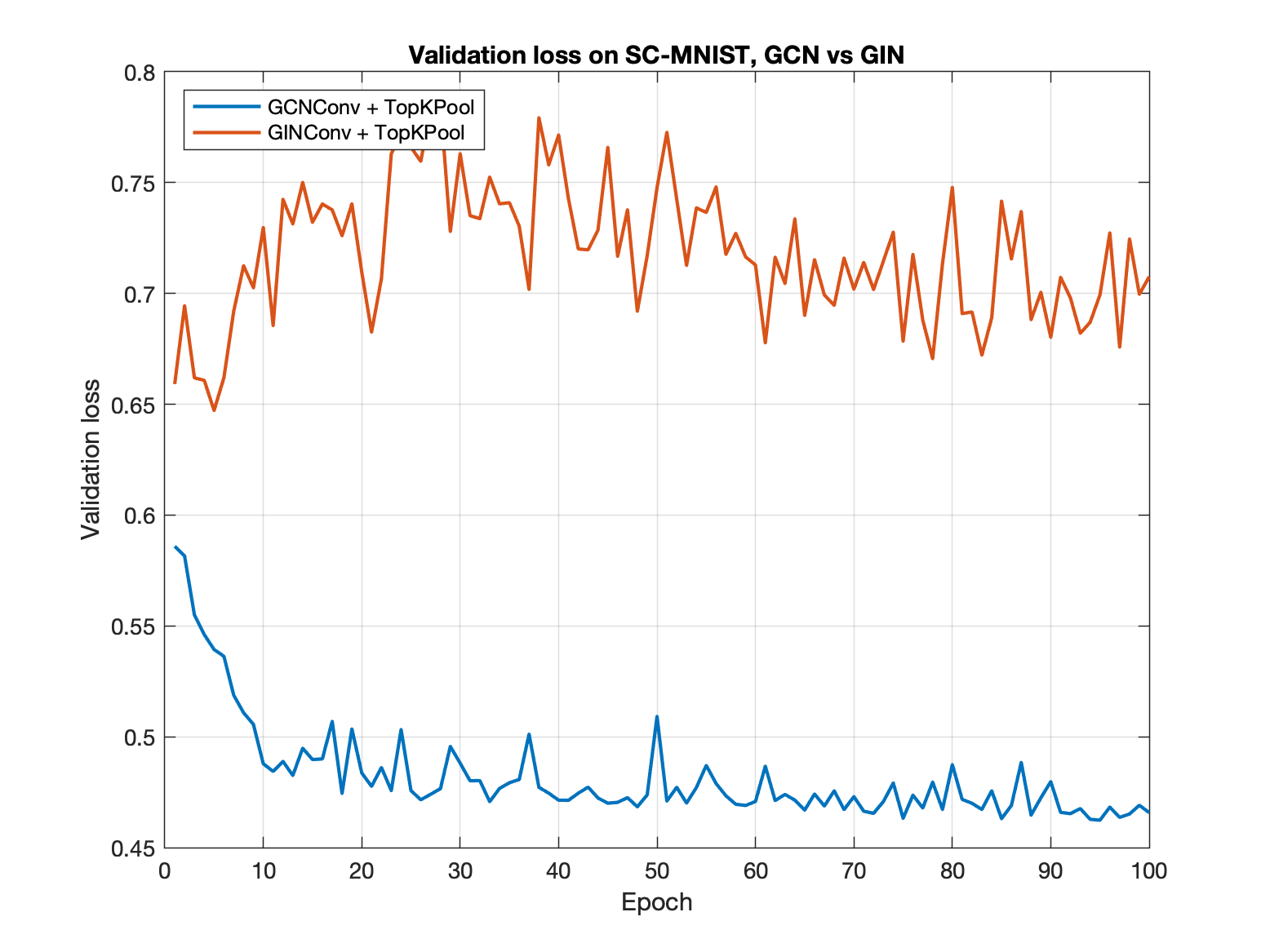}
  \includegraphics[width=.3\textwidth]{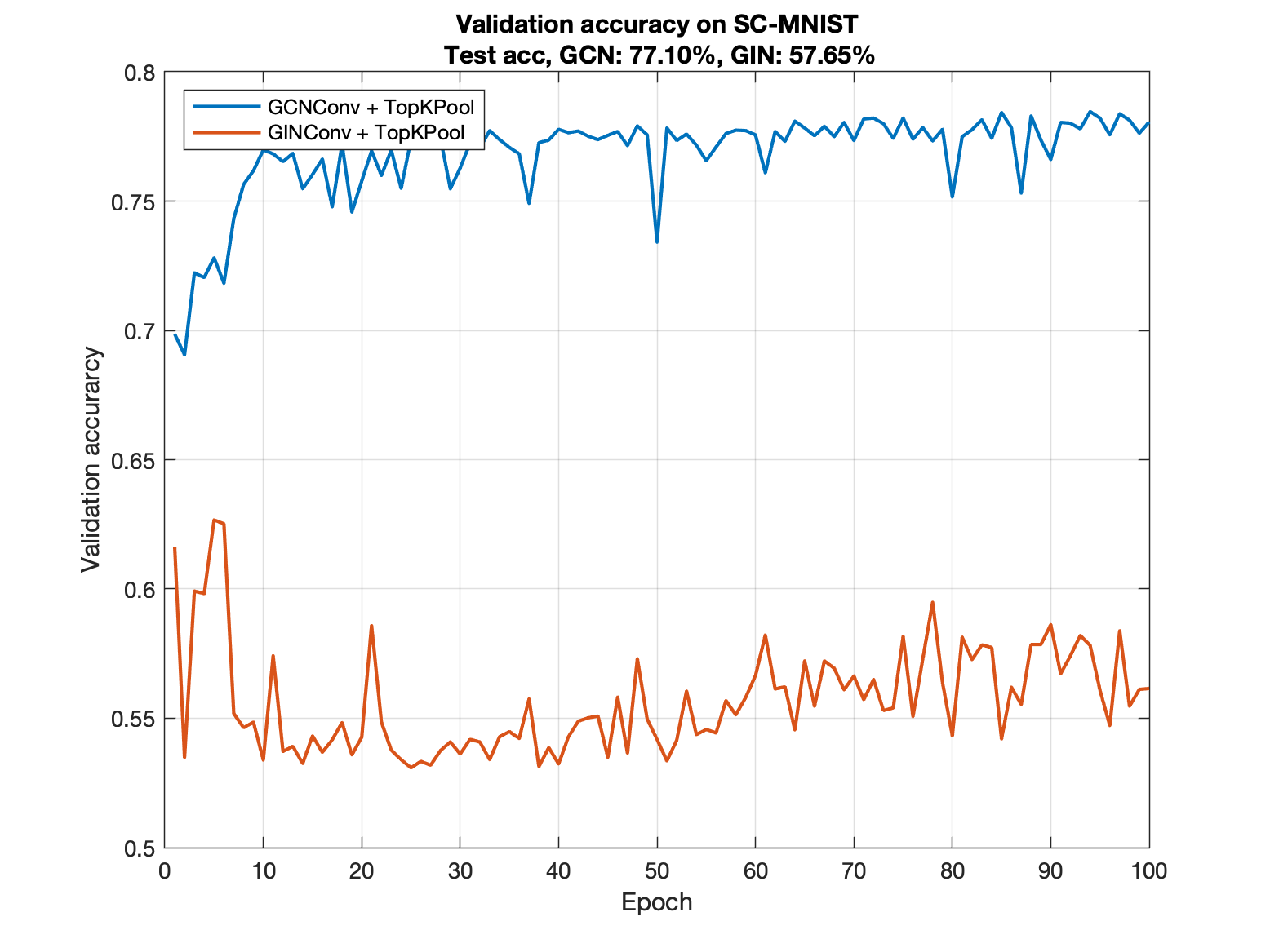}
\end{tabular}\\[2mm]
% This one is for the ordered complex
MNIST FCC - predict a bar of some length 
\begin{tabular}{ccc}
  \includegraphics[width=.3\textwidth]{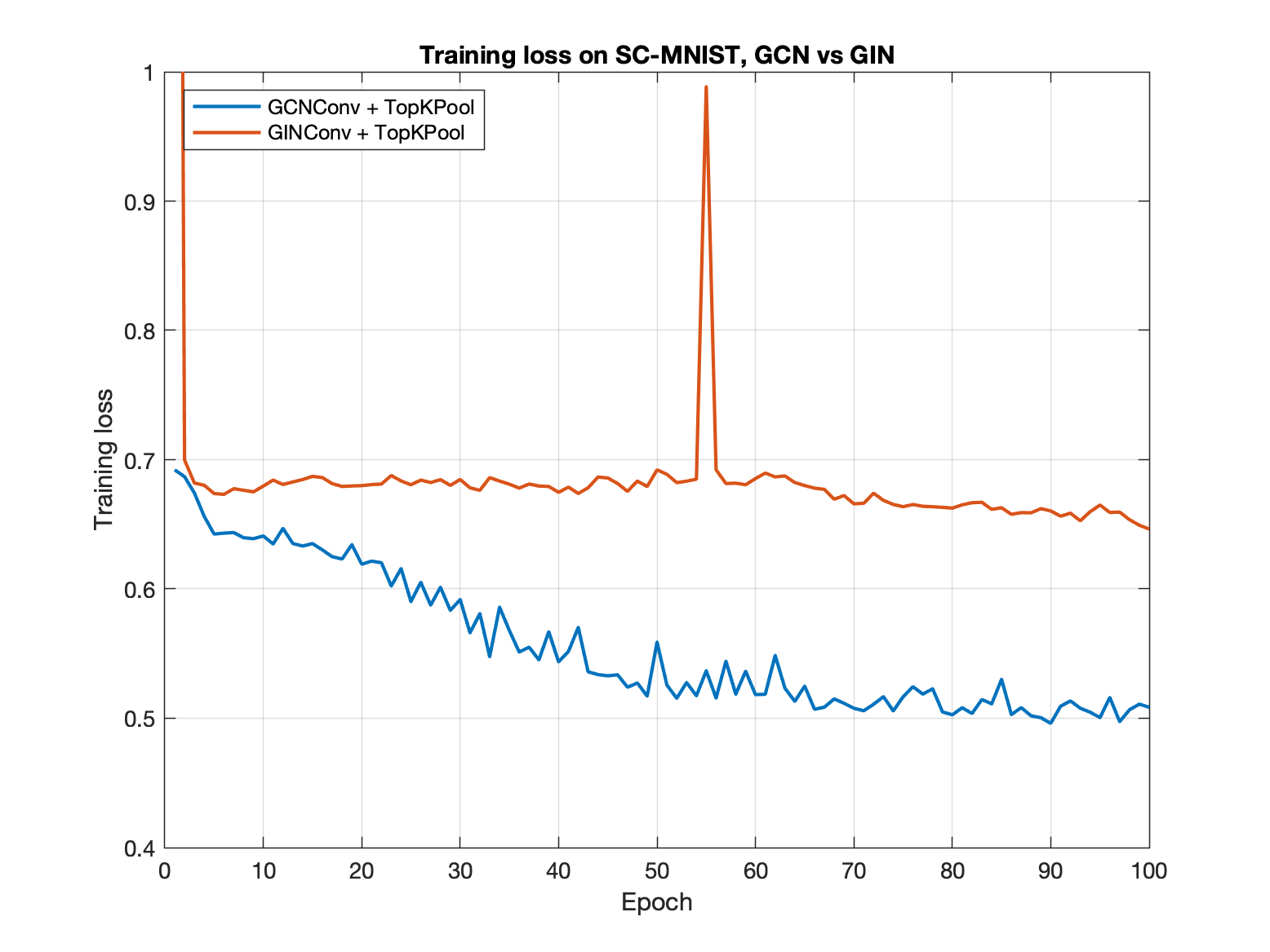}
  \includegraphics[width=.3\textwidth]{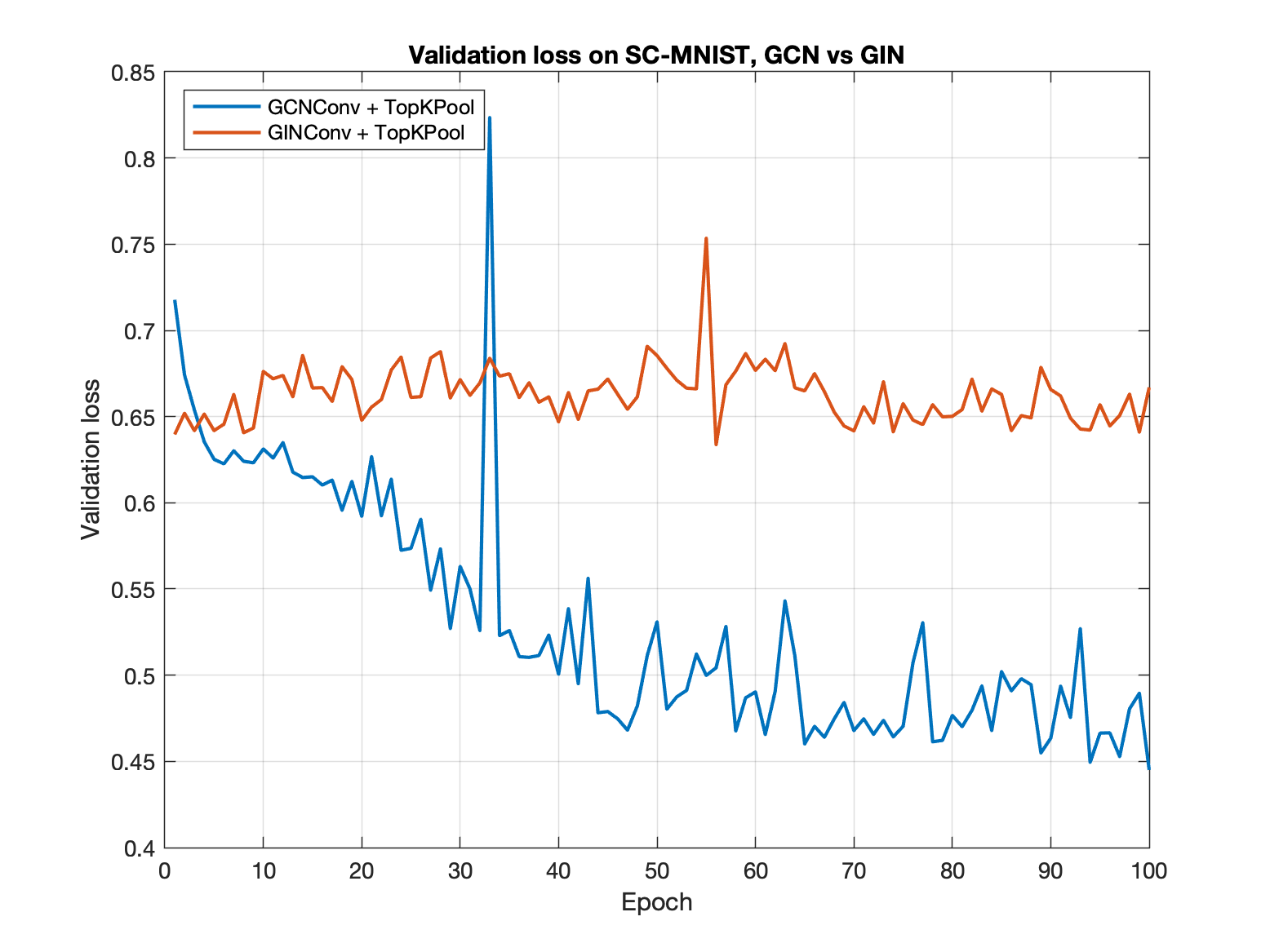}
  \includegraphics[width=.3\textwidth]{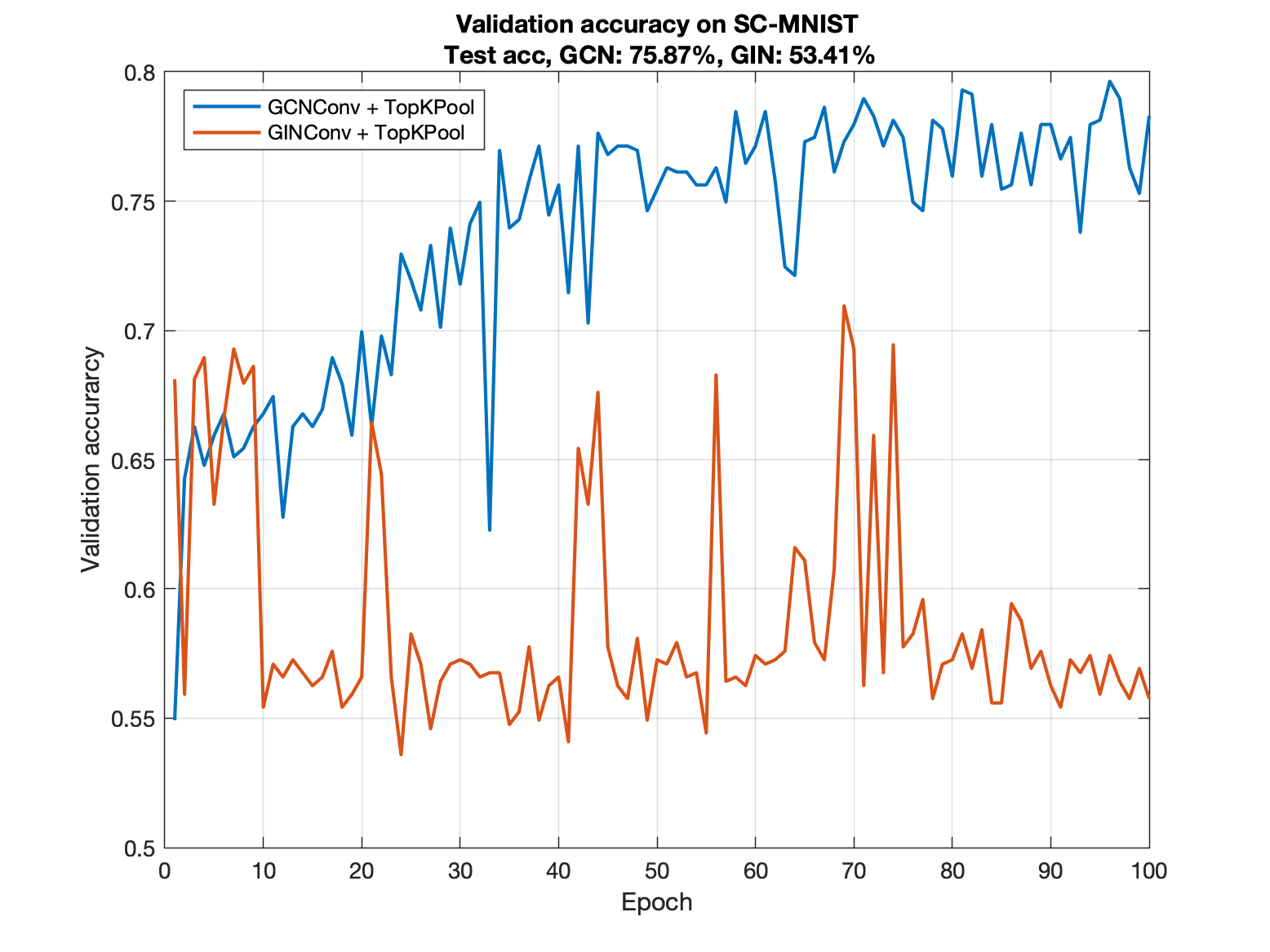}
\end{tabular}
    \caption{Training loss, and validation loss and accuracy of GCN and GIN models, during training for the task of mapping a cubical complex matrix to a number 0 or 1, where 0 or 1 indicates whether at least one bar in the corresponding diagram has length between 0.1 and 0.3. Top panel: the results for CC-MNIST. 
    Bottom panel: the results for FCC-MNIST. In total, 59,986 graphs, samples for training, validation and test taking up 80\%, 10\% and 10\% respectively.}
    \label{fig:MNIST_gnn_0.1_0.3}
\end{figure}

Figure~\ref{fig:MNIST_gnn_0.1_0.3} 
shows the curves of training loss and validation loss and validation accuracy for the two GNN models. 
Here the loss function is the usual multi-class cross-entropy (negative log-likelihood). It shows that the GCN model has much better performance than the GIN model. The GCN model achieves the test accuracy of around $75\%$ on both \textbf{CC-MNIST} and \textbf{FCC-MNIST}. The GIN model only has test accuracy about $51\%$ and $53\%$ on the two data sets.

\subsection{Mapping images to tropical coordinates of their persistence diagrams}
\label{regression_experiment}

The CNN model for MNIST and CIFAR-10 in Section~\ref{sec:regress_tropical}.3 uses early stopping and learning rate reduction strategies for training with patience 30 and 5 respectively. 
The relative test MSE's for tropical coordinates (1)--(5) in Figure~\ref{fig:mnist-cifar-regress} are for MNIST $0.048468, 0.000465, 0.010336, 0.00273689, 0.0017963$ and for CIFAR-10  $0.00398112, 0.00636546, 0.00352925, 0.0026194, 0.00218928$. 

\begin{figure}[h]
    \centering
MNIST - Histogram for tropical coordinates\\
\begin{tabular}{ccccc}
    \includegraphics[width=.16\textwidth]{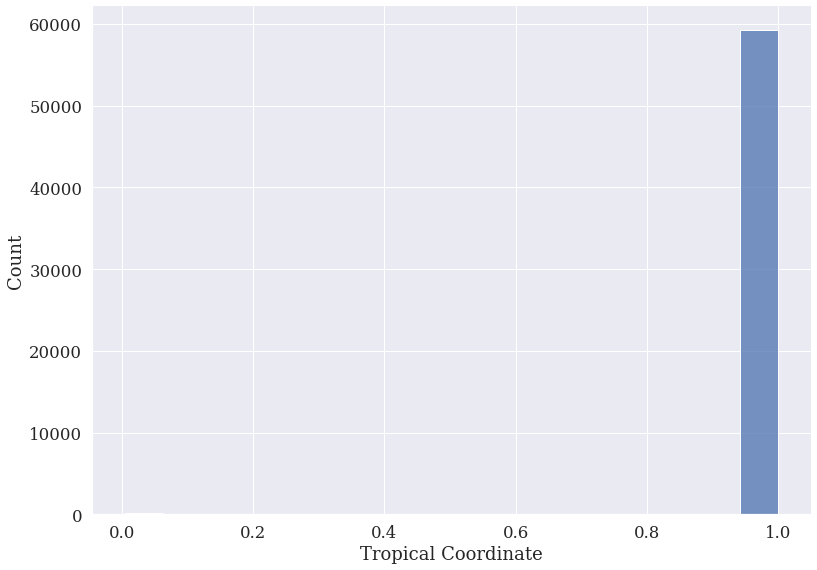}
    &\includegraphics[width=.16\textwidth]{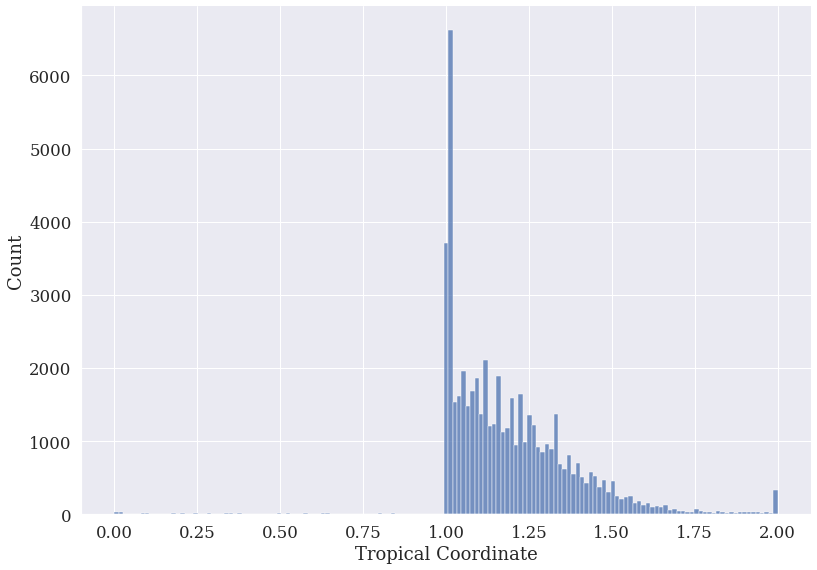}
    &\includegraphics[width=.16\textwidth]{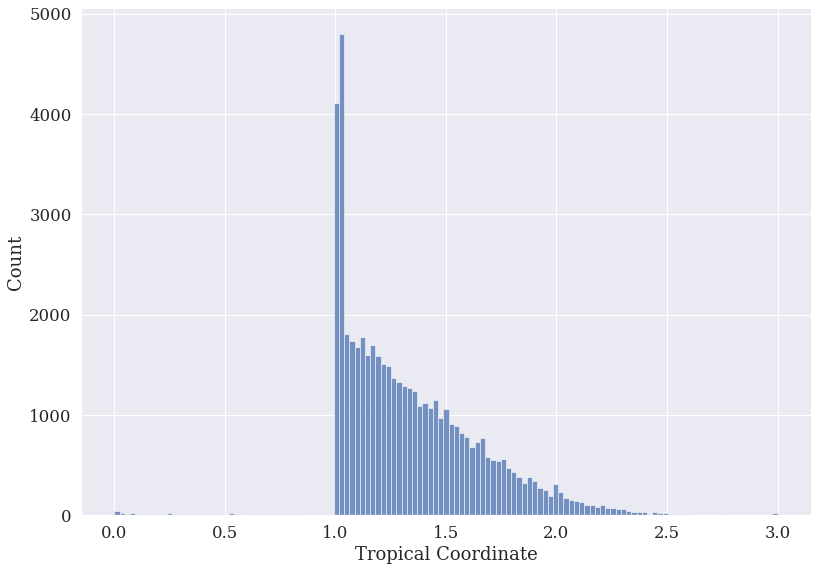}
    &\includegraphics[width=.16\textwidth]{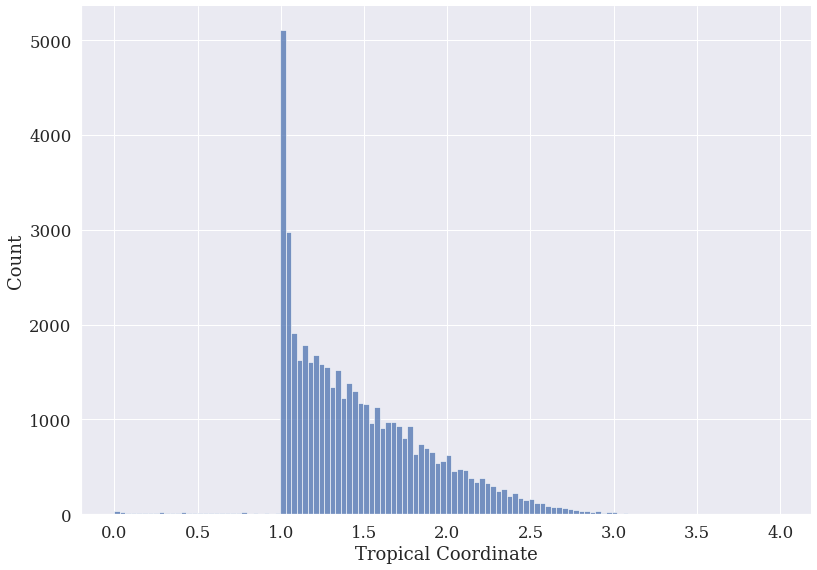}
    &\includegraphics[width=.16\textwidth]{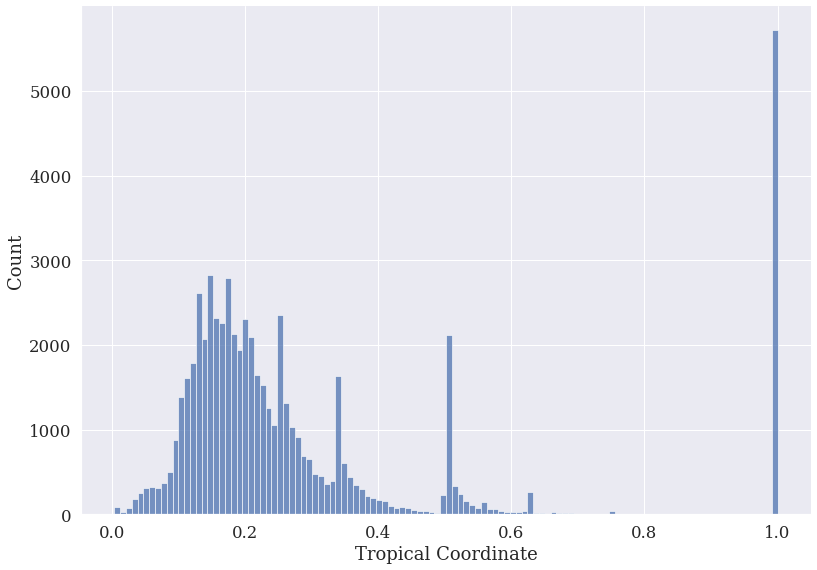} 
\end{tabular}
\vspace{2mm}
MNIST -  Distribution of tropical coordinate for each image class
\begin{tabular}{ccccc}
    \includegraphics[width=.16\textwidth]{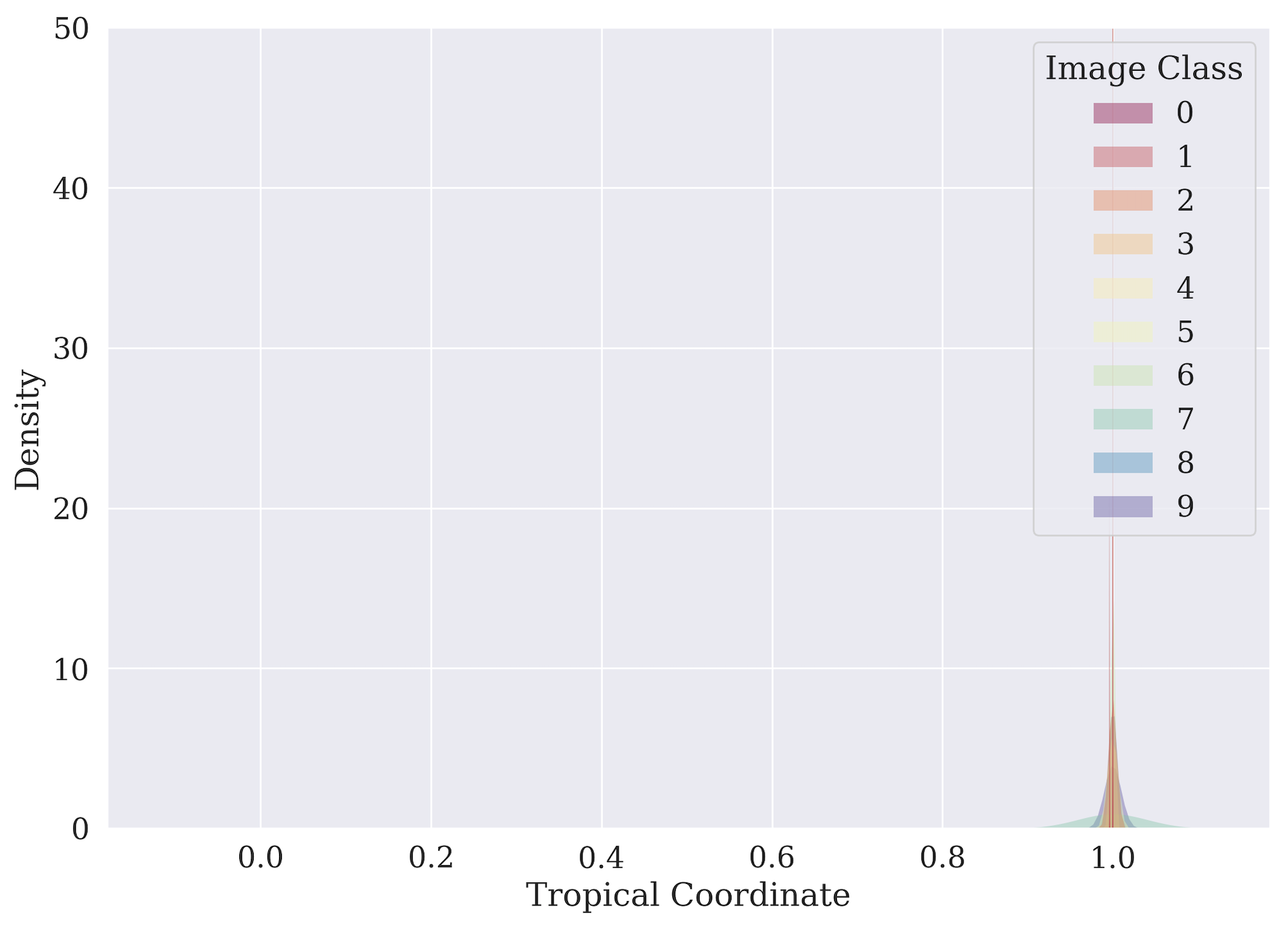}
    &\includegraphics[width=.16\textwidth]{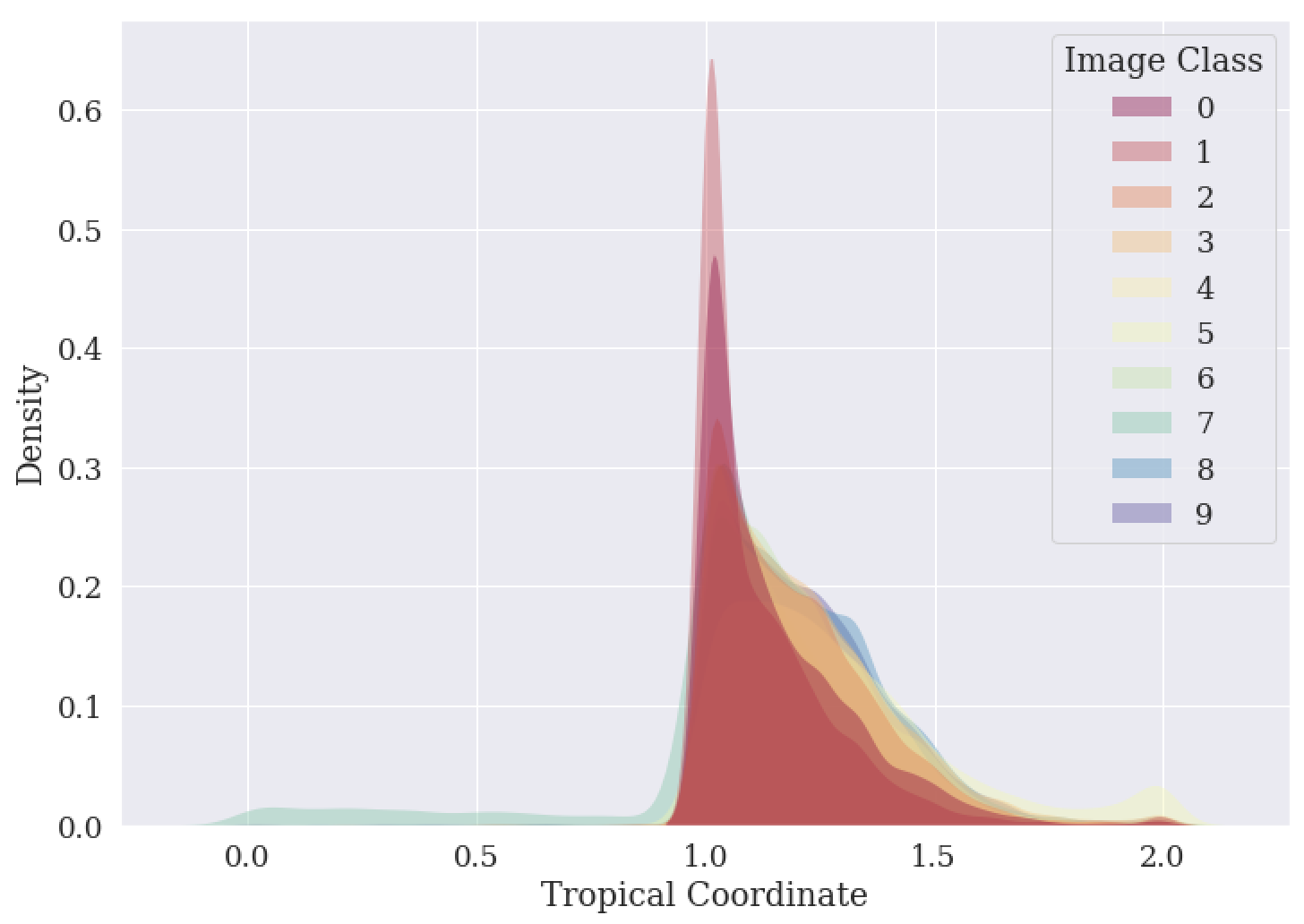}
    &\includegraphics[width=.16\textwidth]{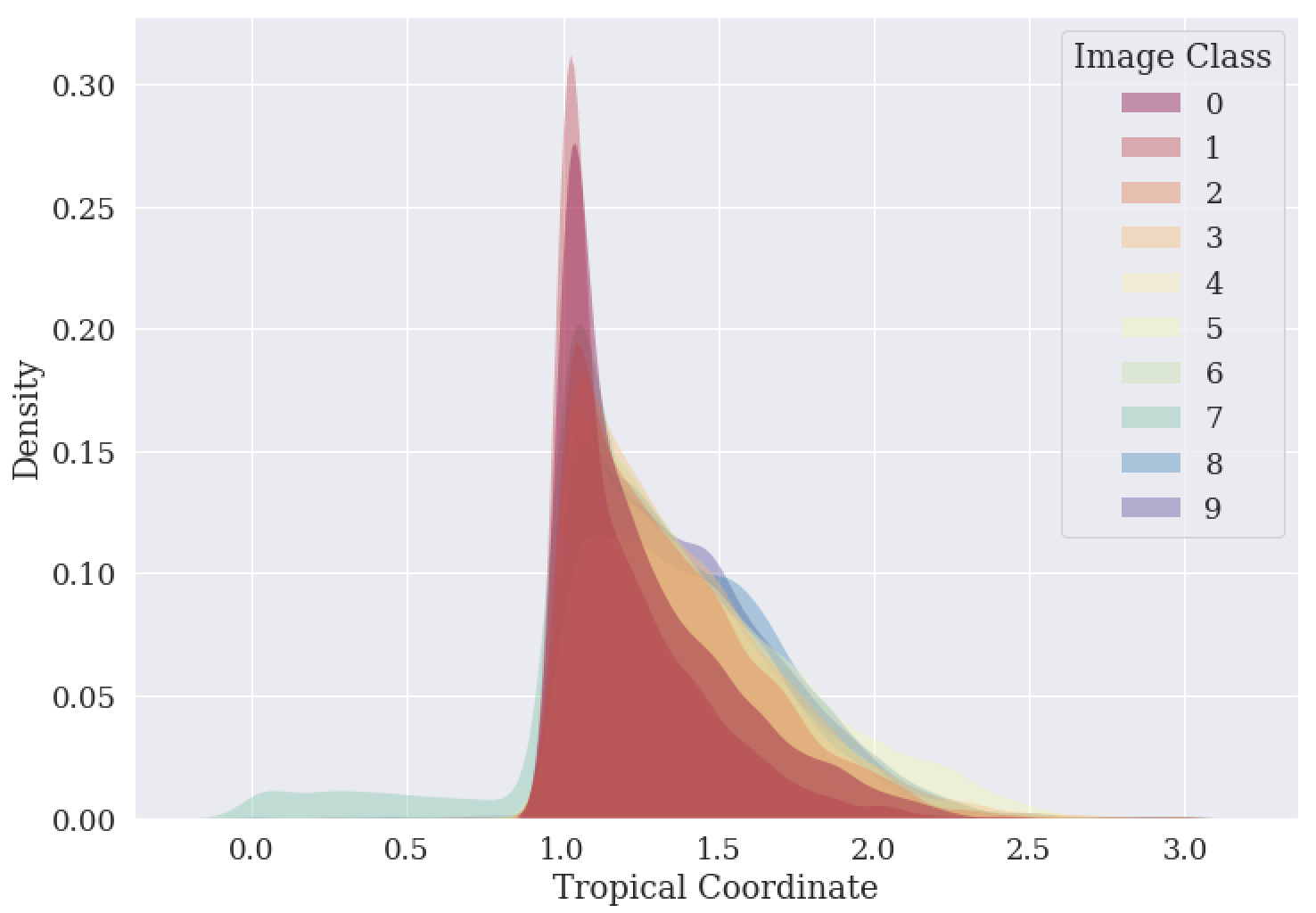}
    &\includegraphics[width=.16\textwidth]{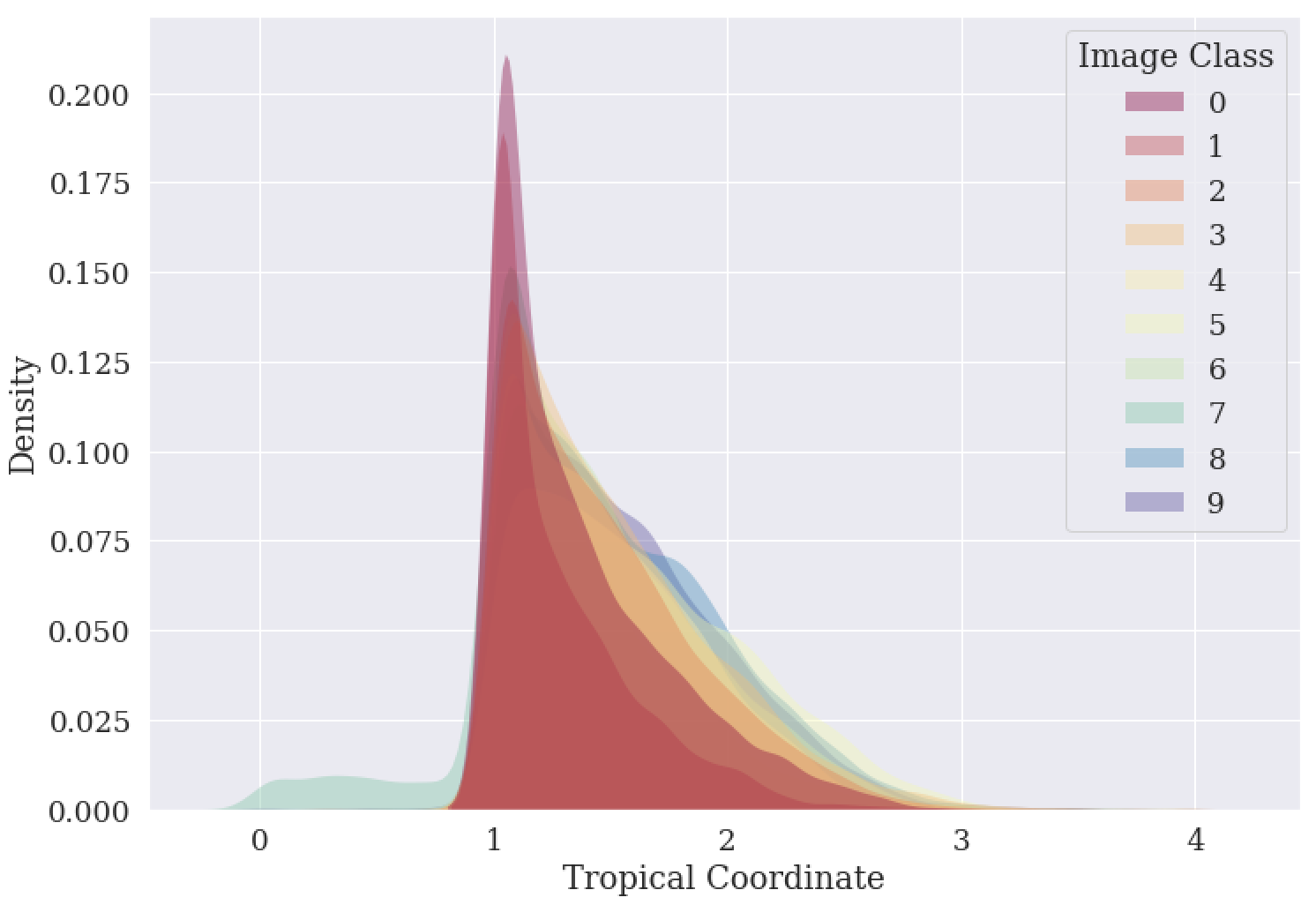}
    &\includegraphics[width=.16\textwidth]{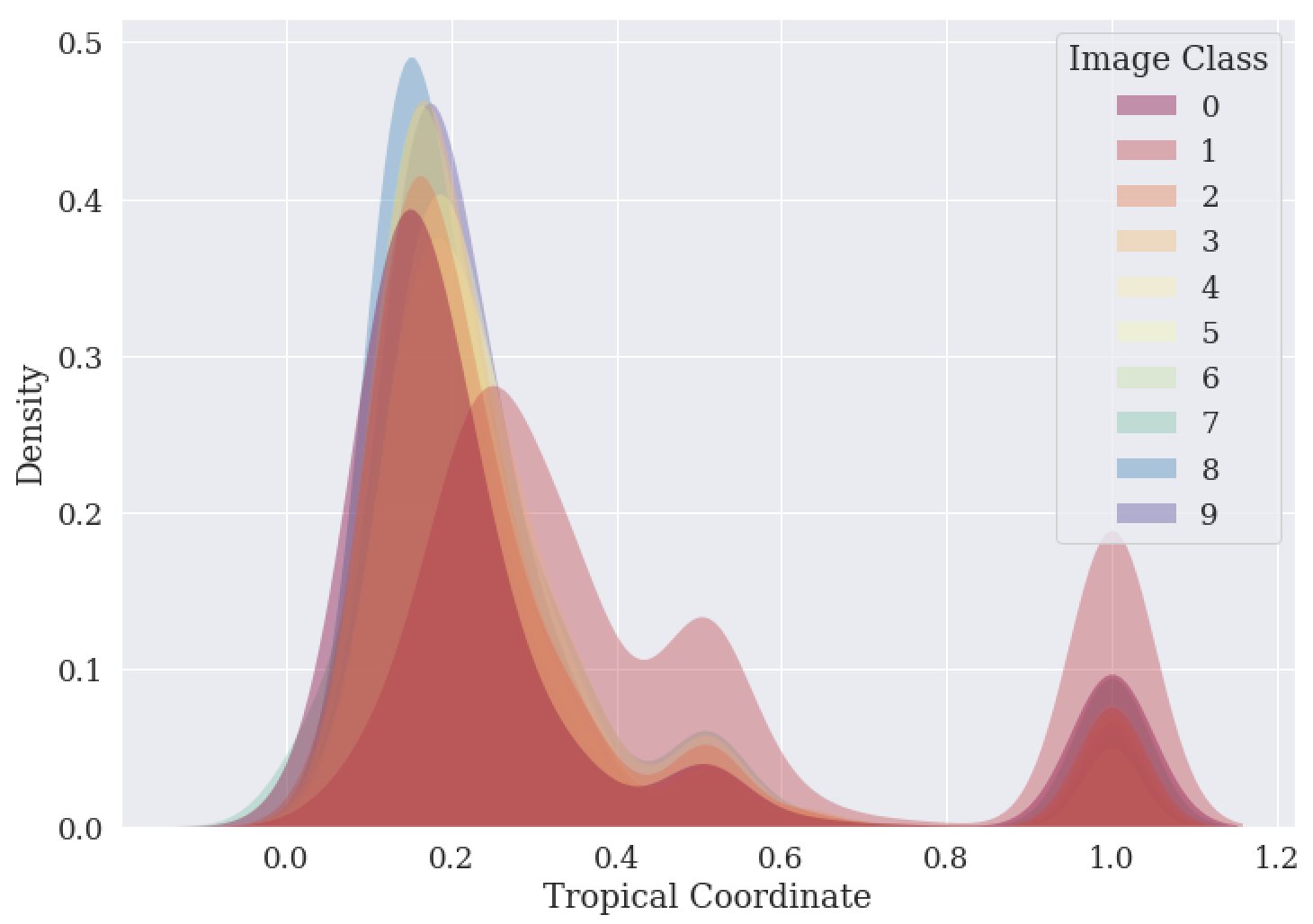} 
\end{tabular}
\vspace{2mm}
CIFAR-10 - Histogram for tropical coordinates\\
\begin{tabular}{ccccc}
    \includegraphics[width=.16\textwidth]{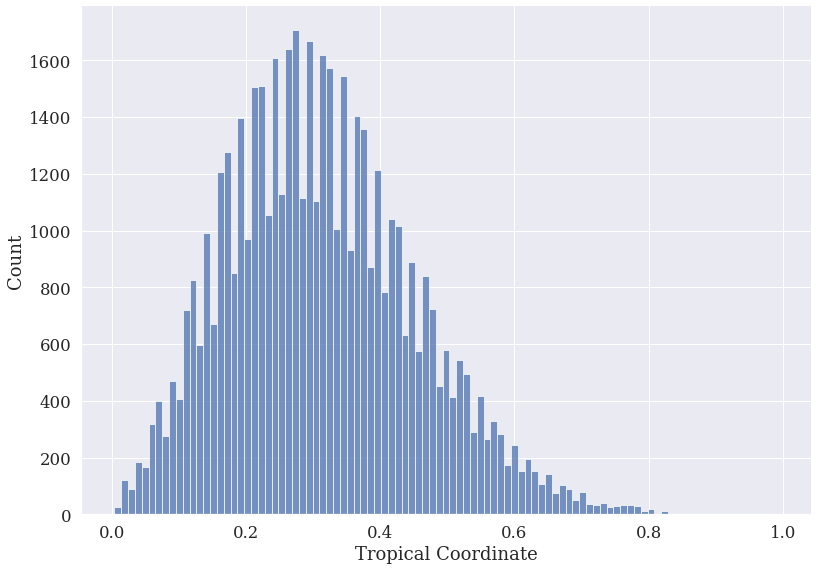}
    &\includegraphics[width=.16\textwidth]{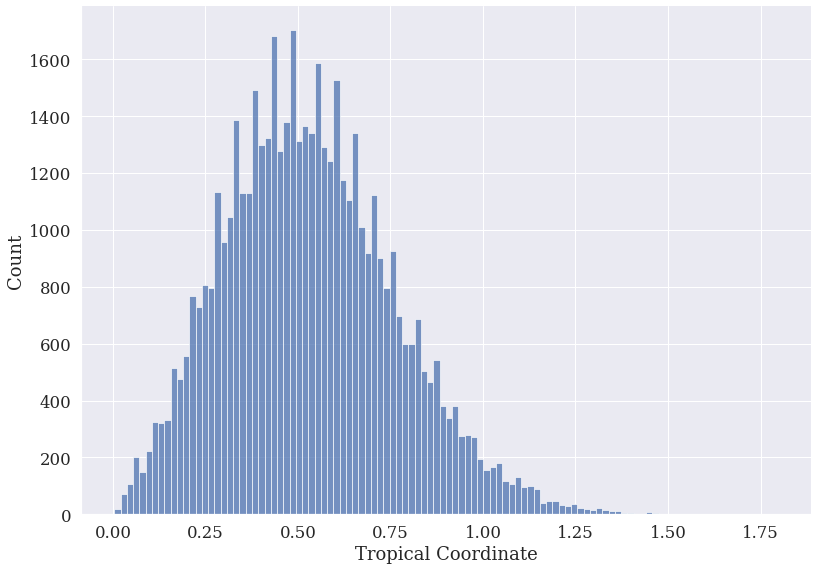}
    &\includegraphics[width=.16\textwidth]{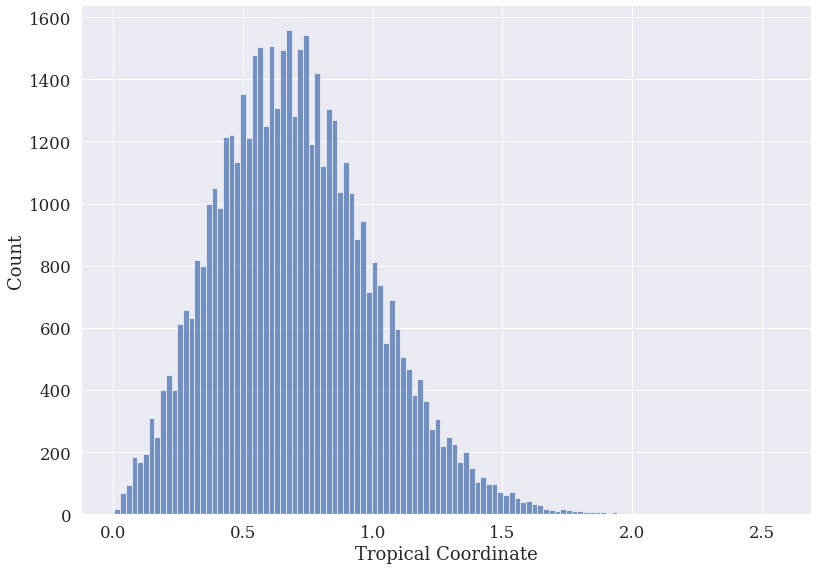}
    &\includegraphics[width=.16\textwidth]{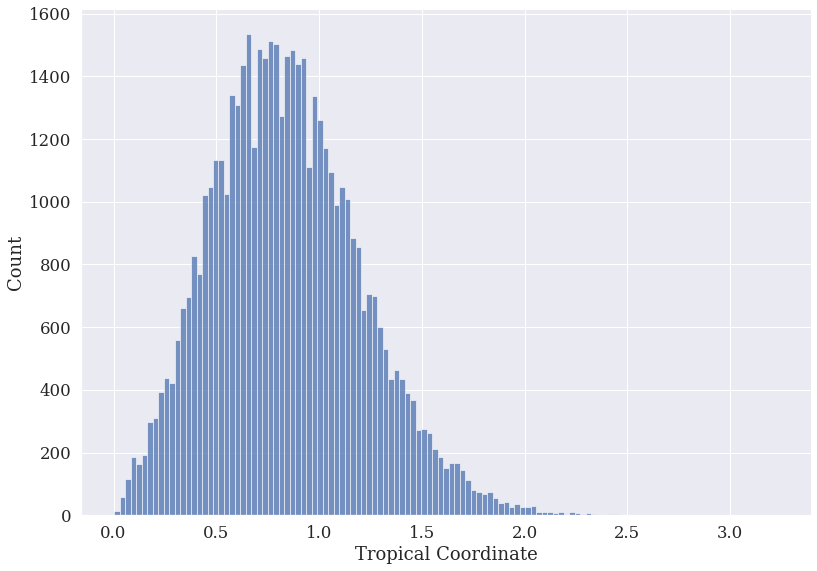}
    &\includegraphics[width=.16\textwidth]{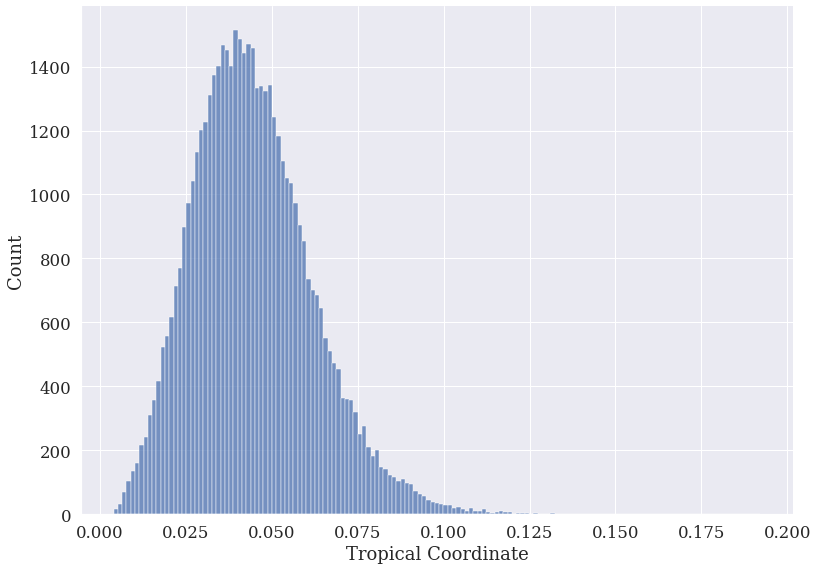} 
\end{tabular}
\vspace{2mm}
CIFAR-10 -  Distribution of tropical coordinate for each image class
\begin{tabular}{ccccc}
    \includegraphics[width=.16\textwidth]{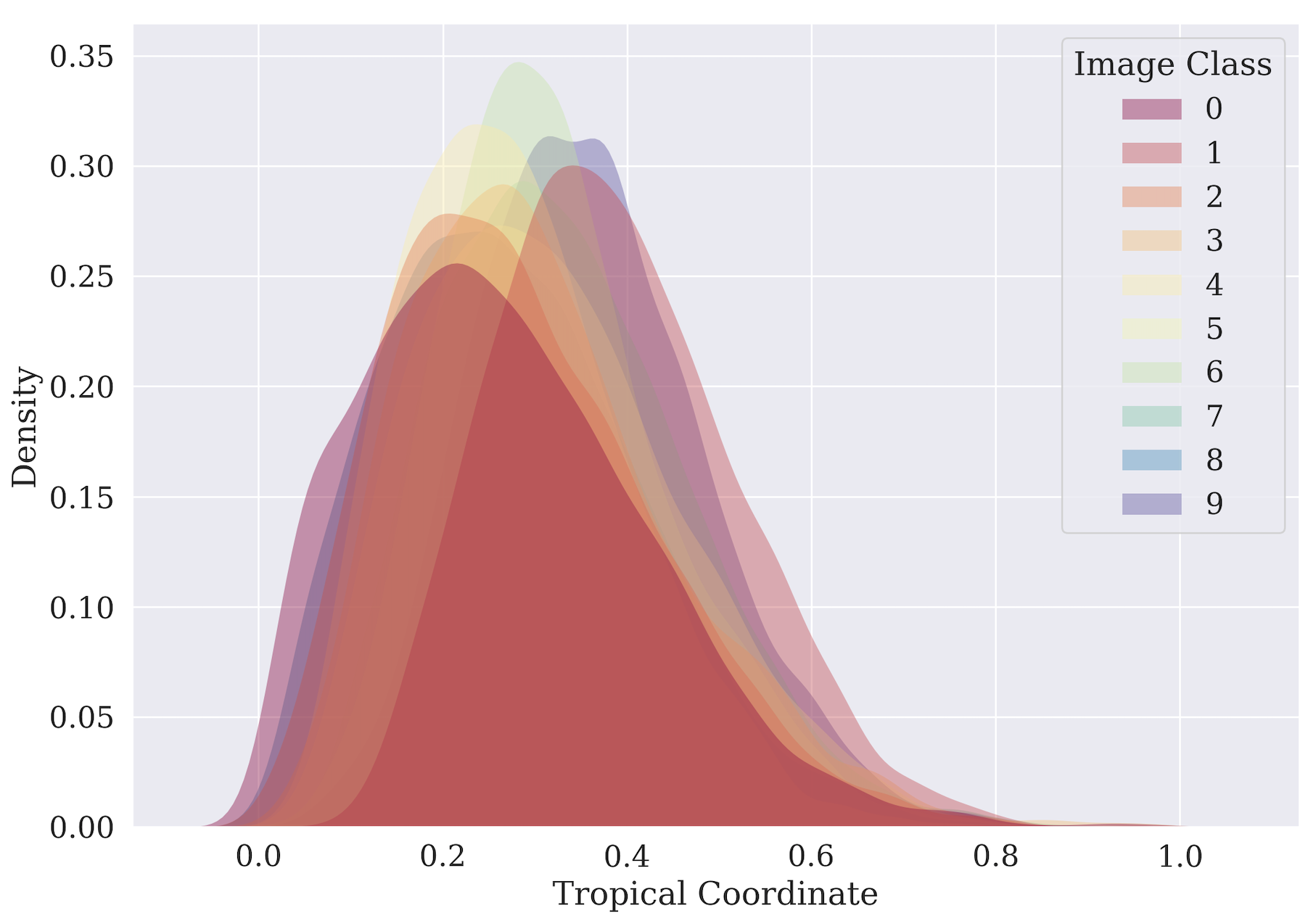}
    &\includegraphics[width=.16\textwidth]{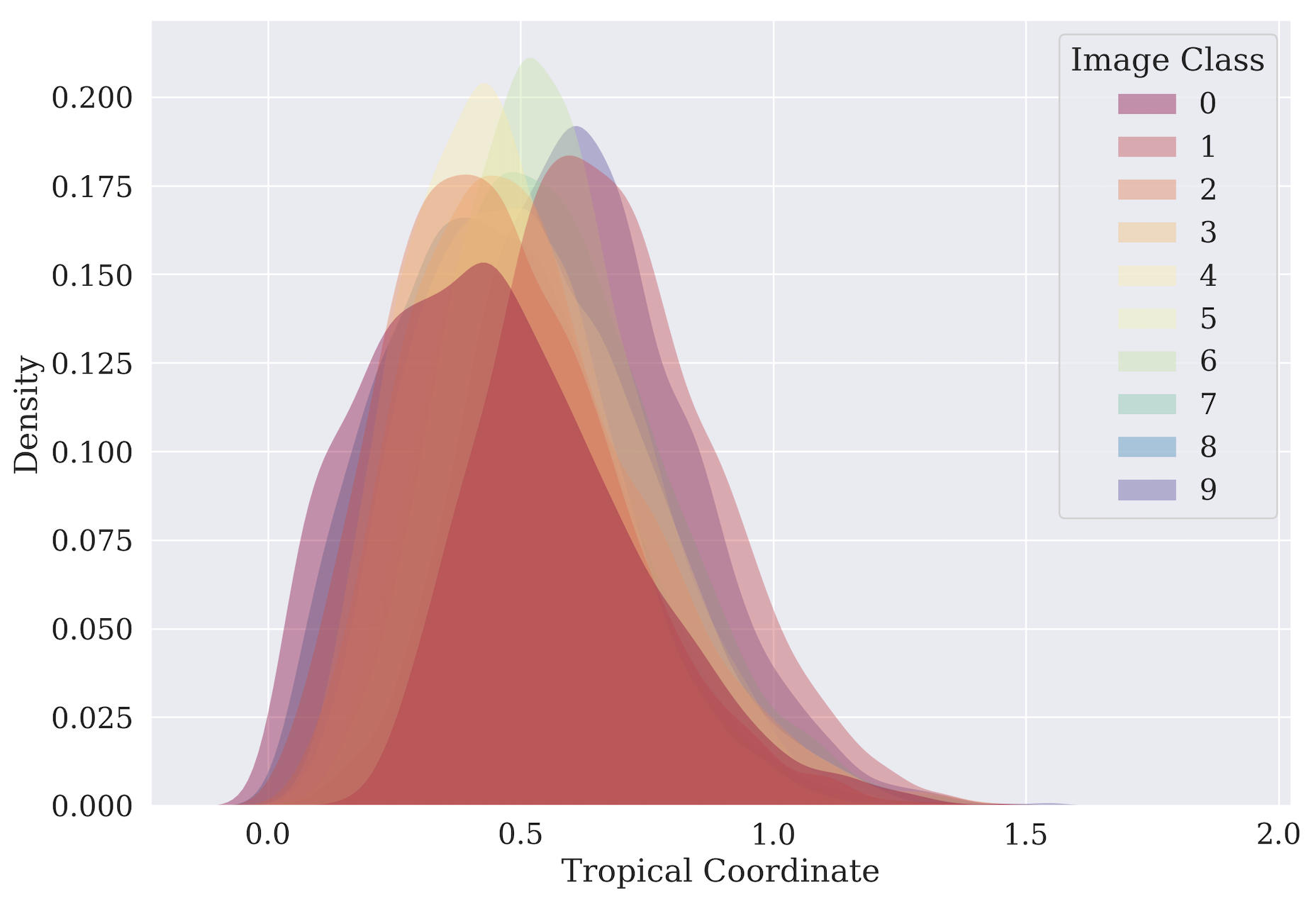}
    &\includegraphics[width=.16\textwidth]{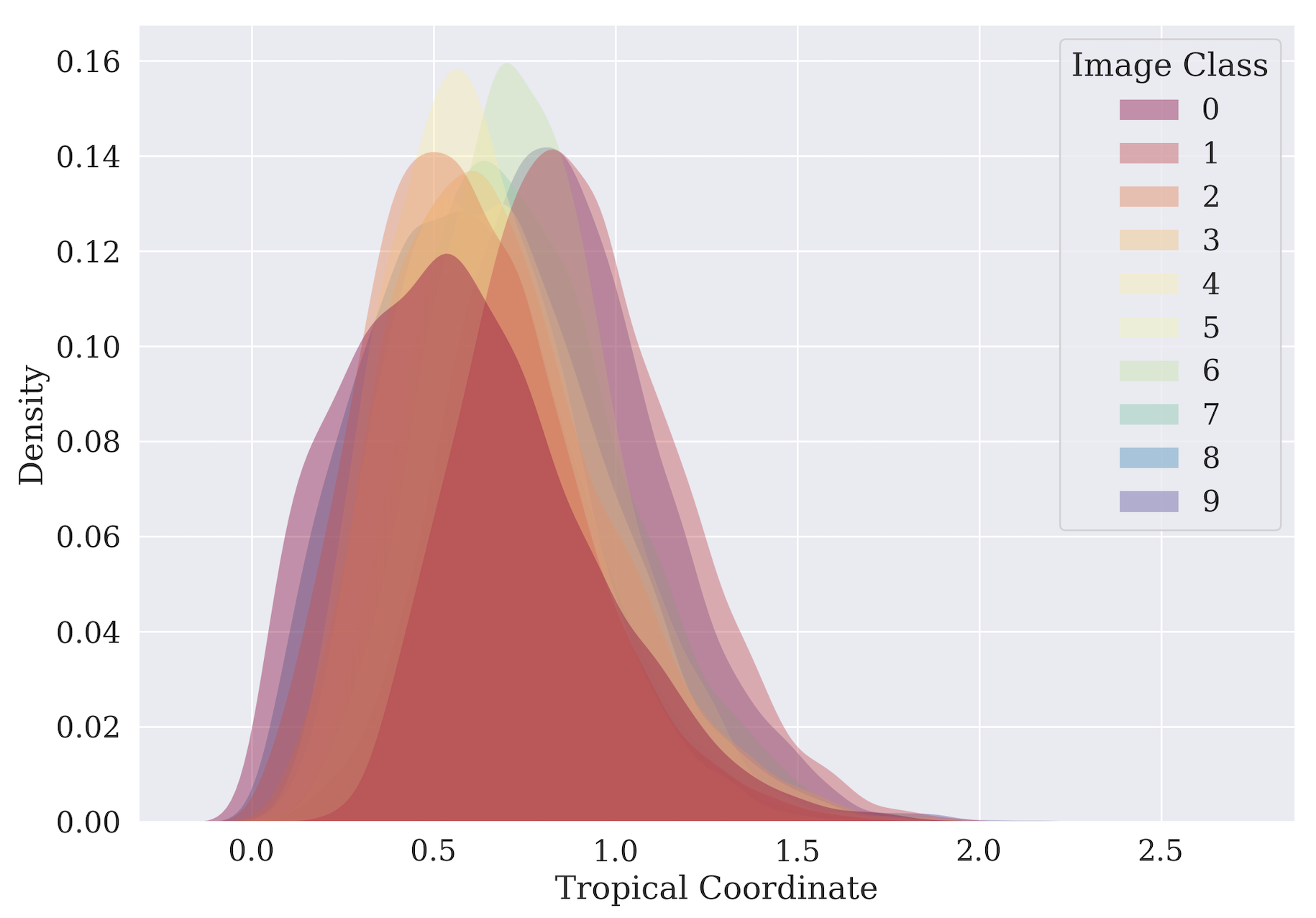}
    &\includegraphics[width=.16\textwidth]{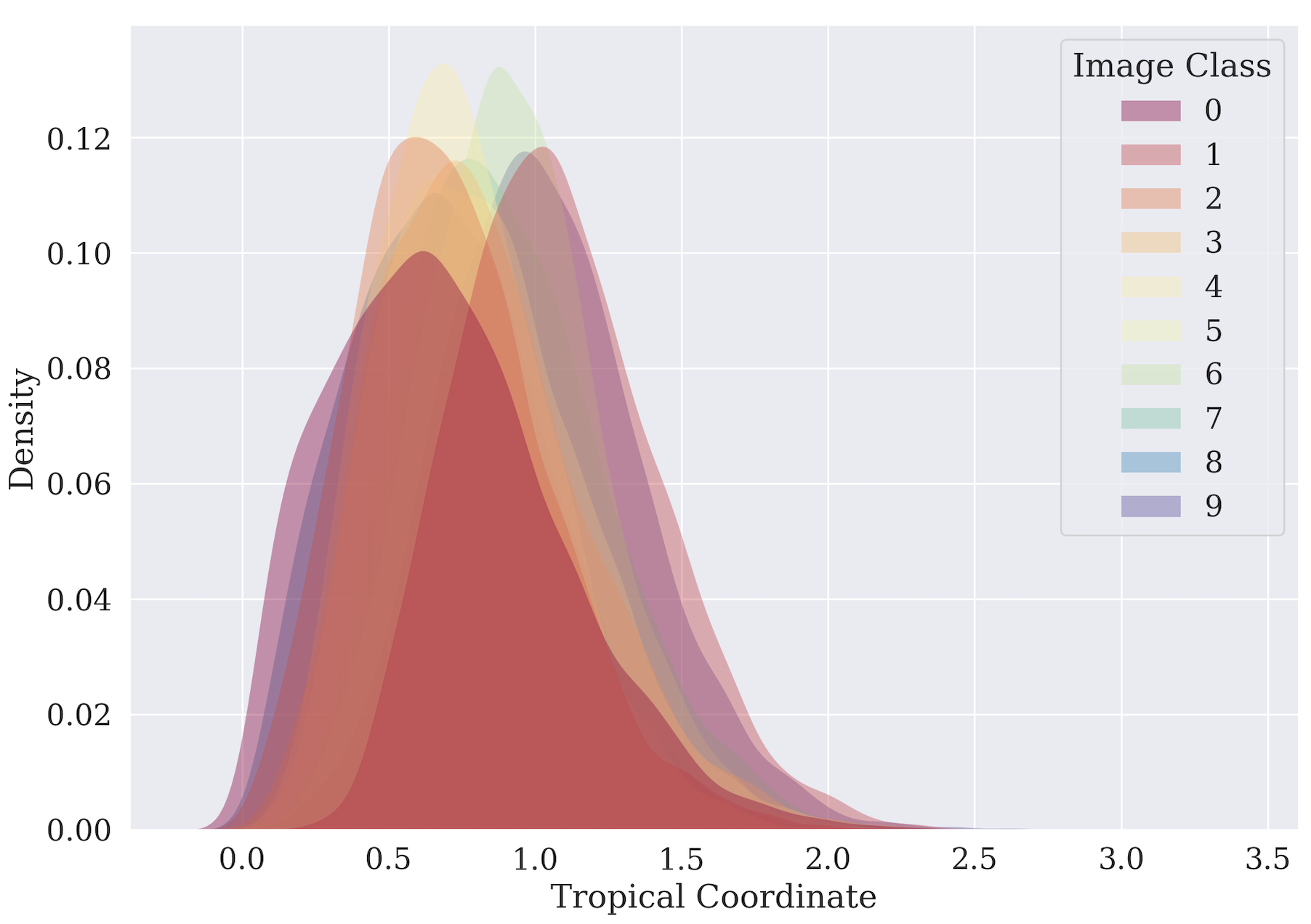}
    &\includegraphics[width=.16\textwidth]{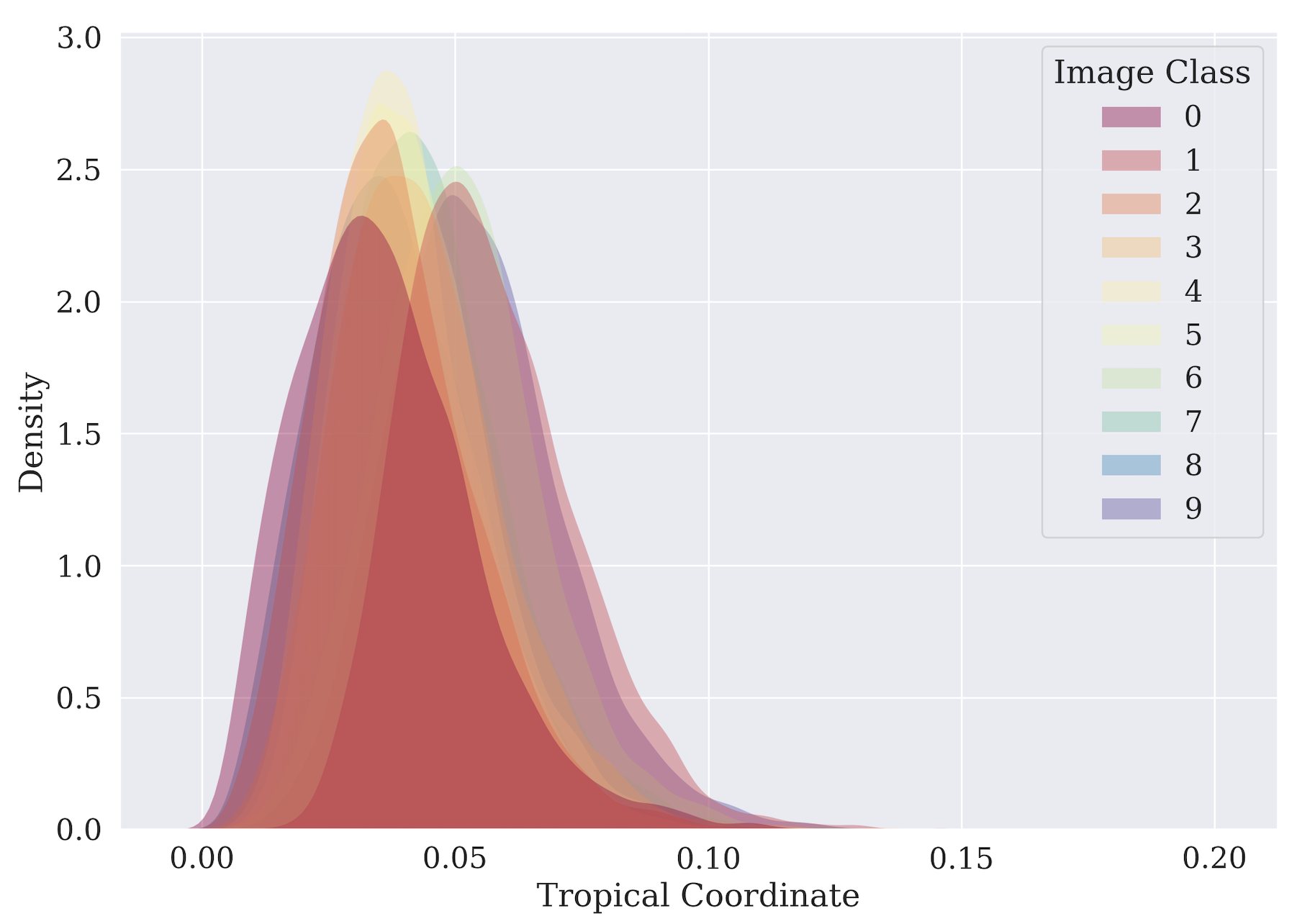} 
\end{tabular}
    \caption{For MNIST and CIFAR-10: First row: Histograms for the value of four tropical coordinates and mean bar length. Second row: Breakdown of the above histogram into the 10 different image classes of the data set.}
    \label{fig:distribution-tropical}
\end{figure}

The distributions of the number of bars and the tropical coordinates for MNIST and CIFAR-10 are shown in Figure~\ref{fig:distribution-tropical}. 

\subsection{Computational time}
\label{app:compute-time} 
We  compare the time it takes to obtain a persistence diagram by traditional methods versus the time it takes to obtain a persistence diagram feature by evaluating a trained neural network. For the traditional methods, we measure the wall-time seconds that it takes to compute persistence diagrams for every image using the library GUDHI, and take averages over the entire data sets. Thus, these times provide a lower bound on the time that it would take to compute a persistence diagram feature from an image using standard methods.
For the NN method, we save the parameters of the trained model and use it to predict the persistence diagram feature from the test image data. The GPU time is recorded for computing the persistence diagram features for the whole test data set, which contains around 60 batches, each of which has 100 samples. It then gives the average time for computing the persistence diagram feature for one batch by the CNN model.
Of course, the neural network needs to be trained first, but once this has been done, it can produce approximate label values for new data very quickly in comparison with traditional methods. 

\end{document}